\documentclass{article}
\PassOptionsToPackage{dvipsnames}{xcolor}
\usepackage{amssymb}
\usepackage[utf8]{inputenc}
\usepackage{main}
\usepackage[nopatch=footnote]{microtype}
\usepackage{subcaption}
\usepackage{graphicx}
\usepackage{times}
\usepackage{latexsym}
\usepackage{amsmath}
\usepackage{float}
\usepackage{footnote}
\usepackage{pifont}
\usepackage{enumitem}
\usepackage{bm}
\usepackage{arydshln}
\usepackage{booktabs}
\usepackage{multicol}
\usepackage{multirow}
\usepackage{color} 
\usepackage{colortbl}
\usepackage{bbding}
\usepackage{makecell}
\usepackage{mathtools}
\usepackage{imakeidx}
\usepackage{longtable}
\usepackage{wrapfig}
\usepackage{dirtree}
\makeindex
\usepackage{arydshln}
\usepackage{lipsum}
\usepackage{natbib}
\usepackage[edges]{forest}
\usepackage[normalem]{ulem}
\definecolor{mydarkblue}{rgb}{0,0.08,0.45}
\usepackage[colorlinks=true,linkcolor=mydarkblue,citecolor=mydarkblue,filecolor=mydarkblue,urlcolor=mydarkblue]{hyperref}
\usepackage{CJKutf8}
\usepackage{awesomebox} 
\usepackage{bbding}
\usepackage[most]{tcolorbox}
\usepackage{booktabs}
\usepackage{geometry}
\definecolor{wkblue}{rgb}{0.2, 0.3, 0.6}
\definecolor{meta-color}{rgb}{0.5, 0.5, 0.5}
\usepackage{amsmath}
\usepackage{enumitem}
\usepackage[ruled,vlined]{algorithm2e}
\usepackage{amsmath}
\usepackage{amssymb}
\usepackage{setspace}
\usepackage{fontawesome5}
\usepackage{tcolorbox}
\tcbset{
  aibox/.style={
    width=\linewidth,
    top=8pt,
    bottom=4pt,
    colback=blue!6!white,
    colframe=black,
    colbacktitle=black,
    enhanced,
    center,
    attach boxed title to top left={yshift=-0.1in,xshift=0.15in},
    boxed title style={boxrule=0pt,colframe=white,},
  }
}

\newtcolorbox{AIbox}[2][]{aibox,title=#2,#1}

\definecolor{brickred}{RGB}{178, 34, 34}       
\definecolor{emojiblue}{RGB}{77, 177, 255}    
\definecolor{pastelpurple}{RGB}{173, 151, 204} 
\definecolor{forestgreen}{RGB}{34, 139, 34}
\definecolor{darkred}{RGB}{139,0,0}

\newcommand{\blackcross}{\ding{55}}
\newcommand{\blackcheck}{\ding{51}}

\newcommand{\brickcheck}{\textcolor{brickred}{\ding{51}}}
\newcommand{\brickcross}{\textcolor{brickred}{\ding{55}}}
\newcommand{\redcross}{\textcolor{red}{\ding{55}}}
\newcommand{\greencheck}{\textcolor{forestgreen}{\ding{51}}}

\newcommand{\fire}{\textcolor{brickred}{\faFire}}  
\newcommand{\ice}{\textcolor{emojiblue}{\faSnowflake}}  

\newcommand{\ruleicon}{{\color{RoyalBlue}\faRuler}}  
\newcommand{\modelicon}{{\color{Plum}\faRobot}}  
\newcommand{\outcomeicon}{{\color{Maroon}\faBullseye}}  
\newcommand{\processicon}{{\color{ForestGreen}\faChartLine}}  

\usepackage{tikz}
\usetikzlibrary{arrows.meta,positioning,calc,shapes,shadows, shapes.misc}
\usepackage[framemethod=tikz]{mdframed}
\definecolor{bgblue}{RGB}{245,243,253}
\definecolor{ttblue}{RGB}{91,194,224}

\mdfdefinestyle{mystyle}{%
  rightline=true,
  innerleftmargin=10,
  innerrightmargin=10,
  outerlinewidth=3pt,
  topline=false,
  rightline=true,
  bottomline=false,
  skipabove=\topsep,
  skipbelow=\topsep
}

\newtcolorbox{myboxi}[1][]{
  breakable,
  title=#1,
  colback=red!5,
  colbacktitle=red!5,
  coltitle=black,
  fonttitle=\bfseries,
  bottomrule=0pt,
  toprule=0pt,
  leftrule=2pt,
  rightrule=2pt,
  titlerule=0pt,
  arc=0pt,
  outer arc=0pt,
  colframe=red,
}

\newtcolorbox{myboxnote}[1][]{
  breakable,
  title=#1,
  colback=orange!0,
  colbacktitle=orange!0,
  coltitle=black,
  fonttitle=\bfseries,
  bottomrule=0pt,
  toprule=0pt,
  leftrule=2pt,
  rightrule=2pt,
  titlerule=0pt,
  arc=0pt,
  outer arc=0pt,
  colframe=orange,
}

\newtcolorbox{myboxii}[1][]{
  breakable,
  freelance,
  title=#1,
  colback=white,
  colbacktitle=white,
  coltitle=black,
  fonttitle=\bfseries,
  bottomrule=0pt,
  boxrule=0pt,
  colframe=white,
  overlay unbroken and first={
  \draw[red!75!black,line width=3pt]
    ([xshift=5pt]frame.north west) -- 
    (frame.north west) -- 
    (frame.south west);
  \draw[red!75!black,line width=3pt]
    ([xshift=-5pt]frame.north east) -- 
    (frame.north east) -- 
    (frame.south east);
  },
  overlay unbroken app={
  \draw[red!75!black,line width=3pt,line cap=rect]
    (frame.south west) -- 
    ([xshift=5pt]frame.south west);
  \draw[red!75!black,line width=3pt,line cap=rect]
    (frame.south east) -- 
    ([xshift=-5pt]frame.south east);
  },
  overlay middle and last={
  \draw[red!75!black,line width=3pt]
    (frame.north west) -- 
    (frame.south west);
  \draw[red!75!black,line width=3pt]
    (frame.north east) -- 
    (frame.south east);
  },
  overlay last app={
  \draw[red!75!black,line width=3pt,line cap=rect]
    (frame.south west) --
    ([xshift=5pt]frame.south west);
  \draw[red!75!black,line width=3pt,line cap=rect]
    (frame.south east) --
    ([xshift=-5pt]frame.south east);
  },
}



\usepackage{fancyhdr} 
\usepackage{blindtext} 

\DeclareCaptionFont{black}{\color{black}}

\definecolor{myblue}{rgb}{0.9, 0.1, 0.94}
\definecolor{mygreen}{rgb}{0.64, 0.56, 0.88}
\definecolor{myyellow}{rgb}{0.68, 0.6, 0.1}
\definecolor{fancygreen}{rgb}{0.33, 0.68, 0.20}
\definecolor{salmon}{rgb}{0.94, 0.52, 0.49}
\definecolor{tablegreen}{rgb}{0.82, 0.94, 0.75}
\definecolor{tableblue}{rgb}{0.81, 0.90, 0.94}
\definecolor{tablered}{rgb}{0.97, 0.85, 0.85}
\definecolor{tableorange}{rgb}{0.96, 0.85, 0.81}

\newenvironment{itemize*}%
 {\leftmargini=10pt\begin{itemize}%
  \setlength{\itemsep}{0pt}%
  \setlength{\parskip}{0pt}%
  }%
 {\end{itemize}}
\newenvironment{enumerate*}%
 {\begin{enumerate}%
  \setlength{\itemsep}{0pt}%
  \setlength{\parskip}{0pt}}%
 {\end{enumerate}}

\usepackage{listings}

\newcommand\JSONnumbervaluestyle{\color{blue}}
\newcommand\JSONstringvaluestyle{\color{red}}

\newif\ifcolonfoundonthisline

\makeatletter

\lstdefinestyle{json}
{
  showstringspaces    = false,
  keywords            = {false,true},
  alsoletter          = 0123456789.,
  morestring          = [s]{"}{"},
  stringstyle         = \ifcolonfoundonthisline\JSONstringvaluestyle\fi,
  MoreSelectCharTable =%
    \lst@DefSaveDef{`:}\colon@json{\processColon@json},
  basicstyle          = \ttfamily,
  keywordstyle        = \ttfamily\bfseries,
}

\newcommand\processColon@json{%
  \colon@json%
  \ifnum\lst@mode=\lst@Pmode%
    \global\colonfoundonthislinetrue%
  \fi
}

\lst@AddToHook{Output}{%
  \ifcolonfoundonthisline%
    \ifnum\lst@mode=\lst@Pmode%
      \def\lst@thestyle{\JSONnumbervaluestyle}%
    \fi
  \fi
  \lsthk@DetectKeywords%
}

\lst@AddToHook{EOL}%
  {\global\colonfoundonthislinefalse}

\makeatother

\usepackage{etoolbox}
\usepackage{natbib}
\usepackage{url}
\newcounter{bibcount}
\makeatletter
\patchcmd{\@lbibitem}{\item[}{\item[\hfil\stepcounter{bibcount}{[\thebibcount]}}{}{}
\renewcommand\NAT@bibsetup%
  [1]{\setlength{\leftmargin}{\bibhang}\setlength{\itemindent}{-\parindent}%
      \setlength{\itemsep}{\bibsep}\setlength{\parsep}{\z@}}
\makeatother

\author{%
Shijie Xia$^{1,2,3}$\space\space\space Yiwei Qin$^{3}$\space\space\space Xuefeng Li$^{1,2,3}$\space\space\space Yan Ma$^{3}$\space\space\space Run-Ze Fan$^{3}$ \\ \textbf{ Steffi Chern$^{3}$ \space\space\space Haoyang Zou$^{1,2,3}$\space\space\space   
Fan Zhou$^{1,2,3}$ \space\space\space Xiangkun Hu$^{2,3}$ \space\space\space Jiahe Jin$^{1,2,3}$} \\
\textbf{Yanheng He$^{1,2,3}$ \space\space\space Yixin Ye$^{1,2,3}$ \space\space\space  Yixiu Liu$^{1,2,3}$ \space\space\space Pengfei Liu$^{1,2,3}$\thanks{~~Corresponding author}}
\\
$^{1}$Shanghai Jiao Tong University, $^{2}$SII, $^{3}$Generative AI Research Lab (GAIR)\\
 }

\begin{document}

\title{Generative AI Act II: \\ Test Time Scaling Drives Cognition Engineering}

\maketitle
\thispagestyle{fancy}
\fancyhead{}
\lhead{\includegraphics[height=0.67cm]{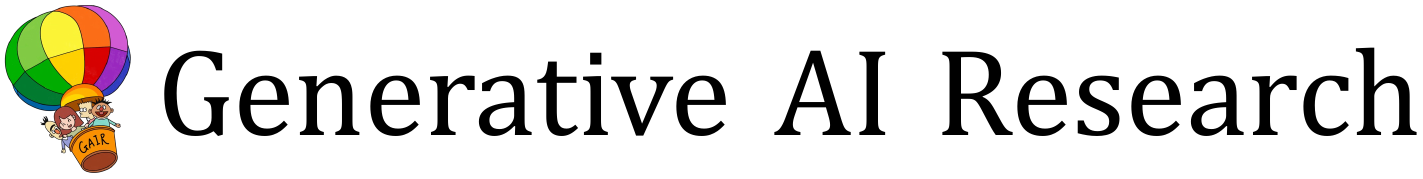}}
\renewcommand{\headrulewidth}{0pt}
\setlength{\headsep}{0mm}

\begin{abstract}

The first generation of Large Language Models—what might be called ``\emph{Act I}" of generative AI (2020-2023)—achieved remarkable success through massive parameter and data scaling, yet exhibited fundamental limitations such as knowledge latency, shallow reasoning, and constrained cognitive processes. During this era, \emph{prompt engineering} emerged as our primary interface with AI, enabling dialogue-level communication through natural language. We now witness the emergence of ``\textbf{\emph{Act II}}" (2024-present), where models are transitioning from knowledge-retrieval systems (in latent space) to thought-construction engines through test-time scaling techniques. This new paradigm establishes a mind-level connection with AI through language-based thoughts.
In this paper, we clarify the conceptual foundations of \textbf{\emph{cognition engineering}} and explain why this moment is critical for its development. We systematically break down these advanced approaches through comprehensive tutorials and optimized implementations, democratizing access to cognition engineering and enabling every practitioner to participate in AI's second act. We provide a regularly updated collection of papers on test-time scaling in the \href{https://github.com/GAIR-NLP/cognition-engineering}{GitHub Repository}.

\end{abstract}

\begin{figure}[htbp]
    \centering
    \includegraphics[width=0.8\textwidth]{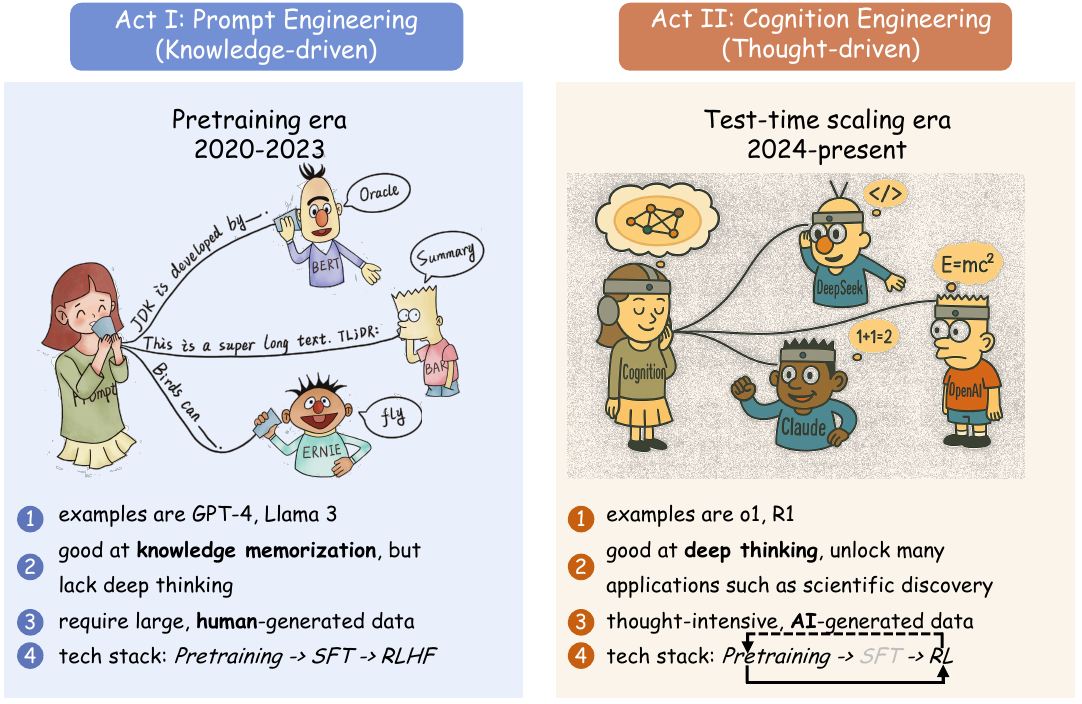}
    \label{fig:two_acts}
\end{figure}

{\centering
 {\color{blue}\textit{Who Might Benefit from This Paper?}
\par}}

\begin{itemize}[ leftmargin=2.5em, labelsep=0.5em, itemsep=-3pt]

   \item[] \includegraphics[height=1em]{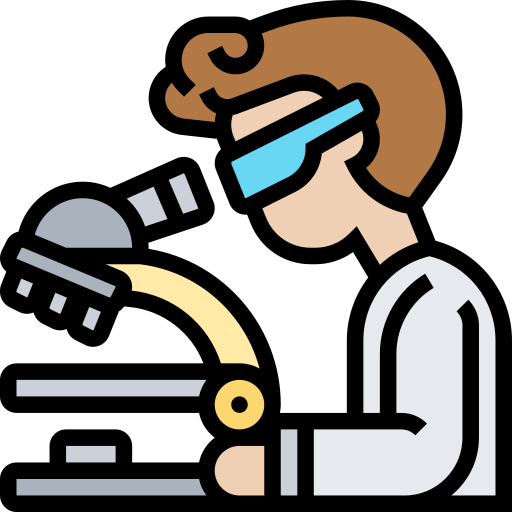} 
 \textbf{Researchers:} Open problems and design challenges for field advancement (e.g., \S\ref{sec:application}, \S\ref{sec:future_direction}).
 
    \item[]  \includegraphics[height=1em]{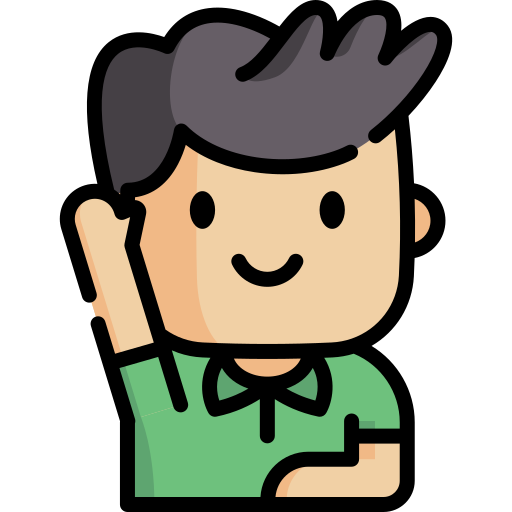} \textbf{Students \& Newcomers:} Cognition engineering tutorials and code examples (e.g., \S\ref{sec:tutorial}).
    
    \item[] \includegraphics[height=1em]{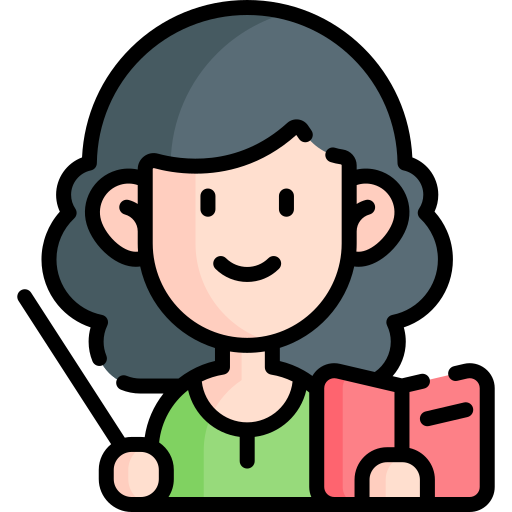}     \textbf{Educators:} Structured teaching resources with guidance for test-time scaling (e.g., \S\ref{sec:tts}, \S\ref{sec:training_strategies_tts}).

   \item[] \includegraphics[height=1em]{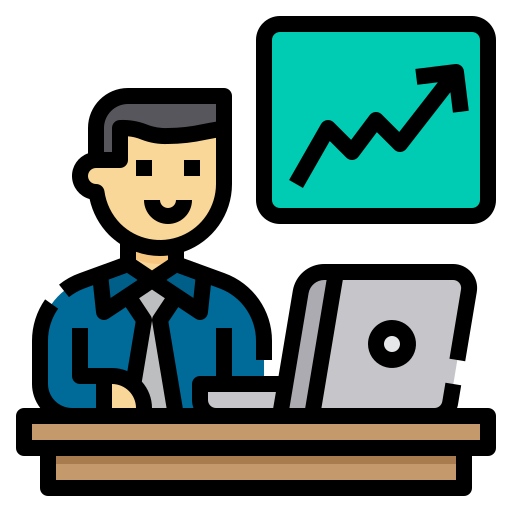} 
 \textbf{Investors \& Decision Makers:} Enhanced vision through Act I/II framework, providing deep, holistic cognition (e.g., \S\ref{sec:introduction}).

\end{itemize}

\newpage

\pagestyle{fancy}
\lhead{\rightmark}
\renewcommand{\headrulewidth}{0.7pt}
\setlength{\headsep}{5mm}

\section*{Three Scaling Phases}

\begin{figure}[htbp]
    \centering
    \includegraphics[width=0.9\textwidth]{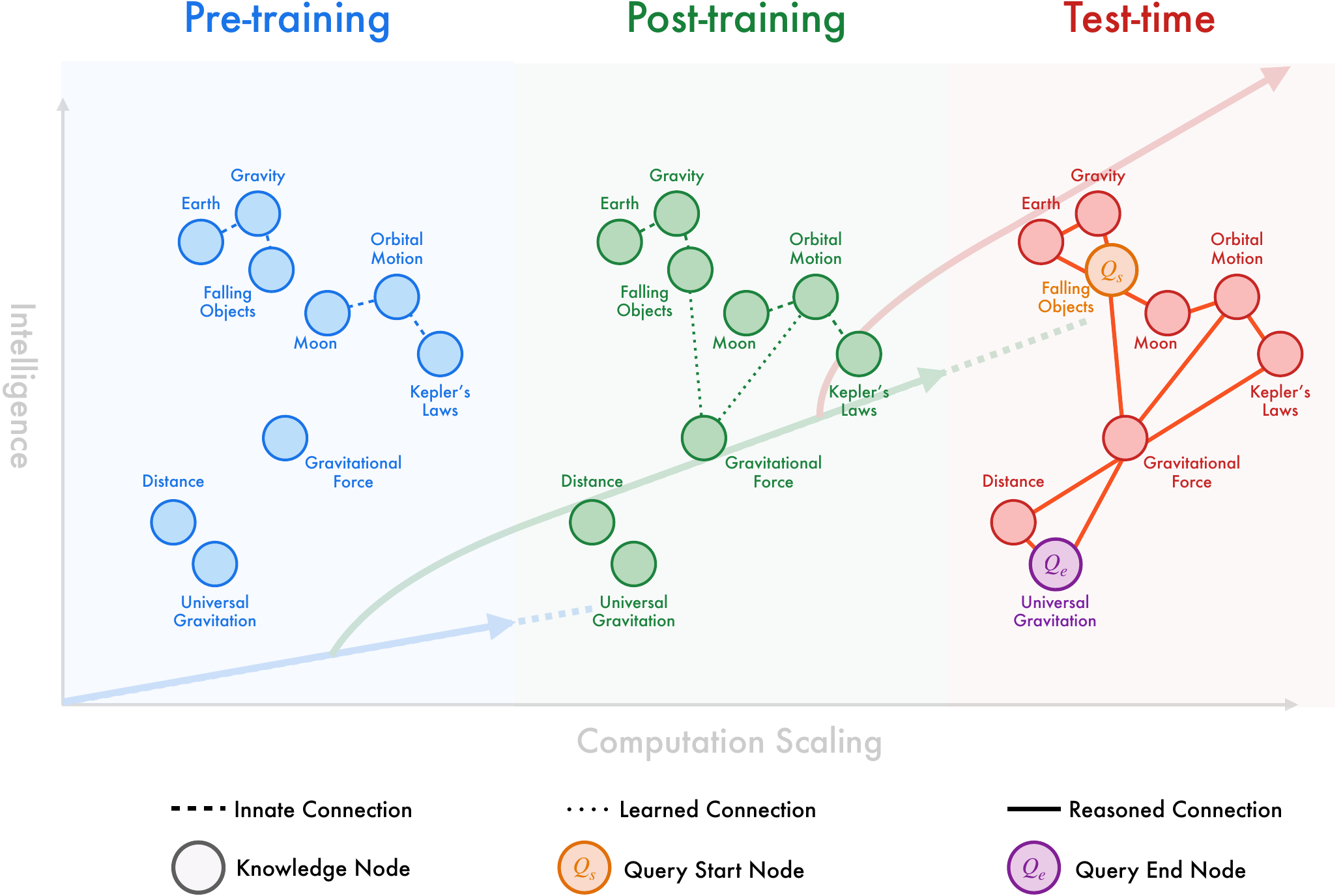}
    \caption{The three scaling phases illustrated as a progression of knowledge representation. Pre-training scaling (blue) forms isolated knowledge islands with fundamental physics concepts connected by limited innate associations. Post-training scaling (green) densifies these islands with more sophisticated learned connections between related concepts. Test-time scaling (red) enables dynamic reasoning pathway formation between previously disconnected concepts through extended computation, facilitating multi-hop inference across the entire knowledge space. 
    \textbf{\textcolor{black}{Test-time scaling builds bridges between knowledge islands, connecting distant nodes that remain isolated during pre-training and conventional post-training.}
    }
    }
    \label{fig:three_scaling_law}
\end{figure}

Before proceeding to the main body of our paper, we present a hypothesis regarding how the three scaling phases shape the cognitive abilities of models and highlight the significance of test-time scaling in this process.

\paragraph{Stage 1: Pre-training Scaling - Formation of Knowledge Islands}

The foundational phase where intelligence emerges through increased model size and training data volume, establishing basic knowledge acquisition capabilities. During pre-training scaling, we observe the formation of distinct ``knowledge islands'' - specialized domains where physics concepts form loosely connected clusters. These concepts are present but connections between them are limited and primarily represent innate relationships, shown as dashed blue lines. The model has acquired these physics concepts but hasn't yet established robust connections between them, limiting its ability to perform complex reasoning across the knowledge graph.

\paragraph{Stage 2: Post-training Scaling - Knowledge Densification}

The post-training scaling phase demonstrates how further fine-tuning densifies knowledge representation. The same physics concepts become more richly connected through learned connections (shown as dotted green lines). The network of physics knowledge becomes more integrated as post-training creates pathways between previously isolated concepts. We see a moderate increase in connections between different physics principles, enabling more sophisticated associations. However, these connections primarily link closely related concepts, still lacking the ability to establish comprehensive reasoning pathways between \textbf{\emph{distant knowledge nodes}} that would require multi-step inference.

\paragraph{Stage 3: Test-time Scaling - Cognitive Pathway Formation}

The final stage represents the paradigm shift enabled by test-time scaling or ``long thinking.'' This breakthrough approach allows the model to establish robust reasoning pathways (shown as solid red lines) between previously weakly connected physics concepts. The diagram illustrates how queries starting at specific nodes ($Q_s$) can now trace sophisticated reasoning paths through the knowledge graph, culminating in comprehensive answers at query end nodes ($Q_e$). Through extended computation time at inference, the model explores deeper search spaces within its physics knowledge representation, connecting concepts through multi-hop reasoning. The test-time phase shows a fully integrated understanding of gravitational principles where concepts like Universal Gravitation can be meaningfully connected to specific applications like Orbital Motion or Falling Objects.

\paragraph{Conclusion}

This integrated view demonstrates how computational scaling directly shapes knowledge representation and reasoning abilities in physics domains. Pre-training scaling builds the foundational physics knowledge, post-training scaling refines connections between related concepts, but only test-time scaling enables the complex cross-domain reasoning that characterizes advanced scientific thinking. The progression shown fundamentally reframes AI advancement - not merely as an accumulation of more data or parameters, but as the development of cognitive capabilities enabling models to navigate the full complexity of physical principles through principled reasoning processes. Test-time scaling represents the critical frontier in this progression, where models transition from knowledge repositories to systems capable of deep scientific insight through extended deliberation - similar to how human experts solve complex physics problems through sustained thought.

\newpage

\section*{The Practitioner's Roadmap: How to Apply Test-Time Scaling to your Applications?}

\begin{figure}[H]
    \centering
    \includegraphics[width=0.9\linewidth]{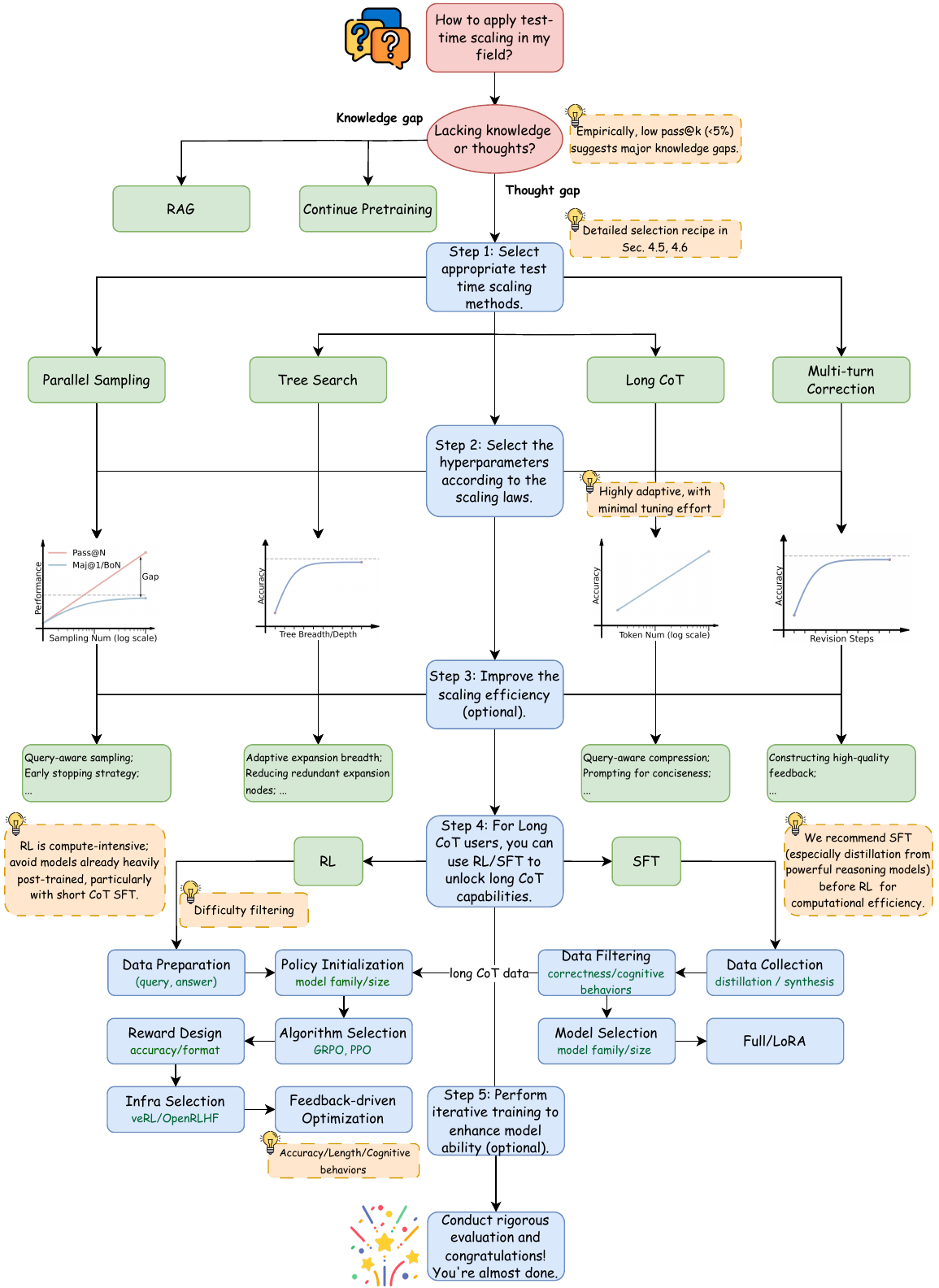}
    \caption{Workflow for applying test-time scaling in a specific domain. For more details, please refer to the main paper.}
    \label{fig:workflow_tts}
\end{figure}

\newpage

\begin{spacing}{1.1}
\tableofcontents
\end{spacing}
\newpage

\section{Introduction} \label{sec:introduction}

In recent years, Large Language Models (LLMs) such as GPT~\cite{openai2024gpt4technicalreport}, LLaMA~\cite{grattafiori2024llama3herdmodels,touvron2023llama2openfoundation}, and Claude~\cite{anthropic2024claude3} have emerged as powerful knowledge management tools through extensive pre-training and fine-tuning processes. These models, trained on vast corpora of human-generated text, have successfully organized and systematized accumulated human knowledge. Through a paradigm defined by scaling pre-training data, computation, and model parameters~\cite{kaplan2020scalinglawsneurallanguage}, these systems can engage in natural language conversations, retrieve information, generate content, and answer questions across diverse domains~\cite{zhao2025surveylargelanguagemodels,wang2024mmluprorobustchallengingmultitask,zheng2023judging}. 
This first generation of LLMs—what we might call ``Act I'' of generative AI—introduced a fundamental shift in human-AI interaction. 
The cornerstone of this first act has been ``\emph{prompt engineering}''---the art of crafting inputs that guide models toward desired outputs~\cite{liu2021pretrainpromptpredictsystematic,sahoo2025systematicsurveypromptengineering}. This innovation enabled humans to communicate with AI systems using natural language for the first time, dramatically lowering the barriers to human-machine interaction.
Act I focused primarily on gathering and organizing existing knowledge through ever-larger models trained on increasingly vast datasets.
However, despite their impressive capabilities, Act I models exhibit several significant limitations:
    (i) \textbf{knowledge latency}: These models primarily learn high-frequency information that has had time to accumulate in their training data, leaving them with limited understanding of emerging knowledge and concepts~\cite{Huang_2025}.
    (ii) \textbf{shallow reasoning}: While capable of basic logical inferences, they struggle with problems requiring multi-step, deep reasoning processes~\cite{zhang2024carefulexaminationlargelanguage,mirzadeh2024gsmsymbolicunderstandinglimitationsmathematical,Kambhampati_2024}.
    (iii) \textbf{limited thought processes}: They fail to demonstrate human-like depth of thought, particularly when confronting novel or open-ended questions~\cite{wu2024reasoningrecitingexploringcapabilities}.
These constraints have kept Act I models primarily confined to knowledge retrieval and simple reasoning tasks, still considerably distant from achieving artificial general intelligence (AGI). Just as knowledge alone is insufficient for human intelligence development, merely amassing information has proven inadequate for AI systems to approach human-like intelligence---they must also develop the capacity for deep thinking and reasoning~\cite{newell1959report,newell1972human}.

Recently, the AI field has witnessed a profound paradigm shift. A new technical approach centered on \textbf{\emph{``test-time scaling''}} is redefining the boundaries of what LLMs can achieve~\cite{openai_o1_system_card,deepseekai2025deepseekr1incentivizingreasoningcapability,snell2024scalingllmtesttimecompute}, inaugurating the second act of generative AI---\textbf{\emph{cognition engineering}}.

\begin{quote}
    \emph{Cognition Engineering is the systematic and constructive development of AI thinking capabilities through test-time scaling paradigms that transcend traditional pretraining approaches. This methodology represents the deliberate cultivation of deep cognitive processes in artificial systems through both \textbf{human cognitive} pattern distillation and \textbf{AI-driven discovery} (e.g., reinforcement learning).}
\end{quote}

At its core, cognition engineering sits at the intersection of two fundamental concepts:
``\emph{\textbf{Cognition}}'' in this context refers not merely to knowledge acquisition but to deep cognition---the ability to perform complex reasoning, engage in deliberate thinking, connect disparate concepts, and generate novel insights. It encompasses the meta-cognitive processes~\cite{metcalfe1994metacognition} that allow for understanding \emph{not just ``what'' but ``why'' and ``how''}---the very essence of human intellectual advancement.
``\emph{\textbf{Engineering}}'' here signifies a constructive approach rather than a purely emergent one.  It moves beyond the limitations of mere scaling toward a more intentional construction of cognitive capabilities through targeted interventions in both training methodologies (e.g., reinforcement learning) and inference optimization (e.g., extending inference-time computation).

Cognition engineering represents a comprehensive technological paradigm shift in LLM development. From the inference perspective, it transitions from crafting prompt templates that retrieve knowledge from LLMs to designing test-time scaling strategies that conduct deeper and more comprehensive searches through knowledge spaces. This evolution demands rigorous analysis of test-time scaling strategy components, characteristics, and efficiency, which underscores the necessity for a structured engineering approach.
From the training perspective, cognition engineering redirects computational resources from knowledge-focused pre-training toward developing deep thinking abilities through techniques like learning on human cognition data and reinforcement learning. The shift implies in Act II, the cognitive exchange becomes bidirectional. Not only can humans teach AI systems how to approach complex problems, but AI systems can also autonomously discover novel cognitive patterns and reasoning pathways through techniques like reinforcement learning. For example, we have already witnessed moments reminiscent of AlphaGo's famous ``Move 37,'' where AI demonstrates thinking approaches that transcend human intuition yet ultimately prove effective~\cite{MasteringGameGo}. These AI-discovered cognitive strategies have the potential to enrich human understanding, opening new research and innovation pathways. This bidirectional cognitive exchange marks our entry into a new era of intelligence symbiosis.

This paper will explore in depth the definition, technical foundations, and application prospects of cognition engineering. First, we will clarify the conceptual connotations of cognition engineering (\S\ref{sec:definition_cognition_engineering}) and explain why now is the critical moment for its development~(\S\ref{sec:why_the_moment}). Next, we will analyze in detail the technical foundations of test-time scaling~(\S\ref{sec:tts}) and various training strategies for it~(\S\ref{sec:training_strategies_tts}). We will then examine the systemic changes cognition engineering brings to AI research and the applications that have already emerged~(\S\ref{sec:application}). Finally, we will analyze the  profound implications beyond the technical implementation~(\S\ref{sec:implications}), discuss the infrastructure~(\S\ref{sec:infra}), and identify several future directions for cognition engineering~(\S\ref{sec:future_direction}). We also provide a practical tutorial for implementing test-time scaling with code examples~(\S\ref{sec:tutorial}).
Through these discussions, we aim to outline the contours of generative AI's second act and provide researchers and practitioners with a framework for thinking in this new paradigm.

\section{What -- Cognition Engineering Definition}\label{sec:definition_cognition_engineering}

The term \emph{cognition engineering} represents a significant conceptual shift in artificial intelligence development. To understand the essence of this emerging field, we can utilize DIKW (Data-Information-Knowledge-Wisdom) pyramid theory~\cite{zeleny1987management,ackoff1989data} as a conceptual framework and explore how cognition engineering enables the leap from knowledge to wisdom.

\begin{figure}[htbp]
    \centering
    \includegraphics[width=0.6\textwidth]{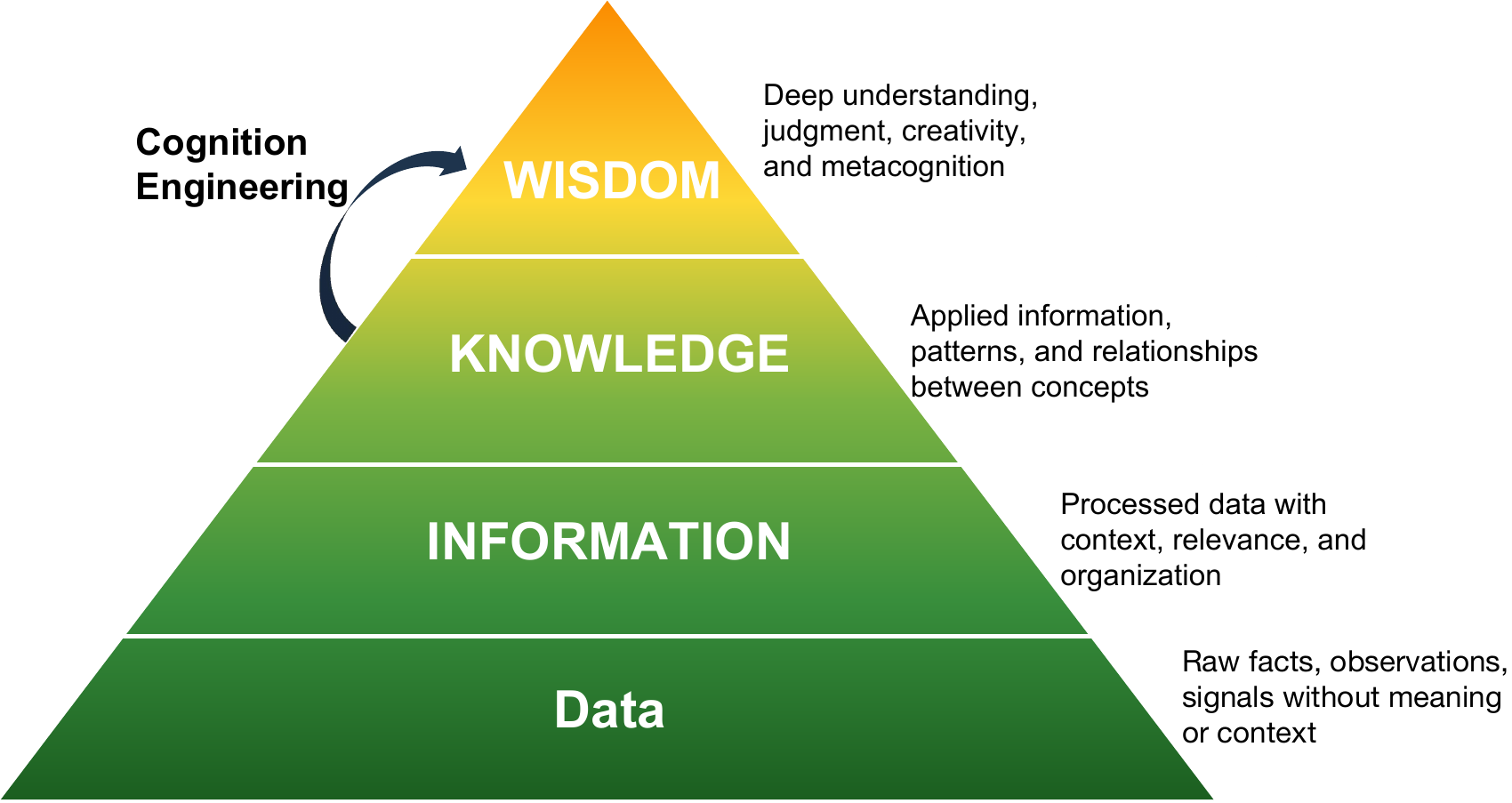}
    \caption{The DIKW pyramid and its relationship to cognition engineering paradigm.}
    \label{fig:dikw}
\end{figure}

\subsection{Cognition}

The DIKW theory portrays cognitive process as a hierarchical transformation: from raw data to contextualized information, then to applicable knowledge, and ultimately to profound wisdom. This framework offers deep insights for understanding cognition engineering. At the data level, we encounter raw facts and observations devoid of inherent meaning; the information level consists of processed and organized data imbued with context and structure; the knowledge level manifests as understanding and application of information, including mastery of rules, patterns, and relationships; while the wisdom level embodies deep comprehension of knowledge, involving judgment, creativity, and metacognitive abilities. Traditional AI systems primarily operated at the data and information levels, whereas first-generation LLMs achieved significant breakthroughs at the knowledge level. Cognition engineering represents the crucial step toward advancing to the wisdom level.

In psychology and cognitive science, cognition refers to the complex mental processes through which organisms acquire and process information, form knowledge, and apply it to problem-solving~\cite{von1995cognitive,nunez2019happened}. However, what cognition engineering pursues is not merely this basic cognitive capability, but the wisdom-level cognition described by DIKW—the ability to understand deep principles, engage in creative thinking, and demonstrate judgment. This deep cognition concerns not only \textbf{\emph{``knowing what"}} (i.e., knowledge) but also \textbf{\emph{``knowing why"}} and \textbf{\emph{``knowing how"}} (i.e., wisdom).
Cognition is characterized by deep thinking—the ability to engage in multi-layered, complex reasoning exploring multiple pathways to resolution—alongside metacognitive capabilities that allow reflection on one's own thought processes. It encompasses creative reasoning that connects knowledge across domains to generate novel insights, cognitive adaptability that applies existing patterns to new contexts, and conceptual abstraction that extracts higher-order principles from concrete instances. These capabilities collectively form the core of human intelligence and are the foundation of humanity's continuous advancement in scientific discovery and technological innovation.

\subsection{Engineering Methodology}

Engineering, as a methodology, is essentially an organized approach to designing, building, and optimizing systems to solve specific problems. 
Within the DIKW framework, engineering methods can be viewed as the process by which humans consciously guide systems to ascend from the data level to the wisdom level. In cognition engineering, this emphasis on intentional construction manifests through targeted interventions in both training methodologies and inference optimization, rather than relying exclusively on scaling approaches.

\subsection{Cognition Engineering}

Combining the concepts of cognition and engineering, and viewed through the lens of DIKW theory, we can define cognition engineering more profoundly:

\begin{quote}
\emph{Cognition Engineering is a systematic methodology that constructs and optimizes AI systems' ability to ascend from the knowledge to wisdom levels of the DIKW pyramid through specific design patterns, training strategies, and computational allocations. It enables AI systems to engage in deep thinking, complex reasoning, and creative problem-solving, exhibiting cognitive characteristics similar to human wisdom-level traits.}
\end{quote}

The key distinction between cognition engineering and traditional LLMs development approaches lies in its methodological characteristics:

\begin{itemize}
    
    \item \textbf{From behavior imitation to thought imitation}: Traditional models primarily learn by imitating human output behaviors, remaining at the knowledge level of DIKW; cognition engineering focuses on imitating human thought processes, directly addressing the cognitive characteristics of the wisdom level.
    
    \item \textbf{From static knowledge to dynamic wisdom}: Traditional models have relatively fixed capabilities after training, whereas cognition engineering emphasizes AI systems' dynamic thinking abilities during inference, allowing them to adjust thinking depth and resource allocation based on problem complexity.
    
    \item \textbf{From knowledge retrieval to knowledge creation}: Traditional models primarily retrieve and combine existing knowledge, while cognition engineering aims to enable AI systems to generate new insights and discoveries through deep thinking, realizing the creative characteristics of the wisdom level in DIKW.
\end{itemize}




\section{Why \& Why Now -- Technical Foundation} \label{sec:why_the_moment}

\subsection{The Necessity of Cognition Engineering}

The rise of cognition engineering is not coincidental but a direct response to the ``wisdom gap'' encountered by AI development in the DIKW pyramid. Despite significant advances in knowledge retrieval, content generation, and basic reasoning, LLMs still exhibit notable shortcomings at the wisdom level:

\paragraph{Limitations in Complex Reasoning}

Current models perform poorly on problems requiring multi-step deep reasoning~\cite{zhang2024carefulexaminationlargelanguage,mirzadeh2024gsmsymbolicunderstandinglimitationsmathematical,Kambhampati_2024}. Even the most advanced models struggle with reliable mathematical proofs, complex scientific problem-solving, or multidimensional analysis~\cite{yang2024formalmathematicalreasoningnew,rein2023gpqagraduatelevelgoogleproofqa}. These tasks require models to decompose problems into sub-problems, explore multiple possible reasoning paths, and conduct deep logical analysis—capabilities beyond what scaling pre-training data alone can achieve.

\paragraph{Challenges in Knowledge Updating and Creation}

Pre-trained models' knowledge is fixed at the end of training, unable to automatically adapt to new developments and changes. More importantly, they struggle to generate \emph{truly original insights or discoveries—the essence of scientific discovery} is not merely understanding known facts but proposing new hypotheses, designing experimental methods, and drawing new conclusions from results. This knowledge creation ability requires going beyond simple knowledge retrieval and pattern recognition.

\paragraph{Elevated Application Requirements}

As AI applications expand from simple tasks to complex decision-making, scientific research, and creative work, demands for AI systems' wisdom-level capabilities also increase~\cite{deepresearch,cuaopenai}. Users are no longer satisfied with answers based on statistical patterns (\emph{knowledge level}); they desire thoughtful analysis, multi-perspective considerations, and innovative insights (\emph{wisdom level}) from AI.

\subsection{Three Pillars}
Cognition engineering emerges at this specific moment due to multiple technological breakthroughs reaching maturity simultaneously. These breakthroughs collectively create the necessary conditions
enabling AI to progress from knowledge management to deep cognitive capabilities. The rise of cognition engineering stems from three key technological pillars:

\subsubsection{Knowledge Foundation}

The first enabling foundation for cognition engineering is the fundamental transformation in how LLMs acquire knowledge. Modern foundation models have not only achieved exponential growth in training data volume (such as Llama 2's 2 trillion token training scale~\cite{touvron2023llama2openfoundation}) but more importantly, have experienced a qualitative transformation.
Pre-training data has evolved from simple web-scraped text to carefully curated knowledge corpora~\cite{shao2024deepseekmathpushinglimitsmathematical,zhou2025programmingexampleliftingpretraining,wang2024mathpilebilliontokenscalepretrainingcorpus,yangQwen25MathTechnicalReport2024}. These datasets now integrate scientific literature and technical documentation with mathematical textbooks and problem sets, multi-language programming code repositories, and structured knowledge from specialized domains, forming a much richer knowledge ecosystem than previously available. This comprehensive knowledge foundation is a necessary prerequisite for cognition engineering---without this extensive embedded knowledge, models would lack the raw materials required for deep thinking.

\subsubsection{Test-time Scaling Foundation}

The second critical pillar enabling cognition engineering is the fundamental reconceptualization of how computational resources are allocated during the inference phase---what we term ``Test-Time Scaling.''
Traditional inference approaches were constrained by fixed output lengths and single-pass generation paradigms. Recently, a series of technical breakthroughs has significantly extended models' reasoning capabilities. Chain-of-Thought (CoT) prompting~\cite{wei2023chainofthoughtpromptingelicitsreasoning} methods encourage models to perform step-by-step reasoning like human problem-solving processes, clearly articulating intermediate steps. Tree search~\cite{yao2023treethoughtsdeliberateproblem,haoReasoningLanguageModel2023,fengAlphazerolikeTreeSearchCan2024} allow for systematic exploration of multiple reasoning pathways simultaneously, rather than being confined to a single line of thinking. Self-correction and verification techniques~\cite{deepseekai2025deepseekr1incentivizingreasoningcapability,kumar2024traininglanguagemodelsselfcorrect,qu2024recursiveintrospectionteachinglanguage} further enhance these capabilities, enabling models to evaluate their own reasoning, identify potential errors, and refine their approaches---mimicking human metacognitive processes. These innovations collectively provide what can be understood as a ``cognitive workspace'' where models can systematically explore their knowledge---similar to how humans need scratch paper to solve complex problems or require time to ``think deeply.''

\subsubsection{Self-Training Foundation}

The third pillar of cognition engineering is advanced self-training methodologies. Developing sophisticated cognitive abilities in models exclusively through expert human cognition data encounters inherent scaling limitations. Self-training technologies not only provide an alternative pathway to elicit cognitive capabilities but also create opportunities for superhuman performance through AI self-discovery strategies.
As demonstrated by DeepSeek-R1~\cite{deepseekai2025deepseekr1incentivizingreasoningcapability} and subsequent research~\cite{gandhi2025cognitivebehaviorsenableselfimproving,yu2025dapoopensourcellmreinforcement}, training with reinforcement learning using verifiable rewards enables models to master complex cognitive behaviors including reflection, backtracking, and verification when solving challenging problems. Through this process, models learn to dynamically allocate computational resources according to problem complexity, effectively internalizing test-time scaling techniques. Additionally, iterative self-training on reasoning trajectories generated through test-time scaling methods facilitates continuous improvement~\cite{zelikmanSTaRBootstrappingReasoning2022,fengAlphazerolikeTreeSearchCan2024,xiong2025selfrewardingcorrectionmathematicalreasoning}, allowing AI systems to progressively enhance their problem-solving abilities.

\subsection{From Theory to Practice: The Road Ahead}
The theoretical foundations are now in place, but translating these foundations into practical implementations requires navigating a complex landscape of specific methods and approaches. The most immediate and promising avenue for realizing cognition engineering in practice is through test-time scaling, a family of techniques that optimize how models allocate computational resources during inference to achieve deeper reasoning. These methods serve as the practical bridge between the theoretical promise of cognition engineering and its real-world implementation. By understanding and refining these techniques, we can begin the systematic construction of AI systems that truly think rather than merely predict.

In the following section, we delve into the specific mechanisms that enable test-time scaling, examining how different approaches address the fundamental challenge of extending and deepening AI reasoning processes. This exploration will reveal not just the technical details of these methods, but also their cognitive implications and their roles in the broader cognition engineering paradigm.

\section{How – Part I: Test-Time Scaling Methods} \label{sec:tts}

Given a query $q$ and a generator $g$, the test-time scaling method can be abstracted to a search strategy $M$ that guides the generator $g$ to find the optimal response:
    \begin{equation}
         y \sim M(.|q,g,\phi)
    \end{equation}
where $\phi$ represents any additional inputs such as scoring functions $v$ (also known as value functions, reward models, or verifiers\footnote{These words share subtle differences in context and we choose the most appropriate term for each method.}) and hyperparameters of the strategy.

\paragraph{Scaling laws} For any test-time scaling method, there exist corresponding scaling dimensions $\lambda$ within $\phi$ that directly determine the computation cost during inference. The scaling laws of $M$ describe the relationship between $\lambda$  and performance.

\paragraph{Scaling efficiency}  Given a computation budget\footnote{It can be measured by FLOPs, running time, token numbers, etc.} $C$, we define an abstract function $f: C \times M \rightarrow \mathbb{R}$ that maps from the computation budget $C$ and the test-time scaling method $M$ to performance. The scaling efficiency measures this performance relative to the computation budget:
\begin{equation}
    \text{efficiency} = \frac{f(C, M)}{C}
\end{equation}

The high-level strategies to improve efficiency of $M$ can be categorized into:\footnote{In this section, we do not consider the model compression techniques like model quantization or inference acceleration from infrastructure aspects as they are orthogonal to the method design.} 1) \emph{Optimizing individual test-time scaling methods:} This involves carefully selecting and tuning components within computation budget constraints, or leveraging additional training-time compute to optimize models specifically for test-time scaling; 2) \emph{Combining multiple test-time scaling methods:} This includes simultaneously combining multiple methods or selecting appropriate test-time scaling methods according to different contexts.

In the following sections, we will investigate four primary test-time scaling methods: parallel sampling (\S\ref{sec:parallel_sampling}), tree search (\S\ref{sec:tree_search}), multi-turn correction (\S\ref{sec:multi-turn_correction}), and long CoT (\S\ref{sec:long_cot}). For each test-time scaling method, we will cover the construction method, the scaling laws, and how to improve the scaling efficiency from the individual optimization aspect. Furthermore, we compare these test-time scaling methods across multiple dimensions (\S\ref{sec:comparisons_tts}) and discuss how to effectively combine them for enhanced performance (\S\ref{sec:ensemble_tts}).

\subsection{Parallel Sampling}\label{sec:parallel_sampling}

\subsubsection{Key Components}
\begin{figure}
    \centering
    \includegraphics[width=1\linewidth]{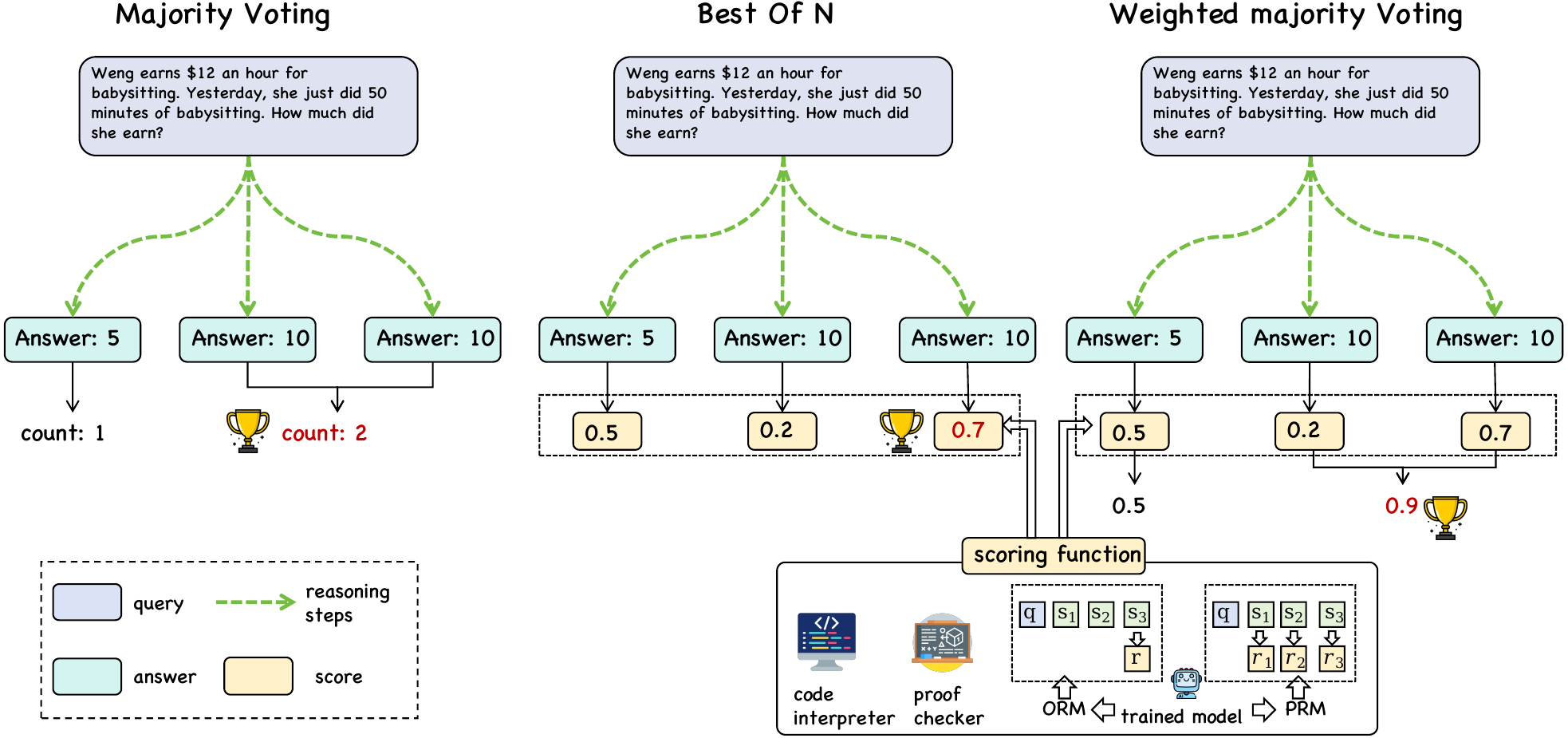}
    \caption{Illustration of parallel sampling selection methods: Best-of-N (F1), Majority voting (F2), and Combined strategy (F3).}
    \label{fig:parallel_sampling:illustration}
\end{figure}

The parallel sampling algorithm samples a set of candidate responses $\mathcal{Y} = \{y_i\}_{i=1}^N$ independently from the generator for the same query, where N is the sampling number, and selects the targeted response or answer from them. This approach can be conceptualized as a global search within the knowledge space~\cite{snell2024scalingllmtesttimecompute}. The selection methods are as follows:
\begin{itemize}
    \item \textbf{F1: Best-of-N (BoN).} This method utilizes a scoring function $v$ to evaluate each response and selects the one with the highest score:
    \begin{equation}
         y^* = \arg \max_{\tilde{y} \in \mathcal{Y}} v(\tilde{y})
    \end{equation}
   
   The scoring function $v$ can be external tools that directly verify the effectiveness of the response, such as code interpreters~\cite{liCompetitionLevelCodeGeneration2022, chenCodeTCodeGeneration2022} or math proof checkers~\cite{brownLargeLanguageMonkeys2024}. For tasks lacking verification tools, $v$ can be a specialized trained model. For instance, \citet{cobbe2021trainingverifierssolvemath} train the outcome reward model (ORM) to score the entire response, while \citet{lightman2023letsverifystepstep,uesato2022solvingmathwordproblems} train the process reward model (PRM) to score each step in the response and apply an aggregation function to determine the overall response score. Self-Certainty~\cite{kang2025scalablebestofnselectionlarge} eliminates the need for an additional reward model by leveraging the generator's inherent probability distribution for scoring.
   
    \item \textbf{F2: Majority voting.} Majority voting (or self-consistency~\cite{wangSelfConsistencyImprovesChain2023}) selects the most frequent answer from the candidates:
    \begin{align}
    y^* &= \arg \max_{\tilde{y} \in \mathcal{Y}} \sum_{\hat{y} \in \mathcal{Y} } g(\tilde{y}, \hat{y}) \\
    g(\tilde{y}, \hat{y}) &= \begin{cases}
    1 & \text{if } \tilde{y} \text{ is equivalent to } \hat{y}, \\
    0 & \text{otherwise,}
    \end{cases}
    \end{align}
    where $g$ is an automatic grading function that first extracts the answers from the responses and checks for equivalence. While this method is lightweight, the requirement for easy answer equivalence comparison limits its applicability for open-ended tasks. Universal Self-Consistency~\cite{chen2023universalselfconsistencylargelanguage} employs LLMs themselves to select the most consistent answer among multiple candidates, though the limited context window size of models still presents challenges for large sampling numbers.
    \item \textbf{F3: Combining voting and scoring strategy.} The scoring strategy can help select targeted low-frequency responses but heavily depends on the reliability of the scoring function, whereas the voting strategy offers greater robustness but has a more fixed upper bound. This combined method leverages advantages from both approaches for more robust selection~\cite{sun2024easytohardgeneralizationscalablealignment}. For example, weighted majority voting~\cite{uesato2022solvingmathwordproblems,liu2023improvinglargelanguagemodel} re-ranks answer clusters according to the sum of the scores in each cluster and selects the answer cluster with the highest score:
\begin{equation}
        y^* = \arg \max_{\tilde{y} \in \mathcal{Y}} \sum_{\hat{y} \in \mathcal{Y}} g(\tilde{y}, \hat{y})v(\hat{y})
\end{equation}
\end{itemize}

Figure~\ref{fig:parallel_sampling:illustration} illustrates these selection methods.

\begin{figure}
    \centering
    \begin{subfigure}[t]{0.24\textwidth}
        \includegraphics[width=\textwidth]{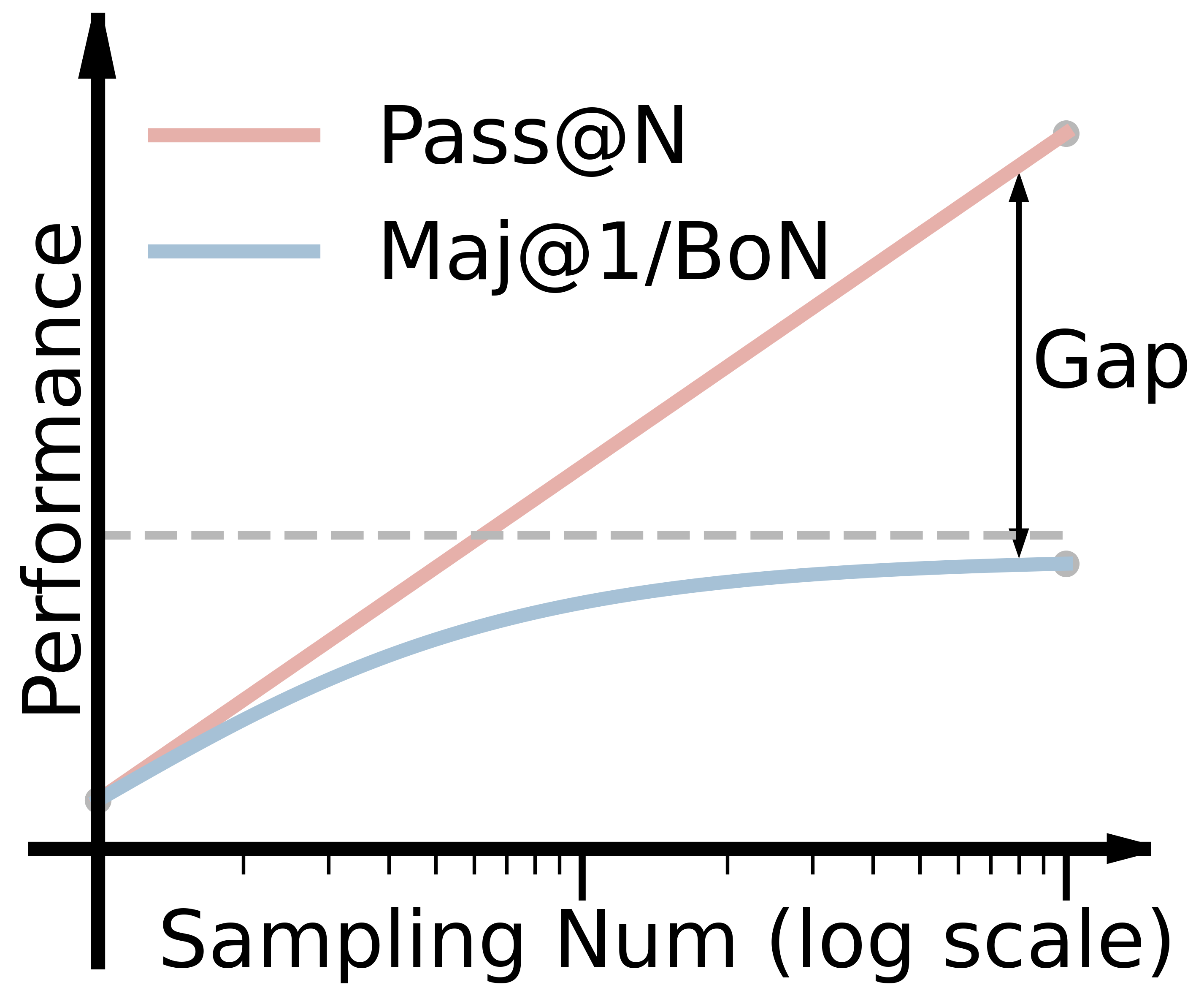}
        \caption{Parallel Sampling~(\S\ref{sec:parallel_sampling:scaling_factor})}
        \label{fig:scaling_factor_parallel_sampling}
    \end{subfigure}
    \hfill
    \begin{subfigure}[t]{0.24\textwidth}
        \includegraphics[width=\textwidth]{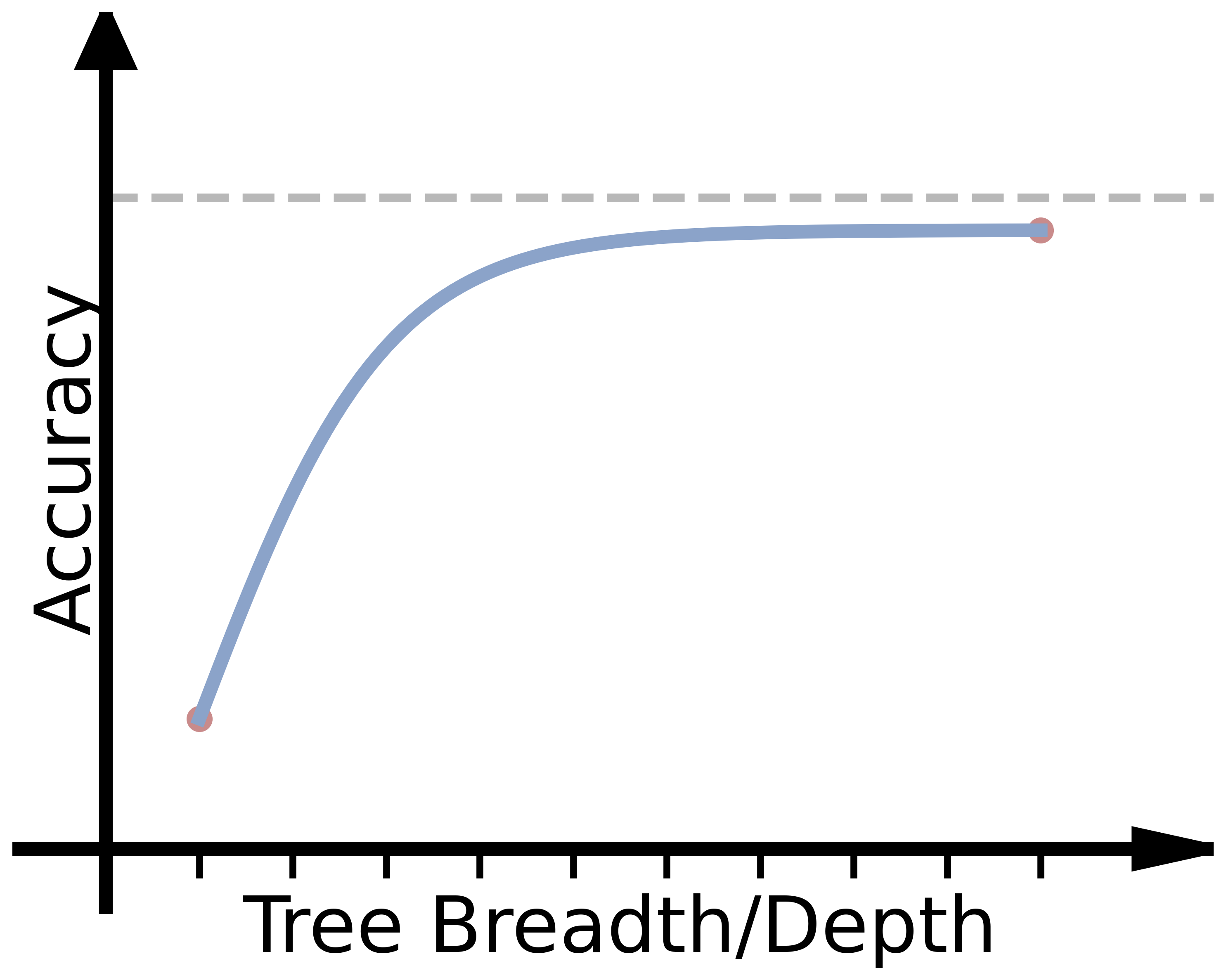}
        \caption{Tree Search~(\S\ref{sec:tree_search:scaling_factor})}
        \label{fig:scaling_factor_symbolic_tree}
    \end{subfigure}
    \hfill
    \begin{subfigure}[t]{0.24\textwidth}
        \includegraphics[width=\textwidth]{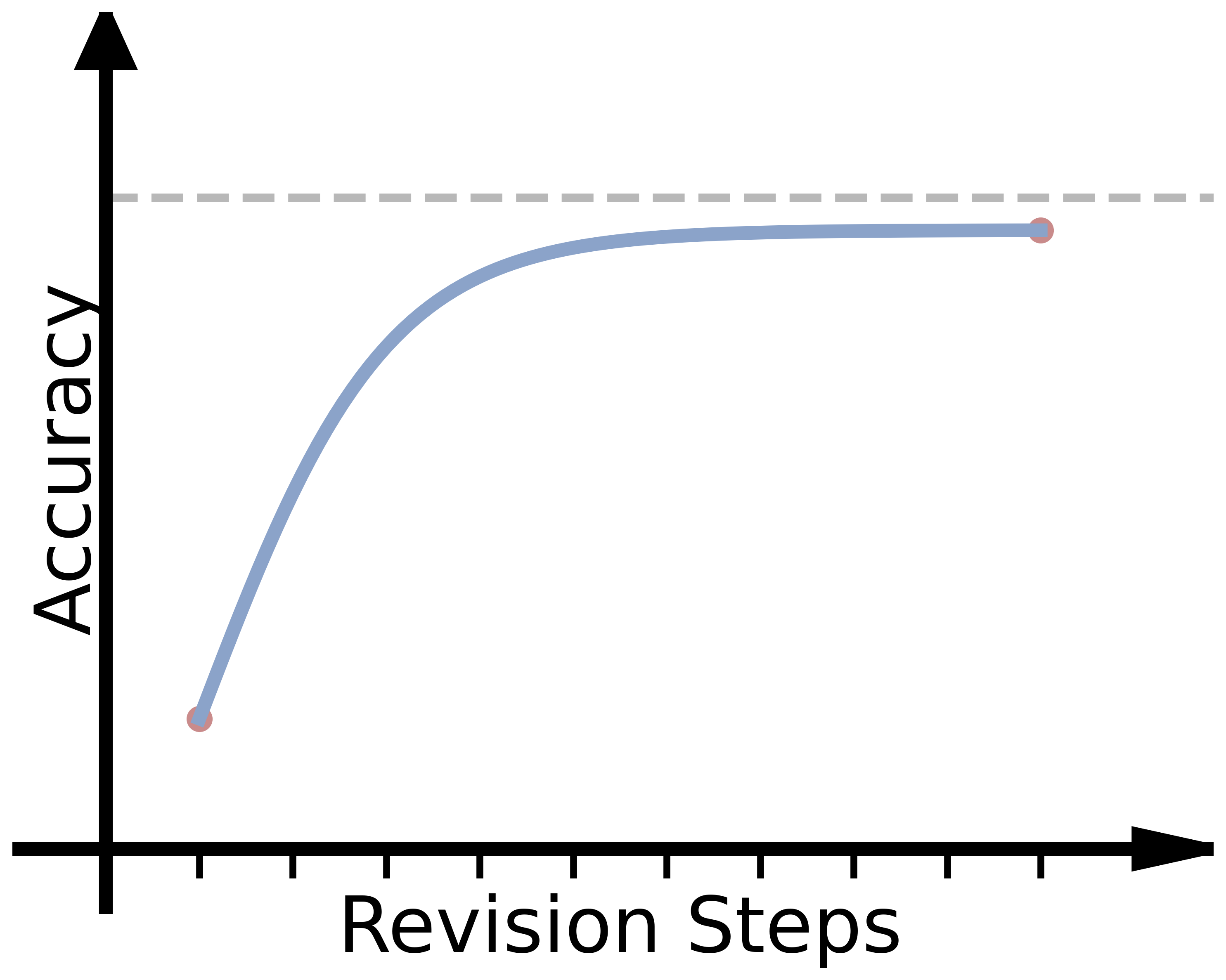}
       \caption{Multi-turn correction~(\S\ref{sec:multi-turn_correction:scaling_factor})}
        \label{fig:scaling_factor_multi_turn}
    \end{subfigure}
    \hfill
    \begin{subfigure}[t]{0.24\textwidth}
        \includegraphics[width=\textwidth]{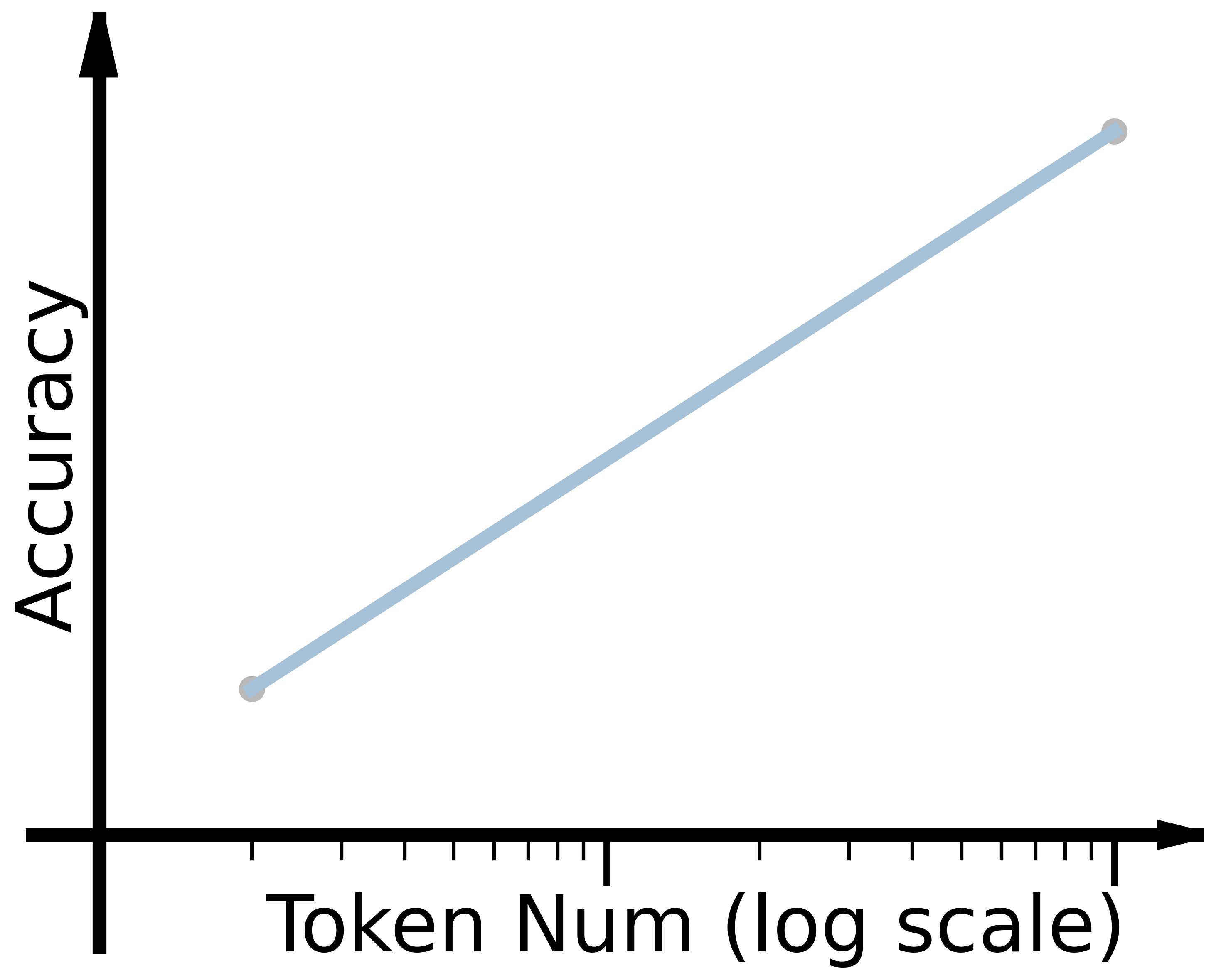}
        \caption{Long CoT~(\S\ref{sec:long_cot:scaling_factor})}
        \label{fig:scaling_factor_long_cot}
    \end{subfigure}
    \caption{The relationship between scaling dimensions and performance for each test-time scaling method.}
    \label{fig:scaling_factors}
\end{figure}

 \subsubsection{Scaling Laws} \label{sec:parallel_sampling:scaling_factor}

The main scaling dimension in parallel sampling is the sampling number N. We investigate the relationship between N and various performance metrics. Specifically, we focus on two types of metrics: Pass@N, which represents the probability of generating at least one correct response among N candidates, and metrics such as Maj@1 or BoN, which measure the practical performance of parallel sampling.
\paragraph{The monotonic growth relationship between N and Pass@N} ~\citet{brownLargeLanguageMonkeys2024} investigate the relationship between N and Pass@N across different models and tasks. The Pass@N grows steadily with sampling numbers. Moreover, the
relationship between the two is often log-linear as demonstrated in Figure~\ref{fig:scaling_factor_parallel_sampling}, similar to the training time scaling law observed~\cite{kaplan2020scalinglawsneurallanguage}.
\paragraph{Scaling Pass@N does not translate to real-world performance improvements}
Although the continuous improvement of Pass@N with increased sampling numbers is promising, there remains a gap between this metric and true performance. This gap exists for several reasons. First, performance improvements can only be realized when appropriate tools exist to select the correct response from the sample set. However, perfect verifiers do not exist for most tasks. As~\citet{brownLargeLanguageMonkeys2024} observe, when using ArmoRM-Llama3-8B-v0.1~\cite{wang2024interpretablepreferencesmultiobjectivereward} as the scoring model, a significant disparity emerges between Pass@N and practical metrics like Maj@1 or BoN (see Figure~\ref{fig:scaling_factor_parallel_sampling}).
Second, the verifiers themselves can be hacked. Code may pass unit tests but fail with additional test cases~\cite{stroeblInferenceScalingFLaws2024}, or mathematical solutions may reach correct answers through incorrect reasoning~\cite{xia2025evaluatingmathematicalreasoningaccuracy}, leading to the false positive problems.
~\citet{stroeblInferenceScalingFLaws2024} observe that the false positive rate increases as the Pass@1 accuracy decreases in code tasks, concluding that this imposes an upper bound on the accuracy of resampling-based inference scaling, even with infinite computational resources. For practical application methods, such as majority voting or scoring methods, performance tends to saturate \cite{brownLargeLanguageMonkeys2024,wuInferenceScalingLaws2024,li2024agentsneed} and may even degrade as the number of samples increases~\cite{chenAreMoreLLM2024} due to imperfect verifiers.

\tikzstyle{my-box}=[
    rectangle,
    draw=black,
    rounded corners,
    text opacity=1,
    minimum height=1.5em,
    minimum width=5em,
    inner sep=2pt,
    align=center,
    fill opacity=.5,
    line width=0.8pt,
]

\colorlet{parallel-fill}{blue!15}
\colorlet{parallel-border}{blue!50}
\colorlet{symbolic-fill}{green!15}
\colorlet{symbolic-border}{green!50}
\colorlet{multiTurn-fill}{orange!15}
\colorlet{multiTurn-border}{orange!50}
\colorlet{longCoT-fill}{purple!15}
\colorlet{longCoT-border}{purple!50}

\tikzstyle{parallel-leaf}=[
    rectangle,
    draw=none,
    rounded corners,
    text opacity=1,
    minimum height=1.5em,
    text=black,
    align=left, 
    text centered=false, 
    font=\scriptsize,
    fill=parallel-fill,
    text width=6em,  
]

\tikzstyle{symbolic-leaf}=[
    rectangle,
    draw=none,
    rounded corners,
    text opacity=1,
    minimum height=1.5em,
    text=black,
    align=left, 
    text centered=false, 
    font=\scriptsize,
    fill=symbolic-fill,
    text width=12em,  
]

\tikzstyle{multiTurn-leaf}=[
    rectangle,
    draw=none,
    rounded corners,
    text opacity=1,
    minimum height=1.5em,
    text=black,
    align=left, 
    text centered=false, 
    font=\scriptsize,
    fill=multiTurn-fill,
    text width=12em,  
]

\tikzstyle{longCoT-leaf}=[
    rectangle,
    draw=none,
    rounded corners,
    text opacity=1,
    minimum height=1.5em,
    text=black,
    align=left, 
    text centered=false, 
    font=\scriptsize,
    fill=longCoT-fill,
    text width=12em,  
]

\begin{figure*}[tb!]
    \centering
        \begin{forest}
            forked edges,
                for tree={
                grow=east,
                reversed=true,
                anchor=base west,
                parent anchor=east,
                child anchor=west,
                base=center,
                font=\small,
                rectangle,
                draw=black,
                rounded corners,
                align=center,
                text centered,
                minimum width=4em,
                edge+={darkgray, line width=1pt},
                s sep=3pt,
                line width=0.8pt,
                ver/.style={rotate=90, child anchor=north, parent anchor=south, anchor=center, minimum width=19em, fill=gray!10},
                leaf/.style={text width=15em, font=\scriptsize,  draw=none, inner xsep=8pt}, %
                leaf2/.style={text width=10em, font=\scriptsize,  draw=none, inner xsep=8pt} %
            },
            where level=1{text width=5.5em,font=\scriptsize}{},
            where level=2{text width=9.5em, align=center, font=\scriptsize}{},
            where level=3{font=\scriptsize}{},
            where level=4{font=\scriptsize}{},
            [
                \textbf{Improving Scaling Efficiency},ver
                [
                    Parallel \\ Sampling~(\S\ref{sec:parallel_sampling:efficiency}), fill=parallel-fill, draw=parallel-border
                    [
                        Query-aware sampling, fill=parallel-fill, draw=parallel-border
                        [
                         {DSC~\citenumber{wangMakeEveryPenny2024}; Chen et al.~\citenumber{chenAreMoreLLM2024}}, leaf, fill=parallel-fill, draw=none
                        ]
                    ]
                    [
                        Early stopping strategy, fill=parallel-fill, draw=parallel-border
                        [
                         {Adaptive-Consistency~\citenumber{aggarwalLetsSampleStep2023}; ESC~\citenumber{liEscapeSkyhighCost2024}; \\  Speculative Rejection~\citenumber{sun2024fastbestofndecodingspeculative}; \\RASC~\citenumber{wanDynamicSelfConsistencyLeveraging2024}; Self-Calibration~\citenumber{huang2025efficienttesttimescalingselfcalibration}}, leaf, fill=parallel-fill, draw=none
                        ]
                    ]
                    [
                        Tradeoff between sampling\\ number and model size , fill=parallel-fill, draw=parallel-border
                        [
                             {Brown et al.~\citenumber{brownLargeLanguageMonkeys2024}; Wu et al.~\citenumber{wuInferenceScalingLaws2024}}, leaf, fill=parallel-fill, draw=none
                        ]
                    ]
                    [
                        Improving the precision \\of response selection, fill=parallel-fill, draw=parallel-border
                        [
                             {Lightman et al.~\citenumber{lightman2023letsverifystepstep}; Sun et al.~\citenumber{sun2024easytohardgeneralizationscalablealignment};\\ MAV~\citenumber{lifshitz2025multiagentverificationscalingtesttime}; SC-GenRM~\citenumber{singhi2025solveverifycomputeoptimalproblem}}, leaf, fill=parallel-fill, draw=none
                        ]
                    ]
                    [
                        Inference-aware fine-tuning, fill=parallel-fill, draw=parallel-border
                        [
                             {BoN-Aware~\citenumber{chowInferenceAwareFineTuningBestofN2024}}, leaf, fill=parallel-fill, draw=none
                        ]
                    ]
                ]
                [
                    Tree \\ Search~(\S\ref{sec:tree_search:efficiency}), fill=symbolic-fill, draw=symbolic-border
                    [
                        Selecting appropriate tree\\ search algorithms, fill=symbolic-fill, draw=symbolic-border
                        [
                             {PG-TD~\citenumber{zhangPlanningLargeLanguage2023}; $\text{ToolChain}^*$~\citenumber{zhuangToolChainEfficientAction2023}}, leaf, fill=symbolic-fill, draw=none
                        ]
                    ]
                    [
                        Reducing the overhead\\ of value functions, fill=symbolic-fill, draw=symbolic-border
                        [
                             {AlphaLLM~\citenumber{tianSelfImprovementLLMsImagination2024b}; rStar-Math~\citenumber{guanRStarMathSmallLLMs2025}}, leaf, fill=symbolic-fill, draw=none
                        ]
                    ]
                    [
                        Adaptive expansion breadth, fill=symbolic-fill, draw=symbolic-border
                        [
                             {LiteSearch~\citenumber{wang2024litesearchefficacioustreesearch}; REBASE~\citenumber{wuInferenceScalingLaws2024}}, leaf, fill=symbolic-fill, draw=none
                        ]
                    ]
                    [
                        Reducing redundant \\expansion nodes, fill=symbolic-fill, draw=symbolic-border
                        [
                             {FETCH~\citenumber{wang2025dontlosttreesstreamlining}; ETS~\citenumber{hooper2025etsefficienttreesearch}}, leaf, fill=symbolic-fill, draw=none
                        ]
                    ]
                ]
                [
                   Multi-turn \\ Correction~(\S\ref{sec:multi-turn_correction:efficiency}), fill=multiTurn-fill, draw=multiTurn-border
                    [
                        Constructing high-quality \\ feedback, fill=multiTurn-fill, draw=multiTurn-border
                        [
                             {
                             Chiang and Lee~\citenumber{chiang2023largelanguagemodelsalternative}; G-Eval~\citenumber{liu2023gevalnlgevaluationusing};\\
                             BSM~\citenumber{saha2024branchsolvemergeimproveslargelanguage}; Varshney et al.~\citenumber{varshney2023stitchtimesavesnine};\\
                             IoE~\citenumber{li2024confidencemattersrevisitingintrinsic};  Self-Correction~\citenumber{welleckGeneratingSequencesLearning2022};\\
                             REFINER~\citenumber{paulREFINERReasoningFeedback2024};  AutoMathCritique~\citenumber{xiEnhancingLLMReasoning2024}
                             }, leaf, fill=multiTurn-fill, draw=none
                        ]
                    ]
                    [
                        Improving the refinement\\ ability of LLMs, fill=multiTurn-fill, draw=multiTurn-border
                        [
                             {RISE~\citenumber{qu2024recursiveintrospectionteachinglanguage}; SCoRe~\citenumber{kumar2024traininglanguagemodelsselfcorrect}}, leaf, fill=multiTurn-fill, draw=none
                        ]
                    ]
                ]
                [
                    Long CoT~(\S\ref{sec:long_cot:efficiency}), fill=longCoT-fill, draw=longCoT-border
                    [
                        Prompting for conciseness, fill=longCoT-fill, draw=longCoT-border
                        [
                            {CCoT~\citenumber{nayab2025concisethoughtsimpactoutput}; CoD~\citenumber{xu2025chaindraftthinkingfaster}; BreakChain~\citenumber{ding2024breakchainlargelanguage}; SoT~\citenumber{aytes2025sketchofthoughtefficientllmreasoning}}, leaf, fill=longCoT-fill, draw=none
                        ]
                    ]
                    [
                        Finetuning on compressed\\ responses using heuristic methods, fill=longCoT-fill, draw=longCoT-border
                        [
                            {Dualformer~\citenumber{su2024dualformercontrollablefastslow}; ICoT-SI~\citenumber{deng2024explicitcotimplicitcot}; SPIRIT~\citenumber{cui2025stepwiseperplexityguidedrefinementefficient};\\ TokenSkip~\citenumber{xia2025tokenskipcontrollablechainofthoughtcompression}; C3oT~\citenumber{kang2024c3otgeneratingshorterchainofthought};\\DistillSystem2To1~\citenumber{yu2024distilling21}}, leaf, fill=longCoT-fill, draw=none
                        ]
                    ]
                    [
                        Query-aware compression, fill=longCoT-fill, draw=longCoT-border
                        [
                            Learning on trajectories with \\predefined optimal length, fill=longCoT-fill, draw=longCoT-border, text width=8.8em
                            [
                                 {Deepseek-R1~\citenumber{deepseekai2025deepseekr1incentivizingreasoningcapability}; TALE~\citenumber{han2025tokenbudgetawarellmreasoning}; \\ TOPS~\citenumber{yang2025thinkingoptimalscalingtesttimecompute}; DAST~\citenumber{shen2025dastdifficultyadaptiveslowthinkinglarge}; Z1~\citenumber{yu2025z1efficienttesttimescaling}}, leaf2, fill=longCoT-fill, draw=none
                            ]
                        ]
                        [
                            Self-training, fill=longCoT-fill, draw=longCoT-border
                            [
                                SFT, fill=longCoT-fill, draw=longCoT-border
                                [
                                     {Kimi K1.5~\citenumber{MoonshotAI}; LMSkip~\citenumber{liu2024languagemodelslearnskip}; \\ Munkhbat et al.~\citenumber{munkhbat2025selftrainingelicitsconcisereasoning} }, leaf2, fill=longCoT-fill, draw=none
                                ]
                            ]
                            [
                                DPO, fill=longCoT-fill, draw=longCoT-border
                                [
                                     {Kimi K1.5~\citenumber{MoonshotAI}; Overthinking~\citenumber{chen2025think23overthinkingo1like}; \\Sky-T1-32B-Flash~\citenumber{liEscapeSkyhighCost2024}}, leaf2, fill=longCoT-fill, draw=none
                                ]
                            ]
                            [
                                RL, fill=longCoT-fill, draw=longCoT-border
                                [
                                     {L1~\citenumber{aggarwal2025l1controllinglongreasoning}; Kimi K1.5~\citenumber{MoonshotAI};  \\ O1-Pruner~\citenumber{luo2025o1prunerlengthharmonizingfinetuningo1like}; Concise RL~\citenumber{fatemi2025concisereasoningreinforcementlearning};\\  Arora and Zanette~\citenumber{arora2025traininglanguagemodelsreason}; MRL~\citenumber{qu2025optimizingtesttimecomputemeta} }, leaf2, fill=longCoT-fill, draw=none
                                ]
                            ]
                        ]
                        [
                            Query router, fill=longCoT-fill, draw=longCoT-border, text width=8.8em
                            [
                                 {System-1.x~\citenumber{saha2024system1xlearningbalancefast}; Dynasor~\citenumber{fu2024efficientlyservingllmreasoning}}, leaf2, fill=longCoT-fill, draw=none
                            ]
                        ]
                    ]
                    [
                        Model merging, fill=longCoT-fill, draw=longCoT-border
                        [
                             {CoT-Valve~\citenumber{ma2025cotvalvelengthcompressiblechainofthoughttuning}; Kimi K1.5~\citenumber{MoonshotAI}; Wu et al.~\citenumber{wu2025unlockingefficientlongtoshortllm}}, leaf, fill=longCoT-fill, draw=none
                        ]
                    ]
                    [
                        Compressing the \\ intermediate state, fill=longCoT-fill, draw=longCoT-border
                        [
                             {InftyThink~\citenumber{yan2025inftythinkbreakinglengthlimits}; MCoT~\citenumber{yang2025markovchainthoughtefficient}; \\ AnLLMs~\citenumber{pang2024anchorbasedlargelanguagemodels}; LightThinker~\citenumber{zhang2025lightthinkerthinkingstepbystepcompression}}, leaf, fill=longCoT-fill, draw=none
                        ]
                    ]
                    [
                        Reasoning in the latent space, fill=longCoT-fill, draw=longCoT-border
                        [
                             {ICoT-KD~\citenumber{deng2023implicitchainthoughtreasoning}; Coconut~\citenumber{hao2024traininglargelanguagemodels};  CODI~\citenumber{shen2025codicompressingchainofthoughtcontinuous};  \\ Recurrence~\citenumber{geiping2025scalingtesttimecomputelatent}; SoftCoT~\citenumber{xu2025softcotsoftchainofthoughtefficient}; CCoT~\citenumber{cheng2024compressedchainthoughtefficient}}, leaf, fill=longCoT-fill, draw=none
                        ]
                    ]
                ]
            ]
        \end{forest}
    \caption{Overview of methods of improving scaling efficiency.}
    \label{fig:improving_scaling_efficiency}
\end{figure*}
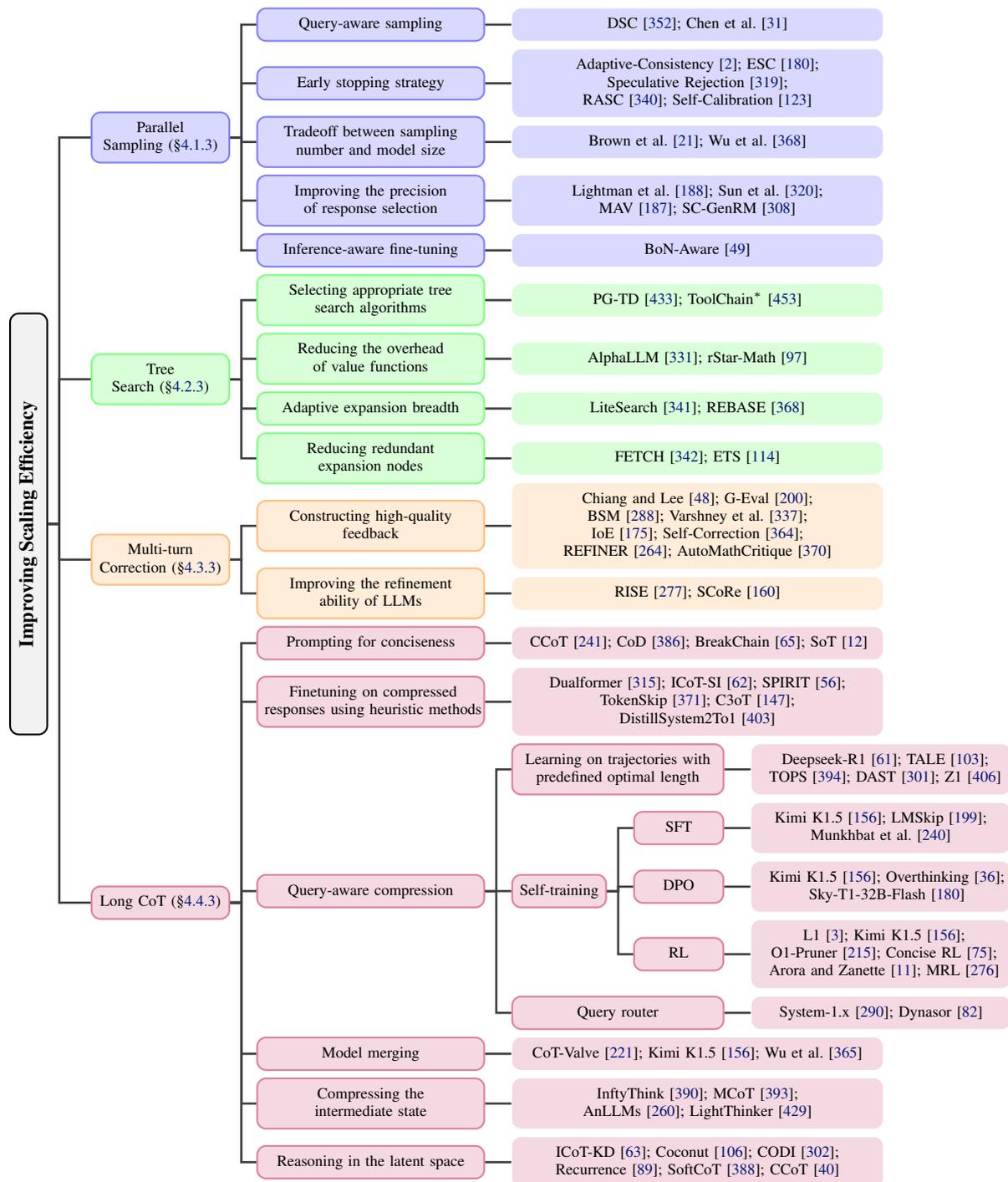
\subsubsection{Improving Scaling Efficiency} \label{sec:parallel_sampling:efficiency}

The strategies to improve the scaling efficiency of parallel sampling are as follows.
\paragraph{Query-aware sampling}
Applying a fixed sampling number for all queries is not optimal, as difficult problems require more sampling while easier ones need fewer. This line of methods employs adaptive sampling numbers for different queries based on difficulty to improve sampling efficiency. \citet{chenAreMoreLLM2024} categorize queries into easy and difficult cases according to model uncertainty and apply different sampling numbers accordingly. DSC~\cite{wangMakeEveryPenny2024} prompts the model to rank query difficulty and distributes sampling numbers based on this ranking.

\paragraph{Early stopping strategy} This method estimates the quality of responses during the sampling process and decides when to stop sampling early by utilizing prior knowledge or model estimation. It includes terminating sampling when observed answers are identical or fit predefined distributions within a small window size~\cite{aggarwalLetsSampleStep2023,liEscapeSkyhighCost2024,wanDynamicSelfConsistencyLeveraging2024}, or training the generator itself to estimate the confidence of the response and stopping once a high-confidence response is observed~\cite{huang2025efficienttesttimescalingselfcalibration}. Moreover, Speculative Rejection~\cite{sun2024fastbestofndecodingspeculative} and ST-BoN~\cite{wang2025samplingefficienttesttimescalingselfestimating} propose sampling responses in parallel and halting the decoding of responses with low reward model scores or self-estimation consistency scores to improve efficiency.

\paragraph{Tradeoff between sampling number and model size}
Given a fixed inference computation budget, there exists a tradeoff between using larger models with fewer samples versus smaller models with more samples, considering the different computational costs across model sizes. \citet{brownLargeLanguageMonkeys2024} observe that larger models perform better on code tasks while smaller models are more effective for mathematical tasks. \citet{wuInferenceScalingLaws2024} further find that for mathematical tasks, while smaller models are optimal, they also saturate earlier as the inference computation increases.

\paragraph{Improving the precision of response selection}
Considering the importance of effective selection mechanisms, several works focus on improving their precision in response selection.~\citet{lightman2023letsverifystepstep} find that PRM is superior to ORM in BoN settings.~\citet{sun2024easytohardgeneralizationscalablealignment} find that the performance of weighted majority voting is superior to majority voting or BoN when using large sampling numbers. MAV~\cite{lifshitz2025multiagentverificationscalingtesttime} employs multiple verifiers to assess response quality and achieves better performance than a single verifier  when the total computation budget for generator and verifiers is high.

\paragraph{Inference-aware fine-tuning}
\citet{chowInferenceAwareFineTuningBestofN2024} overcome the non-differentiable argmax operator within BoN sampling and develop BoN-Aware fine-tuning to directly optimize parallel sampling performance.

\subsection{Tree Search} \label{sec:tree_search}
\begin{table}[tb!]
    \centering
    \scriptsize
\caption{An organization of works on tree search. This table includes inference-only approaches, while work combining training strategies is discussed in \S\ref{sec:iterative_self-reinforced_learning}. Under \textbf{Value Function}, \textbf{E1} denotes self-evaluation, \textbf{E2} denotes specialized trained models, \textbf{E3} denotes likelihood of actions, \textbf{E4} denotes self-consistency score, \textbf{E5} denotes roll out.}

    \begin{tabular}{lcccc}
        \toprule
        \textbf{Work} & \textbf{Application} & \textbf{Search Space} & \textbf{Value Function} & \textbf{Search Algorithm} \\
        \midrule
        \rowcolor[rgb]{ .949,  .949,  .949} Pangu~\cite{guDontGenerateDiscriminate2023} & Knowledge Base QA & Step & E2 & Beam Search \\

        PG-TD~\cite{zhangPlanningLargeLanguage2023} & Code & Token & E5 & MCTS \\

         \rowcolor[rgb]{ .949,  .949,  .949} ToT~\cite{yao2023treethoughtsdeliberateproblem} & Game of 24, Writing, Crosswords & Step & E1 & BFS, DFS \\

        GuidedDecoding~\cite{xieSelfEvaluationGuidedBeam} & Math & Step & E1 & Beam Search \\

        \rowcolor[rgb]{ .949,  .949,  .949} RAP~\cite{haoReasoningLanguageModel2023} & Reasoning & Step & E1, E3, E4 & MCTS \\

         PPO-MCTS~\cite{liuDontThrowAway2024} & Alignment & Token & E2 & MCTS \\

      \rowcolor[rgb]{ .949,  .949,  .949}  LATS~\cite{zhouLanguageAgentTree2024} & Programming, Reasoning & Step & E1, E4, E5 & MCTS \\

         $\text{ToolChain}^*$~\cite{zhuangToolChainEfficientAction2023} & Tool-use, Reasoning & Step & E1, E3, E4 & A* \\

      \rowcolor[rgb]{ .949,  .949,  .949}   MindStar~\cite{kangMindStarEnhancingMath2024} & Math & Step & E2 & BFS \\
 
         $\text{Q}^*$~\cite{wangImprovingMultistepReasoning2024b} & Math, Code & Step & E2, E5 & A* \\

       \rowcolor[rgb]{ .949,  .949,  .949} LiteSearch~\cite{wang2024litesearchefficacioustreesearch} & Math & Step & E2 & BFS \\

          MCTSr~\cite{zhangAccessingGPT4Level2024a} & Math & Solution & E1 & MCTS \\
  
    \rowcolor[rgb]{ .949,  .949,  .949}     REBASE~\cite{wuInferenceScalingLaws2024} & Math & Step & E2 & BFS \\
 
        SearchAgent~\cite{koh2024treesearchlanguagemodel} & Web agents & Step & E1 & A* \\

   \rowcolor[rgb]{ .949,  .949,  .949}      rStar~\cite{qiMutualReasoningMakes2024a} & Math & Step & E4, E5 & MCTS \\

        PLANSEARCH~\cite{wangPlanningNaturalLanguage2024} & Code & Step & - & Beam Search \\

    \rowcolor[rgb]{ .949,  .949,  .949}    RethinkMCTS~\cite{li2024rethinkmctsrefiningerroneousthoughts} & Code & Step & E1, E5 & MCTS \\
    
         $\text{SC-MCTS}^*$~\cite{gaoInterpretableContrastiveMonte2024} & Blocksworld & Step & E1, E3 & MCTS \\
   
    \rowcolor[rgb]{ .949,  .949,  .949}     LLaMA-Berry~\cite{zhang2024llamaberrypairwiseoptimizationo1like} & Math & Solution & E2 & MCTS \\
       
         ETS~\cite{hooper2025etsefficienttreesearch} & Math & Step & E2 & BFS \\

        \bottomrule
    \end{tabular}
    \label{tab:tree_search_summary}
\end{table}
\subsubsection{Key Components} \label{sec:tree_search:key_components}
The tree search method frames problems as searches over tree structures. Guided by a specific tree search algorithm, the generator searches in the search space $S$ and explore different problem-solving approaches, usually accompanied by value functions to assess node values in $S$. This framework enhances the model's deliberate planning ability. We detail each component below.

\paragraph{Search space}
The search space defines the granularity of tree nodes, which significantly impacts search efficiency. It can be categorized as follows:
\begin{itemize}
\item \textbf{S1: Token.} Token-level search increases the optimality of candidate solutions but also incurs high computational costs due to its fine-grained nature. This approach is suitable for scenarios with low tolerance for even minor errors in individual tokens. PG-TD~\citep{zhangPlanningLargeLanguage2023} implements the Monte Carlo Tree Search (MCTS) algorithm for code tasks at the token level, as even minor changes in code may cause errors. PPO-MCTS~\citep{liuDontThrowAway2024} employs token-level search methods to improve the helpfulness and harmlessness of responses.

\item \textbf{S2: Step.} Step-level search balances search granularity and efficiency, making it the most common approach. The definition of ``step'' varies across different tasks. It can be sentences in solutions for reasoning problems~\citep{yao2023treethoughtsdeliberateproblem,xieSelfEvaluationGuidedBeam,haoReasoningLanguageModel2023}, actions in a simulated world~\citep{guDontGenerateDiscriminate2023,zhuangToolChainEfficientAction2023}, lines of code~\citep{wangPlanningNaturalLanguage2024}, or proposed plans or hypotheses~\citep{yao2023treethoughtsdeliberateproblem,wangCPLCriticalPlan2024, wangHYPOTHESISSEARCHINDUCTIVE2024}.

\item \textbf{S3: Solution.} Solution-level search\footnote{It is noted in some works~\citep{zeng2024scalingsearchlearningroadmap} that they consider parallel sampling as a solution-level tree search. However, for parallel sampling, the solution tree nodes only involve the original solutions and do not involve node expansion behaviors, which are vital for tree search. Therefore, we do not merge it.} considers the expansion of tree nodes as updates to the whole response through actions like critique and revision~\citep{zhangAccessingGPT4Level2024a,zhang2024llamaberrypairwiseoptimizationo1like}. It overlaps with the multi-turn correction framework discussed later, and we also consider it as an ensemble of the two methods.
\end{itemize}

\paragraph{Value function} The value function estimates the value of candidate nodes for further pruning or exploitation. The popular methods to construct value functions are as follows:
\begin{itemize}
\item \textbf{E1: Self-evaluation.} This method directly instructs the generator to evaluate node values through well-crafted prompts. ToT~\citep{yao2023treethoughtsdeliberateproblem} proposes to value each node independently or vote across nodes. ~\citet{xieSelfEvaluationGuidedBeam} design prompts in the form of multiple-choice questioning to better calibrate model predictions.

\item \textbf{E2: Specialized trained models.} To reduce evaluation noise, this method utilizes specialized trained LLMs for evaluation~\citep{guDontGenerateDiscriminate2023, kangMindStarEnhancingMath2024}. This introduces process reward models (PRMs) for evaluating reasoning steps and token-level value functions for more fine-grained evaluation~\citep{lee2024tokensupervisedvaluemodelsenhancing}. For the PRM, it can be trained through the following approaches: 
  1) Human annotations: \citet{lightman2023letsverifystepstep, uesato2022solvingmathwordproblems} employ human labelers to label the correctness of each step. This method is costly and still cannot avoid noise in the training data; 
  2) Monte Carlo sampling: to achieve autonomous data annotation, this method employs Monte Carlo sampling that rolls out multiple completions from the current steps and estimates the rate leading to correct results~\citep{wangMathShepherdVerifyReinforce2024b, wang2024multistepproblemsolvingverifier, havrilla2024glorewhenwhereimprove, luo2024improvemathematicalreasoninglanguage}. To improve sampling efficiency, OmegaPRM~\citep{luo2024improvemathematicalreasoninglanguage} utilizes binary search for error locating and integrates data collection into the search process; 
  3) From ORM to PRM: To avoid the high cost of training a PRM, this method aims to derive a PRM from an ORM. AutoPSV~\citep{lu2024autopsvautomatedprocesssupervisedverifier} utilizes the ORM to automatically generate process annotations for each reasoning step by detecting its own confidence variations and thus uses the data for PRM training. \citet{yuan2024freeprocessrewardsprocess} theoretically demonstrate that a PRM can automatically derive from an ORM through simple reward parameterization. In the PRM utilization phase, \citet{setlur2024rewardingprogressscalingautomated} demonstrate that process reward for a step should be advantages calculated by the difference of adjacent step values. For token-level value functions, these can directly come from the value functions in the post-training phase~\citep{liuDontThrowAway2024} or training on data from Monte Carlo sampling~\citep{lee2024tokensupervisedvaluemodelsenhancing}.

\item \textbf{E3: Likelihood of actions.} The likelihood-based approach utilizes the generator's probability of conducting a specific action (i.e., the tree node) to estimate the node value~\citep{haoReasoningLanguageModel2023,gaoInterpretableContrastiveMonte2024}.

\item \textbf{E4: Self-consistency score.} The frequency of intermediate nodes can represent the model's confidence in them, thus being used for evaluation~\citep{qiMutualReasoningMakes2024a, zhuangToolChainEfficientAction2023, zhouLanguageAgentTree2024}. LATS~\citep{zhouLanguageAgentTree2024} combines the self-generated LLM score and the self-consistency score for node value. rStar~\citep{qiMutualReasoningMakes2024a} utilizes the self-consistency score as the reward for the terminal node.

\item \textbf{E5: Roll out.} In algorithms like MCTS, the intermediate node value can be estimated by rollout and further updated based on backup from terminal states~\citep{qiMutualReasoningMakes2024a}. The reward for terminal states can come from external tools like code interpreters or the aforementioned evaluation methods.
\end{itemize}

\begin{figure}
    \centering
    \includegraphics[width=1\linewidth]{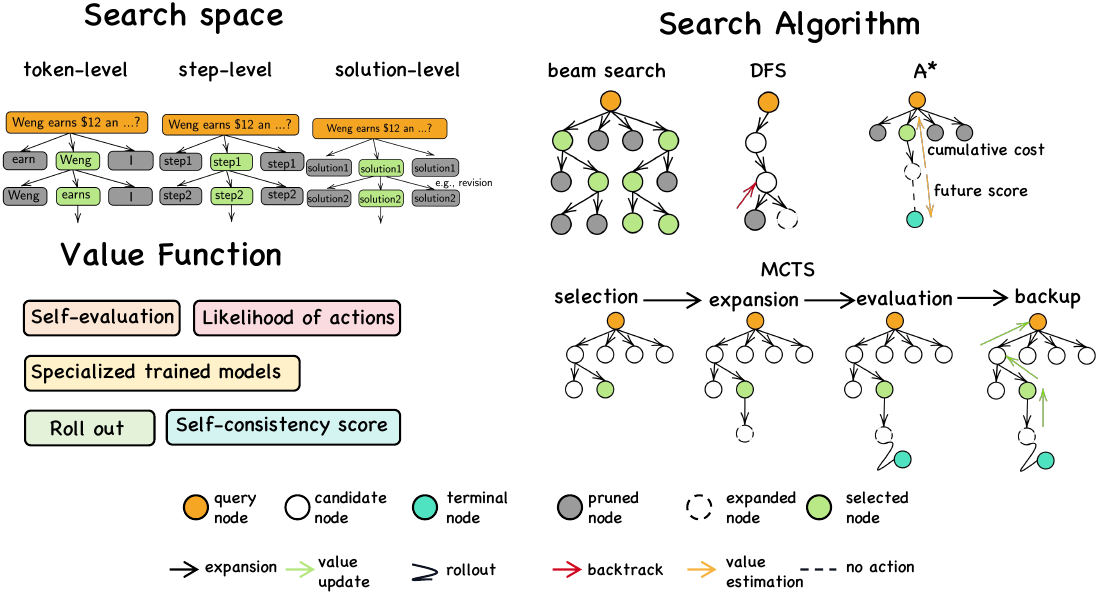}
    \caption{Illustration of key components of tree search.}
    \label{fig:tree_search:illustration}
\end{figure}

\paragraph{Search Algorithm} The search algorithm defines the operational rules for tree nodes. It can be instantiated as follows:
\begin{itemize}
    \item \textbf{A1: Breadth-first Search (BFS).} The widely used versions of BFS algorithm include beam search and A* algorithm. For the beam search algorithm, it generates $k$ candidate nodes at each layer and selects the most promising $m$ nodes from them based on node values~\citep{guDontGenerateDiscriminate2023,yao2023treethoughtsdeliberateproblem,xieSelfEvaluationGuidedBeam}. For the A* algorithm~\citep{hart1968formal}, it calculates the sum of the cumulative cost (i.e., the cost from the root node to the current node) and the future score (i.e., the cost of the path from the current node to the goal) in the search process and always selects the node with minimum value. $\text{ToolChain}^*$~\citep{zhuangToolChainEfficientAction2023} mainly relies on the heuristic function to calculate the score, and $\text{Q}^*$~\citep{wangImprovingMultistepReasoning2024b} utilizes the process reward model and roll out method for estimation.
    
    \item \textbf{A2: Depth-first Search (DFS).} The DFS algorithm explores the most promising node first until the node is no longer promising or the
final output is reached, then backtracks to the parent node to explore alternative thoughts~\citep{yao2023treethoughtsdeliberateproblem}. \citet{longLargeLanguageModel2023} implement the DFS algorithm in the sudoku puzzle, where a checker module utilizes the sudoku rules to check the validity of the partial solution and a controller controls the backtracking behaviors of LLMs.
    
    \item \textbf{A3: Monte Carlo Tree Search (MCTS).} A line of work proposes using the advanced MCTS algorithm to improve the planning ability of LLMs~\citep{zhangPlanningLargeLanguage2023,haoReasoningLanguageModel2023,liuDontThrowAway2024}, considering its success in AlphaGo~\citep{MasteringGameGo}. The key implementation of MCTS lies in four processes: selection, expansion, evaluation, and backup (see Figure~\ref{fig:tree_search:illustration} for the illustration). 
    The selection phase traverses the tree from the root, iteratively selecting the most promising child nodes by balancing exploration and exploitation, until reaching an underexplored node. Widely used methods like Upper Confidence Bound applied on Trees Algorithm (UCT)~\citep{kocsis2006bandit} and Predictor + UCT~\citep{rosin2011multi} (PUCT) guide this choice, favoring nodes with high values while adjusting for visit frequency to prevent overconcentration on heavily explored nodes. The expansion operation grows the tree by adding one or more new child nodes to the underexplored node based on possible actions from the current node's state. The evaluation process evaluates a new child node by methods described in previous value function paragraph, such as rollouts to a terminal state or direct evaluation with LLMs. In backup, the evaluation result is propagated upward along the selected path, updating the values and visit counts of all nodes traversed.
\end{itemize}

Table~\ref{tab:tree_search_summary} presents an organization of works on tree search based on the established taxonomy. Additionally, Table~\ref{tab:application_tts} presents more works applying tree search across various domains.

\subsubsection{Scaling Laws} \label{sec:tree_search:scaling_factor}
Empirical results show that performance can be further enhanced by scaling the breadth and depth of tree search. This includes increasing the number of rollouts in the MCTS algorithm~\cite{zhangPlanningLargeLanguage2023,liuDontThrowAway2024,zhangAccessingGPT4Level2024a,qiMutualReasoningMakes2024a}, the beam size in beam search~\cite{yao2023treethoughtsdeliberateproblem,xieSelfEvaluationGuidedBeam}, and the step limitations in the A* algorithm~\cite{zhuangToolChainEfficientAction2023}. \citet{snell2024scalingllmtesttimecompute} analyze the scaling behavior and finds that performance will eventually saturate. This may be due to the model struggling to produce diverse nodes as the number of sampled candidates increases. Additionally, \citet{kangMindStarEnhancingMath2024} find that increasing the model size of the PRM employed in the search algorithm can enhance performance. This highlights the importance of improving the reliability of value functions through extra training time or test-time compute.

\subsubsection{Improving Scaling Efficiency}
\label{sec:tree_search:efficiency}
The strategies to improve the scaling efficiency of tree search are as follows:

\paragraph{Selecting appropriate tree search algorithms} The characteristics of different tree search algorithms make them suitable for different tasks. For code generation tasks, PG-TD~\cite{zhangPlanningLargeLanguage2023} compares MCTS using public test cases for terminal state evaluation against simple beam search, finding that MCTS achieves significantly better performance given the same computation time. $\text{ToolChain}^*$~\cite{zhuangToolChainEfficientAction2023} demonstrates that the A* algorithm is more time-efficient than MCTS and other alternatives in API call tasks. 

\paragraph{Reducing the overhead of value functions} For algorithms like MCTS, reliable value functions traditionally rely on multiple rollouts, which incur significant computational costs. ALPHALLM~\cite{tianSelfImprovementLLMsImagination2024b} employs a smaller language model as a fast rollout policy to reduce computational overhead. rStar-Math~\cite{guanRStarMathSmallLLMs2025} introduces a two-phase approach: first estimating node values through rollouts, then using this data to train a separate value function that replaces rollouts in subsequent iterations.

\paragraph{Adaptive expansion breadth} Traditional beam search algorithms fix the expansion breadth of each node to a constant value, which may not optimally balance exploration and exploitation. LiteSearch~\cite{wang2024litesearchefficacioustreesearch} allocates expansion breadth based on node value and depth, encouraging exploration on high-value nodes and at the beginning of the search process. This approach helps achieve higher token efficiency than beam search and DFS. REBASE~\cite{wuInferenceScalingLaws2024} takes a similar strategy by defining trajectory collection requirements and dynamically allocating expansion breadth at each depth based on node value and remaining collection needs, resulting in higher efficiency compared to traditional MCTS algorithms.

\paragraph{Reducing redundant expansion nodes} In the node expansion process, there may exist nodes with semantically equivalent content, leading to unnecessary exploration costs. FETCH~\cite{wang2025dontlosttreesstreamlining} and ETS~\cite{hooper2025etsefficienttreesearch} merge semantically similar nodes using agglomerative clustering of text embeddings obtained from a fine-tuned model, achieving higher token efficiency.

\subsection{Multi-turn Correction} \label{sec:multi-turn_correction}
\begin{table}[tb!]
    \centering
    \scriptsize
\caption{An organization of works on multi-turn correction. \textbf{Fine-tuning} indicates whether the method requires additional training. \brickcheck~in the \textbf{Self-feedback} and \textbf{Self-refinement} columns represents that it shares the same parameters with the initial generator but is prompted with different roles.}
\begin{tabular}{lcccccc}
    \toprule
   \multirow{2}{*}{\textbf{Work}} & 
   \multicolumn{2}{c}{\textbf{Feedback}} &
   \multicolumn{2}{c}{\textbf{Refinement}} & \multirow{2}{*}{\textbf{Fine-tuning}} \\
   \cmidrule(lr){2-3} \cmidrule(lr){4-5}
  & \textbf{Self-feedback} & \textbf{External} & \textbf{Self-refinement} & \textbf{External} & \\
    \midrule
    \rowcolor[rgb]{ .949,  .949,  .949} Self-Correction~\cite{welleckGeneratingSequencesLearning2022} & \blackcross & \blackcross  & \blackcross & Trained LM & \blackcheck \\
    Self-refine~\cite{madaanSELFREFINEIterativeRefinement} & \blackcheck & \blackcross &  \blackcheck & \blackcross & \blackcross \\
    \rowcolor[rgb]{ .949,  .949,  .949} Reflexion~\cite{shinnReflexionLanguageAgents2023a} & \brickcheck & Game Envs; Interpreter; Oracle & \blackcheck & \blackcross & \blackcross \\
    RCI~\cite{kim2023languagemodelssolvecomputer} & \blackcheck & Oracle & \blackcheck & \blackcross & \blackcross \\
    \rowcolor[rgb]{ .949,  .949,  .949} Self-Debug~\cite{chen2023teachinglargelanguagemodels} & \blackcheck & Interpreter & \blackcheck & \blackcross & \blackcross \\
    Baldur~\cite{first2023baldurwholeproofgenerationrepair} & \blackcross & Proof checker & \blackcross & Trained LM & \blackcheck \\
    \rowcolor[rgb]{ .949,  .949,  .949} REFINER~\cite{paulREFINERReasoningFeedback2024} & \blackcross & Trained LM & \blackcross  & Trained LM & \blackcheck \\
    LLM-Debate~\cite{du2023improvingfactualityreasoninglanguage} & \brickcheck & \blackcross & \brickcheck & \blackcross & \blackcross \\
    \rowcolor[rgb]{ .949,  .949,  .949} MAD~\cite{liang2024encouragingdivergentthinkinglarge} & \brickcheck & \blackcross & \brickcheck & \blackcross & \blackcross \\
    CRITIC~\cite{gou2024criticlargelanguagemodels} & \blackcheck & Search engine; Interpreter & \blackcheck & \blackcross & \blackcross \\
    \rowcolor[rgb]{ .949,  .949,  .949} CoVe~\cite{dhuliawala2023chainofverificationreduceshallucinationlarge} & \blackcheck & \blackcross & \blackcheck & \blackcross & \blackcross \\
    RISE~\cite{qu2024recursiveintrospectionteachinglanguage} & \blackcross & \blackcross & \blackcheck & \blackcross & \blackcheck \\
    \rowcolor[rgb]{ .949,  .949,  .949} IHR~\cite{qiuPHENOMENALPUZZLINGTESTING2024} & \blackcross & Interpreter & \blackcheck & \blackcross & \blackcross \\
    SCoRe~\cite{kumar2024traininglanguagemodelsselfcorrect} & \blackcross & \blackcross & \blackcheck & \blackcross & \blackcheck \\
    \rowcolor[rgb]{ .949,  .949,  .949} AutoMathCritique~\cite{xiEnhancingLLMReasoning2024} & \blackcross & Trained LM & \blackcross & Trained LM & \blackcheck \\
    DARS~\cite{li2025headsbetteronedualmodel} & \blackcross &  Trained LM & \blackcross &  Trained LM   & \blackcheck\\    
    \bottomrule
\end{tabular}
    \label{tab:multi_turn_correction_summary}
\end{table}

\subsubsection{Key Components}
Multi-turn correction aims to improve response quality through iterative revision. It consists of an initial generator $g_{0}$ that proposes the initial response, a feedback model $f$ that generates feedback for the latest response, and a refinement model $g$ that revises the response given the interaction history \cite{welleck2024decodingmetagenerationinferencetimealgorithms}:
\begin{align}
y^{0} &\sim g_{0}(y|x) \\
z^{t} &\sim f(z|x,y^{(<t)},z^{(<t)}) \\
y^{t} &\sim g(y|x,y^{(<t)},z^{(\leq t)})
\end{align}
where $x$ represents the query, $y^{t}$ represents the response at timestep $t$, and $z^{t}$ represents the feedback at timestep $t$. The system outputs the final response when a stopping condition is met. The feedback generation stage can be omitted, resulting in direct refinement of the initial response~\cite{welleckGeneratingSequencesLearning2022,kamoi-etal-2024-llms}.

Multi-turn correction imitates human reflection and refinement cognitive processes.  The core design of it lies in constructing reliable feedback signals and refinement models to improve response quality. Feedback sources can be categorized as follows~\cite{pan2023automaticallycorrectinglargelanguage}:
\begin{itemize}
\item \textbf{F1: Self-feedback.} The initial generator $g_0$ and the feedback model can share a single language model, resulting in self-feedback. For example, Self-Debug~\cite{chen2023teachinglargelanguagemodels} instructs $g_0$ to explain code line by line and generate execution traces as feedback signals. Self-Refine~\cite{madaanSELFREFINEIterativeRefinement} incentivizes $g_0$ to generate feedback using reflective prompts. Moreover, $g_0$ can be prompted with different roles to encourage divergent thinking~\cite{du2023improvingfactualityreasoninglanguage,liang2024encouragingdivergentthinkinglarge,khanDebatingMorePersuasive2024}, a technique known as ``multi-agent debate.''

\item \textbf{F2: External feedback.} The feedback can come from external sources to $g_0$, including: 1) external tools: such as code interpreters~\cite{chen2023teachinglargelanguagemodels,gou2024criticlargelanguagemodels,shinnReflexionLanguageAgents2023a}, proof checkers~\cite{first2023baldurwholeproofgenerationrepair}, game simulators~\cite{shinnReflexionLanguageAgents2023a}; 2) external knowledge~\cite{gou2024criticlargelanguagemodels,zhao2023verifyandeditknowledgeenhancedchainofthoughtframework}; 3) oracle labels: such as ground truth answers to math problems~\cite{shinnReflexionLanguageAgents2023a}, though these are not guaranteed to be available in real-world applications~\cite{huang2024largelanguagemodelsselfcorrect}; 4) specialized trained models~\cite{paulREFINERReasoningFeedback2024,xiEnhancingLLMReasoning2024}.
\end{itemize}

The refinement model can also be instantiated similarly to the feedback models, including self-refining the response~\cite{madaanSELFREFINEIterativeRefinement,shinnReflexionLanguageAgents2023a}, or using a specialized trained model~\cite{welleckGeneratingSequencesLearning2022}. Specifically, Self-Refine~\cite{madaanSELFREFINEIterativeRefinement} instantiates the initial generator, feedback model, and refinement model with the same language model. Table~\ref{tab:multi_turn_correction_summary} presents an organization of works on multi-turn correction based on the established taxonomy. Furthermore, Table~\ref{tab:application_tts} showcases more studies that implement multi-turn correction techniques across diverse application domains.

Research demonstrates that with reliable external feedback, multi-turn correction significantly enhances model performance across diverse tasks~\cite{kamoi-etal-2024-llms}. However, this approach has faced criticism because high-quality external feedback is often unavailable in real-world scenarios~\cite{huang2024largelanguagemodelsselfcorrect}. In the \emph{intrinsic self-correction} setting, where a model critiques and revises its own responses without external feedback, empirical studies indicate that LLMs generally struggle to generate reliable critiques and revisions, particularly in planning~\cite{valmeekamCanLargeLanguage2023,stechlyGPT4DoesntKnow2023} and reasoning~\cite{huang2024largelanguagemodelsselfcorrect,tyen2024llmsreasoningerrorscorrect} tasks, leading to little or no performance gains. It has also been observed that self-biases can amplify during the self-correction process~\cite{xuPridePrejudiceLLM2024}.

\begin{figure}[t]
    \centering
    \includegraphics[width=1\linewidth]{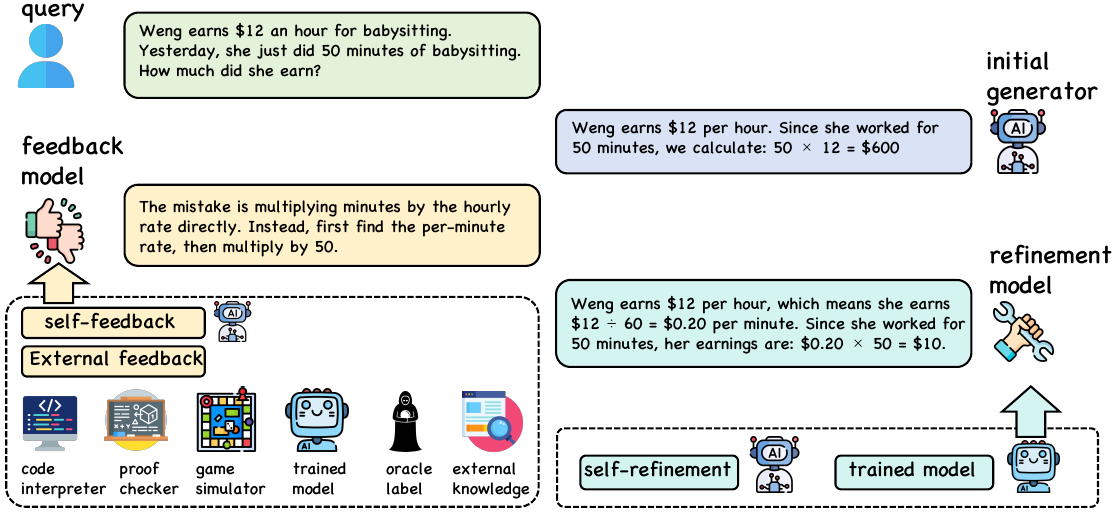}
    \caption{Illustration of key components of multi-turn correction.}
    \label{fig:multi-turn_correction:illustration}
\end{figure}
\subsubsection{Scaling Laws} \label{sec:multi-turn_correction:scaling_factor}
In tasks with reliable external feedback or correctors, performance can be further improved by scaling the number of revision steps until it finally saturates~\cite{welleckGeneratingSequencesLearning2022,madaanSELFREFINEIterativeRefinement,du2023improvingfactualityreasoninglanguage,qiuPHENOMENALPUZZLINGTESTING2024}. In intrinsic self-correction settings where the model lacks critique ability, increasing the revision steps can harm performance~\cite{welleckGeneratingSequencesLearning2022,huang2024largelanguagemodelsselfcorrect}. This limitation can be addressed through additional training to improve self-correction ability~\cite{qu2024recursiveintrospectionteachinglanguage,snell2024scalingllmtesttimecompute}.~\citet{snell2024scalingllmtesttimecompute} observe that after fine-tuning the model to improve its correction ability, performance grows steadily as revision steps increase and eventually saturates, even beyond the revision number used during training. This highlights that additional investment in training compute before deploying multi-turn correction can expand the ceiling of scaling test time.

\subsubsection{Improving Scaling Efficiency}\label{sec:multi-turn_correction:efficiency}

As discussed above, the effectiveness of multi-turn correction is constrained by the reliability of feedback and the model's refinement capabilities. To enhance scaling efficiency, efforts should concentrate on developing high-quality feedback mechanisms or improving the model's refinement abilities.

\paragraph{Constructing high-quality feedback} Reliable feedback is often limited to specific task types. Several strategies can improve feedback quality for broader applications: 1) Using reference-free LLM-based evaluation metrics with human-written evaluation criteria~\cite{chiang2023largelanguagemodelsalternative,liu2023gevalnlgevaluationusing}; 2) Employing task-specific decomposition~\cite{saha2024branchsolvemergeimproveslargelanguage} to break down complex verification into manageable subtasks; 3) Leveraging confidence estimation through generation probabilities~\cite{varshney2023stitchtimesavesnine} or prompting techniques~\cite{li2024confidencemattersrevisitingintrinsic}; 4) Fine-tuning models specifically for feedback~\cite{welleckGeneratingSequencesLearning2022,paulREFINERReasoningFeedback2024,xiEnhancingLLMReasoning2024}.

\paragraph{Improving the refinement ability of LLMs} Instead of focusing solely on constructing high-quality feedback, this line of work directly improves the refinement ability of models through additional training-time compute without the feedback models. RISE~\cite{qu2024recursiveintrospectionteachinglanguage} generates synthetic multi-turn correction data by concatenating the incorrect response before the final correct response and fine-tunes the model on these examples to improve its refinement ability. SCoRe~\cite{kumar2024traininglanguagemodelsselfcorrect} further identifies the behavior collapse issues in the SFT-based method and proposes multi-turn RL training with carefully designed rewards for different turns.

\subsection{Long CoT} \label{sec:long_cot}

\subsubsection{Key Components}
CoT prompting~\cite{wei2023chainofthoughtpromptingelicitsreasoning,nye2021workscratchpadsintermediatecomputation} instructs models to generate human-readable explanations of how problems are solved. This approach can help improve models' representational complexity~\cite{merrill2024expressivepowertransformerschain,nowak2025representationalcapacityneurallanguage} and significantly enhances their performance in reasoning tasks~\cite{wei2023chainofthoughtpromptingelicitsreasoning}. Current models like ChatGPT or Llama 3.1 default to CoT when presented with reasoning problems~\cite{sprague2024cotcotchainofthoughthelps}. Despite its widespread application, the reasoning process in CoT is usually shallow and linear, revealing limitations in complex cognitive capabilities~\cite{Kambhampati_2024,chen2025reasoningerasurveylong}. Recently, models like OpenAI o1~\cite{openai_o1_system_card} or Deepseek R1~\cite{deepseekai2025deepseekr1incentivizingreasoningcapability} have advanced the traditional CoT into long CoT, which incorporates more sophisticated thinking patterns and extended responses. The cognitive patterns present in long CoT but typically less observed in traditional CoT are as follows.

\begin{itemize}
    \item \textbf{Reflection:} The model develops metacognitive abilities~\cite{metcalfe1994metacognition} to assess the correctness and rationality of its own responses. For example, the model may pause its reasoning by outputting ``wait'' when it detects potential issues.
    \item \textbf{Backtracking:} When the model detects an error in its response, it can return to previous steps and revise them. This capability is vital for long-horizon planning problems, such as sudoku and code-breaking. In these problems, the model must find optimal solutions among multiple possibilities, and since the initial solution is not guaranteed to be correct, the model needs to employ trial and error.
    \item \textbf{Verification:} The model learns to recheck both individual steps and complete solutions, which enhances the robustness of its problem-solving approach.
    \item \textbf{Divergent thinking:} When the model recognizes that the current solution cannot solve the problem or leads to an obviously wrong answer, it can employ divergent thinking to explore alternative solutions, often signaled by transitional phrases like ``alternatively.''
    \item \textbf{Internal thinking:} The model can generate human-like thinking processes beyond explicit problem-solving steps. This enables more fine-grained reasoning before generating each subsequent step, thereby improving its overall performance~\cite{wu2024thinkingllmsgeneralinstruction}.
\end{itemize}

\subsubsection{Scaling Laws}
\label{sec:long_cot:scaling_factor}
Early work demonstrates that extending reasoning steps significantly enhances LLMs' reasoning capabilities~\cite{jin2024impactreasoningsteplength}. In the context of long CoT models, recent studies have identified a positive correlation between token count and model performance. Although not explicitly describing token control methodologies,~\citet{openai_o1_system_card} and~\citet{deepseekai2025deepseekr1incentivizingreasoningcapability} discover that performance increases with token count following a log-linear relationship. More transparent research by~\citet{hou2025advancinglanguagemodelreasoning} and~\citet{muennighoff2025s1simpletesttimescaling} applies response post-processing or decoding techniques to regulate token count, revealing a positive association between response length and performance. Specifically,~\citet{hou2025advancinglanguagemodelreasoning} truncate responses to varying lengths from the beginning and suggest using a summarization model to extract final answers.~\citet{muennighoff2025s1simpletesttimescaling} develop a budget-forcing technique to control token count through the addition or suppression of end-of-thinking token delimiters.

Despite these studies providing substantial evidence for the positive correlation between token number and model performance, debate on the effectiveness of extensive response length remains. These debates primarily stem from observations that shorter response lengths yield higher accuracy than longer responses~\cite{zeng2025revisitingtesttimescalingo1like,ballon2025relationshipreasoningperformancelarge}. This phenomenon may be explained by models generating more tokens for more challenging problems where failure risks are higher, or by approaches chosen in longer responses being more convoluted than those in shorter responses, thus increasing the likelihood of failure~\cite{fatemi2025concisereasoningreinforcementlearning}.

\subsubsection{Improving Scaling Efficiency} \label{sec:long_cot:efficiency}

Although long CoT endows models with deep thinking abilities, it can lead to overthinking problems. For instance, models might generate hundreds of tokens for simple questions like ``2+3=5''~\cite{chen2025think23overthinkingo1like}, where the correct answer is reached early but followed by unnecessary reasoning. Furthermore, CoT-based methods operate in the language space, allocating similar computational resources to each token regardless of its importance. This uniform allocation is suboptimal since some tokens, like those maintaining text coherence, require minimal planning, while others crucial to the reasoning process demand more intensive processing~\cite{hao2024traininglargelanguagemodels}. We detail techniques to resolve these issues below.\footnote{It is noted that some of the work focuses on traditional CoT instead of long CoT, but we include them considering their easy generalization.}

\paragraph{Prompting for conciseness} This approach directly instructs models to limit response tokens to a specific number~\cite{nayab2025concisethoughtsimpactoutput,xu2025chaindraftthinkingfaster} or capture only essential information~\cite{ding2024breakchainlargelanguage,aytes2025sketchofthoughtefficientllmreasoning} through prompting. Although straightforward to implement, its effectiveness is limited to simple tasks, and LLMs cannot strictly adhere to token number restrictions~\cite{muennighoff2025s1simpletesttimescaling,aggarwal2025l1controllinglongreasoning}.

\paragraph{Finetuning on compressed responses using heuristic methods}
This approach first compresses CoT responses using heuristic methods and then finetunes on them. The heuristic compression techniques include directly removing intermediate steps \cite{su2024dualformercontrollablefastslow,deng2024explicitcotimplicitcot}, assessing token importance in CoT through perplexity\cite{cui2025stepwiseperplexityguidedrefinementefficient} or a specifically trained model~\cite{xia2025tokenskipcontrollablechainofthoughtcompression} to retain only the most relevant tokens, and leveraging advanced models like GPT-4 to reconstruct CoT sequences while preserving essential information and eliminating redundancy~\cite{kang2024c3otgeneratingshorterchainofthought}. The effectiveness of this method heavily depends on the design of the heuristic compression techniques, limiting its generalizability across tasks. For example, C3oT~\cite{kang2024c3otgeneratingshorterchainofthought} finds that training directly on GPT-4-compressed data significantly degrades task performance, necessitating the inclusion of original uncompressed data during training.

\paragraph{Query-aware compression} The compression limitation of response length varies based on query type~\cite{lee2025llmscompresschainofthoughttoken,arora2025traininglanguagemodelsreason}, as difficult problems require more tokens while easier ones require fewer. This method aims to approach the limitation in a query-aware way, which helps improve token efficiency while maintaining or improving the models' adaptivity in computational resource allocation. The methods are as follows:

\begin{itemize}
    \item \textbf{Learning on trajectories with predefined optimal length:} This approach first determines optimal length explicitly and trains on them. The reference for the optimal length can be based on task types. For example, in the supervised fine-tuning phase of DeepSeek-R1~\cite{deepseekai2025deepseekr1incentivizingreasoningcapability}, for reasoning tasks they collect long CoT responses for training, while for non-reasoning tasks they collect CoT responses for certain tasks and even no-CoT responses for simpler queries. These approaches help the model learn to switch reasoning modes based on the query type. Beyond this, other estimations for the optimal length can be based on search~\cite{han2025tokenbudgetawarellmreasoning,yang2025thinkingoptimalscalingtesttimecompute}, prompting~\cite{han2025tokenbudgetawarellmreasoning}, or query difficulty estimated by sampling~\cite{shen2025dastdifficultyadaptiveslowthinkinglarge}. These selected optimal-length trajectories can be used for further SFT or DPO training.

    \item \textbf{Self-training:} Instead of predefining the optimal length, this method first rolls out trajectories from the generator and incentivizes the model to achieve fewer tokens while maintaining accuracy through self-training, which can be considered an on-policy optimal-length estimation. The training methods can be: 1) SFT: This approach generates multiple responses for each question and selects the shorter correct ones for supervised fine-tuning~\cite{MoonshotAI,munkhbat2025selftrainingelicitsconcisereasoning,liu2024languagemodelslearnskip}; 2) DPO: This method uses the long-CoT model to generate multiple response samples, selecting the shorter correct solution as the positive sample while treating longer responses as negative samples. These positive-negative pairs form the pairwise preference data used for preference learning. For preference data construction,~\citet{chen2025think23overthinkingo1like} find that choosing responses including two solving attempts that reach the correct answer as positive examples performs best. Sky-T1-32B-Flash~\cite{liEscapeSkyhighCost2024} employs multiple preference data construction methods to avoid accuracy drops while reducing reasoning length. For the training algorithm,~\citet{chen2025think23overthinkingo1like} empirically demonstrate that SimPO~\cite{meng2024simposimplepreferenceoptimization} performs better than DPO.
    3) RL: This approach adds a length penalty in the reward function to reduce response length~\cite{aggarwal2025l1controllinglongreasoning,MoonshotAI,luo2025o1prunerlengthharmonizingfinetuningo1like,arora2025traininglanguagemodelsreason} or designs dense reward in the intermediate steps~\cite{qu2025optimizingtesttimecomputemeta}. For example, L1~\cite{aggarwal2025l1controllinglongreasoning} adds a length control factor in the RL reward function to train the model to adhere to the length given in the prompt or not exceed the maximum length. Furthermore, MRL~\cite{qu2025optimizingtesttimecomputemeta} measures progress at each intermediate generation episode through on-policy rollouts and develops corresponding SFT and RL methods for maximizing dense rewards based on the progress. While there is no explicit length-relevant factor in the algorithm, it helps the model balance exploration and exploitation in the content of CoT and improve token efficiency.

    \item \textbf{Query router:} This method classifies queries as difficult or easy and handles them differently by applying different types of models~\cite{saha2024system1xlearningbalancefast} or different computation budgets of the same model~\cite{fu2024efficientlyservingllmreasoning}. For example, System-1.x~\cite{saha2024system1xlearningbalancefast} trains a controller that decomposes a planning problem into sub-goals and classifies them as easy or hard to be solved by either the System-1 planner or the System-2 planner.
\end{itemize}

\paragraph{Model merging} This method combines a long-CoT model with a short-CoT model to create a new model without additional training. CoT-Valve~\cite{ma2025cotvalvelengthcompressiblechainofthoughttuning} manipulates the weights between the parameters of the two models to achieve varying lengths.

\paragraph{Compressing the intermediate state} In the response generation process, the storage overhead of the KV cache increases linearly with the context length for the Transformer architecture. This line of work aims to compress intermediate steps into a shorter form and reason starting from it, continuing the compressing and generation process in the decoding phase. It helps to reduce the number of tokens stored in the context window, thereby lowering memory overhead and computational costs. This includes compressing intermediate steps into a summary~\cite{yan2025inftythinkbreakinglengthlimits}, a subquestion~\cite{yang2025markovchainthoughtefficient}, or a special token~\cite{pang2024anchorbasedlargelanguagemodels,zhang2025lightthinkerthinkingstepbystepcompression} through specific training and corresponding inference strategies.

\paragraph{Reasoning in the latent space}
Switching reasoning from the language space to other spaces like latent space may overcome the restrictions of language and improve token efficiency. This can be achieved by finetuning existing models to possess this capability~\cite{deng2023implicitchainthoughtreasoning,hao2024traininglargelanguagemodels,shen2025codicompressingchainofthoughtcontinuous} or developing new language model architectures capable of implicitly reasoning in latent space~\cite{geiping2025scalingtesttimecomputelatent}.

\begin{table}[tb!]
\centering
\footnotesize
\caption{Comparisons of different test time scaling methods. \colorbox{gray!30}{Gray color} represents the model is optional or can share the same parameters with others. The description of these features is for the standard version.}
\label{tab:Comparisons_test_time_scaling_method}
\begin{tabular}{lccccc}
\toprule
\textbf{Method} & \textbf{Required Model} & \textbf{Controllability} & \textbf{Adaptivity} & \textbf{Training-free}  & \textbf{Compatibility} \\
\midrule
Parallel Sampling & 
\makecell[c]{Generator \\ \colorbox{gray!30}{Scoring function}} & Coarse-grained & Not supported &  \blackcheck & Full \\
\midrule
Tree Search &  \makecell[c]{Generator \\ \colorbox{gray!30}{Value function}} & Coarse-grained & Partial supported  & \blackcheck & Full \\
\midrule
Multi-turn Correction & \makecell[c]{Initial generator\\ \colorbox{gray!30}{Feedback model} \\ \colorbox{gray!30}{Refinement model}} & Coarse-grained & Partial supported  & \blackcheck & Full \\
\midrule
Long CoT  & \makecell[c]{Long-CoT model} & Not supported & Supported & \blackcross & Full \\
\bottomrule
\end{tabular}
\end{table}

\subsection{Comparisons of Test-Time Scaling Methods} \label{sec:comparisons_tts}
For different test-time scaling methods, we summarize their characteristics in Table~\ref{tab:Comparisons_test_time_scaling_method}. Specifically, we focus on the following aspects:

\paragraph{Performance} \emph{What is the optimal test-time scaling method given the same computation budget?} Establishing an absolute ranking of test-time scaling methods is challenging due to the various implementations within each approach and the difficulty in ensuring fair comparisons. For performance ceiling, long CoT methods consistently outperform other test-time scaling approaches that are  based on traditional LLMs, particularly for olympic-level problems~\cite{openai_o1_system_card,deepseekai2025deepseekr1incentivizingreasoningcapability}. Moreover, different test-time scaling methods exhibit distinct advantages for problems of varying difficulty and under different computational constraints. For instance,~\citet{snell2024scalingllmtesttimecompute} empirically demonstrate that beam-search excels on complex questions when operating under limited computation budgets, whereas BoN sampling achieves superior performance on simpler questions when greater computational resources are available. These complementary strengths create opportunities for ensemble methods, which will be discussed in subsequent sections.

\paragraph{Cognitive behaviors} \emph{Which test-time scaling method exhibits the most human-like cognitive behaviors?} Long CoT exhibits the most cognitive behaviors compared to others, including reflection, backtracking, divergent thinking, etc. More importantly, it unifies these cognitive behaviors in the generation process, enabling greater flexibility. Methods like tree search and multi-turn correction rely on external tree search algorithms or predefined multi-turn correction frameworks to endow the model with planning or reflection cognitive capability, limiting their adaptation to specific problems.

\paragraph{Adaptivity} \emph{Can the test-time scaling method allocate different computational resources to different queries?} The degree of adaptivity of a test-time scaling method depends on its stopping condition. In parallel sampling approaches, the standard implementation assigns identical sampling numbers across all queries, resulting in a lack of adaptivity. For tree search and multi-turn correction approaches, different cases exist. One variant of methods stops once reaching predefined hyperparameters (e.g., correction numbers, tree depth) or the answers~\cite{yao2023treethoughtsdeliberateproblem,kangMindStarEnhancingMath2024,snell2024scalingllmtesttimecompute}, thus providing no additional adaptivity from the framework. Another line of methods incorporates verifiers in the stopping condition, such as requiring the quality score of outputs to exceed a given threshold~\cite{wang2024litesearchefficacioustreesearch,welleckGeneratingSequencesLearning2022}, which introduces adaptivity based on the reliability of these verifiers. For example, LiteSearch~\cite{wang2024litesearchefficacioustreesearch} observes that tree search algorithms allocate larger computational resources for harder problems where stopping conditions include verifier values. For long CoT methods, the stopping condition is implicit and inherent in the generation process. Recent studies observe that the long CoT model generates longer responses to more challenging problems~\cite{zeng2025revisitingtesttimescalingo1like}. From the perspective of generalization, long CoT is the most promising approach for differentially allocating computational resources.

\paragraph{Controllability}\emph{Given a computation budget, can it operate within the specified constraints?} For test-time scaling methods with externally controllable  scaling dimensions (e.g., sampling numbers, tree depth, revision steps), coarse-grained controllability can be achieved by mapping the computation budget to specific quantities of these hyperparameters according to empirical estimation~\cite{welleck2024decodingmetagenerationinferencetimealgorithms}. For long CoT, although directly truncating the response to a specific number ensures not exceeding the computation budget, the resulting incomplete response significantly harms performance, making it impractical to consider standard long CoT as a method with controllability. To address this limitation, S1~\cite{muennighoff2025s1simpletesttimescaling} achieves control of response length through the implementation of end-of-thinking token delimiters, while L1~\cite{aggarwal2025l1controllinglongreasoning} develops a reinforcement learning algorithm to achieve precise control over token number with higher token efficiency compared to S1.

\paragraph{Simplicity} \emph{Are the components of the test-time scaling method straightforward to implement?} Methods excluding long CoT usually require additional roles such as evaluators to guide the search process and multiple processes to derive the final solutions. Considering the extra cost to deploy high-quality evaluators for most tasks, this may hinder their practical application. In contrast, the long CoT method eliminates the need for multiple components and is straightforward to implement.

\paragraph{Training-free} \emph{Does the test-time scaling method require additional training?} Methods excluding long CoT can be operated with the traditional LLM directly, while the long CoT ability needs to be elicited with additional training. It is notable that additional training can help improve scaling efficiency across methods, such as enhancing the self-correction ability of models~\cite{kumar2024traininglanguagemodelsselfcorrect,qu2024recursiveintrospectionteachinglanguage} or applying inference-aware fine-tuning to improve computation utilization~\cite{chowInferenceAwareFineTuningBestofN2024,yu2025thinksmarterharderadaptive}.

\paragraph{Compatibility} \emph{Can this method be integrated with other test-time scaling methods?} As will be discussed in \S\ref{sec:ensemble_tts}, all methods can be compatible with each other. Among them, parallel sampling is most easily compatible with others considering the ease of implementing multiple sampling.

Overall, the long CoT test-time scaling method outperforms others with its simplicity, adaptivity, higher ceiling performance, and more complex cognitive behaviors, but it requires additional training to elicit. Moreover, the compatibility and advantages of these test-time scaling methods make it beneficial to comprehensively utilize them together to achieve better performance instead of focusing on a single method.

\begin{figure}[tb!]
\centering
\begin{tikzpicture} 
\def\radius{3.8cm}
\def\nodesize{2cm} 
\node[draw, circle, fill=longCoT-fill,minimum size=\nodesize,
      text width=1.5cm, align=center, font=\footnotesize] (LC) at (0,0) {long CoT};
      
\node[draw, circle, fill=parallel-fill,minimum size=\nodesize,
      text width=1.5cm, align=center, font=\footnotesize] (N1) at ({\radius*cos(90)},{\radius*sin(90)}) {parallel sampling};
      
\node[draw, circle, fill=multiTurn-fill,minimum size=\nodesize,
      text width=1.5cm, align=center, font=\footnotesize] (N2) at ({\radius*cos(210)},{\radius*sin(210)}) {multi-turn correction};
      
\node[draw, circle, fill=symbolic-fill,minimum size=\nodesize, text width=1.5cm, align=center, font=\footnotesize] (N3) at ({\radius*cos(-30)},{\radius*sin(-30)}) {tree search};
      
\draw[{Latex[length=2.5mm]}-{Latex[length=2.5mm]}, thick] (LC) -- (N1) node[midway, sloped, above, font=\tiny,align=center] {~\citenumber{zeng2025revisitingtesttimescalingo1like,cuadron2025dangeroverthinkingexaminingreasoningaction,setlur2025scalingtesttimecomputeverification}};
\draw[{Latex[length=2.5mm]}-{Latex[length=2.5mm]}, thick] (LC) -- (N2) node[midway, sloped, above, font=\tiny,align=center] {RealCritic~\citenumber{tang2025realcriticeffectivenessdrivenevaluationlanguage}};
\draw[{Latex[length=2.5mm]}-{Latex[length=2.5mm]}, thick] (LC) -- (N3) node[midway, sloped, above, font=\tiny,align=center] {SWE-Reasoner~\citenumber{ma2025thinkinglongerlargerenhancing}};
\draw[{Latex[length=3mm]}-{Latex[length=3mm]}, thick] (N1.240) -- (N2.60) node[midway, sloped, above, font=\tiny, text width=4cm,align=center] {Snell et al.~\citenumber{snell2024scalingllmtesttimecompute}; Kumar et al.~\citenumber{kumar2024traininglanguagemodelsselfcorrect};\\SETS~\citenumber{chen2025setsleveragingselfverificationselfcorrection}; Olausson et al.~\citenumber{olausson2024selfrepairsilverbulletcode};PARM~\citenumber{guo2025can}};
\draw[{Latex[length=3mm]}-{Latex[length=3mm]}, thick] (N2.0) -- (N3.180) node[midway, sloped, below, font=\tiny,text width=4cm,align=center] {MCTSr~\citenumber{zhangAccessingGPT4Level2024a}; LLaMA-Berry~\citenumber{zhang2024llamaberrypairwiseoptimizationo1like};\\ MC-NEST~\citenumber{rabby2024mcnestenhancingmathematical}; SPAR~\citenumber{cheng2025sparselfplaytreesearchrefinement};\\RethinkMCTS~\citenumber{li2024rethinkmctsrefiningerroneousthoughts}};
\draw[{Latex[length=3mm]}-{Latex[length=3mm]}, thick] (N3.120) -- (N1.300) node[midway, sloped, above, font=\tiny,text width=4.5cm, align=center] {Snell et al.~\citenumber{snell2024scalingllmtesttimecompute}; DVTS~\citenumber{beeching2024scalingtesttimecompute}; Liu et al.~\citenumber{liu20251bllmsurpass405b};\\TreeBON~\citenumber{qiu2024treebonenhancinginferencetimealignment}; FoT~\citenumber{bi2025forestofthoughtscalingtesttimecompute}};
\end{tikzpicture}
\caption{Ensemble of Test-Time Scaling Methods~(\S\ref{sec:ensemble_tts}). Solid lines represent combination work between two connected methods.}
\label{fig:test_time_scaling_ensemble}
\end{figure}
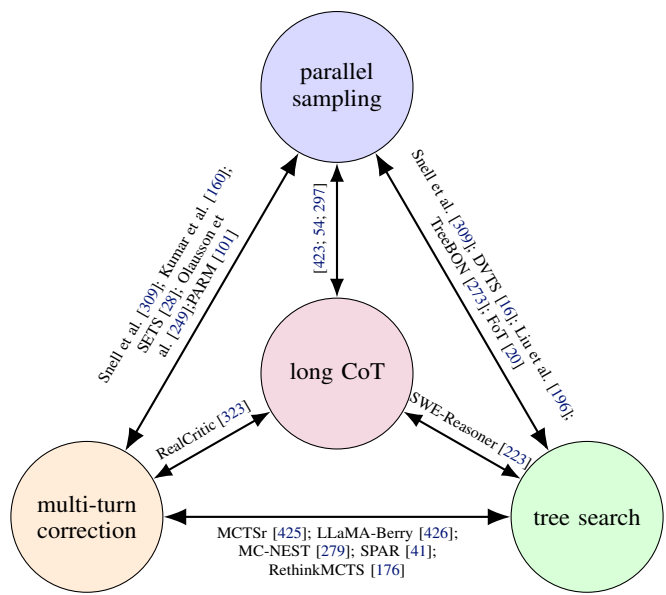

\subsection{Ensemble of Test-Time Scaling Methods} \label{sec:ensemble_tts}
Ensemble methods aim to comprehensively utilize multiple test-time scaling approaches rather than allocating computational resources to a single method, potentially achieving superior performance compared to individual approaches. These include simultaneously combining multiple methods or selecting appropriate test-time scaling methods according to different contexts. Figure~\ref{fig:test_time_scaling_ensemble} presents an organization of works on ensemble methods.

\paragraph{Combining parallel sampling with other methods} 
The simplicity of parallel sampling facilitates compatibility with other test-time scaling methods:

\begin{itemize}
    \item \textbf{Tree search.} Instead of searching along a single tree, parallel sampling can enhance tree search diversity by expanding the initial set of beams into multiple independent subtrees that are searched independently~\cite{beeching2024scalingtesttimecompute,bi2025forestofthoughtscalingtesttimecompute}. Empirical results demonstrate improved tree search performance, especially at large computation budgets. Moreover, tree search algorithms can also help accelerate parallel sampling. For example, TreeBON~\cite{qiu2024treebonenhancinginferencetimealignment} reduces the computational overhead of BoN by using tree search to prune low-quality responses at an early stage.

    \item \textbf{Multi-turn correction.} Parallel sampling functions as a global search by generating responses independently in parallel, whereas multi-turn correction operates as a local search on the initial response~\cite{snell2024scalingllmtesttimecompute}. This complementarity indicates that combining the two methods can yield better performance.~\citet{kumar2024traininglanguagemodelsselfcorrect,chen2025setsleveragingselfverificationselfcorrection} show that allocating a portion of the computation budget to self-correction of the initial response rather than solely increasing the sampling number can achieve higher token efficiency.~\citet{olausson2024selfrepairsilverbulletcode} demonstrate that in the mixed scheme, allocating more of the sampling budget to generating a diverse set of initial candidates is more optimal than carrying out extensive correction.
    
    \item \textbf{Long CoT.} Combining long CoT with parallel sampling methods such as majority voting is straightforward. Recent research has optimized majority voting strategies by considering the overthinking phenomenon in long CoT~\cite{zeng2025revisitingtesttimescalingo1like,cuadron2025dangeroverthinkingexaminingreasoningaction}. Specifically, high overthinking correlates with decreased performance in math tasks or agentic environments. Therefore, integrating metrics that measure the degree of overthinking with voting strategies can outperform both majority voting and single high-computation-cost response generation~\cite{zeng2025revisitingtesttimescalingo1like, cuadron2025dangeroverthinkingexaminingreasoningaction}. Additionally,~\citet{setlur2025scalingtesttimecomputeverification} compare the performance of scaling long-CoT length by budget-forcing~\cite{muennighoff2025s1simpletesttimescaling} against applying the computation for parallel sampling with shorter responses, finding the latter to be more compute-optimal.
\end{itemize}

\paragraph{Combining tree search with multi-turn correction}
This line of work incorporates critique and revision into the tree search algorithm by treating the revision behavior as the action to update the response, at either the solution level~\cite{zhangAccessingGPT4Level2024a,zhang2024llamaberrypairwiseoptimizationo1like,rabby2024mcnestenhancingmathematical,cheng2025sparselfplaytreesearchrefinement} or the step level~\cite{li2024rethinkmctsrefiningerroneousthoughts}. This approach enriches the expansion behavior of tree search and helps achieve better performance than simply revising responses sequentially~\cite{li2024rethinkmctsrefiningerroneousthoughts,cheng2025sparselfplaytreesearchrefinement}.

\paragraph{Combining long CoT with tree search or multi-turn correction}
The content of long CoT implicitly contains branch search processes or self-correction~\cite{xiang20252reasoningllmslearning}. Thus, it can be viewed as a method that internalizes these two approaches. For the combination with multi-turn correction,~\citet{tang2025realcriticeffectivenessdrivenevaluationlanguage} show that o1-mini benefits from self-correction while traditional LLMs perform worse, demonstrating that long CoT models possess strong intrinsic self-correction ability. For tree search, future research should analyze how to define the search space within the long thinking processes.

\paragraph{Adaptive selection of test-time scaling methods} 
Empirical analysis of different test-time scaling methods' performance relative to various factors can help derive optimal test-time scaling methods based on adaptive selection.~\citet{snell2024scalingllmtesttimecompute} find that multi-turn correction methods are better suited for simpler queries, while a certain ratio of parallel sampling and multi-turn correction is appropriate for difficult queries. Moreover, they determine that beam search is more effective for harder questions whereas best-of-N is more effective for easier questions. These findings guide optimal test-time scaling strategies based on query difficulty classifiers.~\citet{liu20251bllmsurpass405b} analyze the relationship between model size and test-time scaling methods to derive an optimal scaling strategy.

\section{How – Part II: Training Strategies for Test-Time Scaling} \label{sec:training_strategies_tts}

As discussed in \S\ref{sec:comparisons_tts}, long CoT demonstrates higher ceiling performance and more complex cognitive behaviors compared to other test-time scaling strategies, though it requires additional training. In this section, we examine methods to elicit the model's long CoT capabilities through two primary approaches: reinforcement learning (\S\ref{sec:scaling_rl}) and supervised fine-tuning (\S\ref{sec:sft}). Additionally, we discuss how to effectively combine test-time scaling techniques with iterative training methodologies to achieve self-improvement (\S\ref{sec:iterative_self-reinforced_learning}).

\subsection{Scaling Reinforcement Learning} \label{sec:scaling_rl}

\begin{table}[tb!]
    \centering
    \scriptsize
    \caption{Summary of recent works on RL scaling. For training algorithms, `PMD' denotes policy mirror descent method. $\text{REINFORCE}^{*}$ denotes REINFORCE-style method. For reward types, \ruleicon~and \modelicon~represent rule-based and model-based rewards respectively, while \outcomeicon~and \processicon~represent outcome and process rewards respectively. `\#D' indicates the query dataset size. `MS' denotes the multi-stage training strategy, including long CoT cold start (LCS), iterative lengthening strategy (IL), and curriculum sampling strategy (CSS). In accuracy (Acc.) and length (Len.) figures, for works presenting multiple figures, we show the common pattern. ``Cog." indicates whether the response contains words indicating cognitive behaviors like ``wait."}
    \setlength{\tabcolsep}{2.2pt}
    \label{tab:summary_scaling_rl_work}  
\vspace{-2.5mm}
\resizebox{0.95\textwidth}{!}{%
    \begin{tabular}{l|cccccc|ccc}
    \toprule
    \textbf{Work} & \makecell[c]{\textbf{Algorithm}\\(\S\ref{sec:scaling_rl:training_algorithm})}  & \makecell[c]{\textbf{Reward}\\(\S\ref{sec:scaling_rl:reward_function})} & \makecell[c]{\textbf{Series}\\(\S\ref{sec:scaling_rl:policy_model_selection})} & \makecell[c]{\textbf{Size}\\(\S\ref{sec:scaling_rl:policy_model_selection})} & \makecell[c]{\textbf{\#D}\\(\S\ref{sec:scaling_rl:training_data_construction})}  & \makecell[c]{\textbf{MS}\\(\S\ref{sec:scaling_rl:multi_stage_training})} & \textbf{Acc.} & \textbf{Len.} & \textbf{Cog.}  \\
    \midrule
    \makecell[l]{Eurus-2-7B-PRIME\\\cite{cui2024process}} & $\text{REINFORCE}^{*}$ & \makecell[c]{\ruleicon \hspace{0.3em} \modelicon\\ \outcomeicon \hspace{0.32em}\processicon} & {\includegraphics[width=0.02\linewidth]{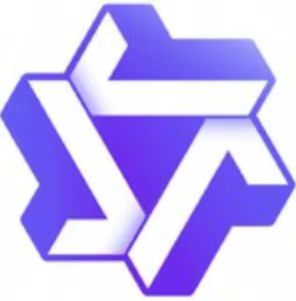}} & 7B & 150K &  \redcross & \raisebox{-0.5\height}{\includegraphics[width=0.09\linewidth]{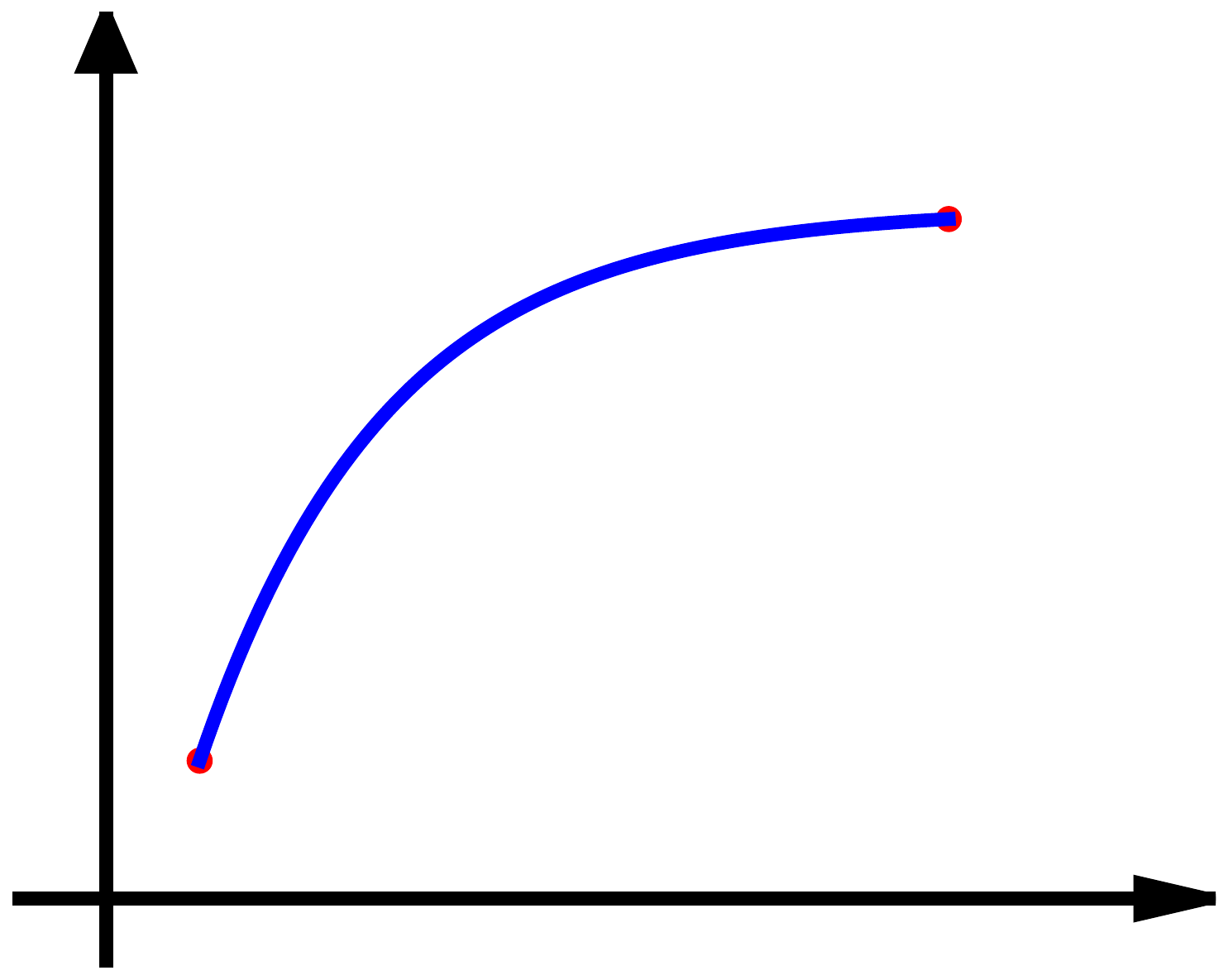}} & - & -   \\
    \midrule
    \makecell[l]{Deepseek-R1-Zero\\\cite{deepseekai2025deepseekr1incentivizingreasoningcapability}} & GRPO & \makecell[c]{\ruleicon\\\outcomeicon} & \raisebox{-0.5\height}{\includegraphics[width=0.02\linewidth]{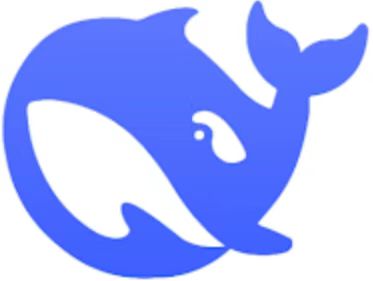}} & 671B & -  & \redcross & \raisebox{-0.5\height}{\includegraphics[width=0.09\linewidth]{table/summary_scaling_rl_work/exponential_saturation_plot.pdf}} & \raisebox{-0.5\height}{\includegraphics[width=0.09\linewidth]{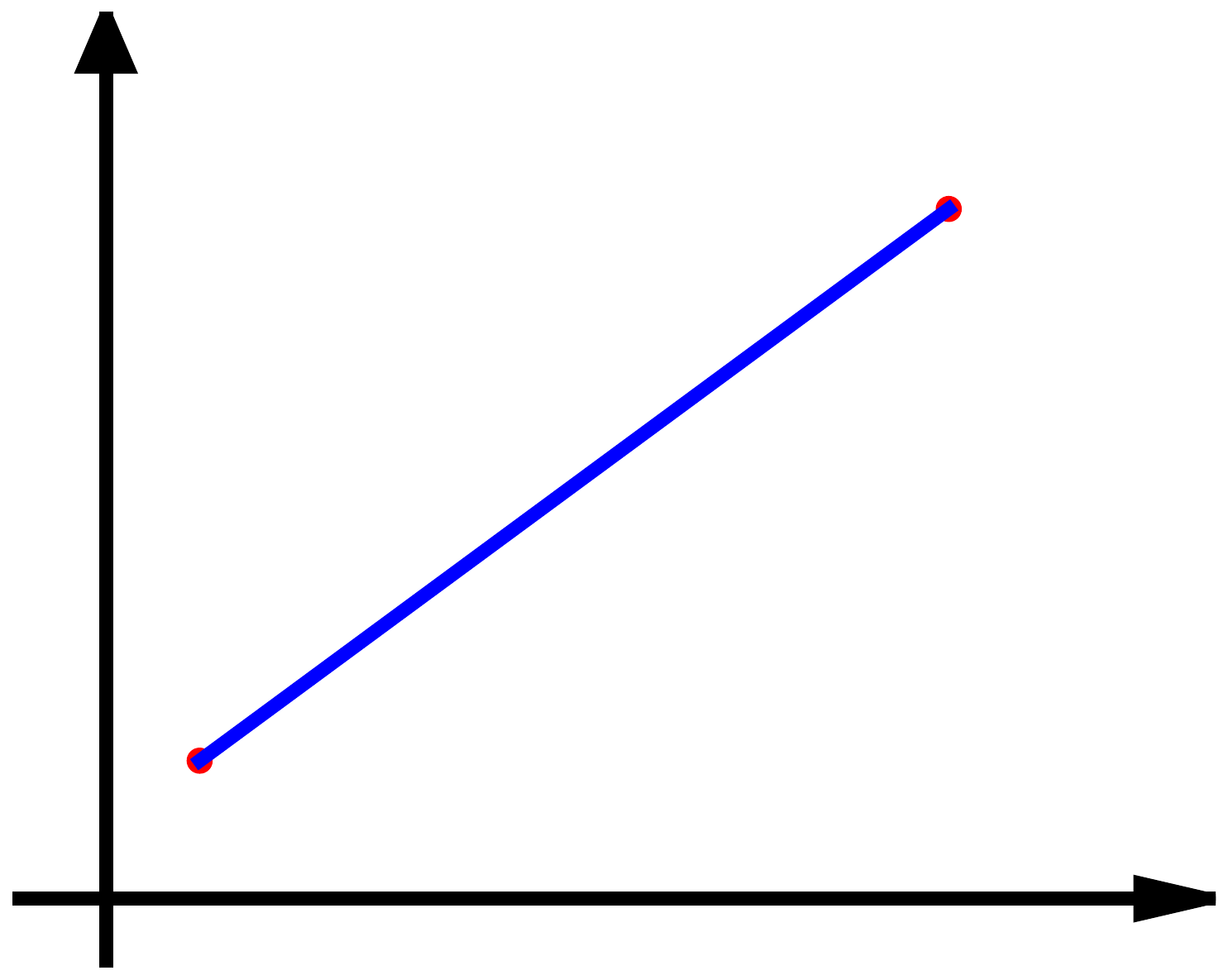}} & \greencheck \\
    \midrule
    \makecell[l]{Kimi k1.5\\\cite{MoonshotAI}}& PMD & \makecell[c]{\ruleicon\\\outcomeicon} &\raisebox{-0.5\height}{\includegraphics[width=0.02\linewidth]{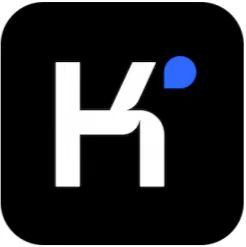}}  & -     & -  & \makecell[c]{LCS\\CSS}  & \raisebox{-0.5\height}{\includegraphics[width=0.09\linewidth]{table/summary_scaling_rl_work/exponential_saturation_plot.pdf}} & \raisebox{-0.5\height}{\includegraphics[width=0.09\linewidth]{table/summary_scaling_rl_work/linear_growth_plot.pdf}}  & - \\
    \midrule
    \makecell[l]{SimpleRL-Zero\\\cite{zeng2025simplerl}} & PPO & \makecell[c]{\ruleicon\\\outcomeicon} & {\includegraphics[width=0.02\linewidth]{table/summary_scaling_rl_work/qwen.png}} & 7B & 8K  & \redcross & \raisebox{-0.5\height}{\includegraphics[width=0.09\linewidth]{table/summary_scaling_rl_work/exponential_saturation_plot.pdf}} & \raisebox{-0.5\height}{\includegraphics[width=0.09\linewidth]{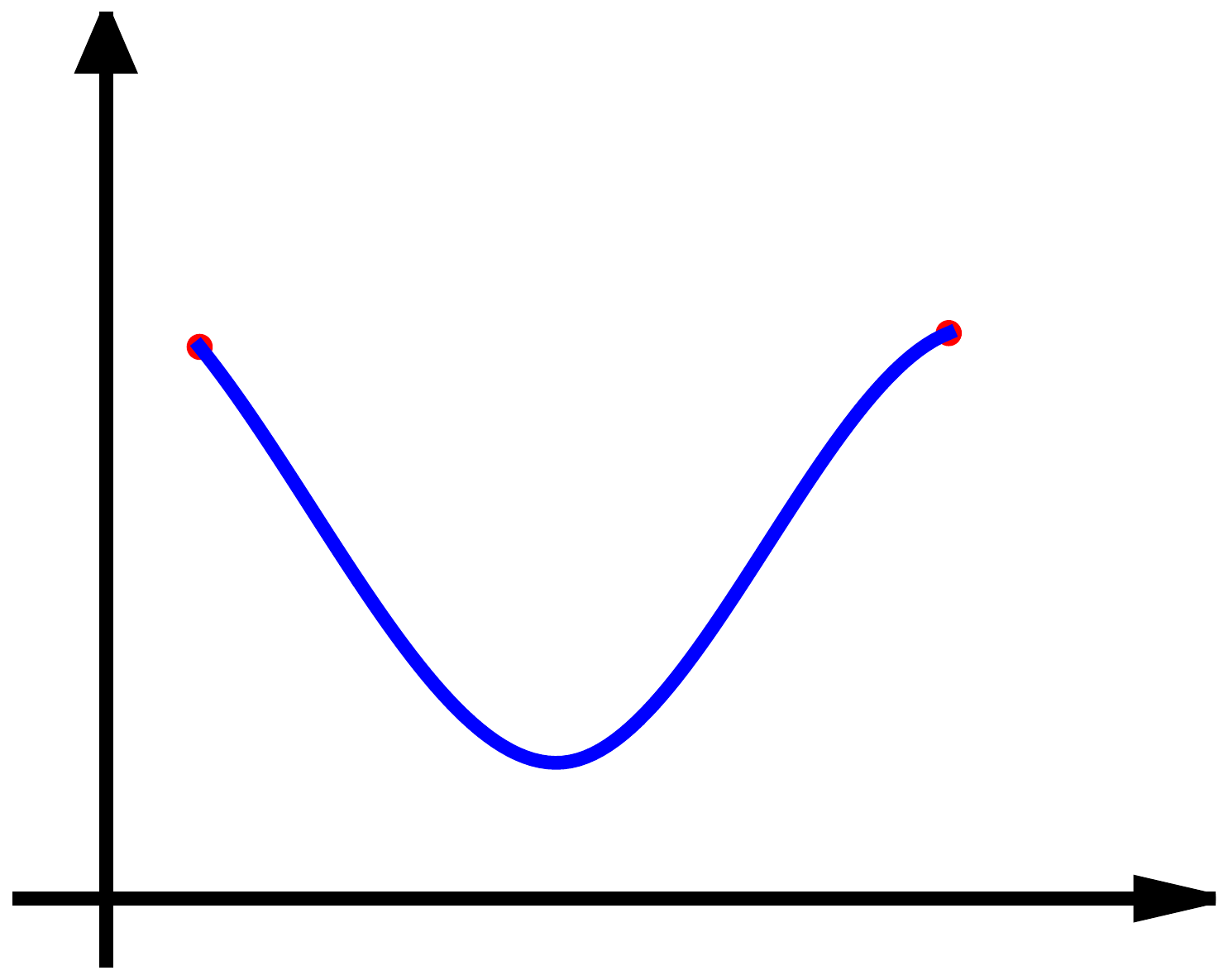}}
    & \greencheck \\
    \midrule
    \makecell[l]{SimpleRL\\\cite{zeng2025simplerl}} & PPO & \makecell[c]{\ruleicon\\\outcomeicon} & \raisebox{-0.5\height}{\includegraphics[width=0.02\linewidth]{table/summary_scaling_rl_work/qwen.png}} & 7B & 8K & LCS & \raisebox{-0.5\height}{\includegraphics[width=0.09\linewidth]{table/summary_scaling_rl_work/exponential_saturation_plot.pdf}} & \raisebox{-0.5\height}{\includegraphics[width=0.09\linewidth]{table/summary_scaling_rl_work/u_shaped_curve_visualization.pdf}}
    & - \\

    \midrule
\makecell[l]{STILL-3-ZERO-32B\\\cite{chen2025empiricalstudyelicitingimproving}} & GRPO & \makecell[c]{\ruleicon\\\outcomeicon} &  {\includegraphics[width=0.02\linewidth]{table/summary_scaling_rl_work/qwen.png}} & 32B & 90K & IL & \raisebox{-0.5\height}{\includegraphics[width=0.09\linewidth]{table/summary_scaling_rl_work/linear_growth_plot.pdf}} & \raisebox{-0.5\height}{\includegraphics[width=0.09\linewidth]{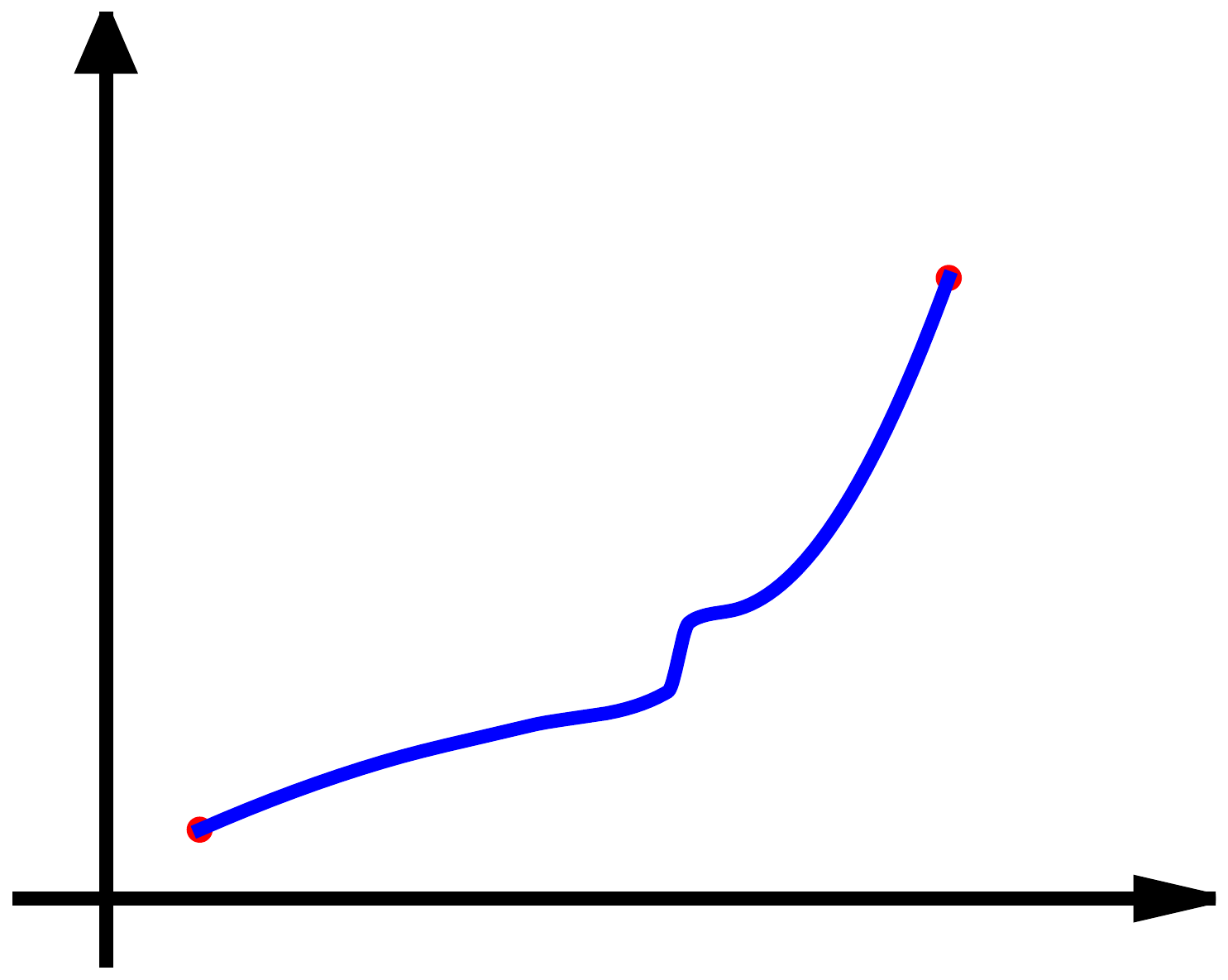}} & \greencheck \\
\midrule
    \makecell[l]{Sea AI Lab\\\cite{liu2025oatzero}} & PPO & \makecell[c]{\ruleicon\\\outcomeicon} & {\includegraphics[width=0.02\linewidth]{table/summary_scaling_rl_work/qwen.png}} & 1.5B & 8K & \redcross & \raisebox{-0.5\height}{\includegraphics[width=0.09\linewidth]{table/summary_scaling_rl_work/exponential_saturation_plot.pdf}} & 
    \raisebox{-0.5\height}{\includegraphics[width=0.09\linewidth]{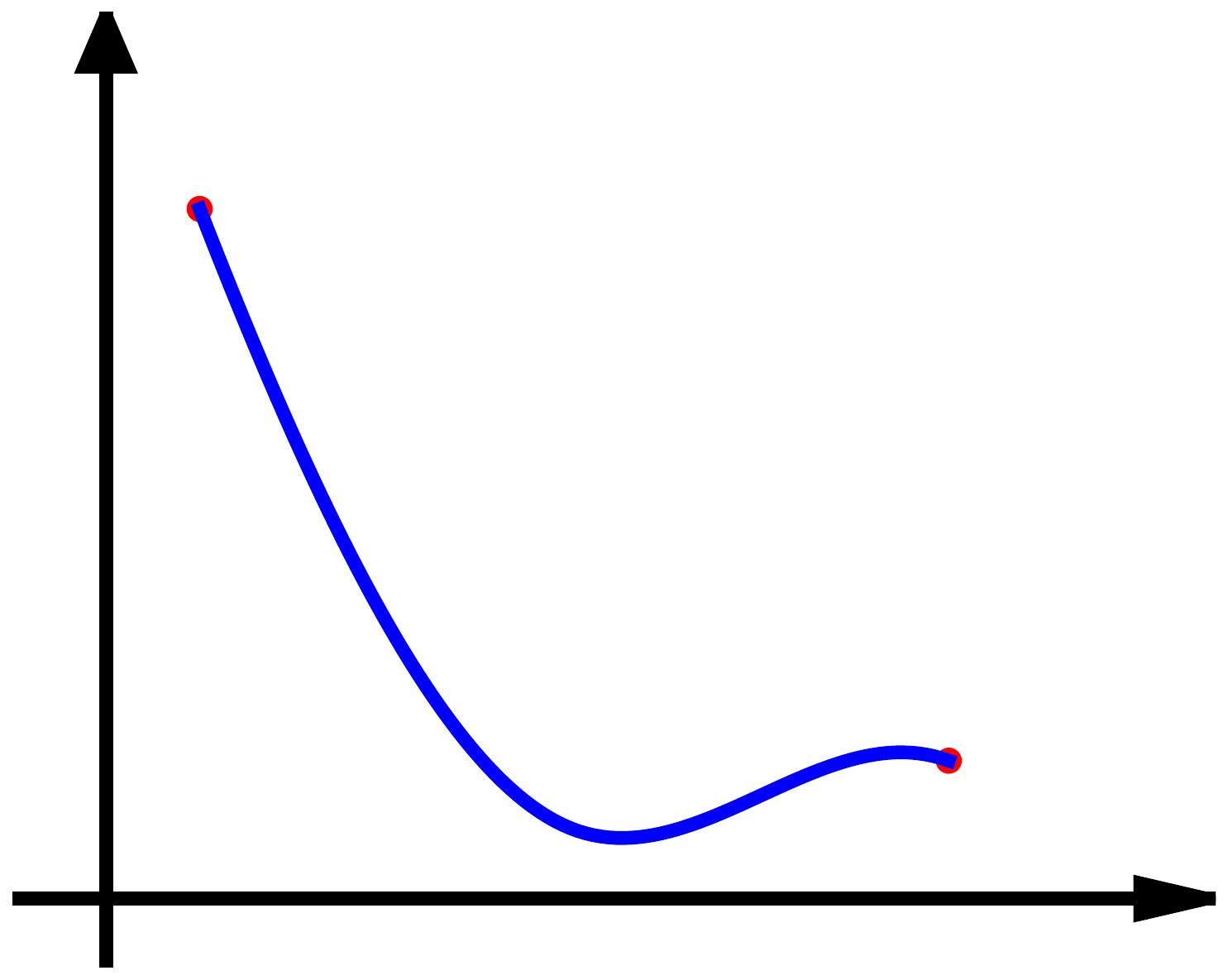}} & \greencheck \\
    \midrule
    \makecell[l]{DeepScaleR-1.5B-Preview\\\cite{deepscaler2025}} & GRPO & \makecell[c]{\ruleicon\\\outcomeicon} & {\includegraphics[width=0.02\linewidth]{table/summary_scaling_rl_work/qwen.png}} & 1.5B & 40K &  IL & \raisebox{-0.5\height}{\includegraphics[width=0.09\linewidth]{table/summary_scaling_rl_work/linear_growth_plot.pdf}} & \raisebox{-0.5\height}{\includegraphics[width=0.09\linewidth]{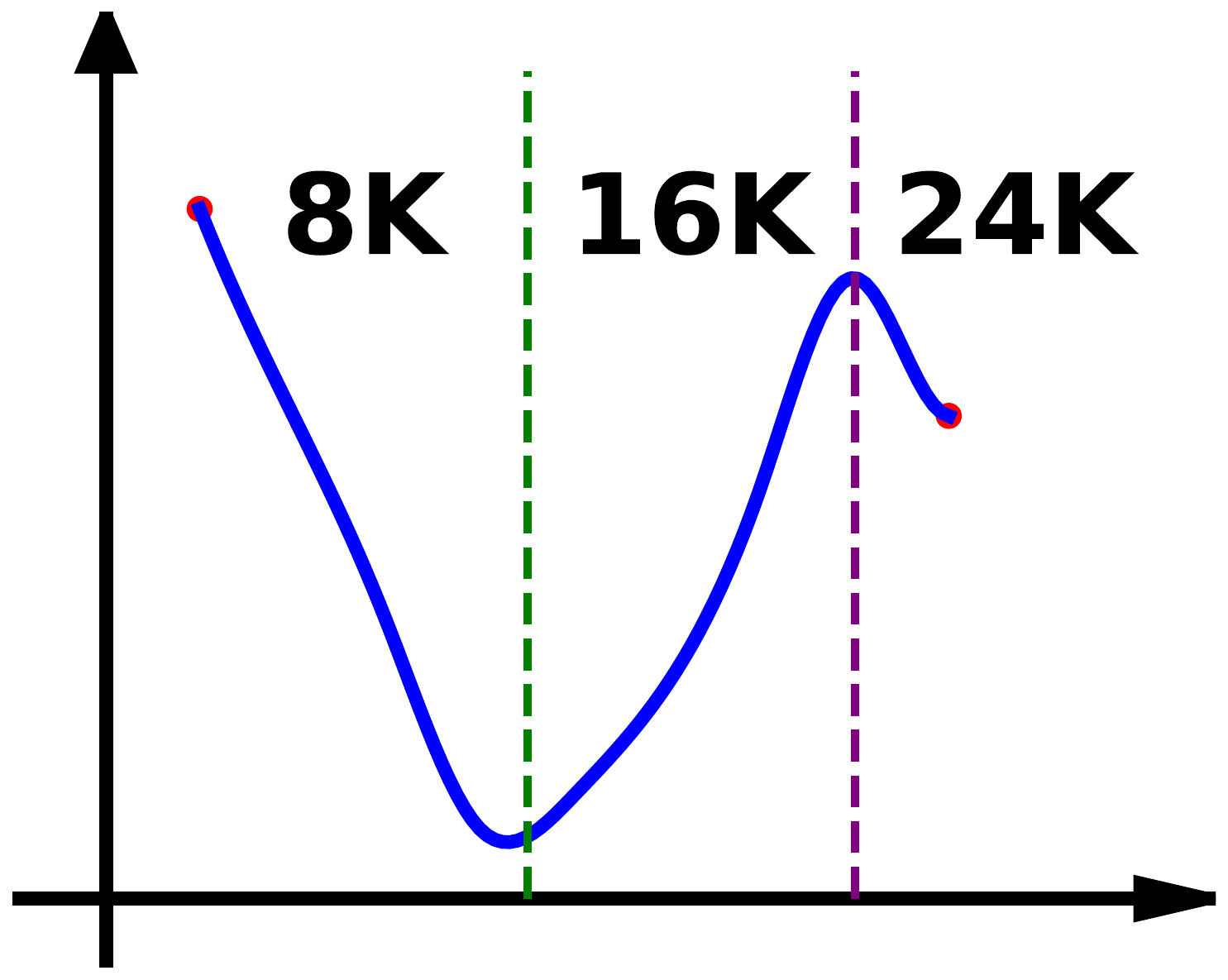}} & - \\
    \midrule
    \makecell[l]{T1\\\cite{hou2025advancinglanguagemodelreasoning}} & $\text{REINFORCE}^{*}$ & \makecell[c]{\ruleicon\\\outcomeicon} & {\includegraphics[width=0.02\linewidth]{table/summary_scaling_rl_work/qwen.png}} & 14B & 30K & LCS & \raisebox{-0.5\height}{\includegraphics[width=0.09\linewidth]{table/summary_scaling_rl_work/exponential_saturation_plot.pdf}} & \raisebox{-0.5\height}{\includegraphics[width=0.09\linewidth]{table/summary_scaling_rl_work/linear_growth_plot.pdf}} & \greencheck \\
    \midrule
    \makecell[l]{DAPO\\\cite{yu2025dapoopensourcellmreinforcement}} & GRPO & \makecell[c]{\ruleicon\\\outcomeicon} & {\includegraphics[width=0.02\linewidth]{table/summary_scaling_rl_work/qwen.png}} &32B &  17K & \redcross & \raisebox{-0.5\height}{\includegraphics[width=0.09\linewidth]{table/summary_scaling_rl_work/exponential_saturation_plot.pdf}} & \raisebox{-0.5\height}{\includegraphics[width=0.09\linewidth]{table/summary_scaling_rl_work/linear_growth_plot.pdf}} & \greencheck \\

    \midrule
    \makecell[l]{LIMR\\\cite{li2025limrrlscaling}} & GRPO & \makecell[c]{\ruleicon\\\outcomeicon} & {\includegraphics[width=0.02\linewidth]{table/summary_scaling_rl_work/qwen.png}} & 7B & 1.4K & \redcross & \raisebox{-0.5\height}{\includegraphics[width=0.09\linewidth]{table/summary_scaling_rl_work/exponential_saturation_plot.pdf}} & \raisebox{-0.5\height}{\includegraphics[width=0.09\linewidth]{table/summary_scaling_rl_work/u_shaped_curve_lower_visualization.pdf}} & - \\
    \midrule
    \makecell[l]{Open-Reasoner-Zero\\\cite{OpenReasonerZero2025}} & PPO & \makecell[c]{\ruleicon\\\outcomeicon} & {\includegraphics[width=0.02\linewidth]{table/summary_scaling_rl_work/qwen.png}} & \makecell[c]{7B\\32B} & 57K & \redcross & \raisebox{-0.5\height}{\includegraphics[width=0.09\linewidth]{table/summary_scaling_rl_work/exponential_saturation_plot.pdf}} & \raisebox{-0.5\height}{\includegraphics[width=0.09\linewidth]{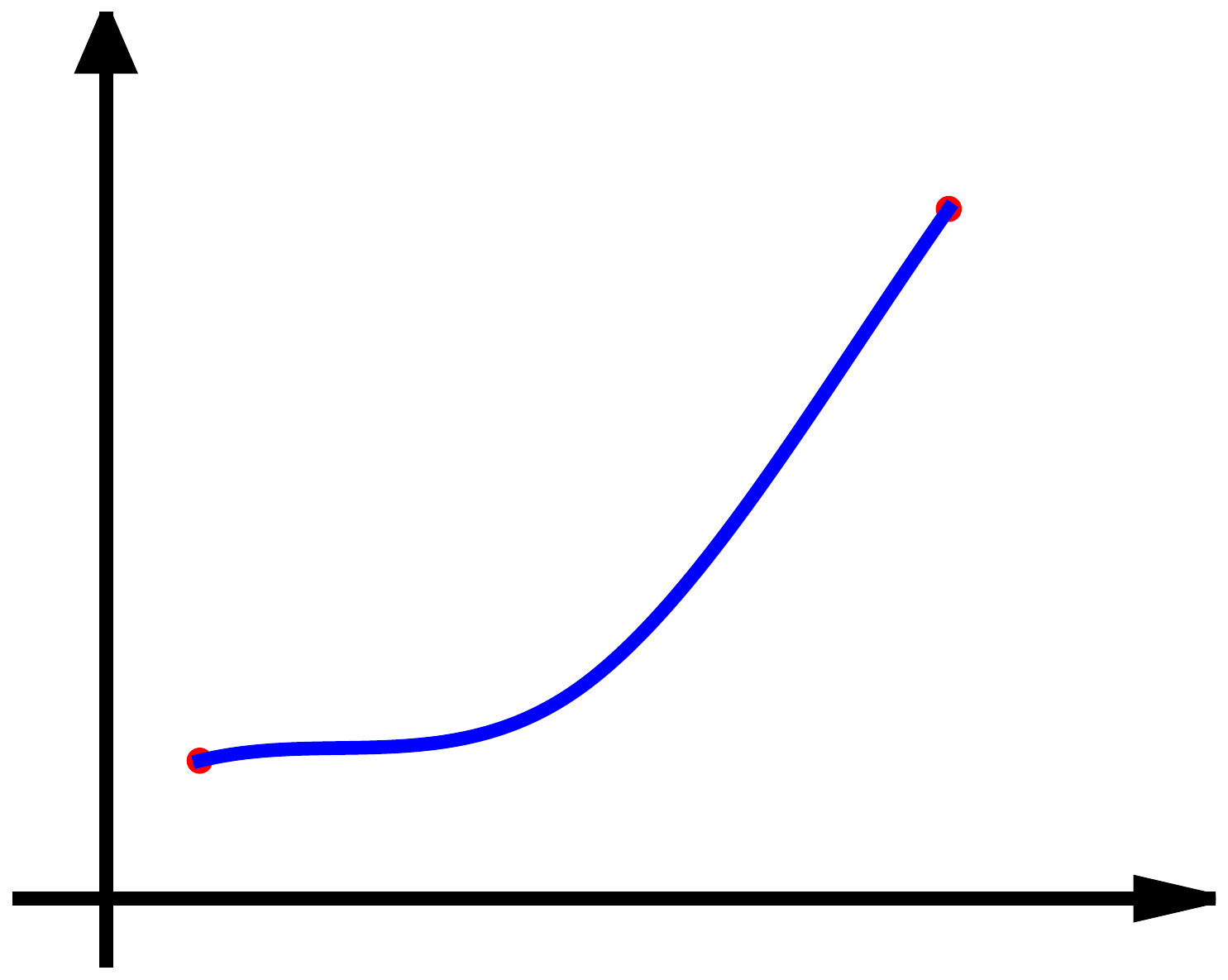}} & \greencheck \\
    \midrule
     \makecell[l]{Logic-RL\\\cite{xie2025logicrlunleashingllmreasoning}} & $\text{REINFORCE}^{*}$ & \makecell[c]{\ruleicon\\\outcomeicon} & {\includegraphics[width=0.02\linewidth]{table/summary_scaling_rl_work/qwen.png}} & 7B & 5K & \redcross & \raisebox{-0.5\height}{\includegraphics[width=0.09\linewidth]{table/summary_scaling_rl_work/linear_growth_plot.pdf}} & \raisebox{-0.5\height}{\includegraphics[width=0.09\linewidth]{table/summary_scaling_rl_work/flat_start_accelerating_growth_curve.pdf}} & \greencheck \\
    \bottomrule
    \end{tabular}
}
\end{table}

Recent research demonstrates that training LLMs through online reinforcement learning with rule-based rewards in tasks like mathematics and code can significantly enhance their reasoning abilities~\cite{deepseekai2025deepseekr1incentivizingreasoningcapability,MoonshotAI}. During the training process, models autonomously learn to master long-CoT test-time scaling methods to solve challenging problems and demonstrate cognitive behaviors including self-reflection and self-correction. This phenomenon has been described as the RL scaling phenomenon\footnote{In the paper, we use ``RL scaling'' to describe the line of work.} or the ``Aha moment.'' We systematically summarize recent works in Table~\ref{tab:summary_scaling_rl_work}. Additionally, Table~\ref{tab:rl_methods_comparison} presents recipes to address common challenges in RL scaling training based on recent studies. In the following sections, we detail the design considerations for each component.

\subsubsection{Training Algorithm} \label{sec:scaling_rl:training_algorithm}

\paragraph{REINFORCE} The REINFORCE~\cite{sutton1999policy} algorithm is a foundational policy gradient method in reinforcement learning that directly optimizes the expected return of a policy through gradient ascent. The algorithm optimizes the policy model $\pi_\theta$ by minimizing the loss:
\begin{align}
\mathcal{L}_{\rm REINFORCE}(\theta) &= -\mathbb{E}_{\tau \sim \pi_\theta}\left[\sum_{t=1}^T G_t \nabla_\theta \log \pi_\theta(a_t|s_t)\right]
\end{align}
where $G_t$ is the discounted cumulative reward from time step $t$. Despite its simplicity, REINFORCE suffers from high variance in gradient estimates.

\paragraph{Proximal Policy Optimization (PPO)} For the PPO algorithm~\cite{schulman2017proximalpolicyoptimizationalgorithms}, it optimizes the policy model by minimizing the loss:
\begin{align}
\mathcal{L}_{\rm PPO}(\theta) &= - \mathbb{E}_{q \sim P(Q), o \sim \pi_{\theta_{\rm old}}(O|q)}\frac{1}{|o|}\sum_{t=1}^{|o|}\min\left(\frac{\pi_\theta(o_t|q,o_{<t})}{\pi_{\theta_{\rm old}}(o_t|q,o_{<t})}A_t,{\rm \text{clip}(\theta)}A_t\right) \\
\text{clip}(\theta) &= \text{clip}\left(\frac{\pi_\theta(o_t|q,o_{<t})}{\pi_{\theta_{\rm old}}(o_t|q,o_{<t})},1-\varepsilon,1+\varepsilon\right)
\end{align}
where $\pi_\theta$ and $\pi_{\theta_{\rm old}}$ are the current and old policy models, and $q,o$ are the sampled questions and outputs. The $\text{clip}(\theta)$ function constrains policy updates to ensure stable training. $A_t$ is the advantage computed by applying GAE~\cite{schulman2018highdimensionalcontinuouscontrolusing} based on the rewards $\{r_{\geq t}\}$ and a learned value function $V_\psi$. The KL penalty can be added to the reward function:
\begin{equation}
r_t = r_\varphi(q, o_{\leq t}) - \beta \log \frac{\pi_\theta(o_t|q, o_{<t})}{\pi_{\rm ref}(o_t|q, o_{<t})}
\end{equation}
where $r_\varphi$ is the reward model, $\pi_{\rm ref}$ is the reference model (initial SFT model), and $\beta$ is the coefficient of the KL penalty.

\paragraph{Group Relative Policy Optimization (GRPO)} The GRPO algorithm~\cite{shao2024deepseekmathpushinglimitsmathematical} directly uses the average reward of multiple parallel sampled responses as the baseline, eliminating the need for additional value function approximation as in PPO. Specifically, for each question $q$, GRPO samples a group of outputs $\{o_1,o_2,\cdots,o_G\}$ from the old policy $\pi_{\theta_{\rm old}}$ and then optimizes the policy model $\pi_\theta$ by minimizing the loss:
\begin{align}
\mathcal{L}_{\rm GRPO}(\theta) &= - \mathbb{E}_{q \sim P(Q), \{o_i\}_{i=1}^G \sim \pi_{\theta_{old}}(O|q)} \nonumber \\
& \frac{1}{G}\sum_{i=1}^G\frac{1}{|o_i|}\sum_{t=1}^{|o_i|}\left[\min\left(\frac{\pi_\theta(o_{i,t}|q,o_{i,<t})}{\pi_{\theta_{\rm old}}(o_{i,t}|q,o_{i,<t})}\hat{A_{i,t}},{\rm \text{clip}(\theta)}\hat{A_{i,t}}\right)-\beta\mathbb{D}_{\rm KL}[\pi_\theta||\pi_{\rm ref}]\right] \\
\text{clip}(\theta) &= \text{clip}\left(\frac{\pi_\theta(o_{i,t}|q,o_{i,<t})}{\pi_{\theta_{\rm old}}(o_{i,t}|q,o_{i,<t})},1-\varepsilon,1+\varepsilon\right) \\
\mathbb{D}_{\rm KL}[\pi_\theta||\pi_{\rm ref}] &= \frac{\pi_{\rm ref}(o_{i,t}|q,o_{i,<t})}{\pi_\theta(o_{i,t}|q,o_{i,<t})} - \log\frac{\pi_{\rm ref}(o_{i,t}|q,o_{i,<t})}{\pi_\theta(o_{i,t}|q,o_{i,<t})} - 1
\end{align}

where $\varepsilon$ and $\beta$ are hyper-parameters, and $\hat{A_{i,t}}$ is the advantage computed using a group of rewards corresponding to the outputs within each group.

\paragraph{REINFORCE++}  REINFORCE++~\cite{hu2025reinforcesimpleefficientapproach} is a variant of the classical REINFORCE algorithm that integrates key optimization techniques from PPO while eliminating the need for a critic network. The algorithm incorporates several enhancements to address the limitations of REINFORCE as follows:
\begin{itemize}
    \item It implements a token-level KL divergence penalty to prevent the policy from deviating too far from the initial model.
    \item It adopts PPO's clipping mechanism to constrain policy updates and maintain stability during training.
    \item It introduces mini-batch updates for improved training efficiency and better convergence rates.
    \item It employs comprehensive reward normalization and clipping to stabilize training by mitigating outliers and constraining reward values within predefined bounds.
    \item It implements advantage normalization using z-score normalization to ensure stable gradients and prevent divergence during training.
\end{itemize}

\begin{table}[tb!]
\centering
\caption{Comparisons of different training algorithms. For the computational overhead, \fire~represents the model needs to be updated, \ice~indicates the model needs to perform inference. `RL Scaling' represents whether RL scaling phenomena have been successfully observed with this algorithm.}\
\label{tab:Comparisons_training_algorithms}
\begin{tabular}{lccccc}
\toprule
\multirow{2}{*}{\textbf{Algorithm}} & \multicolumn{4}{c}{\textbf{Computational Overhead}} & \multirow{2}{*}{\textbf{RL Scaling}} \\
\cmidrule{2-5}
 & \textbf{Policy}  & \textbf{Reward}  & \textbf{Critic} & \textbf{Reference}  & \\
\midrule
REINFORCE & \fire~\ice &\ice & \brickcross & \brickcross & \brickcross  \\
PPO & \fire~\ice  & \ice & \fire~\ice & \ice & \greencheck  \\
GRPO & \fire~\ice & \ice & \brickcross & \ice & \greencheck \\
REINFORCE++ &\fire~\ice & \ice & \brickcross & \ice & \greencheck  \\
\bottomrule
\end{tabular}
\end{table}

\paragraph{Comparisons with different algorithms} We summarize the characteristics of different training algorithms in Table~\ref{tab:Comparisons_training_algorithms}. Regarding computational cost, PPO shows predominant computational cost with four models to be loaded, among which the policy model and the critic model need to perform both inference and training. GRPO and REINFORCE++ eliminate the need for a critic model and achieve higher training stability than REINFORCE~\cite{hu2025reinforcesimpleefficientapproach}. Regarding performance, all algorithms except REINFORCE exhibit the RL scaling phenomenon. For specific performance comparisons,~\citet{hou2024doesrlhfscaleexploring} find that the performance of PPO and GRPO is similar in RLHF settings, while~\citet{xie2025logicrlunleashingllmreasoning} observe that the performance of PPO and REINFORCE++ is superior to GRPO in rule-based reward settings for synthetic logic puzzles. More rigorous and large-scale studies should be conducted to comprehensively evaluate the performance of these algorithms.

\definecolor{primarycolor}{RGB}{60, 60, 60} 
\definecolor{secondarycolor}{RGB}{235, 235, 235}
\definecolor{headingbg}{RGB}{210, 210, 210}
\definecolor{linecol}{RGB}{120, 120, 120}

\newlength{\headingwidth}
\setlength{\headingwidth}{\dimexpr\linewidth-2\tabcolsep\relax}

\newcolumntype{M}[1]{>{\raggedright\arraybackslash\hspace{0pt}}m{#1}}

\begin{table}[tb!]
\centering
\footnotesize
\fontsize{5.8pt}{4.8pt}\selectfont 
\setlength{\extrarowheight}{1pt}
\rowcolors{1}{white}{secondarycolor!70} 

\caption{Recipes to resolve common problems in RL scaling training based on recent studies.}
\label{tab:rl_methods_comparison}

\begin{tabular}{M{0.14\textwidth}M{0.36\textwidth}M{0.30\textwidth}M{0.10\textwidth}}
\toprule[1.5pt]
\rowcolor{white}
\textbf{{Problem to Solve}} & 
\textbf{{Method Overview}} & 
\textbf{{Evidence}} & 
\textbf{{Related Studies}} \\

\midrule
\rowcolor{white}
\multicolumn{4}{l}{\hspace*{-\tabcolsep}\colorbox{headingbg}{\parbox{\headingwidth}{\centering\textbf{\textcolor{primarycolor}{\textsf{TRAINING ALGORITHM}}}}}} \\
\midrule
\textbf{Computational inefficiency in traditional PPO for LLM training} & \textbf{GRPO (Group Relative Policy Optimization):} Eliminates the need for a separate value model by using the average reward of multiple outputs from the same prompt as the baseline for advantage calculation. & Performance comparisons demonstrate computational efficiency while maintaining comparable effectiveness to traditional PPO, particularly well-suited for LLM reward modeling where rewards are often comparative in nature. & GRPO~\citenumber{shao2024deepseekmathpushinglimitsmathematical} \\
\addlinespace
\textbf{Token inefficiency and overthinking in long-form reasoning} & \textbf{Dr.GRPO (Doctor GRPO):} Addresses optimization bias in GRPO by removing response-length normalization and reward standardization, implementing an unbiased policy gradient estimation. & Experimental results show significantly improved token efficiency with better controlled response lengths, effectively mitigating overthinking problems. & Dr.GRPO~\citenumber{liu2025understanding} \\
\addlinespace
\textbf{Instability with varying response lengths in long-form reasoning} & \textbf{DAPO (Decouple Clip and Dynamic Sampling Policy Optimization):} Implements token-level policy gradient calculation, allowing longer sequences to appropriately influence the gradient updates regardless of individual response lengths. & Comparative analysis reveals more stable training dynamics with healthier entropy management and better quality pattern recognition, particularly for handling varying response lengths effectively. & DAPO~\citenumber{yu2025dapoopensourcellmreinforcement} \\
\addlinespace
\textbf{Limited policy exploration due to rigid constraints} & \textbf{GPG (Group Policy Gradient):} Simplifies the policy gradient approach by removing  reference models and policy constraints while maintaining stability through group-level reward normalization. & Comparative experiments demonstrate enhanced exploration capabilities with reduced computational requirements, providing more flexible policy updates. & GPG~\citenumber{chu2025gpgsimplestrongreinforcement} \\

\addlinespace
\textbf{Repetitive or narrow reasoning patterns} & \textbf{Auxiliary entropy bonus:} Incorporates an additive entropy term into the RL loss function to encourage token diversity and prevent deterministic response patterns. & Experimental results show more varied and creative reasoning paths without sacrificing solution accuracy. & T1~\citenumber{hou2025advancinglanguagemodelreasoning} \\
\addlinespace
\textbf{Limitations of fixed reference models} & \textbf{On-policy KL normalization:} Combines KL normalization with Exponential Moving Average (EMA) updates to the reference model. & Dynamic reference model updating allows for more effective RL scaling while maintaining stable training dynamics. & T1~\citenumber{hou2025advancinglanguagemodelreasoning} \\
\addlinespace

\textbf{Value model misalignment with strong prior policies} & \textbf{Value-Pretraining Alignment:} Implements a dedicated pretraining phase for the value model to ensure alignment with strong prior policies before RL begins. & Two-stage convergence pattern shows initial range alignment followed by crucial knowledge injection, preventing collapse in output length for long-CoT tasks. & VC-PPO~\citenumber{yuan2025whatspposcollapselongcot}, VAPO
~\citenumber{yue2025vapoefficientreliablereinforcement} \\
\addlinespace

\textbf{Conflicting variance-bias requirements between value and policy optimization} & \textbf{Decoupled-GAE (Generalized Advantage Estimation):}  Separates the $\lambda$ parameter for value function and policy optimization, allowing unbiased value estimation while maintaining variance reduction benefits for policy updates. & Mathematical analysis and experimental results demonstrate improved convergence rates without introducing additional bias, particularly effective for trajectory-level rewards in long CoT tasks. & VC-PPO~\citenumber{yuan2025whatspposcollapselongcot}, VAPO
~\citenumber{yue2025vapoefficientreliablereinforcement} \\

\addlinespace
\textbf{Limited exploration in constrained policy optimization} & \textbf{KL Divergence Removal:} Eliminates the KL penalty term that constrains policy divergence from the reference model, allowing the reasoning policy to explore more freely. & Experiments reveal significant performance gains when removing constraints on policy distribution shifts during extended reasoning training. & Open-Reasoner-Zero~\citenumber{OpenReasonerZero2025}, DAPO~\citenumber{yu2025dapoopensourcellmreinforcement} \\
\addlinespace
\textbf{Premature deterministic behavior in RL systems} & \textbf{Clip-Higher Strategy:} Decouples lower and higher clipping ranges in PPO to specifically promote exploration of low-probability tokens while maintaining stability. & Asymmetric clipping thresholds effectively counteract entropy collapse and maintain policy diversity throughout extended training. & DAPO~\citenumber{yu2025dapoopensourcellmreinforcement} \\
\addlinespace
\textbf{Ineffective gradient signals in late-stage training} & \textbf{Dynamic Sampling:} Implements an adaptive sampling approach that filters out prompts with accuracy values of exactly 0 or 1 to ensure effective gradient signals. & Comparative training curves demonstrate faster convergence to target performance despite the additional computational overhead of oversampling. & \makecell[l]{DAPO~\citenumber{yu2025dapoopensourcellmreinforcement}, \\  
Bae et al.~\citenumber{bae2025onlinedifficultyfilteringreasoning}} \\
\addlinespace
\textbf{Noisy reward signals from length-truncated samples} & \textbf{Overlong Filtering:} Masks the loss contribution of truncated samples that exceed maximum length to prevent inappropriate penalization of otherwise sound reasoning. & Ablation studies highlight substantial training stability improvements when removing noisy reward signals from length-truncated samples. & DAPO~\citenumber{yu2025dapoopensourcellmreinforcement} \\
\addlinespace
\textbf{Inconsistent advantage estimation across variable-length sequences} & \textbf{Length-Adaptive GAE: } Dynamically adjusts the $\lambda$ parameter in GAE based on sequence length, ensuring balanced TD-error influence for both short and long outputs. & Empirical tests reveal more balanced advantage estimation and improved training stability across sequences of varying lengths, particularly beneficial for long-form reasoning. & VAPO
~\citenumber{yue2025vapoefficientreliablereinforcement} \\
\midrule
\rowcolor{white}
\multicolumn{4}{l}{\hspace*{-\tabcolsep}\colorbox{headingbg}{\parbox{\headingwidth}{\centering\textbf{\textcolor{primarycolor}{\textsf{REWARD DESIGN}}}}}} \\
\midrule
\textbf{Uncontrolled CoT length in reasoning tasks} & \textbf{Cosine Length Reward:} Applies a cosine-based reward shaping that prioritizes shorter, correct CoTs while penalizing short, incorrect ones. & Evaluation across diverse reasoning tasks reveals stabilized CoT length with preserved performance. & Demysitify~\citenumber{yeo2025demystifyinglongchainofthoughtreasoning} \\
\addlinespace
\textbf{Reward hacking in deterministic reasoning tasks} & \textbf{Accuracy+Format Reward:} Combines verification of answer correctness with structured formatting requirements that enforce explicit reasoning within specialized tags. & Rule-based reward systems demonstrate greater resistance to reward hacking than neural alternatives while simplifying the training pipeline. & DeepSeek-R1~\citenumber{deepseekai2025deepseekr1incentivizingreasoningcapability}, SimpleRL~\citenumber{zeng2025simplerl}, T1~\citenumber{hou2025advancinglanguagemodelreasoning}, Logic-RL~\citenumber{xie2025logicrlunleashingllmreasoning}, STILL-3~\citenumber{chen2025empiricalstudyelicitingimproving} \\

\addlinespace
\textbf{Language mixing issues in multilingual environments} & \textbf{Language Consistency Incentive:} Calculates rewards based on the proportion of target language words in the CoT to mitigate language mixing issues. & User studies indicate enhanced readability despite minor performance trade-offs in multilingual contexts. & DeepSeek-R1~\citenumber{deepseekai2025deepseekr1incentivizingreasoningcapability} \\
\addlinespace
\textbf{Model overthinking and verbosity} & \textbf{Overthinking Length Penalty:} Implements a weighted reward mechanism that penalizes excessive response length while preserving correctness to combat model overthinking. & Gradually introduced length penalties resulted in more token-efficient reasoning. & KIMI-K1.5~\citenumber{MoonshotAI}, DAPO~\citenumber{yu2025dapoopensourcellmreinforcement} \\
\addlinespace
\textbf{Inaccurate reward modeling in nuanced domains} & \textbf{Chain-of-Thought RM:} Enhances reward modeling with explicit step-by-step reasoning before final correctness judgment, particularly for domains with nuanced evaluation criteria. & Manual verification confirmed that CoT reward models achieved significantly higher accuracy compared to classic reward models without reasoning steps. & KIMI-K1.5~\citenumber{MoonshotAI} \\

\midrule
\rowcolor{white}
\multicolumn{4}{l}{\hspace*{-\tabcolsep}\colorbox{headingbg}{\parbox{\headingwidth}{\centering\textbf{\textcolor{primarycolor}{\textsf{TRAINING DATA}}}}}} \\
\midrule
\textbf{Resource-constrained RL training environments} & \textbf{High-impact Sample Selection:} Prioritizes training samples based on learning impact measurement. & Results show significant reduction in required training data while maintaining performance. & LIMR~\citenumber{li2025limrrlscaling} \\
\addlinespace
\textbf{Training with noisy web-extracted data} & \textbf{Noise Reduction Filtering:} Employs filtering mechanisms to remove noisy web-extracted data. & Filtered datasets demonstrate improved generalization capabilities on OOD tasks. & Demysitify~\citenumber{yeo2025demystifyinglongchainofthoughtreasoning} \\

\midrule
\rowcolor{white}
\multicolumn{4}{l}{\hspace*{-\tabcolsep}\colorbox{headingbg}{\parbox{\headingwidth}{\centering\textbf{\textcolor{primarycolor}{\textsf{MULTI-STAGE TRAINING}}}}}} \\
\midrule
\textbf{Poor readability and reasoning in direct RL approaches} & \textbf{Cold-start Progression:} Implements a phased training approach beginning with high-quality CoT data fine-tuning before transitioning to large-scale reinforcement learning. & Models with cold-start initialization exhibit enhanced readability and reasoning capabilities compared to direct RL approaches. & DeepSeek-R1~\citenumber{deepseekai2025deepseekr1incentivizingreasoningcapability}, T1~\citenumber{hou2025advancinglanguagemodelreasoning}, DeepscaleR~\citenumber{deepscaler2025}, STILL-3~\citenumber{chen2025empiricalstudyelicitingimproving} \\
\addlinespace
\textbf{Inefficient training with problems of varied difficulty} & \textbf{Strategic Sampling:} Combines curriculum-based progression from simple to complex problems with prioritization of difficult cases where model performance is weakest. & Targeted sampling approaches demonstrated faster convergence and more efficient use of computational resources during training. & KIMI K1.5~\citenumber{MoonshotAI} \\
\addlinespace
\textbf{Inefficient use of context in long-form reasoning} & \textbf{Progressive Context Scaling:} Implements a multi-stage training approach that gradually increases context window size as model performance begins to plateau at each level. & Phased context window expansion demonstrates significant improvements in both computational efficiency and final performance metrics compared to fixed maximum context training. & DeepscaleR~\citenumber{deepscaler2025} \\
\addlinespace
\textbf{Performance gaps on challenging reasoning problems} & \textbf{Targeted Annealing:} Implements a final training phase on specifically mined challenging problems with a linearly decaying learning rate to refine reasoning capabilities. & Enhanced performance metrics on complex reasoning tasks without compromising general capabilities. & Open-Reasoner-Zero~\citenumber{OpenReasonerZero2025} \\

\bottomrule
\end{tabular}
\end{table}

\subsubsection{Reward Function} \label{sec:scaling_rl:reward_function}

The reward types can be categorized according to their source and granularity as follows:
\begin{itemize}
    \item \modelicon~Model-based reward: In traditional RLHF~\cite{ouyang2022traininglanguagemodelsfollow} settings, an explicit reward model is learned from human preference data and guides the optimization process in RL training. The explicit reward model can be omitted by directly training on human preference data, resulting in an implicit reward model~\cite{rafailov2024directpreferenceoptimizationlanguage}.
    \item \ruleicon~Rule-based reward: The term ``rule-based'' represents rewards that are well-defined and can be determined by explicit rules, sometimes also termed verifiable rewards. For example, for math problems with ground truth answers or code tasks with unit tests, response correctness can be easily verified and thus used to construct the reward. This can be further extended to include response format or language consistency. Even when verification is automated using a specialized model to check answer equivalence~\cite{chen2024huatuogpto1medicalcomplexreasoning,MoonshotAI}, we still attribute it to rule-based reward as long as the model's performance closely matches ideal rule verification.
    \item \outcomeicon~Outcome reward: In general settings, the rule-based reward or model-based reward is only given to the last token of the response, termed ``outcome reward.''
   \item \processicon~Process reward: In multi-step reasoning tasks, the outcome reward may not be sufficient to supervise the policy model and help avoid logic errors in the solutions~\cite{shao2024deepseekmathpushinglimitsmathematical,lightman2023letsverifystepstep}. This necessitates more fine-grained rewards for each step, termed ``process reward,'' which are typically calculated in a model-based way. We detail the construction of process reward models in \S\ref{sec:tree_search:key_components}. Besides constructing process reward models, recent work also explores other ways to help achieve more accurate credit assignment. For example,~\citet{kazemnejad2024vineppounlockingrlpotential} replace the value networks in the PPO algorithm with unbiased Monte Carlo-based estimates.~\citet{hwang2024selfexploreenhancingmathematicalreasoning} and~\citet{setlur2024rlincorrectsyntheticdata} introduce MC-based methods to detect key errors in reasoning chains for use as ad-hoc mechanisms in DPO.
\end{itemize}

\begin{figure}[tb!] 
\centering 
\begin{tikzpicture}[ 
    font=\small, 
    align=center, 
    node distance=1cm and 1.5cm, 
    quadrant/.style={draw, minimum width=5cm, minimum height=1.6cm, anchor=north, inner sep=2pt, text width=4.9cm}, 
    gencolor/.style={fill=blue!15}, 
    limcolor/.style={fill=green!15}, 
    rarecolor/.style={fill=orange!15},
    taskbox/.style={draw, rounded corners, fill=gray!10}
] 
\draw[-stealth, thick] (-3.5,2.6) -- (8.5,2.6) node[right] {\textbf{Task} \\ \textbf{Non-verifiability}}; 
\draw[-stealth, thick] (-3.5,2.6) -- (-3.5,-5.5) node[below] {\textbf{Reward Granularity}}; 
\node[taskbox, text width=1.5cm] at (-1.5,2.6) {math; code};
\node[taskbox] at (1,2.6) {search};
\node[taskbox] at (4,2.6) {writing};
\node[taskbox] at (6.5,2.6) {research};
\node[quadrant, limcolor] (rule-outcome) at (0, 1.6) { 
\begin{minipage}[t][3cm][t]{4.8cm} 
\raggedright 
\vspace{0.3cm} 
\begin{itemize}[leftmargin=*] 
\item \textbf{Task type:} Verifiable tasks 
\item \textbf{Advantages:} Mitigates reward hacking 
\item \textbf{Disadvantages:} Restricted to tasks with clear verification criteria 
\end{itemize} 
\end{minipage}}; 
\node[quadrant, gencolor] (model-outcome) at (5.2, 1.6) { 
\begin{minipage}[t][3cm][t]{4.8cm} 
\raggedright 
\vspace{0.3cm} 
\begin{itemize}[leftmargin=*] 
\item \textbf{Task type:} General tasks 
\item \textbf{Advantages:} Universal applicability 
\item \textbf{Disadvantages:} Prone to reward hacking 
\end{itemize} 
\end{minipage}}; 
\node[quadrant, rarecolor] (rule-process) at (0, -1.7) { 
\begin{minipage}[t][3cm][t]{4.8cm} 
\raggedright 
\vspace{0.4cm} 
\begin{itemize}[leftmargin=*] 
\item \textbf{Task type}: Step-verifiable tasks 
\item \textbf{Advantages}: Improves process quality; Mitigates reward hacking 
\item \textbf{Disadvantages}: Extremely limited task types 
\end{itemize} 
\end{minipage}}; 
\node[quadrant, limcolor] (model-process) at (5.2, -1.7) { 
\begin{minipage}[t][3cm][t]{4.8cm} 
\raggedright 
\begin{itemize}[leftmargin=*] 
\item \textbf{Task type:} Multi-step reasoning tasks 
\item \textbf{Advantages:} Improves process quality 
\item \textbf{Disadvantages:} Prone to reward hacking; Less effective in RL than parallel sampling 
\end{itemize} 
\end{minipage}}; 
\node[above=0.1cm of rule-outcome, draw=none, fill=none] (rule-title) {\textbf{\ruleicon~Rule-based Reward}}; 
\node[above=0.1cm of model-outcome, draw=none, fill=none] (model-title) {\textbf{\modelicon~Model-based Reward}}; 
\node[rotate=90, left=0.4cm of rule-outcome, draw=none, fill=none, anchor=center] (outcome-title) {\textbf{\outcomeicon~Outcome Reward}}; 
\node[rotate=90, left=0.4cm of rule-process, draw=none, fill=none, anchor=center] (process-title) {\textbf{\processicon~Process Reward}}; 
\draw[-latex,thick] (rule-title) -- (rule-title |- {0,2.6}); 
\draw[-latex,thick] (model-title) -- (model-title |- {0,2.6}); 
\draw[-latex,thick] (outcome-title) -- (outcome-title -| {-3.5,0}); 
\draw[-latex,thick] (process-title) -- (process-title -| {-3.5,0}); 
\node[below=0.5cm of rule-process, draw=none, fill=none] (legend) {}; 
\node[rectangle, gencolor, minimum width=0.8cm, minimum height=0.5cm, draw] at ($(legend) + (-0.7,0)$) {}; 
\node[anchor=west] at ($(legend) + (-0.3,0)$) {General tasks}; 
\node[rectangle, limcolor, minimum width=0.8cm, minimum height=0.5cm, draw] at ($(legend) + (2.6,0)$) {}; 
\node[anchor=west] at ($(legend) + (3,0)$) {Limited tasks}; 
\node[rectangle, rarecolor, minimum width=0.8cm, minimum height=0.5cm, draw] at ($(legend) + (5.9,0)$) {}; 
\node[anchor=west] at ($(legend) + (6.3,0)$) {Rare tasks}; 
\end{tikzpicture} 
\caption{Comparisons of different reward types. Colors indicate applicable task scope.} 
\label{fig:comparisons_reward_types} 
\end{figure}
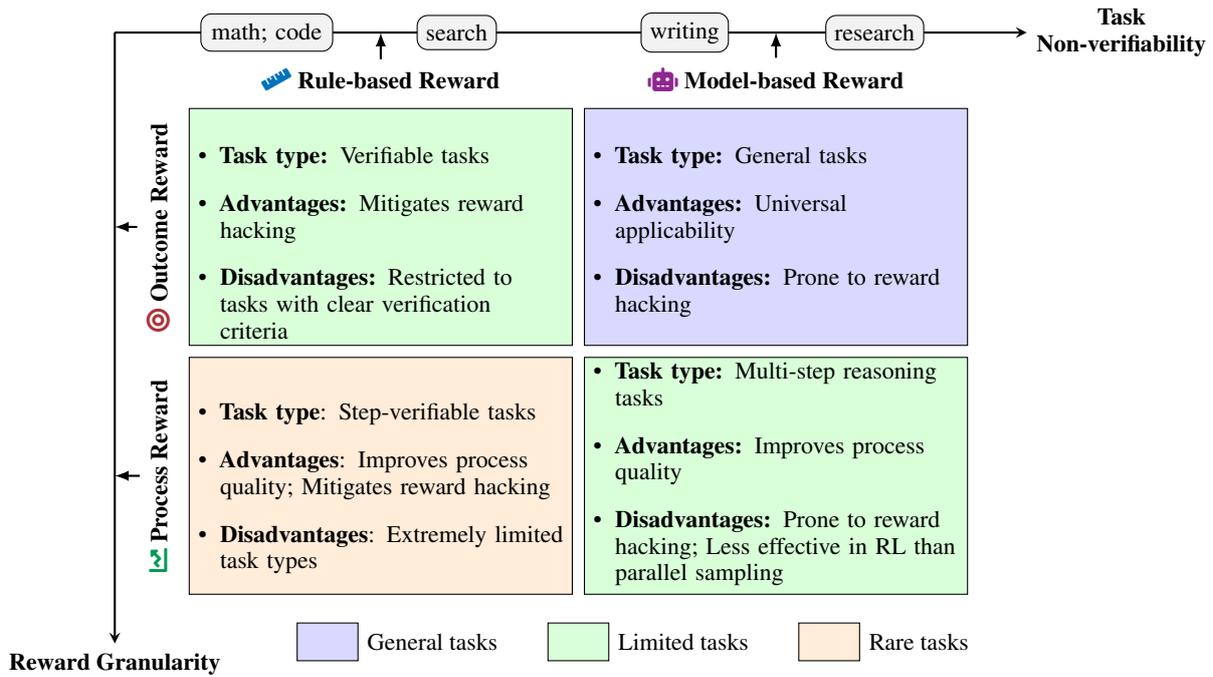

Figure~\ref{fig:comparisons_reward_types} presents a comparison of different reward types. We detail the discussion below.

\paragraph{Rule-based reward vs. model-based reward: The model-based reward can be applied to general tasks but also easily leads to reward hacking problems.} This pipeline of constructing preference data to learn a reward model to proxy human preference can be applied to general tasks, leading to its widespread adoption. However, it has been observed that the reward is an imperfect proxy in the training process. There are two prevailing explanations for this phenomenon~\cite{rafailov2024scalinglawsrewardmodel}: 1) OOD Robustness: the reward function is continuously queried using unseen model samples which are potentially out-of-distribution, and 2) Reward Mis-specification: learned reward functions may exhibit spurious correlations that cause them to prefer unintended behaviors. These issues lead to reward overoptimization problems where, during the training process, while the proxy reward score monotonically increases, the golden reward score will saturate and then decrease~\cite{gao2022scalinglawsrewardmodel}. Although this issue can be alleviated by improving the reward model's capability through increased scale or training data~\cite{ouyang2022traininglanguagemodelsfollow,hou2024doesrlhfscaleexploring} or iteratively retraining the reward model to improve its supervision of the policy model~\cite{shao2024deepseekmathpushinglimitsmathematical}, the phenomenon still exists and hinders the success of large-scale RL~\cite{deepseekai2025deepseekr1incentivizingreasoningcapability}.

\paragraph{Outcome reward vs. process reward: The fine-grained process reward may help improve the RL performance, but also introduces reward hacking problems.} Empirical results show that process rewards can help improve RL performance compared to using only outcome rewards~\cite{cui2024process,shao2024deepseekmathpushinglimitsmathematical}. However, it still faces several challenges: 1) the construction of high-quality training data for process reward models requires significant labor; 2) an imperfect process reward model can be easily hacked. For example,~\citet{gao2024designingeffectiverlreward} find that repeating correct but unnecessary reasoning steps can lead to high rewards from process reward model. Although these issues can be addressed through reward refinement, it complicates the RL pipeline; 3) process rewards show less significant improvements in RL training than in parallel sampling settings. In parallel sampling settings, empirical results show that process reward models significantly outperform outcome reward models~\cite{lightman2023letsverifystepstep,wangMathShepherdVerifyReinforce2024b} in response selection. However, the gain is not as pronounced in RL settings~\cite{gaoInterpretableContrastiveMonte2024,cui2024process,shao2024deepseekmathpushinglimitsmathematical}.

\paragraph{Optimization for rule-based reward} Rule-based rewards for eliciting long CoT reasoning primarily consist of correctness rewards and format rewards for specific tags. While this approach has proven sufficient for RL scaling, it can lead to potential content misalignment problems due to its narrow focus on accuracy. Two main issues arise from this approach. First, it may result in poor readability and inconsistent language use. Deepseek-R1~\cite{deepseekai2025deepseekr1incentivizingreasoningcapability} addresses these challenges by initially fine-tuning their model on thousands of carefully selected long CoT examples. Additionally, it introduces a language consistency reward during RL training to mitigate language misalignment issues. Second, this approach may lead to excessive response length, potentially causing overthinking problems. To address this, Kimi k1.5~\cite{MoonshotAI} implements length penalties in the later training stages, while T1~\cite{hou2025advancinglanguagemodelreasoning} penalizes responses that either exceed the context window size or contain repetitive n-grams.

The success of RL in verifiable tasks demonstrates the importance of robust reward signals. As more research into RL scaling strengthens its theoretical and empirical foundation to facilitate implementation, it decouples the RL training process into two distinct steps: first defining verifiable rewards and then conducting RL training, as partially implemented in OpenAI's Reinforcement Fine-Tuning Service.\footnote{\href{https://openai.com/form/rft-research-program/}{https://openai.com/form/rft-research-program/}} Search-R1~\cite{jin2025searchr1trainingllmsreason} utilizes a simple outcome reward function that verifies the correctness of final answers to conduct RL training and successfully endows LLMs with the ability to autonomously generate search queries during step-by-step reasoning with real-time retrieval, showcasing the power of RL beyond math and code. For future work in fields like open scientific questions, constructing reliable reward signals remains an open challenge and offers significant potential for innovation.

\subsubsection{Policy Model Selection} \label{sec:scaling_rl:policy_model_selection}
The policy model is a prerequisite for successful RL training. The selection criteria can be based on the following aspects:

\paragraph{Model Family}
As shown in Table~\ref{tab:summary_scaling_rl_work}, most RL scaling work utilizes Qwen2.5 as the base model. Recent studies demonstrate that Qwen2.5 exhibits cognitive behaviors such as verification and correction in its problem-solving process before applying RL~\cite{gandhi2025cognitivebehaviorsenableselfimproving,liu2025understanding,liu2025oatzero}, although the model cannot effectively use them. This indicates that the model's pretrained knowledge already contains these thinking patterns.~\citet{gandhi2025cognitivebehaviorsenableselfimproving} investigate this phenomenon based on the observation that Qwen-2.5-3B exhibits substantial gains while Llama-3.2-3B quickly plateaus under identical RL training conditions for the game of Countdown. When Llama is primed with synthetic reasoning traces containing these behaviors or pretrained on cognitive behavioral augmentation data, it shows substantial improvements during RL, matching Qwen's performance trajectory. This highlights the importance of pretraining on corpus containing the cognitive behaviors before conducting RL. Moreover,~\cite{yue2025doesreinforcementlearningreally} find that RL training does not increase the Pass@K score at large K values, indicating that most reasoning abilities manifested in RL-trained models are already possessed by base models. This underscores the importance of base model selection. Further studies should be conducted to analyze the relationship between base model capability and RL training in this new context.

\paragraph{Model Size} While traditional RLHF settings show that larger models gain fewer benefits from reinforcement learning optimization~\cite{gao2022scalinglawsrewardmodel,hou2024doesrlhfscaleexploring}, RL scaling settings demonstrate that larger models achieve higher token efficiency and thus better performance~\cite{MoonshotAI}. The limited success in reproducing DeepSeek-R1-Zero's (671B) scaling behavior in 7B or smaller models for challenging tasks without long CoT cold start further suggests that model size significantly impacts scaling behavior.

\subsubsection{Training Data Construction} \label{sec:scaling_rl:training_data_construction}
The quality and quantity of training data significantly affect the efficiency and upper bound of RL.

\paragraph{Data Quality} Eliminating easy queries that require no further training helps save the unnecessary computation cost of RL as a post-training technique, where query difficulty can be estimated by sampling multiple times from the policy model to calculate the success rate for correct answers~\cite{MoonshotAI,chen2025empiricalstudyelicitingimproving}. Similarly, it is also beneficial to remove problems for which the current model lacks the fundamental capability to solve~\cite{chen2025empiricalstudyelicitingimproving}. From the training perspective, queries that the model consistently answers correctly or incorrectly introduce the gradient-decreasing problem. DAPO~\cite{yu2025dapoopensourcellmreinforcement} proposes a dynamic sampling strategy that over-samples and filters out prompts with accuracies of 1 and 0, observing significant performance gains, which can be considered an online difficulty control method.

\paragraph{Data Quantity} In traditional RLHF settings, scaling the prompt quantity does not lead to significant performance improvements~\cite{hou2024doesrlhfscaleexploring}. However, this conclusion does not hold for RL scaling scenarios. Open-Reasoner-Zero~\cite{OpenReasonerZero2025} investigates the performance discrepancy between a 7.5K MATH~\citep{hendrycks2021measuringmathematicalproblemsolving} training set and their curated 57K prompt set, finding that the larger set leads to continuous scaling in both accuracy and response length, while the smaller set plateaus. Similarly, DeepSeek-R1-Zero observes continuous performance improvements using their large-scale curated dataset~\cite{deepseekai2025deepseekr1incentivizingreasoningcapability}.

\subsubsection{Multi-stage Training} \label{sec:scaling_rl:multi_stage_training}
Training efficiency can be enhanced by employing the following multi-stage training strategy:

\paragraph{Long CoT Cold Start} Fine-tuning on long CoT data before RL training can facilitate subsequent RL improvements~\cite{yeo2025demystifyinglongchainofthoughtreasoning} and mitigate early instability issues during RL training~\cite{deepseekai2025deepseekr1incentivizingreasoningcapability}. Additionally, enhancing the quality of long CoT significantly amplifies RL gains~\cite{yeo2025demystifyinglongchainofthoughtreasoning}. Furthermore,~\citet{li2025cold} demonstrates improved performance by incorporating sparse updates and adaptive termination mechanisms into the SFT loss function, which helps preserve response diversity after training.

\paragraph{Iterative Lengthening Strategy} DeepScaleR-1.5B-Preview~\cite{deepscaler2025} initially restricts the context window size to 8K, during which the model generates shorter responses while training rewards increase. Upon reaching a critical point where model responses begin to lengthen, the context window size is expanded to 16K and subsequently to 24K (see the `DeepScaleR-1.5B-Preview' row in Table~\ref{tab:summary_scaling_rl_work}). This strategy guides controlled response length expansion while reducing computational costs.

\paragraph{Curriculum Sampling Strategy} When allocating a restricted computation budget in the initial training phase to very challenging problems, this often yields few correct samples, resulting in lower training efficiency. To address this limitation, the curriculum sampling strategy begins with training on simpler tasks before progressively advancing to more complex ones. Kimi K1.5~\cite{MoonshotAI} reports enhanced performance by implementing this curriculum sampling strategy, leveraging their training dataset that naturally incorporates grade and difficulty labels. Similarly, logic-RL~\cite{xie2025logicrlunleashingllmreasoning} examines the utility of this approach but finds that improvements are not substantial in logic puzzles tasks, concluding that it is necessary to balance the complexity of staged training against potential performance gains.

\begin{table}[tb!]
    \centering
    \scriptsize
\caption{An organization of long CoT resource.}
    \begin{tabular}{l|cccccc}
        \toprule
        \textbf{Work} & \textbf{Application} & \textbf{Type} & \textbf{Source} & \textbf{Quantity} &   \textbf{Modality} &  \textbf{Link} \\
        \midrule

        O1 Journey–Part 1~\citenumber{o1journey} & Math & Synthesize & GPT-4o  & 0.3K & Text & \href{https://github.com/GAIR-NLP/O1-Journey}{\faGithub} \href{https://huggingface.co/datasets/GAIR/o1-journey}{\includegraphics[height=1em]{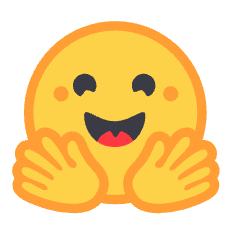}}  \\

        \midrule
        Marco-o1~\citenumber{zhao2024marcoo1openreasoningmodels} & Reasoning & Synthesize & Qwen2-7B-Instruct &  10K   & Text   & \href{https://github.com/AIDC-AI/Marco-o1}{\faGithub}  \\
        \midrule

        STILL-2~\citenumber{min2024imitateexploreselfimprovereproduction} & Math, Code, Science, Puzzle & Distillation & \makecell[c]{DeepSeek-R1-Lite-Preview \\ QwQ-32B-preview} & 5K & Text &  \href{https://github.com/RUCAIBox/Slow_Thinking_with_LLMs}{\faGithub} \href{https://huggingface.co/datasets/RUC-AIBOX/long_form_thought_data_5k}
        {\includegraphics[height=1em]{figure/huggingface.png}} \\
        \midrule
        RedStar-math~\citenumber{xu2025redstardoesscalinglongcot} & Math & Distillation & QwQ-32B-preview & 4K & Text & \href{https://huggingface.co/datasets/RedStar-Reasoning/math_dataset}
        {\includegraphics[height=1em]{figure/huggingface.png}}  \\
        \midrule
        RedStar-code~\citenumber{xu2025redstardoesscalinglongcot} & Code & Distillation & QwQ-32B-preview & 16K & Text & \href{https://huggingface.co/datasets/RedStar-Reasoning/code_dataset}
        {\includegraphics[height=1em]{figure/huggingface.png}}  \\        
        \midrule
        RedStar-multimodal~\citenumber{xu2025redstardoesscalinglongcot} & Math & Distillation & QwQ-32B-preview & 12K & \makecell[c]{Vision \\ Text} & \href{https://huggingface.co/datasets/RedStar-Reasoning/multimodal_dataset}{\includegraphics[height=1em]{figure/huggingface.png}}  \\  
        \midrule
         S1K~\citenumber{muennighoff2025s1simpletesttimescaling} & Math, Science, Code & Distillation &  Gemini Flash Thinking  & 1K  & Text & \href{https://github.com/simplescaling/s1}{\faGithub} \href{https://huggingface.co/datasets/simplescaling/s1K}{\includegraphics[height=1em]{figure/huggingface.png}} \\

        \midrule
        S1K-1.1~\citenumber{muennighoff2025s1simpletesttimescaling} & Math, Science, Code & Distillation & DeepSeek R1  & 1K  & Text & \href{https://github.com/simplescaling/s1}{\faGithub} \href{https://huggingface.co/datasets/simplescaling/s1K-1.1}{\includegraphics[height=1em]{figure/huggingface.png}} \\
        
        \midrule
        LIMO~\citenumber{ye2025limoreasoning} & Math & Distillation & \makecell[c]{DeepSeek R1 \\ DeepSeekR1-Distill-Qwen-32B} & 0.8K & Text & \href{https://github.com/GAIR-NLP/LIMO}{\faGithub} \href{https://huggingface.co/datasets/GAIR/LIMO}{\includegraphics[height=1em]{figure/huggingface.png}} \\
        \midrule
        OpenThoughts-114k~\citenumber{openthoughts} & Math, Code, Science, Puzzle & Distillation & DeepSeek R1 & 114K & Text & \href{https://github.com/open-thoughts/open-thoughts}{\faGithub} \href{https://huggingface.co/datasets/open-thoughts/OpenThoughts-114k}{\includegraphics[height=1em]{figure/huggingface.png}} \\
        \midrule
        OpenR1-Math-220k~\citenumber{openr1} & Math & Distillation & DeepSeek R1 & 220K & Text & \href{https://github.com/huggingface/open-r1}{\faGithub} \href{https://huggingface.co/datasets/open-r1/OpenR1-Math-220k}{\includegraphics[height=1em]{figure/huggingface.png}} \\
        \midrule
        OpenThoughts2-1M~\citenumber{openthoughts} &  Math, Code, Science, Puzzle & Distillation & DeepSeek R1 & 1M & Text & \href{https://github.com/open-thoughts/open-thoughts}{\faGithub} \href{https://huggingface.co/datasets/open-thoughts/OpenThoughts2-1M}{\includegraphics[height=1em]{figure/huggingface.png}} \\
        
        \midrule
         CodeForces-CoTs~\citenumber{openthoughts} &  Code & Distillation & DeepSeek R1 & 47K & Text & \href{https://github.com/huggingface/open-r1}{\faGithub} \href{https://huggingface.co/datasets/open-r1/codeforces-cots}{\includegraphics[height=1em]{figure/huggingface.png}}
        \\
        \midrule
        Sky-T1-17k~\citenumber{li2025llmseasilylearnreason}  & Math, Code, Science, Puzzle & Distillation  & QwQ-32B-Preview & 17K & Text & \href{https://github.com/NovaSky-AI/SkyThought}{\faGithub} \href{https://huggingface.co/datasets/NovaSky-AI/Sky-T1_data_17k}{\includegraphics[height=1em]{figure/huggingface.png}} \\
        \midrule
        $S^2R$~\citenumber{ma2025s2rteachingllmsselfverify} & Math & Synthesize & Qwen2.5-Math-7B &3K & Text & \href{https://github.com/NineAbyss/S2R}{\faGithub} \href{https://huggingface.co/datasets/S2R-data/S2R-dataset}{\includegraphics[height=1em]{figure/huggingface.png}}   \\
        \midrule
        R1-Onevision~\citenumber{yang2025r1onevisionadvancinggeneralizedmultimodal} & \makecell[c]{Science, Math, General}  & Distillation & DeepSeek R1 &155K & \makecell[c]{Vision \\ Text} & \href{https://github.com/Fancy-MLLM/R1-Onevision}{\faGithub} \href{https://huggingface.co/datasets/Fancy-MLLM/R1-Onevision}{\includegraphics[height=1em]{figure/huggingface.png}}\\
        \midrule
       OpenO1-SFT~\citenumber{openo1} & \makecell[c]{Math, Code}  & Synthesize & - & 77K & Text & \href{https://github.com/Open-Source-O1/Open-O1}{\faGithub} \href{https://huggingface.co/datasets/O1-OPEN/OpenO1-SFT}{\includegraphics[height=1em]{figure/huggingface.png}} 
        \\ 
        \midrule
       Medical-o1~\citenumber{chen2024huatuogpto1medicalcomplexreasoning} & Medical  & Distillation & Deepseek R1 & 25K & Text & \href{https://github.com/FreedomIntelligence/HuatuoGPT-o1}{\faGithub} \href{https://huggingface.co/datasets/FreedomIntelligence/medical-o1-reasoning-SFT}{\includegraphics[height=1em]{figure/huggingface.png}} 
        \\ 
        \midrule

        O1 Journey–Part 3~\citenumber{huang2025o1replicationjourney} & Medical & Distillation & o1-preview   & 0.5K & Text & \href{https://github.com/SPIRAL-MED/Ophiuchus}{\faGithub} \href{https://huggingface.co/datasets/SPIRAL-MED/o1-journey-Ophiuchus}{\includegraphics[height=1em]{figure/huggingface.png}}  \\
        \midrule
        
        SCP-116K~\citenumber{lu2025scp116khighqualityproblemsolutiondataset} & Math, Science & Distillation & Deepseek R1  & 116K & Text & \href{https://github.com/AQA6666/SCP-116K-open}{\faGithub} \href{https://huggingface.co/datasets/EricLu/SCP-116K}{\includegraphics[height=1em]{figure/huggingface.png}}  \\
        \midrule
        open-r1-multimodal~\citenumber{open-r1-multimodal} & \makecell[c]{Math} & Distillation & GPT-4o  & 8K & \makecell[c]{Vision \\ Text} & \href{https://github.com/EvolvingLMMs-Lab/open-r1-multimodal}{\faGithub} \href{https://huggingface.co/datasets/lmms-lab/multimodal-open-r1-8k-verified}{\includegraphics[height=1em]{figure/huggingface.png}}  \\
        \midrule
        Vision-R1-cold~\citenumber{huang2025vision} & \makecell[c]{Science, Math, General} & Distillation & Deepseek R1  & 200K &  \makecell[c]{Vision \\ Text} & \href{https://github.com/Osilly/Vision-R1}{\faGithub} \href{https://huggingface.co/datasets/Osilly/Vision-R1-cold}{\includegraphics[height=1em]{figure/huggingface.png}}  \\
        \midrule
        \makecell[l]{MMMU-Reasoning-Distill-\\Validation~\citenumber{mmmu_reasoning_distill_validation}} & \makecell[c]{Science, Math, General} & Distillation & Deepseek R1  & 0.8K &  \makecell[c]{Vision \\ Text} &  \href{https://www.modelscope.cn/datasets/modelscope/MMMU-Reasoning-Distill-Validation}{\includegraphics[height=1em]{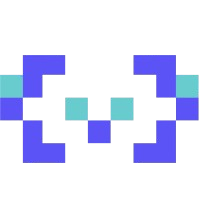}}  \\
        \midrule
        \makecell[l]{Clevr-CoGenT~\citenumber{chen2025r1v}} & \makecell[c]{Vision Counting} & Distillation & Deepseek R1  & 37.8K &  \makecell[c]{Vision \\ Text}& \href{https://github.com/Deep-Agent/R1-V}{\faGithub} \href{https://huggingface.co/datasets/MMInstruction/Clevr_CoGenT_TrainA_R1}{\includegraphics[height=1em]{figure/huggingface.png}}  \\
        \midrule
        \makecell[l]{VL-Thinking\citenumber{vl-thinking2025}} & \makecell[c]{Science, Math, General} & Distillation & Deepseek R1  & 158K &  \makecell[c]{Vision \\ Text} & \href{https://github.com/UCSC-VLAA/VL-Thinking}{\faGithub} \href{https://huggingface.co/datasets/UCSC-VLAA/VL-Thinking}{\includegraphics[height=1em]{figure/huggingface.png}}  \\
        \midrule
        \makecell[l]{Video-R1\citenumber{feng2025video}} & \makecell[c]{Video} & Distillation & Qwen2.5-VL-72B  & 158K &  \makecell[c]{Vision \\ Text} & \href{https://github.com/tulerfeng/Video-R1}{\faGithub} \href{https://huggingface.co/datasets/Video-R1/Video-R1-data}{\includegraphics[height=1em]{figure/huggingface.png}}  \\
        \midrule
        Embodied-Reasoner~\citenumber{zhang2025embodied} & \makecell[c]{Embodied AI} & Synthesize & GPT-4o  & 9K &  \makecell[c]{Vision \\ Text }  &\href{https://github.com/zwq2018/embodied_reasoner}{\faGithub} \href{https://huggingface.co/datasets/zwq2018/embodied_reasoner}{\includegraphics[height=1em]{figure/huggingface.png}}  \\

        \midrule
        OpenCodeReasoning~\citenumber{ahmad2025opencodereasoning} & Code & Distillation & DeepSeek R1 & 736K & Text & \href{https://huggingface.co/datasets/nvidia/OpenCodeReasoning}{\includegraphics[height=1em]{figure/huggingface.png}}\\

        \midrule
        SafeChain~\citenumber{jiang2025safechainsafetylanguagemodels} & \makecell[c]{Safety} & Distillation & Deepseek R1  & 40K & Text & \href{https://github.com/uw-nsl/safechain}{\faGithub} \href{https://huggingface.co/datasets/UWNSL/SafeChain}{\includegraphics[height=1em]{figure/huggingface.png}}  \\

        \midrule
        KodCode~\citenumber{xu2025kodcode} & Code & Distillation & DeepSeek R1 & 2.8K & Text & \href{https://github.com/KodCode-AI/kodcode}{\faGithub} \href{https://huggingface.co/datasets/KodCode/KodCode-V1-SFT-R1}{\includegraphics[height=1em]{figure/huggingface.png}} \\

        \bottomrule
    \end{tabular}
    \label{tab:long_cot_resource}
\end{table}
\subsection{Supervised Fine-tuning} \label{sec:sft}
Recent work demonstrates that long CoT test-time scaling behavior can be elicited through simple supervised fine-tuning on similar data~\cite{muennighoff2025s1simpletesttimescaling,ye2025limoreasoning}. This approach is promising given its simpler training process and higher data efficiency compared to RL-based methods. Table~\ref{tab:long_cot_resource} presents an organization of long CoT resources. We detail the core design considerations of SFT-based methods as follows:

\paragraph{Training data source} The data sources can be categorized based on synthesized data or distillation from existing long CoT models. The trajectory synthesis method includes directly collecting trajectories from tree or graph search processes~\cite{lehnert2024abetterplanningtransformers,gandhi2024streamsearchsoslearning,ye2024physicslanguagemodels22} in tasks like path finding or formal logic problems. However, the limited tasks and cognitive behaviors hinder its wider application. Another line of work first solves problems using test-time scaling methods such as tree search processes and multi-turn correction, then translates the search history into thorough exploration trajectories~\cite{zhao2024marcoo1openreasoningmodels,ma2025s2rteachingllmsselfverify}. For example, Journey Learning~\cite{o1journey} proposes first guiding LLMs to solve problems using tree search, then using another LLM to translate the backtracking or evaluation steps into natural language to form the trajectory. However, the lack of logical coherence and diversity in these trajectories limits their performance. \citet{xiEnhancingLLMReasoning2024} transfer multi-turn correction processes into self-talk data, but this approach struggles to generalize to challenging problems due to LLMs' limitations in critique. In contrast to the complexity of synthesis methods, the distillation method directly extracts trajectories from open-source long CoT models, such as Deepseek R1 or QwQ~\cite{ye2025limoreasoning,muennighoff2025s1simpletesttimescaling,li2025llmseasilylearnreason}. While this method is cost-effective and performs well compared to synthesis methods, the nature of distillation makes it difficult for student models to surpass their teacher models~\cite{o1journeypart2}.

\paragraph{Training data quality} The quality of long-CoT data significantly determines its effectiveness in eliciting model reasoning ability. It comprises both query quality and response quality. For queries, they should be challenging to the base model and cover diverse domains. LIMO~\cite{ye2025limoreasoning} applies a multi-stage filtration process and retains queries that are challenging even for state-of-the-art reasoning models. S1~\cite{muennighoff2025s1simpletesttimescaling} maintains difficult queries based on model performance and reasoning trace length while covering diverse subjects. For responses, they can be post-filtered through answer checkers and code interpreters. For example, OpenThoughts-114k~\cite{openthoughts} with verifiers for filtering demonstrates higher performance than OpenThoughts-Unverified-173k, and the precision of verifiers affects the performance. Regarding response content, \citet{li2025llmseasilylearnreason} find that the global reasoning structure matters more than local content details through perturbation experiments.

\paragraph{Training data quantity} Empirical results show that scaling data quantity does not bring expected performance improvements relative to computational cost. S1~\cite{muennighoff2025s1simpletesttimescaling} compares the curated 1k dataset with the 59k-full dataset and finds that performance gains are limited. Similarly, the modest performance gap between OpenThoughts-114k~\cite{openthoughts} and carefully curated 1k datasets further supports this observation. LIMO~\cite{ye2025limoreasoning} attributes this phenomenon to the Less-Is-More Reasoning Hypothesis, which suggests that the training data primarily serves to elicit sophisticated reasoning capabilities inherent in the model rather than to teach new knowledge.

\paragraph{Training methods} Whether the self-correction and backtracking ability can be learned through parameter-efficient fine-tuning such as LoRA~\cite{hu2021loralowrankadaptationlarge} is still under exploration. \citet{li2025llmseasilylearnreason} compare the performance of LoRA fine-tuning with full parameter fine-tuning and find that the performance is close. This observation contradicts the conclusion of \citet{ye2024physicslanguagemodels22} that the self-correction pattern cannot be learned with LoRA fine-tuning, though the latter only conduct experiments on the synthesized dataset using GPT2-small. More studies should be conducted to verify the effectiveness of LoRA fine-tuning.

\paragraph{Base models} The performance gain of different base models from long CoT fine-tuning varies significantly~\cite{li2025llmseasilylearnreason}. \citet{li2025smallmodelsstrugglelearn} find small models ($\leq$ 3B parameters) do not consistently benefit from long CoT reasoning and instead perform better when fine-tuned on shorter reasoning chains. They attribute this to the limited domain knowledge of small models and demonstrate that models with more domain knowledge perform better than those without. Future work should quantitatively analyze the relationship between performance and characteristics of base models.

Although the SFT-based method is easier to implement and more cost-effective compared to RL-based methods, it has several potential limitations. First, the success of the SFT method in eliciting long CoT reasoning ability largely depends on existing open-source long CoT models trained through RL, highlighting the reliance on the teacher models. This characteristic suggests that SFT and RL methods should be combined for higher data efficiency. For example, in the training process of Deepseek-R1, they employ multi-stage training methods with interleaved RL and SFT training, fully utilizing the advantages of both approaches. Second, the SFT-based method is often criticized for memorizing fixed patterns rather than achieving true generalization~\cite{chu2025sftmemorizesrlgeneralizes,mirzadeh2024gsmsymbolicunderstandinglimitationsmathematical,zhang2024carefulexaminationlargelanguage}. Although empirical results show this criticism may not always hold true as models after small-data SFT can still improve their performance in other subjects and domains~\cite{ye2025limoreasoning}, future studies should carefully analyze the relationship between SFT training steps and generalization ability.

\subsection{Iterative Self-reinforced Learning} \label{sec:iterative_self-reinforced_learning}

The trajectories generated from test-time scaling methods can be utilized to optimize the policy model through offline methods such as SFT or DPO, thereby achieving self-improvement. This framework functions as a self-reinforcing cycle where data is first generated, then leveraged for learning, after which the original policy model is replaced by its enhanced iteration. We term this training paradigm \textbf{iterative self-reinforced learning (ISRL)}. An organization of relevant works is presented in Table~\ref{tab:iterative_self_reinforced_learning}. The core steps of the algorithm are detailed as follows:
\begin{table}[tb!]
\centering
\scriptsize
\caption{An organization of works on iterative self-reinforced learning. \textbf{IT} denotes whether iterative training is involved. Under \textbf{Sampling}, \textbf{Query} denotes whether new queries are synthesized, \textbf{Response} denotes the sampling method. For the \textbf{Scoring} column, \textbf{GT} denotes ground truth, \textbf{CI} denotes code interpreter, \textbf{MV} denotes majority voting, \textbf{RM} denotes model-based reward model or LLM-as-a-judge. Under the \textbf{Selection \& Update}, \textbf{Algorithm} represents the training algorithm for the policy model, \textbf{Model} represents the training model of each turn (\textbf{Orig.} denotes the original model, \textbf{Curr.} denotes the current model in the iteration). \textbf{Data} represents the source of training data (\textbf{Curr.} represents data from the current turn, \textbf{Orig.} represents the original data, \textbf{Prev.} represents data from all previous turns). \textbf{RM} represents whether the reward model gets updated in the process.}
\begin{tabular}{lcccccccc}
\toprule
\multirow{2}{*}{\textbf{Work}} & \multirow{2}{*}{\textbf{IT}} & \multicolumn{2}{c}{\textbf{Sampling}} & \multirow{2}{*}{\textbf{Scoring}} & \multicolumn{4}{c}{\textbf{Selection \& Update}} \\
\cmidrule{3-4} \cmidrule{6-9}
& & \textbf{Query} & \textbf{Response} & & \textbf{Algorithm} & \textbf{Model} & \textbf{Data} & \textbf{RM} \\
\midrule
\rowcolor[rgb]{ .949,  .949,  .949} STaR~\cite{zelikmanSTaRBootstrappingReasoning2022} & \blackcheck &  \blackcross & Parallel & GT & SFT & Orig. & Curr. & \blackcross \\
P3~\cite{haluptzok2023languagemodelsteachprogram} &  \blackcheck & \blackcheck & Parallel & CI & SFT & Curr. & Curr. & \blackcross \\
\rowcolor[rgb]{ .949,  .949,  .949} LMSI~\cite{huangLargeLanguageModels2023} & \blackcross  & \blackcheck & Parallel & MV & SFT & Curr. & Curr. & \blackcross \\
RAFT~\cite{dongRAFTRewardRAnked2023} & \blackcheck & \blackcross & Parallel & RM & SFT & Curr. & Curr. & \blackcross \\
\rowcolor[rgb]{ .949,  .949,  .949} RFT~\cite{yuanScalingRelationshipLearning2023} & \blackcross & \blackcross & Parallel & GT & SFT & Orig. & Curr.; Orig.   & \blackcross \\
ReST~\cite{gulcehreReinforcedSelfTrainingReST2023} & \blackcheck & \blackcross & Parallel & RM & SFT & Curr. & Curr.; Prev.  & \blackcross \\
\rowcolor[rgb]{ .949,  .949,  .949} $\text{ReST}^{\text{EM}}$~\cite{singhHumanDataScaling2024} & \blackcheck & \blackcross & Parallel & GT & SFT & Orig. & Curr. & \blackcross \\
Self-Rewarding~\cite{yuanSelfRewardingLanguageModels2024} & \blackcheck & \blackcheck & Parallel & RM & DPO & Curr. & Curr. & \blackcheck\\
\rowcolor[rgb]{ .949,  .949,  .949} V-STaR~\cite{hosseiniVSTaRTrainingVerifiers2024b} & \blackcheck & \blackcross & Parallel &  GT  & SFT & Orig. &  Curr.; Prev. & \blackcheck \\
IRPO~\cite{pangIterativeReasoningPreference2024a} & \blackcheck & \blackcross & Parallel & GT & DPO & Curr. & Curr. & \blackcross \\
\rowcolor[rgb]{ .949,  .949,  .949} Qwen2.5-MATH~\cite{yangQwen25MathTechnicalReport2024} & \blackcheck & \blackcross & Parallel & GT; RM  & SFT & - & - & \blackcheck \\
Process-SelfRewarding~\cite{zhang2025processbasedselfrewardinglanguagemodels} & \blackcheck & \blackcross & Parallel &  RM  & DPO &Curr. & Curr. & \blackcheck \\
\rowcolor[rgb]{ .949,  .949,  .949} TS-LLM~\cite{fengAlphazerolikeTreeSearchCan2024} & \blackcheck &\blackcross & MCTS & GT & SFT & Curr. & Curr. & \blackcheck \\
ALPHALLM~\cite{tianSelfImprovementLLMsImagination2024b} & \blackcheck & \blackcross & MCTS & RM & SFT & Curr. & Curr. & \blackcheck \\
\rowcolor[rgb]{ .949,  .949,  .949} AlphaMath~\cite{chen2024alphamathzeroprocesssupervision} & \blackcheck & \blackcross & MCTS & GT & SFT & Curr. & Curr. & \blackcheck \\
$\text{ReST-MCTS}^*$~\cite{zhangReSTMCTSLLMSelfTraining2024a} & \blackcheck & \blackcross & MCTS & GT & SFT & Curr. & Curr. & \blackcheck \\
\rowcolor[rgb]{ .949,  .949,  .949} MCTS-IPL~\cite{xieMonteCarloTree2024} & \blackcheck & \blackcross &MCTS & GT; RM& DPO  & Curr. & Curr. & \blackcross \\
CPL~\cite{wangCPLCriticalPlan2024} & \blackcheck & \blackcross & MCTS & GT; RM & SFT; Step-APO & Curr. & Curr. & \blackcheck \\
\rowcolor[rgb]{ .949,  .949,  .949} SRA-MCTS~\cite{xu2024sramctsselfdrivenreasoningaugmentation} & \blackcross & \blackcross & MCTS & RM & SFT & Curr. & Curr. & \blackcheck \\
rStar-Math~\cite{guanRStarMathSmallLLMs2025} & \blackcheck & \blackcross & MCTS & GT & SFT & Curr. & Curr. & \blackcheck \\
\rowcolor[rgb]{ .949,  .949,  .949} Xiong et al. ~\cite{xiong2025selfrewardingcorrectionmathematicalreasoning} & \blackcross & \blackcross & \makecell[c]{Multi-turn \\ correction} & GT & SFT & Curr. & Curr. & \blackcross \\
$\mu\text{CODE}$~\cite{jain2025multiturncodegenerationsinglestep} & \blackcheck & \blackcross & \makecell[c]{Multi-turn \\ correction} & RM & SFT & Curr. & Curr. & \blackcheck \\
\rowcolor[rgb]{ .949,  .949,  .949} SPaR~\cite{cheng2025sparselfplaytreesearchrefinement} & \blackcheck & \blackcross & Ensemble & RM & DPO & Curr. & Curr. & \blackcheck \\
SWE-Reasoner~\cite{ma2025thinkinglongerlargerenhancing} & \blackcross & \blackcross & \makecell{Long CoT} & GT & SFT & Curr. & Curr. & \blackcross \\

\bottomrule
\end{tabular}
\label{tab:iterative_self_reinforced_learning}
\end{table}

\paragraph{Sampling} First, responses are sampled from the policy model using a controlled sampling function, which can be implemented with the aforementioned test time scaling methods, including parallel sampling~\cite{zelikmanSTaRBootstrappingReasoning2022,gulcehreReinforcedSelfTrainingReST2023,dongRAFTRewardRAnked2023}, tree search~\cite{fengAlphazerolikeTreeSearchCan2024,tianSelfImprovementLLMsImagination2024b,zhangReSTMCTSLLMSelfTraining2024a}, and multi-turn correction~\cite{xiong2025selfrewardingcorrectionmathematicalreasoning,jain2025multiturncodegenerationsinglestep}. To enhance the diversity and quality of candidate responses, STaR~\cite{zelikmanSTaRBootstrappingReasoning2022} provides the correct answer as a hint to guide the generation of rationales when the model fails to solve the problem independently. $\text{ReST-MCTS}^*$~\cite{zhangReSTMCTSLLMSelfTraining2024a} empirically demonstrates that step-level tree search outperforms parallel sampling, as the step-level search improves the quality of intermediate reasoning steps. Additionally, other research explores methods to increase query diversity, such as using few-shot prompting to synthesize new problems~\cite{haluptzok2023languagemodelsteachprogram,yuanSelfRewardingLanguageModels2024}.

\paragraph{Scoring} The sampled responses can be scored using the following methods: 1) for tasks like math and code, the generated solution can be verified against ground truth answers~\cite{zelikmanSTaRBootstrappingReasoning2022} or validated with unit tests~\cite{huangLargeLanguageModels2023}; 2) for general tasks, an off-the-shelf reward model can be utilized to score the responses~\cite{dongRAFTRewardRAnked2023}, or the policy model itself can serve as a judge~\cite{yuanSelfRewardingLanguageModels2024}; 3) for tree search algorithms, accompanying scores from the sampling process help select the correct solution or construct preference pairs~\cite{guanRStarMathSmallLLMs2025,zhangChainPreferenceOptimization2024,xieMonteCarloTree2024}; 4) majority voting. The majority voting strategy can help determine the correct answer when the ground truth is unavailable~\cite{huangLargeLanguageModels2023}.

\paragraph{Selection and Update} The response pool accompanied by scores is further selected and utilized to update the policy model and optionally the reward model. The policy model can be updated using SFT~\cite{zelikmanSTaRBootstrappingReasoning2022,gulcehreReinforcedSelfTrainingReST2023,singhHumanDataScaling2024} or DPO~\cite{yuanSelfRewardingLanguageModels2024,pangIterativeReasoningPreference2024a}. The reward model can also be updated, such as updating the process reward model on labels generated by rollouts in the tree search process~\cite{fengAlphazerolikeTreeSearchCan2024,zhangReSTMCTSLLMSelfTraining2024a} or training an outcome reward model on generated positive and negative samples~\cite{hosseiniVSTaRTrainingVerifiers2024b,yangQwen25MathTechnicalReport2024}. For example, V-star~\cite{hosseiniVSTaRTrainingVerifiers2024b} trains an ORM to utilize negative samples generated in the sampling process and uses it for reranking during inference time. During the update process, the policy model for next iteration data generation can be fine-tuned from either the initial model or the model in the current iteration. The training data can originate from various sources: the current turn, the initial dataset, or accumulated data from all previous turns.

Although ISRL is promising considering its data efficiency empowered by offline methods and the advantage of not requiring expert demonstrations, empirical results show that the rate of improvement tends to plateau or even decline slightly after few iterations~\citep{wu2024progressregressselfimprovementreversal}. This phenomenon contrasts with RL scaling methods where model performance improves monotonically. One major distinction between the two algorithms is that ISRL is closer to off-policy sampling.
As demonstrated by~\citet{tajwar2024preferencefinetuningllmsleverage}, a higher degree of on-policy sampling leads to better performance. This is further supported by the observation of ~\citet{shao2024deepseekmathpushinglimitsmathematical} that the performance of online rejection sampling fine-tuning (RFT) is comparable to RFT in the early stage of training but gains a significant advantage in the later stage. Moreover, for algorithms that employ SFT for policy updates, they do not utilize negative gradients for pushing down certain responses. Empirical results show that including negative gradients in policy updates leads to significantly better performance compared to using positive gradients alone, especially in reasoning tasks~\citep{MoonshotAI}.

\newcommand{\rl}[1]{{\fboxsep=1pt\colorbox{yellow!25}{#1}}}
\newcommand{\sft}[1]{{\fboxsep=1pt\colorbox{blue!15}{#1}}}
\newcommand{\iterative}[1]{{\fboxsep=1pt\colorbox{green!15}{#1}}}

\begin{table}[tb!]
  \centering
  \tiny
    \caption{Application of Test Time Scaling in different domains. Under the \textbf{Long CoT}, \textit{italics} represents traditional CoT work, \rl{Yellow color} represents using RL tech, \sft{Purple color} represents using SFT tech. \iterative{Green color} represents combining test time scaling with iterative training, i.e., the iterative self-reinforced learning in the paper.}
\label{tab:application_tts}
  \begin{tabular}{lllll}
    \toprule
    \makecell[c]{\textbf{Application}}  
    & \makecell[c]{\textbf{Parallel} \textbf{Sampling}} & \makecell[c]{\textbf{Tree Search}} & \makecell[c]{\textbf{Multi-turn} \textbf{Correction}} & \makecell[c]{\textbf{Long CoT}}  \\
    \midrule
    \textbf{Mathematics} & 
    \makecell[l]{Self-Consistency~\citenumber{wangSelfConsistencyImprovesChain2023};\\ 
    ORM~\citenumber{cobbe2021trainingverifierssolvemath};\\ 
    PRM800K~\citenumber{lightman2023letsverifystepstep};\\
    Math-shepherd~\citenumber{wangMathShepherdVerifyReinforce2024b};\\ 
    \iterative{STaR}~\citenumber{zelikmanSTaRBootstrappingReasoning2022};\\
    \iterative{V-STaR}~\citenumber{hosseiniVSTaRTrainingVerifiers2024b};\\ 
    \iterative{IRPO}~\citenumber{pangIterativeReasoningPreference2024a};\\
    \iterative{Process-Self-Rewarding}~\citenumber{zhang2025processbasedselfrewardinglanguagemodels} } & 
     \makecell[l]
    {GPT-f~\citenumber{polu2020generativelanguagemodelingautomated};
    HTPS~\citenumber{lample2022hypertreeproofsearchneural};\\
    BFS-Prover~\citenumber{xin2025bfsproverscalablebestfirsttree};
    ToT~\citenumber{yao2023treethoughtsdeliberateproblem};\\
    MindStar~\citenumber{kangMindStarEnhancingMath2024};
    RAP~\citenumber{haoReasoningLanguageModel2023};\\ 
    $\text{Q}^*$~\citenumber{wangImprovingMultistepReasoning2024b};
    Self-Evaluation Guided~\citenumber{xieSelfEvaluationGuidedBeam};\\
    \iterative{TS-LLM}~\citenumber{fengAlphazerolikeTreeSearchCan2024};
    LiteSearch~\citenumber{wang2024litesearchefficacioustreesearch};\\
    \iterative{ALPHALLM}~\citenumber{tianSelfImprovementLLMsImagination2024b};
    \iterative{AlphaMath}~\citenumber{chen2024alphamathzeroprocesssupervision};\\
    MCTSr~\citenumber{zhangAccessingGPT4Level2024a};
    \iterative{$\text{ReST-MCTS}^*$}~\citenumber{zhangReSTMCTSLLMSelfTraining2024a};\\
    REBASE~\citenumber{wuInferenceScalingLaws2024};
    \iterative{rStar-Math}~\citenumber{guanRStarMathSmallLLMs2025};\\
    LLaMA-Berry~\citenumber{zhang2024llamaberrypairwiseoptimizationo1like}} & 
    \makecell[l]{RISE~\citenumber{qu2024recursiveintrospectionteachinglanguage};\\ 
    SCoRe~\citenumber{kumar2024traininglanguagemodelsselfcorrect};\\
    AutoMathCritique~\citenumber{xiEnhancingLLMReasoning2024}}
     & 
     \makecell[l]{\rl{Openai o1}~\citenumber{openo1};
     \sft{O1 Journey–Part1}~\citenumber{o1journey};\\ 
     \sft{O1 Journey-Part2}~\citenumber{o1journeypart2};
     \sft{STILL-2}~\citenumber{min2024imitateexploreselfimprovereproduction};\\
     \rl{T1}~\citenumber{hou2025advancinglanguagemodelreasoning};
     \rl{Deepseek-R1}~\citenumber{deepseekai2025deepseekr1incentivizingreasoningcapability};\\
     \rl{Kimi k1.5}~\citenumber{MoonshotAI};
     \rl{SimpleRL}~\citenumber{zeng2025simplerl};\\
     \sft{S1}~\citenumber{muennighoff2025s1simpletesttimescaling};
     \sft{LIMO}~\citenumber{ye2025limoreasoning};\\
     \rl{Demystifying}~\citenumber{yeo2025demystifyinglongchainofthoughtreasoning};
     \rl{LIMR}~\citenumber{li2025limrrlscaling};\\
     \rl{DeepScaleR}~\citenumber{deepscaler2025};
     \rl{QwQ}~\citenumber{qwq32b};\\
     \rl{DAPO}~\citenumber{yu2025dapoopensourcellmreinforcement};
     \rl{\textit{Eurus-2-7B-PRIME}}~\citenumber{cui2024process};\\
     \rl{STILL-3}~\citenumber{chen2025empiricalstudyelicitingimproving};
     \rl{Open-Reasoner-Zero}~\citenumber{OpenReasonerZero2025};\\
     \rl{VAPO}~\citenumber{yue2025vapoefficientreliablereinforcement};
     \rl{Open-RS}~\citenumber{dang2025reinforcementlearningreasoningsmall}}\\
    \midrule
    \textbf{Code}
    & \makecell[l]{MBR-EXEC~\citenumber{shi2022naturallanguagecodetranslation};\\CodeT~\citenumber{chenCodeTCodeGeneration2022};\\S$^*$~\citenumber{li2025s};\\AlphaCode~\citenumber{liCompetitionLevelCodeGeneration2022};\\AlphaCode2~\citenumber{AlphaCode2Technical}}
    & \makecell[l]{PG-TD~\citenumber{zhangPlanningLargeLanguage2023}; o1-Coder~\citenumber{zhang2024o1codero1replicationcoding};\\$\text{Q}^*$~\citenumber{wangImprovingMultistepReasoning2024b}; SWE-Reasoner~\citenumber{ma2025thinkinglongerlargerenhancing}; \\PLANSEARCH~\citenumber{wangPlanningNaturalLanguage2024};\\RethinkMCTS~\citenumber{li2024rethinkmctsrefiningerroneousthoughts};\\SRA-MCTS~\citenumber{xu2024sramctsselfdrivenreasoningaugmentation}}
    & \makecell[l]{Reflexion~\citenumber{shinnReflexionLanguageAgents2023a};\\Self-Debug~\citenumber{chen2023teachinglargelanguagemodels};\\CRITIC~\citenumber{gou2024criticlargelanguagemodels};\\IHR~\citenumber{qiuPHENOMENALPUZZLINGTESTING2024};
    \\Self-Repair~\citenumber{olausson2024selfrepairsilverbulletcode};
    \\STOP~\citenumber{zelikman2024self}}
    & \makecell[l]{
    \rl{Deepseek-R1}~\citenumber{deepseekai2025deepseekr1incentivizingreasoningcapability}; \rl{Kimi k1.5}~\citenumber{MoonshotAI};\\
    \rl{SWE-RL}~\citenumber{wei2025swerladvancingllmreasoning}; \rl{SWE-Gym}~\citenumber{pan2024training};\\
    \rl{OpenAI o1}~\citenumber{openai_o1_system_card}; \rl{QwQ-preview}~\citenumber{qwq-32b-preview};\\ 
    \rl{QwQ}~\citenumber{qwq32b}; \sft{SWE-Reasoner}~\citenumber{ma2025thinkinglongerlargerenhancing};\\
    \sft{OpenCodeReasoning}~\citenumber{ahmad2025opencodereasoning}; \rl{ToRL}~\citenumber{li2025torlscalingtoolintegratedrl}; \\
    \rl{DeepCoder}~\citenumber{deepcoder2025}; \rl{Seed-Thinking-v1.5}~\citenumber{seedthinking2025}
    }
    \\
    \midrule
    \textbf{Multimodality}  
    & \makecell[l]{URSA~\citenumber{luo2025ursa};\\\iterative{PARM}~\citenumber{guo2025can}}
    & \makecell[l]{VisVM~\citenumber{visvm};\\LLaVA-CoT~\citenumber{llavacot};\\
    Mulberry~\citenumber{yao2024mulberry};\\LlamaV-o1~\citenumber{llamavo1};\\Video-T1~\citenumber{liu2025video}}
    & \makecell[l]{\iterative{$R^3V$}~\citenumber{r3v};\\\iterative{Insight-V}~\citenumber{insightv};\\\iterative{PARM}~\citenumber{guo2025can};\\GoT~\citenumber{fang2025got}}
    & \makecell[l]{\sft{\textit{MAmmoTH-VL}}~\citenumber{guo2024mammoth}; \sft{Virgo}~\citenumber{du2025virgo};\\\rl{QVQ-72B-Preview~\citenumber{qvq-72b-preview}};\\\rl{Open-R1-Multimodal}~\citenumber{open-r1-multimodal};
    \\\rl{R1-Multimodal-Journey}~\citenumber{r1-multimodal-journey};
    \rl{R1V}~\citenumber{chen2025r1v};\\\rl{LMM-R1}~\citenumber{peng2025lmmr1}; \rl{VLM-R1}~\citenumber{shen2025vlmr1};\\\rl{R1-Video}~\citenumber{wang-2025-open-r1-video}; \sft{R1-Onevision}~\citenumber{yang2025r1onevisionadvancinggeneralizedmultimodal};\\\rl{MM-Eureka}~\citenumber{meng2025mm}; \rl{Vision-R1}~\citenumber{huang2025vision};\\\rl{VisualThinker-R1-Zero}~\citenumber{zhou2025r1zerosahamomentvisual}; \rl{MAYE}~\citenumber{MAYE};\\\rl{Visual-RFT}~\citenumber{shen2025vlmr1}; \rl{Seg-Zero}~\citenumber{liu2025seg};\\\rl{vsGRPO}~\citenumber{liao2025improved}; \rl{Video-R1}~\citenumber{feng2025video}; \sft{\textit{MVoT}}~\citenumber{li2025imagine};\\\rl{Kimi-VL}~\citenumber{kimi2024vl}; \rl{Kimi k1.5}~\citenumber{MoonshotAI};\\ \rl{O3/O4-mini}~\citenumber{o3_o4_mini,thinking_with_images}}  \\
    \midrule
    \textbf{Agent} 
    & 
    \makecell[c]{-}
    & \makecell[l]{Agent Q~\citenumber{putta2024agentqadvancedreasoning};\\ToT~\citenumber{yao2023treethoughtsdeliberateproblem};\\SearchAgent~\citenumber{koh2024treesearchlanguagemodel}}
    & \makecell[l]{Reflexion~\citenumber{shinnReflexionLanguageAgents2023a};\\ \iterative{Agent Q}~\citenumber{putta2024agentqadvancedreasoning};\\ \iterative{Agent-Eval-Refine}~\citenumber{pan2024autonomousevaluationrefinementdigital}}
    & \makecell[l]{\textit{ReAct}~\citenumber{yao2023reactsynergizingreasoningacting};\\ \rl{Deep Research}~\citenumber{deepresearch};\\ \rl{SWE-RL}~\citenumber{wei2025swerladvancingllmreasoning}; \rl{\textit{Operator}}~\citenumber{cuaopenai};\\ \sft{\textit{UI-TRARS}}~\citenumber{qin2025uitarspioneeringautomatedgui}; \textit{\sft{PC Agent}}~\citenumber{he2024pcagentsleepai};\\\rl{DeepResearcher}~\citenumber{zheng2025deepresearcherscalingdeepresearch}; \\\rl{\textit{SWEET-RL}}~\citenumber{zhou2025sweetrltrainingmultiturnllm}; \rl{Claude 3.7 Sonnet}~\citenumber{claude3.7sonnet}}
    
    \\
    \midrule
    \textbf{Embodied AI} &  \makecell[c]{-} &  \makecell[c]{-} & \makecell[l]{Inner Monologue~\citenumber{huang2022inner};\\ REFLECT~\citenumber{liu2023reflect};\\KnowNo~\citenumber{renrobots}} & \makecell[l]{\sft{\textit{Embodied-CoT}}~\citenumber{michal2024robotic}; \sft{\textit{CoA}}~\citenumber{li2024improving};\\ \sft{\textit{SpatialCoT}}~\citenumber{liu2025spatialcot}; \sft{\textit{RAD}}~\citenumber{clark2025action};\\ \rl{Cosmos-Reason1}~\citenumber{azzolini2025cosmos};\\ \sft{\textit{Gemini Robotics}}~\citenumber{team2025gemini}; \sft{CoT-VLA}~\citenumber{zhao2025cot};\\ \sft{Embodied-Reasoner}~\citenumber{zhang2025embodied}} \\
    \midrule
    \textbf{Safety} 
     & \makecell[l]
     {SelfCheckGPT~\citenumber{manakul2023selfcheckgpt};\\SRG~\citenumber{wang2025SRG}}
    & \makecell[l]{\iterative{STAIR~\citenumber{zhang2025stair}};\\C-MCTS~\citenumber{dinesh2023cmcts};\\\iterative{HaluSearch}~\citenumber{cheng2025halusearch};\\{InferenceGuard}~\citenumber{ji2025inferenceguard};\\{ARGS}~\citenumber{khanov2024args}}
    & \makecell[l]{MART~\citenumber{ge2024mart};\\Combat Adv. Attacks~\citenumber{chern2024combatadversarial};\\Improve Factuality~\citenumber{du2023improvingfactualityreasoninglanguage};\\{Multi-expert Prompting}~\citenumber{long2024multiexpert};\\DebateGPT~\citenumber{subramaniam2024debategpt}}
    & \makecell[l]{\rl{Deliberate Alignment}~\citenumber{guan2025deliberate};\\\sft{SafeChain}~\citenumber{jiang2025safechainsafetylanguagemodels};\\{Chain-of-Verification}~\citenumber{dhuliawala2023chainofverificationreduceshallucinationlarge};\\\sft{MoTE}~\citenumber{liu2024mixture}}
    \\
    \midrule
    \textbf{RAG}  & CoRAG~\citenumber{wang2025chainofretrievalaugmentedgeneration} & \makecell[l]{AirRAG~\citenumber{feng2025airragactivatingintrinsicreasoning}; \\CoRAG~\citenumber{wang2025chainofretrievalaugmentedgeneration}}  &  \makecell[c]{-} & \makecell[l]{\textit{IterDRAG}~\citenumber{yue2024inferencescalinglongcontextretrieval}; \textit{Plan*RAG}~\citenumber{verma2025planragefficienttesttimeplanning};\\ \rl{DeepRAG}~\citenumber{guan2025deepragthinkingretrievalstep}; Search-o1~\citenumber{li2025searcho1agenticsearchenhancedlarge};\\AirRAG~\citenumber{feng2025airragactivatingintrinsicreasoning}; \sft{\textit{Auto-RAG}}~\citenumber{yu2024autoragautonomousretrievalaugmentedgeneration};\\ \sft{CoRAG}~\citenumber{wang2025chainofretrievalaugmentedgeneration}; \rl{Search-R1}~\citenumber{jin2025searchr1trainingllmsreason};\\\rl{R1-Searcher}~\citenumber{song2025r1}; \rl{ReSearch}~\citenumber{chen2025research};\\\rl{DeepRetrieval}~\citenumber{jiang2025deepretrievalhackingrealsearch}} \\
    \midrule
    \textbf{Evaluation} & \makecell[l]{CCE~\citenumber{zhang2025crowdcomparativereasoningunlocking};\\ \iterative{SPCT}~\citenumber{liu2025inferencetimescalinggeneralistreward}} & MCTS-Judge~\citenumber{wang2025mctsjudgetesttimescalingllmasajudge} &  
    \makecell[l]{ChatEval~\citenumber{chan2023chateval}; \\ScaleEval~\citenumber{chern2024scaleeval}} & 
    \makecell[l]{\textit{FActScore}~\citenumber{min2023factscorefinegrainedatomicevaluation}; \textit{FacTool}~\citenumber{chern2023factoolfactualitydetectiongenerative};\\\textit{RefChecker}~\citenumber{hu-etal-2024-knowledge}; \textit{RAGChecker}~\citenumber{ru2024ragchecker};\\\textit{Agent-as-a-Judge}~\citenumber{zhuge2024agentasajudgeevaluateagentsagents}; \textit{EvalPlanner}~\citenumber{saha2025learningplanreason};\\ Kim et al.~\citenumber{kim2025scalingevaluationtimecomputereasoning}} \\
    \bottomrule
  \end{tabular}
  
\end{table}
\section{How’s Progress – Application So Far} \label{sec:application}

In this section, we examine the systemic changes in AI research that cognition engineering driven by test-time scaling brings and the applications that have already emerged.

\subsection{Mathematics}
\definecolor{parallelColor}{RGB}{65, 105, 225}  
\definecolor{treeSearchColor}{RGB}{34, 139, 34}  
\definecolor{correctionColor}{RGB}{220, 20, 60}  
\definecolor{longCoTColor}{RGB}{148, 0, 211}  
\definecolor{islColor}{RGB}{0, 0, 0}  

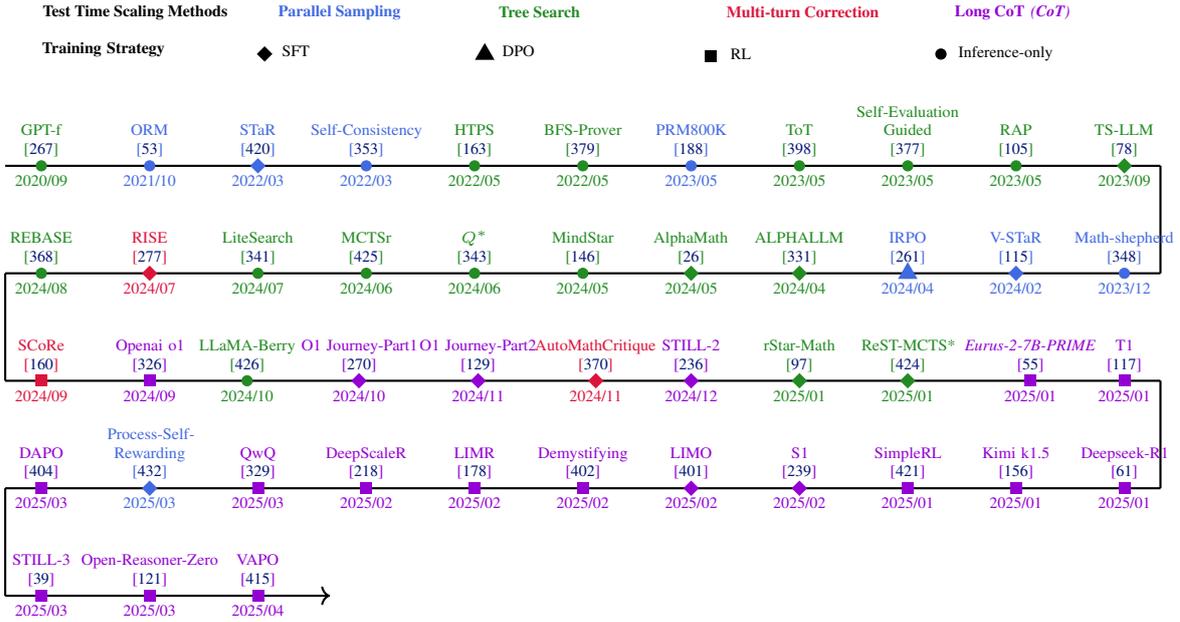
\begin{figure}[ht]
\centering
\begin{tikzpicture}[scale=0.95]
    \def\rowspacing{1.5} 
    \def\paperspacing{1.5} 
    \def\papersperrow{11} 
    
    \def\startx{0.5} 
    \def\endx{\startx + 10*\paperspacing} 
    
    \node[anchor=north east] at (\endx-0.5,10.5) {
        \begin{tikzpicture}
            \node[font=\tiny, anchor=west] at (-7,0) {\textbf{Test Time Scaling Methods}};
            
            \node[right, font=\tiny, text=parallelColor] at (-3.9,0) {\textbf{Parallel Sampling}};
            
            \node[right, font=\tiny, text=treeSearchColor] at (-1,0) {\textbf{Tree Search}};
            
            \node[right, font=\tiny, text=correctionColor] at (2,0) {\textbf{Multi-turn Correction}};
            
            \node[right, font=\tiny, text=longCoTColor] at (5,0) {\textbf{Long CoT}};

            \node[right, font=\tiny, text=longCoTColor] at (6,0) {\textit{(\textbf{CoT})}};
            
            \node[font=\tiny, anchor=west, text width=3cm] at (-7,-0.5) {\textbf{Training Strategy}};
            
            \node[diamond, fill=black, inner sep=1.5pt] at (-3.9,-0.5) {};
            \node[right, font=\tiny] at (-3.85,-0.52) {SFT};
            
            \node[regular polygon, regular polygon sides=3, fill=black, inner sep=1.5pt] at (-1,-0.5) {};
            \node[right, font=\tiny] at (-0.95,-0.52) {DPO};
            
            \node[regular polygon, regular polygon sides=4, fill=black, inner sep=1.5pt] at (2,-0.5) {};
            \node[right, font=\tiny] at (2.05,-0.55) {RL};

            \node[circle, regular polygon sides=4, fill=black, inner sep=1.5pt] at (5,-0.5) {};
            \node[right, font=\tiny] at (5.05,-0.55) {Inference-only};
            
        \end{tikzpicture}
    };
    
    \draw[thick] (0,8) -- (\endx+0.5,8);
    
    \draw[thick] (\endx+0.5,8) -- (\endx+0.5,8-\rowspacing);
    
    \draw[thick] (\endx+0.5,8-\rowspacing) -- (0,8-\rowspacing);
    
    \draw[thick] (0,8-\rowspacing) -- (0,8-2*\rowspacing);
    
    \draw[thick] (0,8-2*\rowspacing) -- (\endx+0.5,8-2*\rowspacing);
    
    \draw[thick] (\endx+0.5,8-2*\rowspacing) -- (\endx+0.5,8-3*\rowspacing);
    
    \draw[thick] (\endx+0.5,8-3*\rowspacing) -- (0,8-3*\rowspacing);

    \draw[thick] (0,8 -3*\rowspacing) -- (0,8-4*\rowspacing);
    \draw[thick,->] (0,8-4*\rowspacing) -- (3*\paperspacing,8-4*\rowspacing);
    
    
    \filldraw[treeSearchColor] (\startx,8) circle (2pt) 
        node[above, text width=1.8cm, align=center, font=\tiny] {GPT-f \\\citenumber{polu2020generativelanguagemodelingautomated}}
        node[below, font=\tiny] {2020/09};
    
    \filldraw[parallelColor] (\startx+\paperspacing,8) circle (2pt) 
        node[above, text width=1.8cm, align=center, font=\tiny] {ORM \\\citenumber{cobbe2021trainingverifierssolvemath}}
        node[below, font=\tiny] {2021/10};
    
    \filldraw[parallelColor] (\startx+2*\paperspacing,8) node[diamond, fill=parallelColor, inner sep=1.5pt] {} 
        node[above, text width=1.8cm, align=center, font=\tiny] {STaR \\\citenumber{zelikmanSTaRBootstrappingReasoning2022}}
        node[below, font=\tiny] {2022/03};
    
    \filldraw[parallelColor] (\startx+3*\paperspacing,8) circle (2pt) 
        node[above, text width=1.8cm, align=center, font=\tiny] {Self-Consistency \\\citenumber{wangSelfConsistencyImprovesChain2023}}
        node[below, font=\tiny] {2022/03};
    
    \filldraw[treeSearchColor] (\startx+4*\paperspacing,8) circle (2pt) 
        node[above, text width=1.8cm, align=center, font=\tiny] {HTPS \\\citenumber{lample2022hypertreeproofsearchneural}}
        node[below, font=\tiny] {2022/05};
    
    \filldraw[treeSearchColor] (\startx+5*\paperspacing,8) circle (2pt) 
        node[above, text width=1.8cm, align=center, font=\tiny] {BFS-Prover \\\citenumber{xin2025bfsproverscalablebestfirsttree}}
        node[below, font=\tiny] {2022/05};
    
    \filldraw[parallelColor] (\startx+6*\paperspacing,8) circle (2pt) 
        node[above, text width=1.8cm, align=center, font=\tiny] {PRM800K \\\citenumber{lightman2023letsverifystepstep}}
        node[below, font=\tiny] {2023/05};
    
    \filldraw[treeSearchColor] (\startx+7*\paperspacing,8) circle (2pt) 
        node[above, text width=1.8cm, align=center, font=\tiny] {ToT \\\citenumber{yao2023treethoughtsdeliberateproblem}}
        node[below, font=\tiny] {2023/05};
    
    \filldraw[treeSearchColor] (\startx+8*\paperspacing,8) circle (2pt) 
        node[above, text width=1.8cm, align=center, font=\tiny] {Self-Evaluation Guided \\\citenumber{xieSelfEvaluationGuidedBeam}}
        node[below, font=\tiny] {2023/05};
    
    \filldraw[treeSearchColor] (\startx+9*\paperspacing,8) circle (2pt) 
        node[above, text width=1.8cm, align=center, font=\tiny] {RAP \\\citenumber{haoReasoningLanguageModel2023}}
        node[below, font=\tiny] {2023/05};
    
    \filldraw[treeSearchColor] (\startx+10*\paperspacing,8) node[diamond, fill=treeSearchColor, inner sep=1.5pt] {} 
        node[above, text width=1.8cm, align=center, font=\tiny] {TS-LLM \\\citenumber{fengAlphazerolikeTreeSearchCan2024}}
        node[below, font=\tiny] {2023/09};
    
    
    \filldraw[parallelColor] (\endx,8-\rowspacing) circle (2pt) 
        node[above, text width=1.8cm, align=center, font=\tiny] {Math-shepherd \\\citenumber{wangMathShepherdVerifyReinforce2024b}}
        node[below, font=\tiny] {2023/12};
    
    \filldraw[parallelColor] (\endx-\paperspacing,8-\rowspacing) node[diamond, fill=parallelColor, inner sep=1.5pt] {} 
        node[above, text width=1.8cm, align=center, font=\tiny] {V-STaR \\\citenumber{hosseiniVSTaRTrainingVerifiers2024b}}
        node[below, font=\tiny] {2024/02};
    
    \filldraw[parallelColor] (\endx-2*\paperspacing,8-\rowspacing) node[regular polygon, regular polygon sides=3, fill=parallelColor, inner sep=1.5pt] {} 
        node[above, text width=1.8cm, align=center, font=\tiny] {IRPO \\\citenumber{pangIterativeReasoningPreference2024a}}
        node[below, font=\tiny] {2024/04};
    
    \filldraw[treeSearchColor] (\endx-3*\paperspacing,8-\rowspacing) node[diamond, fill=treeSearchColor, inner sep=1.5pt] {} 
        node[above, text width=1.8cm, align=center, font=\tiny] {ALPHALLM \\\citenumber{tianSelfImprovementLLMsImagination2024b}}
        node[below, font=\tiny] {2024/04};
    
    \filldraw[treeSearchColor] (\endx-4*\paperspacing,8-\rowspacing) node[diamond, fill=treeSearchColor, inner sep=1.5pt] {} 
        node[above, text width=1.8cm, align=center, font=\tiny] {AlphaMath \\\citenumber{chen2024alphamathzeroprocesssupervision}}
        node[below, font=\tiny] {2024/05};
    
    \filldraw[treeSearchColor] (\endx-5*\paperspacing,8-\rowspacing) circle (2pt) 
        node[above, text width=1.8cm, align=center, font=\tiny] {MindStar \\\citenumber{kangMindStarEnhancingMath2024}}
        node[below, font=\tiny] {2024/05};
    
    \filldraw[treeSearchColor] (\endx-6*\paperspacing,8-\rowspacing) circle (2pt) 
        node[above, text width=1.8cm, align=center, font=\tiny] {$Q^*$ \\\citenumber{wangImprovingMultistepReasoning2024b}}
        node[below, font=\tiny] {2024/06};
    
    \filldraw[treeSearchColor] (\endx-7*\paperspacing,8-\rowspacing) circle (2pt) 
        node[above, text width=1.8cm, align=center, font=\tiny] {MCTSr \\\citenumber{zhangAccessingGPT4Level2024a}}
        node[below, font=\tiny] {2024/06};
    
    \filldraw[treeSearchColor] (\endx-8*\paperspacing,8-\rowspacing) circle (2pt) 
        node[above, text width=1.8cm, align=center, font=\tiny] {LiteSearch \\\citenumber{wang2024litesearchefficacioustreesearch}}
        node[below, font=\tiny] {2024/07};
    
    \filldraw[correctionColor] (\endx-9*\paperspacing,8-\rowspacing) node[diamond, fill=correctionColor, inner sep=1.5pt] {}
        node[above, text width=1.8cm, align=center, font=\tiny] {RISE \\\citenumber{qu2024recursiveintrospectionteachinglanguage}}
        node[below, font=\tiny] {2024/07};
    
    \filldraw[treeSearchColor] (\endx-10*\paperspacing,8-\rowspacing) circle (2pt) 
        node[above, text width=1.8cm, align=center, font=\tiny] {REBASE \\\citenumber{wuInferenceScalingLaws2024}}
        node[below, font=\tiny] {2024/08};
    
    
    \filldraw[correctionColor] (\startx,8-2*\rowspacing) node[regular polygon, regular polygon sides=4, fill=correctionColor, inner sep=1.5pt]{}
        node[above, text width=1.8cm, align=center, font=\tiny] {SCoRe \\\citenumber{kumar2024traininglanguagemodelsselfcorrect}}
        node[below, font=\tiny] {2024/09};

    \filldraw[longCoTColor] (\startx +\paperspacing,8-2*\rowspacing) node[regular polygon, regular polygon sides=4, fill=longCoTColor, inner sep=1.5pt]{}
        node[above, text width=1.8cm, align=center, font=\tiny] {Openai o1 \\\citenumber{openo1}}
        node[below, font=\tiny] {2024/09};

    \filldraw[treeSearchColor] (\startx+2*\paperspacing -0.15,8-2*\rowspacing) circle (2pt) 
        node[above, text width=1.8cm, align=center, font=\tiny] {LLaMA-Berry \\\citenumber{zhang2024llamaberrypairwiseoptimizationo1like}}
        node[below, font=\tiny] {2024/10};
    
    \filldraw[longCoTColor] (\startx+3*\paperspacing - 0.1,8-2*\rowspacing) node[diamond, fill=longCoTColor, inner sep=1.5pt] {} 
        node[above, text width=1.8cm, align=center, font=\tiny] {O1 Journey-Part1 \\\citenumber{o1journey}}
        node[below, font=\tiny] {2024/10};
    
    \filldraw[longCoTColor] (\startx+4*\paperspacing+0.05 ,8-2*\rowspacing) node[diamond, fill=longCoTColor, inner sep=1.5pt] {}  
        node[above, text width=1.8cm, align=center, font=\tiny] {O1 Journey-Part2 \\\citenumber{o1journeypart2}}
        node[below, font=\tiny] {2024/11};
    
    \filldraw[correctionColor] (\startx+5*\paperspacing + 0.18,8-2*\rowspacing) node[diamond, fill=correctionColor, inner sep=1.5pt] {}
        node[above, text width=1.8cm, align=center, font=\tiny] {AutoMathCritique \\\citenumber{xiEnhancingLLMReasoning2024}}
        node[below, font=\tiny] {2024/11};
    
    \filldraw[longCoTColor] (\startx+6*\paperspacing,8-2*\rowspacing) node[diamond, fill=longCoTColor, inner sep=1.5pt] {}  
        node[above, text width=1.8cm, align=center, font=\tiny] {STILL-2 \\\citenumber{min2024imitateexploreselfimprovereproduction}}
        node[below, font=\tiny] {2024/12};
    
    \filldraw[treeSearchColor] (\startx+7*\paperspacing,8-2*\rowspacing) node[diamond, fill=treeSearchColor, inner sep=1.5pt] {} 
        node[above, text width=1.8cm, align=center, font=\tiny] {rStar-Math \\\citenumber{guanRStarMathSmallLLMs2025}}
        node[below, font=\tiny] {2025/01};
    
    \filldraw[treeSearchColor] (\startx+8*\paperspacing,8-2*\rowspacing) node[diamond, fill=treeSearchColor, inner sep=1.5pt] {} 
        node[above, text width=1.8cm, align=center, font=\tiny] {ReST-MCTS* \\\citenumber{zhangReSTMCTSLLMSelfTraining2024a}}
        node[below, font=\tiny] {2025/01};

    \filldraw[longCoTColor] (\startx+9*\paperspacing + 0.2,8-2*\rowspacing) node[regular polygon, regular polygon sides=4, fill=longCoTColor, inner sep=1.5pt] {} 
        node[above, text width=1.8cm, align=center, font=\tiny] {\textit{Eurus-2-7B-PRIME} \\\citenumber{cui2024process}}
        node[below, font=\tiny] {2025/01};
        
    \filldraw[longCoTColor] (\startx+10*\paperspacing,8-2*\rowspacing) node[regular polygon, regular polygon sides=4, fill=longCoTColor, inner sep=1.5pt] {} 
        node[above, text width=1.8cm, align=center, font=\tiny] {T1 \\\citenumber{hou2025advancinglanguagemodelreasoning}}
        node[below, font=\tiny] {2025/01};
    
    \filldraw[longCoTColor] (\endx ,8-3*\rowspacing) node[regular polygon, regular polygon sides=4, fill=longCoTColor, inner sep=1.5pt] {} 
        node[above, text width=1.8cm, align=center, font=\tiny] {Deepseek-R1 \\\citenumber{deepseekai2025deepseekr1incentivizingreasoningcapability}}
        node[below, font=\tiny] {2025/01};
    
    \filldraw[longCoTColor] (\endx -\paperspacing,8-3*\rowspacing) node[regular polygon, regular polygon sides=4, fill=longCoTColor, inner sep=1.5pt] {} 
        node[above, text width=1.8cm, align=center, font=\tiny] {Kimi k1.5 \\\citenumber{MoonshotAI}}
        node[below, font=\tiny] {2025/01};
    
    
    \filldraw[longCoTColor] (\endx -2*\paperspacing,8-3*\rowspacing) node[regular polygon, regular polygon sides=4, fill=longCoTColor, inner sep=1.5pt] {} 
        node[above, text width=1.8cm, align=center, font=\tiny] {SimpleRL \\\citenumber{zeng2025simplerl}}
        node[below, font=\tiny] {2025/01};
    
    \filldraw[longCoTColor] (\endx -3*\paperspacing,8-3*\rowspacing) node[diamond, fill=longCoTColor, inner sep=1.5pt] {}  
        node[above, text width=1.8cm, align=center, font=\tiny] {S1 \\\citenumber{muennighoff2025s1simpletesttimescaling}}
        node[below, font=\tiny] {2025/02};
    
    \filldraw[longCoTColor] (\endx-4* \paperspacing,8-3*\rowspacing) node[diamond, fill=longCoTColor, inner sep=1.5pt] {} 
        node[above, text width=1.8cm, align=center, font=\tiny] {LIMO \\\citenumber{ye2025limoreasoning}}
        node[below, font=\tiny] {2025/02};

    \filldraw[longCoTColor] (\endx-5*\paperspacing,8-3*\rowspacing) node[regular polygon, regular polygon sides=4, fill=longCoTColor, inner sep=1.5pt] {} 
    node[above, text width=1.8cm, align=center, font=\tiny] {Demystifying \\\citenumber{yeo2025demystifyinglongchainofthoughtreasoning}}
    node[below, font=\tiny] {2025/02};

    \filldraw[longCoTColor] (\endx-6*\paperspacing,8-3*\rowspacing) node[regular polygon, regular polygon sides=4, fill=longCoTColor, inner sep=1.5pt] {} 
    node[above, text width=1.8cm, align=center, font=\tiny] {LIMR \\\citenumber{li2025limrrlscaling}}
    node[below, font=\tiny] {2025/02};

    \filldraw[longCoTColor] (\endx-7*\paperspacing,8-3*\rowspacing) node[regular polygon, regular polygon sides=4, fill=longCoTColor, inner sep=1.5pt] {} 
        node[above, text width=1.8cm, align=center, font=\tiny] {DeepScaleR \\\citenumber{deepscaler2025}}
        node[below, font=\tiny] {2025/02};

    \filldraw[longCoTColor] (\endx-8*\paperspacing,8-3*\rowspacing) node[regular polygon, regular polygon sides=4, fill=longCoTColor, inner sep=1.5pt] {} 
        node[above, text width=1.8cm, align=center, font=\tiny] {QwQ \\\citenumber{qwq32b}}
        node[below, font=\tiny] {2025/03};
        
    \filldraw[parallelColor] (\endx-9*\paperspacing,8-3*\rowspacing) node[diamond, fill=parallelColor, inner sep=1.5pt] {} 
        node[above, text width=1.8cm, align=center, font=\tiny] {Process-Self-Rewarding \\\citenumber{zhang2025processbasedselfrewardinglanguagemodels}}
        node[below, font=\tiny] {2025/03};
    
    \filldraw[longCoTColor] (\endx-10*\paperspacing,8-3*\rowspacing) node[regular polygon, regular polygon sides=4, fill=longCoTColor, inner sep=1.5pt] {} 
        node[above, text width=1.8cm, align=center, font=\tiny] {DAPO \\\citenumber{yu2025dapoopensourcellmreinforcement}}
        node[below, font=\tiny] {2025/03};


    \filldraw[longCoTColor] (\startx,8-4*\rowspacing) node[regular polygon, regular polygon sides=4, fill=longCoTColor, inner sep=1.5pt] {} 
        node[above, text width=1.8cm, align=center, font=\tiny] {STILL-3 \\\citenumber{chen2025empiricalstudyelicitingimproving}}
        node[below, font=\tiny] {2025/03};

    \filldraw[longCoTColor] (\startx + \paperspacing,8-4*\rowspacing) node[regular polygon, regular polygon sides=4, fill=longCoTColor, inner sep=1.5pt] {} 
        node[above, text width=1.8cm, align=center, font=\tiny] {Open-Reasoner-Zero \\\citenumber{OpenReasonerZero2025}}
        node[below, font=\tiny] {2025/03};

    \filldraw[longCoTColor] (\startx + 2*\paperspacing,8-4*\rowspacing) node[regular polygon, regular polygon sides=4, fill=longCoTColor, inner sep=1.5pt]{}
    node[above, text width=1.8cm, align=center, font=\tiny] {VAPO \\\citenumber{yue2025vapoefficientreliablereinforcement}}
    node[below, font=\tiny] {2025/04};

\end{tikzpicture}
\caption{Works of applying test-time scaling methods in the math field.}
\label{fig:timeline-wrapped-math}
\end{figure}
Mathematical reasoning is crucial for resolving complex problems and making informed decisions~\cite{hendrycks2021measuringmathematicalproblemsolving,xia2025evaluatingmathematicalreasoningaccuracy}. Research in AI for mathematics (AI4Math) has developed along two complementary paths: natural language reasoning focusing on questions with verifiable answers, and formal language reasoning utilizing formal systems like Lean~\cite{de2015lean} and Isabelle~\cite{nipkow2002isabelle} for automatic formal theorem proving. For questions with verifiable answers, the easy verification characteristic makes it reliable to construct feedback signals for search and learning, facilitating the wide application of test-time scaling methods to enhance reasoning abilities. These include parallel sampling~\cite{cobbe2021trainingverifierssolvemath,wangSelfConsistencyImprovesChain2023}, tree search~\cite{fengAlphazerolikeTreeSearchCan2024,chen2024alphamathzeroprocesssupervision,haoReasoningLanguageModel2023}, multi-turn correction~\cite{kumar2024traininglanguagemodelsselfcorrect,qu2024recursiveintrospectionteachinglanguage}, and long CoT~\cite{deepseekai2025deepseekr1incentivizingreasoningcapability,openai_o1_system_card}. Notably, powered by long CoT, DeepSeek-R1 achieves a score of 79.8 on the American Invitational Mathematics Examination (AIME), significantly outperforming traditional models without long CoT and approaching competitive human performance. For formal language reasoning, formal systems make the reasoning process verifiable and provide signals for tree search~\cite{polu2020generativelanguagemodelingautomated,lample2022hypertreeproofsearchneural,xin2024deepseekproverv15harnessingproofassistant,xin2025bfsproverscalablebestfirsttree} or multi-turn correction~\cite{first2023baldurwholeproofgenerationrepair}. Breakthrough systems like AlphaProof~\cite{alphaproof} and AlphaGeometry~\cite{alphageometry} demonstrate that combining neural networks with formal methods and proof checkers can achieve unprecedented mathematical reasoning abilities.

Despite these successes, there still exists room for improvement. In natural language reasoning, while training data accumulation is substantial, the difficulty in strictly verifying reasoning process correctness means solutions generated by LLMs may contain logical errors or lack rigor in intermediate steps~\cite{lightman2023letsverifystepstep,xia2025evaluatingmathematicalreasoningaccuracy}. For formal language reasoning, while it ensures reasoning process verifiability, the lack of training data compared to natural language limits its development. Future work can focus on unifying the advantages of formal and natural languages for more robust model development. Moreover, although LLMs with strong reasoning and cognition abilities have made progress on exam problems and competition tasks, applications to more advanced domains such as mathematical research remain relatively unexplored~\cite{yang2024formalmathematicalreasoningnew}. This necessitates not merely enhanced model capabilities but also novel evaluation frameworks to assess these competencies.

\begin{AIbox}{Future Direction for Mathematics}
\begin{itemize}
    \item Unify the advantages of formal and natural languages to develop more robust reasoning models that combine the verifiability of formal systems with the rich training data available in natural language.
    \item Expand applications of LLMs with strong reasoning capabilities beyond exam problems toward more advanced domains such as mathematical research, developing novel evaluation frameworks to assess these higher-level competencies.
\end{itemize}
\end{AIbox}

\subsection{Code}
\definecolor{parallelColor}{RGB}{65, 105, 225}  
\definecolor{treeSearchColor}{RGB}{34, 139, 34}  
\definecolor{correctionColor}{RGB}{220, 20, 60}  
\definecolor{longCoTColor}{RGB}{148, 0, 211}  
\definecolor{islColor}{RGB}{0, 0, 0}  

\begin{figure}[h]
\centering
\begin{tikzpicture}[scale=0.95]
    \def\rowspacing{1.5} 
    \def\paperspacing{1.5} 
    \def\papersperrow{11} 
    
    \def\startx{0.5} 
    \def\endx{\startx + 10*\paperspacing} 
    
    \node[anchor=north east] at (\endx-0.5,10.5) {
       \begin{tikzpicture}
            \node[font=\tiny, anchor=west] at (-7,0) {\textbf{Test Time Scaling Methods}};
            
            \node[right, font=\tiny, text=parallelColor] at (-3.9,0) {\textbf{Parallel Sampling}};
            
            \node[right, font=\tiny, text=treeSearchColor] at (-1,0) {\textbf{Tree Search}};
            
            \node[right, font=\tiny, text=correctionColor] at (2,0) {\textbf{Multi-turn Correction}};
            
            \node[right, font=\tiny, text=longCoTColor] at (5,0) {\textbf{Long CoT}};

            \node[right, font=\tiny, text=longCoTColor] at (6,0) {\textit{(\textbf{CoT})}};
            
            \node[font=\tiny, anchor=west, text width=3cm] at (-7,-0.5) {\textbf{Training Strategy}};
            
            \node[diamond, fill=black, inner sep=1.5pt] at (-3.9,-0.5) {};
            \node[right, font=\tiny] at (-3.85,-0.52) {SFT};
            
            \node[regular polygon, regular polygon sides=3, fill=black, inner sep=1.5pt] at (-1,-0.5) {};
            \node[right, font=\tiny] at (-0.95,-0.52) {DPO};
            
            \node[regular polygon, regular polygon sides=4, fill=black, inner sep=1.5pt] at (2,-0.5) {};
            \node[right, font=\tiny] at (2.05,-0.55) {RL};

            \node[circle, regular polygon sides=4, fill=black, inner sep=1.5pt] at (5,-0.5) {};
            \node[right, font=\tiny] at (5.05,-0.55) {Inference-only};
            
        \end{tikzpicture}
    };
    
    \draw[thick] (0,8) -- (\endx+0.5,8);

    \draw[thick] (\endx+0.5,8) -- (\endx+0.5,8-\rowspacing);

    \draw[thick] (\endx+0.5,8-\rowspacing) -- (0,8-\rowspacing);
    
    \draw[thick] (0,8-\rowspacing) -- (0,8-2*\rowspacing);
    
    \draw[thick,->] (0,8-2*\rowspacing) -- (\endx+0.5-2.5*\paperspacing,8-2*\rowspacing);
    
    
    
    \filldraw[parallelColor] (\startx,8) circle (2pt) 
        node[above, text width=1.8cm, align=center, font=\tiny] {AlphaCode \\\citenumber{liCompetitionLevelCodeGeneration2022}}
        node[below, font=\tiny] {2022/02};

    \filldraw[parallelColor] (\startx +\paperspacing,8) circle (2pt) 
        node[above, text width=1.8cm, align=center, font=\tiny] {MBR-
EXEC \\\citenumber{shi2022naturallanguagecodetranslation}}
        node[below, font=\tiny] {2022/04};

    \filldraw[parallelColor] (\startx+ 2*\paperspacing,8) circle (2pt) 
        node[above, text width=1.8cm, align=center, font=\tiny] {CodeT \\\citenumber{chenCodeTCodeGeneration2022}}
        node[below, font=\tiny] {2022/07};
    \filldraw[treeSearchColor] (\startx+3*\paperspacing,8) circle (2pt) 
        node[above, text width=1.8cm, align=center, font=\tiny] {PG-TD \\\citenumber{zhangPlanningLargeLanguage2023}}
        node[below, font=\tiny] {2023/03};
    \filldraw[correctionColor] (\startx+4*\paperspacing,8) circle (2pt) 
        node[above, text width=1.8cm, align=center, font=\tiny] {Reflexion \\\citenumber{shinnReflexionLanguageAgents2023a}}
        node[below, font=\tiny] {2023/03};
    \filldraw[correctionColor] (\startx+5*\paperspacing,8) circle (2pt) 
        node[above, text width=1.8cm, align=center, font=\tiny] {Self-Debug \\\citenumber{chen2023teachinglargelanguagemodels}}
        node[below, font=\tiny] {2023/04};
    \filldraw[correctionColor] (\startx+6*\paperspacing,8) node[circle, fill=correctionColor, inner sep=1.5pt] {} 
        node[above, text width=1.8cm, align=center, font=\tiny] {CRITIC \\\citenumber{gou2024criticlargelanguagemodels}}
        node[below, font=\tiny] {2023/05};    
    \filldraw[correctionColor] (\startx+7*\paperspacing,8) node[circle, fill=correctionColor, inner sep=1.5pt] {} 
        node[above, text width=1.8cm, align=center, font=\tiny] {Self-Repair \\\citenumber{olausson2024selfrepairsilverbulletcode}}
        node[below, font=\tiny] {2023/07};   
    \filldraw[correctionColor] (\startx+8*\paperspacing,8) node[circle, fill=correctionColor, inner sep=1.5pt] {} 
        node[above, text width=1.8cm, align=center, font=\tiny] {STOP \\\citenumber{zelikman2024self}}
        node[below, font=\tiny] {2023/10};   
    \filldraw[parallelColor] (\startx+9*\paperspacing,8) circle (2pt) 
        node[above, text width=1.8cm, align=center, font=\tiny] {AlphaCode2 \\\citenumber{AlphaCode2Technical}}
        node[below, font=\tiny] {2023/12};
    \filldraw[treeSearchColor] (\startx+10*\paperspacing,8) node[diamond, fill=treeSearchColor, inner sep=1.5pt] {}
        node[above, text width=1.8cm, align=center, font=\tiny] {CRUXEval \\\citenumber{gu2024cruxeval}}
        node[below, font=\tiny] {2024/01};

    \filldraw[treeSearchColor] (\endx-1*\paperspacing,8-\rowspacing) circle (2pt) 
        node[above, text width=1.8cm, align=center, font=\tiny] {Q$^*$ \\\citenumber{wangImprovingMultistepReasoning2024b}}
        node[below, font=\tiny] {2024/06};
    \filldraw[correctionColor] (\endx,8-\rowspacing) node[regular polygon, regular polygon sides=3, fill=correctionColor, inner sep=1.5pt] {} 
        node[above, text width=1.8cm, align=center, font=\tiny] {IHR \\\citenumber{qiuPHENOMENALPUZZLINGTESTING2024}}
        node[below, font=\tiny] {2024/04};
    \filldraw[longCoTColor] (\endx-2*\paperspacing,8-\rowspacing) node[regular polygon, regular polygon sides=4, fill=longCoTColor, inner sep=1.5pt] {} 
        node[above, text width=1.8cm, align=center, font=\tiny] {OpenAI o1 \\\citenumber{openai_o1_system_card}}
        node[below, font=\tiny] {2024/09};
    \filldraw[treeSearchColor] (\endx-3*\paperspacing,8-\rowspacing) circle (2pt) 
        node[above, text width=1.8cm, align=center, font=\tiny] {PLANSEARCH \\\citenumber{wangPlanningNaturalLanguage2024}}
        node[below, font=\tiny] {2024/09};
    \filldraw[treeSearchColor] (\endx-4*\paperspacing,8-\rowspacing) circle (2pt) 
        node[above, text width=1.8cm, align=center, font=\tiny] {RethinkMCTS \\\citenumber{li2024rethinkmctsrefiningerroneousthoughts}}
        node[below, font=\tiny] {2024/09};
    \filldraw[treeSearchColor] (\endx-5*\paperspacing,8-\rowspacing) circle (2pt) 
        node[diamond, fill=treeSearchColor, inner sep=1.5pt]{}
        node[above, text width=1.8cm, align=center, font=\tiny] {SRA-MCTS \\\citenumber{xu2024sramctsselfdrivenreasoningaugmentation}}
        node[below, font=\tiny] {2024/11};
    \filldraw[longCoTColor] (\endx-6*\paperspacing,8-\rowspacing) circle (2pt) 
        node[regular polygon, regular polygon sides=4, fill=longCoTColor, inner sep=1.5pt]{}
        node[above, text width=1.8cm, align=center, font=\tiny] {QwQ-Preview \\\citenumber{qwq-32b-preview}}
        node[below, font=\tiny] {2024/11};
    \filldraw[treeSearchColor] (\endx-7*\paperspacing,8-\rowspacing) circle (2pt) 
        node[regular polygon, regular polygon sides=4, fill=treeSearchColor, inner sep=1.5pt]{}
        node[above, text width=1.8cm, align=center, font=\tiny] {o1-Coder \\\citenumber{zhang2024o1codero1replicationcoding}}
        node[below, font=\tiny] {2024/11};
    \filldraw[longCoTColor] (\endx-8*\paperspacing,8-\rowspacing) circle (2pt) 
        node[regular polygon, regular polygon sides=4, fill=longCoTColor, inner sep=1.5pt]{}
        node[above, text width=1.8cm, align=center, font=\tiny] {SWE-Gym \\\citenumber{pan2024training}}
        node[below, font=\tiny] {2024/11};
    \filldraw[longCoTColor] (\endx-9*\paperspacing,8-\rowspacing) node[regular polygon, regular polygon sides=4, fill=longCoTColor, inner sep=1.5pt] {} 
        node[above, text width=1.8cm, align=center, font=\tiny] {DeepSeek-R1 \\\citenumber{deepseekai2025deepseekr1incentivizingreasoningcapability}}
        node[below, font=\tiny] {2025/01};
    \filldraw[longCoTColor] (\endx-10*\paperspacing,8-\rowspacing) 
        node[regular polygon, regular polygon sides=4, fill=longCoTColor, inner sep=1.5pt]{}
        node[above, text width=1.8cm, align=center, font=\tiny] {Kimi k1.5 \\\citenumber{MoonshotAI}}
        node[below, font=\tiny] {2025/01};

    \filldraw[longCoTColor] (\startx,8-2*\rowspacing) node[diamond, fill=longCoTColor, inner sep=1.5pt] {} 
        node[diamond, fill=longCoTColor, inner sep=1.5pt]{}
        node[above, text width=1.8cm, align=center, font=\tiny] {S1 \\\citenumber{muennighoff2025s1simpletesttimescaling}}
        node[below, font=\tiny] {2025/01};
    \filldraw[parallelColor] (\startx+\paperspacing,8-2*\rowspacing) circle (2pt)
        node[circle, fill=parallelColor, inner sep=1.5pt]{}
        node[above, text width=1.8cm, align=center, font=\tiny] {S$^*$ \\\citenumber{li2025s}}
        node[below, font=\tiny] {2025/02};
    \filldraw[longCoTColor] (\startx+2*\paperspacing,8-2*\rowspacing) circle (2pt)
        node[regular polygon, regular polygon sides=4, fill=longCoTColor, inner sep=1.5pt]{}
        node[above, text width=1.8cm, align=center, font=\tiny] {SWE-RL \\\citenumber{wei2025swerladvancingllmreasoning}}
        node[below, font=\tiny] {2025/02};
    \filldraw[longCoTColor] (\startx+3*\paperspacing,8-2*\rowspacing) node[regular polygon, regular polygon sides=4, fill=longCoTColor, inner sep=1.5pt] {} 
        node[above, text width=1.8cm, align=center, font=\tiny] {QwQ \\\citenumber{qwq32b}}
        node[below, font=\tiny] {2025/03};
    \filldraw[longCoTColor] (\startx+4*\paperspacing,8-2*\rowspacing) node[regular polygon, regular polygon sides=4, fill=longCoTColor, inner sep=1.5pt] {} 
        node[above, text width=1.8cm, align=center, font=\tiny] {ToRL \\\citenumber{li2025torlscalingtoolintegratedrl}}
        node[below, font=\tiny] {2025/03};
    \filldraw[longCoTColor] (\startx+5*\paperspacing,8-2*\rowspacing) node[diamond, fill=longCoTColor, inner sep=1.5pt] {} 
        node[above, text width=1.8cm, align=center, font=\tiny] {SWE-Reasoner\\\citenumber{ma2025thinkinglongerlargerenhancing}}
        node[below, font=\tiny] {2025/03};
    \filldraw[longCoTColor] (\startx+6*\paperspacing,8-2*\rowspacing) circle (2pt)
        node[regular polygon, regular polygon sides=4, fill=longCoTColor, inner sep=1.5pt]{}
        node[above, text width=1.8cm, align=center, font=\tiny] {DeepCoder \\\citenumber{deepcoder2025}}
        node[below, font=\tiny] {2025/04};
    \filldraw[longCoTColor] (\startx+7*\paperspacing,8-2*\rowspacing) circle (2pt)
        node[regular polygon, regular polygon sides=4, fill=longCoTColor, inner sep=1.5pt]{}
        node[above, text width=1.8cm, align=center, font=\tiny] {Seed-Thinking-v1.5 \\\citenumber{seedthinking2025}}
        node[below, font=\tiny] {2025/04};

\end{tikzpicture}
\caption{Works of applying test-time scaling methods in the code field.}
\label{fig:timeline-wrapped-code}
\end{figure}
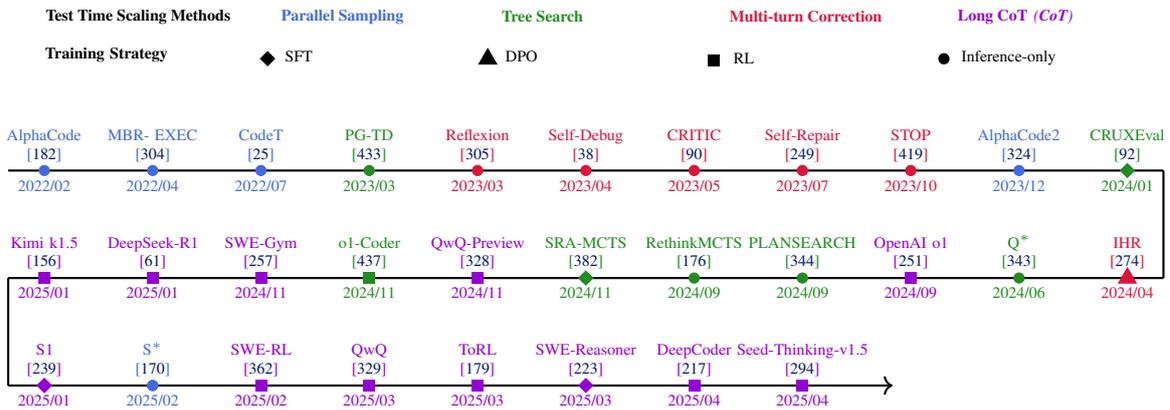
The swift emergence of coding capabilities in language models—exemplified by Codex~\citep{chen2021evaluating} and AlphaCode~\citep{liCompetitionLevelCodeGeneration2022,AlphaCode2Technical}, has transformed software development and boosted productivity. Several studies~\citep{fu2022gptroadmap,ma2023training,shao2024deepseekmathpushinglimitsmathematical} even suggest that code enhances model intelligence. Moreover, coding is now a core feature of general-purpose foundation models~\citep{deepseekv2}, highlighting its critical role in modern model development.

Previously, approaches in code synthesis and code generation have demonstrated how execution verification enables both scalable and verifiable training and test-time signals, laying the foundation for subsequent advancements~\citep{le2022coderl,chenCodeTCodeGeneration2022,zhu2024deepseek}.  
Additionally, directly prompting LLMs to self-reflect, debug, and generate tests for coding tasks represents another direction of test-time scaling~\citep{shinnReflexionLanguageAgents2023a, chen2023teachinglargelanguagemodels}, which enhances model performance in specific downstream scenarios while also providing feedback for refining training strategies~\citep{gu2024cruxeval}.  
These innovations are believed to be instrumental in developing stronger reasoning models, such as the o1-series, which achieve state-of-the-art performance on elite programming benchmarks like SWE-bench~\citep{jimenez2024swebench} and even human-competitive platforms such as Codeforces.\footnote{\href{https://codeforces.com}{https://codeforces.com}}  
This is further exemplified by o1 and o3 model variants earning gold medals at the 2024 International Olympiad in Informatics (IOI), where reinforcement learning on challenging programming tasks is combined with cognitively aligned human guidance~\citep{el2025competitive}.

Still, critical challenges remain.  
First, while code executability facilitates verification, naive execution poses security risks, necessitating robust sandboxing solutions~\citep{hui2024qwen25coder,liu2024fullstack}.  This also requires the development of infrastructure for the practical deployment of these safety mechanisms.
Second, frequent reflection behaviors—often referred to as ``overthinking''—can degrade performance in certain tasks, as observed in coding agentic benchmarks like Aider.\footnote{\href{https://aider.chat/}{https://aider.chat/}}
Third, the reliability of execution-based feedback remains an open question. While DeepSeek-R1~\cite{deepseekai2025deepseekr1incentivizingreasoningcapability} relies on execution results as feedback signals to achieve superior performance, code that passes unit tests can still fail with additional tests, leading to false positives~\cite{stroeblInferenceScalingFLaws2024}, highlighting a fundamental limitation in execution-based evaluation.
Furthermore, further investigation is also needed to align models with real-world coding tasks, as motivated by SWE-Arena~\citep{swe-arena2025} and Copilot-Arena~\citep{chi2025copilot}. These challenges may even intersect with broader issues such as multi-modal understanding and agentic capabilities which is further discussed in the next few sections.
\begin{AIbox}{Future Direction for Coding}
\begin{itemize}
    \item Go beyond competition-level programming to encompass real-world software development tasks and advance toward automatic code review, debugging, and repo-level optimization.
    \item Expand to support more programming languages and continuously learn newly released libraries to achieve expert-level proficiency.
\end{itemize}
\end{AIbox}

\subsection{Multimodality}

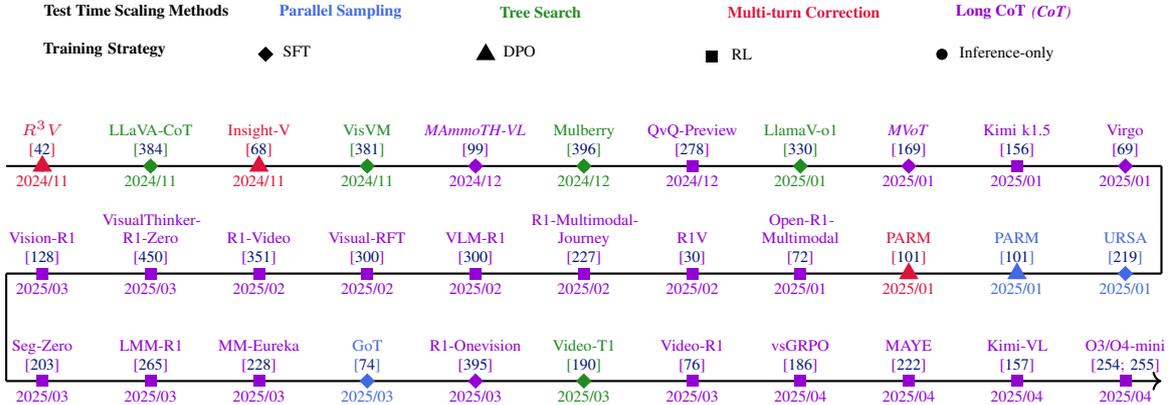
\begin{figure}[h]
\centering
\begin{tikzpicture}[scale=0.95]
    \def\rowspacing{1.5} 
    \def\paperspacing{1.5} 
    \def\papersperrow{11} 
    
    \def\startx{0.5} 
    \def\endx{\startx + 10*\paperspacing} 
    
    \node[anchor=north east] at (\endx-0.5,10.5) {
       \begin{tikzpicture}
            \node[font=\tiny, anchor=west] at (-7,0) {\textbf{Test Time Scaling Methods}};
            
            \node[right, font=\tiny, text=parallelColor] at (-3.9,0) {\textbf{Parallel Sampling}};
            
            \node[right, font=\tiny, text=treeSearchColor] at (-1,0) {\textbf{Tree Search}};
            
            \node[right, font=\tiny, text=correctionColor] at (2,0) {\textbf{Multi-turn Correction}};
            
            \node[right, font=\tiny, text=longCoTColor] at (5,0) {\textbf{Long CoT}};

            \node[right, font=\tiny, text=longCoTColor] at (6,0) {\textit{(\textbf{CoT})}};
            
            \node[font=\tiny, anchor=west, text width=3cm] at (-7,-0.5) {\textbf{Training Strategy}};
            
            \node[diamond, fill=black, inner sep=1.5pt] at (-3.9,-0.5) {};
            \node[right, font=\tiny] at (-3.85,-0.52) {SFT};
            
            \node[regular polygon, regular polygon sides=3, fill=black, inner sep=1.5pt] at (-1,-0.5) {};
            \node[right, font=\tiny] at (-0.95,-0.52) {DPO};
            
            \node[regular polygon, regular polygon sides=4, fill=black, inner sep=1.5pt] at (2,-0.5) {};
            \node[right, font=\tiny] at (2.05,-0.55) {RL};

            \node[circle, regular polygon sides=4, fill=black, inner sep=1.5pt] at (5,-0.5) {};
            \node[right, font=\tiny] at (5.05,-0.55) {Inference-only};
            
        \end{tikzpicture}
    };
    
    \draw[thick] (0,8) -- (\endx+0.5,8);
    
    \draw[thick] (\endx+0.5,8) -- (\endx+0.5,8-\rowspacing);
    
    \draw[thick] (\endx+0.5,8-\rowspacing) -- (0,8-\rowspacing);
    
    \draw[thick] (0,8-\rowspacing) -- (0,8-2*\rowspacing);
    
    \draw[thick, ->] (0,8-2*\rowspacing) -- (\endx+0.5,8-2*\rowspacing);
    
\filldraw[correctionColor] (\startx,8) circle (2pt)
    node[regular polygon, regular polygon sides=3, fill=correctionColor, inner sep=1.5pt] {}
    node[above, text width=1.8cm, align=center, font=\tiny] {$R^3V$\\\citenumber{r3v}}
    node[below, font=\tiny] {2024/11};
    
\filldraw[treeSearchColor] (\startx+\paperspacing,8) circle (2pt) 
    node[diamond, fill=treeSearchColor, inner sep=1.5pt] {} 
    node[above, text width=1.8cm, align=center, font=\tiny] {LLaVA-CoT\\\citenumber{llavacot}}
    node[below, font=\tiny] {2024/11};
    
\filldraw[correctionColor] (\startx+2*\paperspacing,8) 
    node[regular polygon, regular polygon sides=3, fill=correctionColor, inner sep=1.5pt] {}  
    node[above, text width=1.8cm, align=center, font=\tiny] {Insight-V\\\citenumber{insightv}}
    node[below, font=\tiny] {2024/11};
    
\filldraw[treeSearchColor] (\startx+3*\paperspacing,8) circle (2pt)
    node[diamond, fill=treeSearchColor, inner sep=1.5pt] {}
    node[above, text width=1.8cm, align=center, font=\tiny] {VisVM\\\citenumber{visvm}}
    node[below, font=\tiny] {2024/11};

\filldraw[longCoTColor] (\startx+4*\paperspacing,8) circle (2pt)
    node[diamond, fill=longCoTColor, inner sep=1.5pt] {}
    node[above, text width=1.8cm, align=center, font=\tiny] {\textit{MAmmoTH-VL}\\\citenumber{guo2024mammoth}}
    node[below, font=\tiny] {2024/12};

\filldraw[treeSearchColor] (\startx+5*\paperspacing,8) circle (2pt) 
    node[diamond, fill=treeSearchColor, inner sep=1.5pt] {} 
    node[above, text width=1.8cm, align=center, font=\tiny] {Mulberry\\\citenumber{yao2024mulberry}}
    node[below, font=\tiny] {2024/12};

\filldraw[longCoTColor] (\startx+6*\paperspacing,8) circle (2pt)
    node[regular polygon, regular polygon sides=4, fill=longCoTColor, inner sep=1.5pt]{}
    node[above, text width=1.8cm, align=center, font=\tiny] {QvQ-Preview\\\citenumber{qvq-72b-preview}}
    node[below, font=\tiny] {2024/12};
\filldraw[treeSearchColor]  (\startx+7*\paperspacing,8)  node[diamond, fill=treeSearchColor, inner sep=1.5pt] {} 
    node[above, text width=1.8cm, align=center, font=\tiny] {LlamaV-o1 \\\citenumber{llamavo1}}
    node[below, font=\tiny] {2025/01};
    
\filldraw[longCoTColor]  (\startx+8*\paperspacing,8) circle (2pt)     node[diamond, fill=longCoTColor, inner sep=1.5pt] {} 
    node[above, text width=1.8cm, align=center, font=\tiny] {\textit{MVoT} \\\citenumber{li2025imagine}}
    node[below, font=\tiny] {2025/01};

\filldraw[longCoTColor] (\startx+9*\paperspacing,8) node[regular polygon, regular polygon sides=4, fill=longCoTColor, inner sep=1.5pt] {} 
    node[above, text width=1.8cm, align=center, font=\tiny] {Kimi k1.5 \\\citenumber{MoonshotAI}}
    node[below, font=\tiny] {2025/01};
    
\filldraw[longCoTColor] (\startx+10*\paperspacing,8) node[diamond, fill=longCoTColor, inner sep=1.5pt] {}
    node[above, text width=1.8cm, align=center, font=\tiny]{Virgo\\\citenumber{du2025virgo}} 
    node[below, font=\tiny] {2025/01};

\filldraw[parallelColor] (\endx,8-\rowspacing) node[diamond, fill=parallelColor, inner sep=1.5pt] {}
    node[above, text width=1.8cm, align=center, font=\tiny] {URSA\\\citenumber{luo2025ursa}}
    node[below, font=\tiny] {2025/01};
    
\filldraw[parallelColor] (\endx-\paperspacing,8-\rowspacing) node[regular polygon, regular polygon sides=3, fill=parallelColor, inner sep=1.5pt] {} 
    node[above, text width=1.8cm, align=center, font=\tiny] {PARM \\\citenumber{guo2025can}}
    node[below, font=\tiny] {2025/01};
    
\filldraw[correctionColor]  (\endx-2*\paperspacing,8-\rowspacing) node[regular polygon, regular polygon sides=3, fill=correctionColor, inner sep=1.5pt] {} 
    node[above, text width=1.8cm, align=center, font=\tiny] {PARM \\\citenumber{guo2025can}}
    node[below, font=\tiny] {2025/01};
    
\filldraw[longCoTColor]  (\endx-3*\paperspacing,8-\rowspacing) node[regular polygon, regular polygon sides=4, fill=longCoTColor, inner sep=1.5pt] {} 
    node[above, text width=1.8cm, align=center, font=\tiny] {Open-R1-Multimodal \\\citenumber{open-r1-multimodal}}
    node[below, font=\tiny] {2025/01};
    
\filldraw[longCoTColor]  (\endx-4*\paperspacing,8-\rowspacing) node[regular polygon, regular polygon sides=4, fill=longCoTColor, inner sep=1.5pt] {} 
    node[above, text width=1.8cm, align=center, font=\tiny] {R1V \\\citenumber{chen2025r1v}}
    node[below, font=\tiny] {2025/02};
    
\filldraw[longCoTColor] (\endx-5*\paperspacing,8-\rowspacing) circle (2pt) node[regular polygon, regular polygon sides=4, fill=longCoTColor, inner sep=1.5pt] {} 
    node[above, text width=1.8cm, align=center, font=\tiny] {R1-Multimodal-Journey \\\citenumber{r1-multimodal-journey}}
    node[below, font=\tiny] {2025/02};
    
\filldraw[longCoTColor](\endx-6*\paperspacing,8-\rowspacing) circle (2pt) node[regular polygon, regular polygon sides=4, fill=longCoTColor, inner sep=1.5pt] {} 
    node[above, text width=1.8cm, align=center, font=\tiny] {VLM-R1 \\\citenumber{shen2025vlmr1}}
    node[below, font=\tiny] {2025/02};
    
\filldraw[longCoTColor](\endx-7*\paperspacing,8-\rowspacing) circle (2pt) node[regular polygon, regular polygon sides=4, fill=longCoTColor, inner sep=1.5pt] {} 
    node[above, text width=1.8cm, align=center, font=\tiny] {Visual-RFT \\\citenumber{shen2025vlmr1}}
    node[below, font=\tiny] {2025/02};
    
\filldraw[longCoTColor] (\endx-8*\paperspacing,8-\rowspacing) node[regular polygon, regular polygon sides=4, fill=longCoTColor, inner sep=1.5pt] {} 
    node[above, text width=1.8cm, align=center, font=\tiny] {R1-Video \\\citenumber{wang-2025-open-r1-video}}
    node[below, font=\tiny] {2025/02};
    
\filldraw[longCoTColor] (\endx-9*\paperspacing,8-\rowspacing) node[regular polygon, regular polygon sides=4, fill=longCoTColor, inner sep=1.5pt] {} 
    node[above, text width=1.8cm, align=center, font=\tiny] {VisualThinker-R1-Zero \\\citenumber{zhou2025r1zerosahamomentvisual}}
    node[below, font=\tiny] {2025/03};

\filldraw[longCoTColor] (\endx-10*\paperspacing,8-\rowspacing) circle (2pt) node[regular polygon, regular polygon sides=4, fill=longCoTColor, inner sep=1.5pt] {} 
    node[above, text width=1.8cm, align=center, font=\tiny] {Vision-R1 \\\citenumber{huang2025vision}}
    node[below, font=\tiny] {2025/03};
    
\filldraw[longCoTColor] (\startx,8-2*\rowspacing) circle (2pt) node[regular polygon, regular polygon sides=4, fill=longCoTColor, inner sep=1.5pt] {}  
    node[above, text width=1.8cm, align=center, font=\tiny] {Seg-Zero \\\citenumber{liu2025seg}}
    node[below, font=\tiny] {2025/03};
    
\filldraw[longCoTColor] (\startx+1*\paperspacing,8-2*\rowspacing) node[regular polygon, regular polygon sides=4, fill=longCoTColor, inner sep=1.5pt] {} 
    node[above, text width=1.8cm, align=center, font=\tiny] {LMM-R1 \\\citenumber{peng2025lmmr1}}
    node[below, font=\tiny] {2025/03};
    
\filldraw[longCoTColor] (\startx+2*\paperspacing,8-2*\rowspacing) node[regular polygon, regular polygon sides=4, fill=longCoTColor, inner sep=1.5pt] {}  
    node[above, text width=1.8cm, align=center, font=\tiny] {MM-Eureka \\\citenumber{meng2025mm}}
    node[below, font=\tiny] {2025/03};
    
\filldraw[parallelColor] (\startx+3*\paperspacing,8-2*\rowspacing) node[diamond, fill=parallelColor, inner sep=1.5pt] {} 
    node[above, text width=1.8cm, align=center, font=\tiny] {GoT \\\citenumber{fang2025got}}
    node[below, font=\tiny] {2025/03};
    
\filldraw[longCoTColor]  (\startx+4*\paperspacing,8-2*\rowspacing) node[diamond, fill=longCoTColor, inner sep=1.5pt] {} 
    node[above, text width=1.8cm, align=center, font=\tiny] {R1-Onevision \\\citenumber{yang2025r1onevisionadvancinggeneralizedmultimodal}}
    node[below, font=\tiny] {2025/03};
    
\filldraw[treeSearchColor] (\startx+5*\paperspacing,8-2*\rowspacing) node[diamond, fill=treeSearchColor, inner sep=1.5pt] {} 
    node[above, text width=1.8cm, align=center, font=\tiny] {Video-T1 \\\citenumber{liu2025video}}
    node[below, font=\tiny] {2025/03};
\filldraw[longCoTColor] (\startx+6*\paperspacing,8-2*\rowspacing) node[regular polygon, regular polygon sides=4, fill=longCoTColor, inner sep=1.5pt] {}
    node[above, text width=1.8cm, align=center, font=\tiny] {Video-R1 \\\citenumber{feng2025video}}
    node[below, font=\tiny] {2025/03};    
\filldraw[longCoTColor] (\startx+7*\paperspacing,8-2*\rowspacing) node[regular polygon, regular polygon sides=4, fill=longCoTColor, inner sep=1.5pt] {} 
    node[above, text width=1.8cm, align=center, font=\tiny] {vsGRPO \\\citenumber{liao2025improved}}
    node[below, font=\tiny] {2025/04};
    
\filldraw[longCoTColor]  (\startx+8*\paperspacing,8-2*\rowspacing)node[regular polygon, regular polygon sides=4, fill=longCoTColor, inner sep=1.5pt] {} 
    node[above, text width=1.8cm, align=center, font=\tiny] {MAYE \\\citenumber{MAYE}}
    node[below, font=\tiny] {2025/04};
\filldraw[longCoTColor]  (\startx+9*\paperspacing,8-2*\rowspacing)node[regular polygon, regular polygon sides=4, fill=longCoTColor, inner sep=1.5pt] {} 
    node[above, text width=1.8cm, align=center, font=\tiny] {Kimi-VL \\\citenumber{kimi2024vl}}
    node[below, font=\tiny] {2025/04};
\filldraw[longCoTColor]  (\startx+10*\paperspacing,8-2*\rowspacing)node[regular polygon, regular polygon sides=4, fill=longCoTColor, inner sep=1.5pt] {} 
    node[above, text width=1.8cm, align=center, font=\tiny] {O3/O4-mini \\\citenumber{o3_o4_mini,thinking_with_images}}
    node[below, font=\tiny] {2025/04};
\end{tikzpicture}
\caption{Works of applying test-time scaling methods in the multi-modal field.}
\label{fig:timeline-wrapped-multi-modal}
\end{figure}
\paragraph{Test-Time Scaling for Multimodal Understanding}

Vision-Language Models (VLMs)~\cite{vlm_survey} have demonstrated significant potential in tasks involving multimodal understanding and generating textual outputs. Test-time scaling techniques for VLMs can be directly adapted from LLMs since the task outputs are textual.

DeepSeek R1 represents a significant milestone, being the first to fully demonstrate the effectiveness of RL training in scaling test times for LLMs. It effectively divides research on test-time scaling in VLMs into two distinct phases: pre-R1 and post-R1. Before R1, research efforts relied on long CoT distillation~\cite{guo2024mammoth, du2025virgo, llavacot}, tree search~\cite{llavacot, llamavo1, yao2024mulberry, visvm}, multi-turn correction~\cite{r3v, insightv}, and parallel sampling~\cite{luo2025ursa}. For example, LLaVA-CoT~\cite{llavacot} and LlamaV-O1~\cite{llamavo1} enhance reasoning ability by decomposing the reasoning process into explicit sequential steps, such as summarizing the problem, reviewing image content, and performing reasoning. Similarly, Mulberry~\cite{yao2024mulberry} and LLaVA-CoT~\cite{llavacot} employ search algorithms like beam search and MCTS to expand the search space during reasoning. MAmmoTH-VL~\cite{guo2024mammoth} and Virgo~\cite{du2025virgo} scale the output length of smaller models by distilling reasoning chains from larger VLMs, resulting in significant improvements across various visual reasoning tasks. As part of R1's concurrent work, QVQ-72B-Preview~\cite{qvq-72b-preview} and K1.5~\cite{MoonshotAI} have demonstrated the immense potential of RL in VLMs.
After R1, the research community has increasingly explored RL-based training as a strategy to elicit VLMs' test-time scaling ability. Recent efforts primarily fall into two directions: 1) enhancing the depth of multimodal reasoning, by pushing the limits of VLMs' visual problem-solving capabilities—such as in open-r1-multimodal~\cite{open-r1-multimodal}, MM-Eureka~\cite{meng2025mm}, LMM-R1~\cite{peng2025lmmr1}, Vision-R1~\cite{huang2025vision}, VisualThinker-R1-Zero~\cite{zhou2025r1zerosahamomentvisual} and so on; and 2) expanding the breadth of multi-modal tasks, demonstrating the effectiveness of RL training across diverse vision-centric domains, including visual counting~\cite{chen2025r1v}, detection~\cite{liu2025visual,shen2025vlmr1}, segmentation~\cite{liu2025seg}, etc.

Despite promising results, long-CoT-based test-time scaling on VLMs still faces several challenges. First, unlike LLMs, input instructions for VLMs involve multiple modalities including both visual and textual data. Many studies that replicate LLM test-time scaling methods directly on VLMs fail to incorporate rethinking and reflection on visual inputs in the VLM's responses, overlooking the unique characteristics of multimodal understanding tasks. Future research should aim to better integrate test-time scaling with processing different modality inputs to address hallucinations from non-textual inputs~\cite{reducing_vlm_hallucination}, while leveraging the synergies of multimodal inputs to enhance the effectiveness of test-time scaling. Second, conducting training directly on base models remains challenging. Unlike LLMs, VLMs lack strong base models because their training focuses mainly on modality alignment through image-caption pairs and instruction tuning~\cite{llava,llava_15,llavaonevision}, without extensive pre-training on general multimodal corpora, which demands substantial computational and data resources.

\paragraph{Test-Time Scaling for Multimodal Generation} 
Recent breakthroughs such as Gemini’s image-text interleaved generation and GPT-4o’s image editing capabilities have reignited broad interest in multimodal generation. Unlike multimodal understanding tasks, where the output remains textual and test-time scaling techniques can be readily borrowed from LLMs, multimodal generation involves producing non-textual outputs (e.g., images, videos, or audio), thus requiring new approaches to unlock the full potential of test-time scaling in this domain.

Several recent works have explored this frontier. \citet{li2025imagine} enhance MLLM spatial reasoning via the Multimodal Visualization-of-Thought (MVoT) framework, which adopts a long CoT strategy to scale reasoning at test time. PARM~\cite{guo2025can} and MINT~\cite{mint2025} promote multimodal generation through multi-turn correction, aiming to improve image quality by iteratively reflecting on and revising the associated text prompts.
GoT~\cite{fang2025got} unleashes the reasoning capability of MLLMs for visual generation and editing by introducing the Generation Chain-of-Thought framework, which adopts a long CoT strategy for test-time scaling. Expanding beyond images, Video-T1~\cite{liu2025video} extends input modalities to video generation, utilizing a tree search method at inference time to enhance video generation quality. 

Looking ahead, applying test-time scaling to multimodal generation represents a highly promising but underexplored direction. Open questions include identifying the most effective model architecture~\cite{team2024chameleon, zhou2024transfusion, chen2025janus}, designing pretraining and instruction-tuning strategies tailored to non-textual outputs, and improving the computational efficiency of such systems. Moreover, if reinforcement learning is to be extended to multimodal generation, key challenges emerge in defining suitable reward signals and constructing RL infrastructure for non-text modalities.

\begin{AIbox}{Future Direction for Multimodality}
\begin{itemize}
    \item Focus more on applying test-time scaling to vision-centric tasks such as classification and detection, as well as non-text outputs like images and videos.
    \item Integrate more modalities (e.g., audio) and develop stronger native multimodal foundation models to unlock test-time scaling potential.
\end{itemize}
\end{AIbox}

\subsection{Agent}
\definecolor{parallelColor}{RGB}{65, 105, 225}  
\definecolor{treeSearchColor}{RGB}{34, 139, 34}  
\definecolor{correctionColor}{RGB}{220, 20, 60}  
\definecolor{longCoTColor}{RGB}{148, 0, 211}  
\definecolor{islColor}{RGB}{0, 0, 0}  

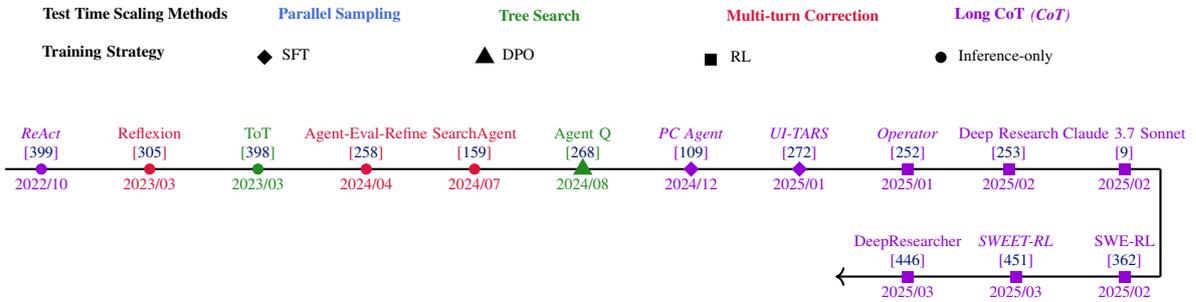
\begin{figure}[h]
\centering
\begin{tikzpicture}[scale=0.95]
    \def\rowspacing{1.5} 
    \def\paperspacing{1.5} 
    \def\papersperrow{11} 
    
    \def\startx{0.5} 
    \def\endx{\startx + 10*\paperspacing} 
    
        \node[anchor=north east] at (\endx-0.5,10.5) {
      \begin{tikzpicture}
            \node[font=\tiny, anchor=west] at (-7,0) {\textbf{Test Time Scaling Methods}};
            
            \node[right, font=\tiny, text=parallelColor] at (-3.9,0) {\textbf{Parallel Sampling}};
            
            \node[right, font=\tiny, text=treeSearchColor] at (-1,0) {\textbf{Tree Search}};
            
            \node[right, font=\tiny, text=correctionColor] at (2,0) {\textbf{Multi-turn Correction}};
            
            \node[right, font=\tiny, text=longCoTColor] at (5,0) {\textbf{Long CoT}};

            \node[right, font=\tiny, text=longCoTColor] at (6,0) {\textit{(\textbf{CoT})}};
            
            \node[font=\tiny, anchor=west, text width=3cm] at (-7,-0.5) {\textbf{Training Strategy}};
            
            \node[diamond, fill=black, inner sep=1.5pt] at (-3.9,-0.5) {};
            \node[right, font=\tiny] at (-3.85,-0.52) {SFT};
            
            \node[regular polygon, regular polygon sides=3, fill=black, inner sep=1.5pt] at (-1,-0.5) {};
            \node[right, font=\tiny] at (-0.95,-0.52) {DPO};
            
            \node[regular polygon, regular polygon sides=4, fill=black, inner sep=1.5pt] at (2,-0.5) {};
            \node[right, font=\tiny] at (2.05,-0.55) {RL};

            \node[circle, regular polygon sides=4, fill=black, inner sep=1.5pt] at (5,-0.5) {};
            \node[right, font=\tiny] at (5.05,-0.55) {Inference-only};
            
        \end{tikzpicture}
    };
    
    \draw[thick] (0,8) -- (\endx+0.5,8);

    \draw[thick] (\endx+0.5,8) -- (\endx+0.5,8-\rowspacing);

    \draw[thick,->] (\endx+0.5,8-\rowspacing) -- (\endx - 4,8-\rowspacing);
    \filldraw[longCoTColor] (\startx,8) circle (2pt) 
        node[above, text width=1.8cm, align=center, font=\tiny] {\textit{ReAct} \\\citenumber{yao2023reactsynergizingreasoningacting}}
        node[below, font=\tiny] {2022/10};
    
    \filldraw[correctionColor] (\startx+\paperspacing,8)  circle (2pt)
        node[above, text width=1.8cm, align=center, font=\tiny] {Reflexion \\\citenumber{shinnReflexionLanguageAgents2023a}}
        node[below, font=\tiny] {2023/03};
    
    \filldraw[treeSearchColor] (\startx+2*\paperspacing,8) circle (2pt) 
        node[above, text width=1.8cm, align=center, font=\tiny] {ToT \\\citenumber{yao2023treethoughtsdeliberateproblem}}
        node[below, font=\tiny] {2023/03};

    \filldraw[correctionColor] (\startx+3*\paperspacing,8) circle (2pt) 
        node[above, text width=1.8cm, align=center, font=\tiny]{}
        node[above, text width=1.8cm, align=center, font=\tiny] {Agent-Eval-Refine\\\citenumber{pan2024autonomousevaluationrefinementdigital}}
        node[below, font=\tiny] {2024/04};

    \filldraw[correctionColor] (\startx+4*\paperspacing,8) circle (2pt) 
        node[above, text width=1.8cm, align=center, font=\tiny]{}
        node[above, text width=1.8cm, align=center, font=\tiny] {SearchAgent\\\citenumber{koh2024treesearchlanguagemodel}}
        node[below, font=\tiny] {2024/07};
    
    \filldraw[treeSearchColor] (\startx+5*\paperspacing,8) node[regular polygon, regular polygon sides=3, fill=treeSearchColor, inner sep=1.5pt] {} 
        node[above, text width=1.8cm, align=center, font=\tiny] {Agent Q \\\citenumber{putta2024agentqadvancedreasoning}}
        node[below, font=\tiny] {2024/08};
    
    \filldraw[longCoTColor] (\startx+6*\paperspacing,8) node[diamond, fill=longCoTColor, inner sep=1.5pt] {} 
        node[above, text width=1.8cm, align=center, font=\tiny] {\textit{PC Agent} \\\citenumber{he2024pcagentsleepai}}
        node[below, font=\tiny] {2024/12};
    
    \filldraw[longCoTColor] (\startx+7*\paperspacing,8) node[diamond, fill=longCoTColor, inner sep=1.5pt] {} 
        node[above, text width=1.8cm, align=center, font=\tiny] {\textit{UI-TARS} \\\citenumber{qin2025uitarspioneeringautomatedgui}}
        node[below, font=\tiny] {2025/01};
    
    \filldraw[longCoTColor] (\startx+8*\paperspacing,8) node[regular polygon, regular polygon sides=4, fill=longCoTColor, inner sep=1.5pt]{}
        node[above, text width=1.8cm, align=center, font=\tiny] {\textit{Operator} \\\citenumber{cuaopenai}}
        node[below, font=\tiny] {2025/01};
    
    \filldraw[longCoTColor] (\startx+9*\paperspacing - 0.1,8) node[regular polygon, regular polygon sides=4, fill=longCoTColor, inner sep=1.5pt]{}
        node[above, text width=1.8cm, align=center, font=\tiny] {Deep Research\\\citenumber{deepresearch}}
        node[below, font=\tiny] {2025/02};
    
    \filldraw[longCoTColor] (\startx+10*\paperspacing,8) node[regular polygon, regular polygon sides=4, fill=longCoTColor, inner sep=1.5pt]{}
        node[above, text width=1.8cm, align=center, font=\tiny] {Claude 3.7 Sonnet \\\citenumber{claude3.7sonnet}}
        node[below, font=\tiny] {2025/02};
    
    \filldraw[longCoTColor] (\endx,8-\rowspacing) node[regular polygon, regular polygon sides=4, fill=longCoTColor, inner sep=1.5pt]{}
        node[above, text width=1.8cm, align=center, font=\tiny] {SWE-RL \\\citenumber{wei2025swerladvancingllmreasoning}}
        node[below, font=\tiny] {2025/02};
    
    
    \filldraw[longCoTColor] (\endx - \paperspacing,8-\rowspacing) node[regular polygon, regular polygon sides=4, fill=longCoTColor, inner sep=1.5pt]{}
        node[above, text width=1.8cm, align=center, font=\tiny] {\textit{SWEET-RL} \\\citenumber{zhou2025sweetrltrainingmultiturnllm}}
        node[below, font=\tiny] {2025/03};

    
    \filldraw[longCoTColor] (\endx - 2*\paperspacing,8-\rowspacing) circle (2pt) 
        node[regular polygon, regular polygon sides=4, fill=longCoTColor, inner sep=1.5pt]{}
        node[above, text width=1.8cm, align=center, font=\tiny] {DeepResearcher \\\citenumber{zheng2025deepresearcherscalingdeepresearch}}
        node[below, font=\tiny] {2025/03};

\end{tikzpicture}
\caption{Works of applying test-time scaling methods in the agent field.}
\label{fig:timeline-wrapped-agent}
\end{figure}
LLM Agent is an autonomous system that leverages LLMs as its cognitive core to automatically perform complex tasks through action execution in dynamic environments~\citep{sumers2024cognitivearchitectureslanguageagents}. Building on prior efforts, the objective of agents is evolving from specific pre-defined workflows to handling more open-ended tasks in complex environments that need decision-making at scale, such as software engineering~\cite{jimenez2024swebench}, deep research~\cite{deepresearch}, and computer use~\cite{anthropic2024models, cuaopenai}. Such tasks often require long-horizon planning to be accomplished with multi-step interaction with the environment, leading to substantial test-time computation. For instance, OpenAI DeepResearch~\cite{deepresearch} demands 5–30 minutes for a single research task, while CUA~\cite{cuaopenai} may require hundreds of steps to complete a computer use task and exhibit clear test-time scaling behavior.

A prerequisite for effective multi-step task execution is the enhancement of decision-making in each step. This necessitates models with advanced reasoning capabilities to perform verification, backtracking, and reflection based on historical trajectories and current environment observations, thereby aligning actions with long-term goals. Numerous approaches have adopted test-time scaling strategies to optimize per-step decision quality. For example, ReAct~\cite{yao2023reactsynergizingreasoningacting} introduces CoT reasoning during action selection. Reflexion~\cite{shinnReflexionLanguageAgents2023a} further advances this paradigm by incorporating explicit feedback signals from prior steps to enable self-correction. Recently, Deep Research has highlighted the potential of scaling the reasoning process in single-step decision making. Powered by an optimized version of the OpenAI o3, it achieves research-analyst-level proficiency in synthesizing online sources into comprehensive reports, and attains 26.6\% accuracy on the challenging Humanity's Last Exam benchmark~\cite{phan2025humanitysexam}, significantly surpassing previous SOTA models.

Regarding training strategies, many methods introduce historical trajectories into SFT training samples to help models learn to handle multi-step histories, and incorporate thinking processes at each step to achieve CoT capabilities~\cite{he2024pcagentsleepai}. Additionally, UI-TARS~\cite{qin2025uitarspioneeringautomatedgui} addresses the limitation of SFT methods that only utilize the corrected steps through DPO methods, thereby enhancing the agent's error correction and post-reflection abilities.
Although the specific implementation of OpenAI Deep Research remains unclear, recent work applies RL for end-to-end training of deep research agents~\cite{zheng2025deepresearcherscalingdeepresearch}.

Despite some production-ready implementations such as GitHub Copilot, most agentic systems remain confined to proof-of-concept demonstrations rather than robust, large-scale deployment. Key barriers include insufficient general model capabilities, the predominance of prompt engineering over specialized agentic training, and context window constraints for long trajectories—particularly for visual observations. Additionally, unlike code or math tasks, many agentic tasks lack well-defined external verifiers, making it challenging to provide reliable rewards in reinforcement learning frameworks. Moreover, the involvement of long CoT in reasoning also introduces the ``Reasoning-Action Dilemma,'' which requires models to carefully balance active engagement with the environment against the need for internal reasoning, highlighting the importance of developing reasoning models that remain effectively grounded in environmental context when applied to agentic tasks~\citep{cuadron2025dangeroverthinkingexaminingreasoningaction}.

\begin{AIbox}{Future Direction for Agents}
\begin{itemize}
    \item Develop robust execution environments that embrace diverse tool use, and craft scalable evaluation frameworks, which pave the way for reinforcement learning in agent training together.
    \item Further explore action scaling as a new scaling dimension—expand agents’ competence by growing the number of interaction steps with the environment.
\end{itemize}
\end{AIbox}

\subsection{Embodied AI}
\definecolor{parallelColor}{RGB}{65, 105, 225}  
\definecolor{treeSearchColor}{RGB}{34, 139, 34}  
\definecolor{correctionColor}{RGB}{220, 20, 60}  
\definecolor{longCoTColor}{RGB}{148, 0, 211}  
\definecolor{islColor}{RGB}{0, 0, 0}  

\begin{figure}[h]
\centering
\begin{tikzpicture}[scale=0.95]
    \def\rowspacing{1.5} 
    \def\paperspacing{1.5} 
    \def\papersperrow{11} 
    
    \def\startx{0.5} 
    \def\endx{\startx + 10*\paperspacing} 
    
    \node[anchor=north east] at (\endx-0.5,10.5) {
      \begin{tikzpicture}
            \node[font=\tiny, anchor=west] at (-7,0) {\textbf{Test Time Scaling Methods}};
            
            \node[right, font=\tiny, text=parallelColor] at (-3.9,0) {\textbf{Parallel Sampling}};
            
            \node[right, font=\tiny, text=treeSearchColor] at (-1,0) {\textbf{Tree Search}};
            
            \node[right, font=\tiny, text=correctionColor] at (2,0) {\textbf{Multi-turn Correction}};
            
            \node[right, font=\tiny, text=longCoTColor] at (5,0) {\textbf{Long CoT}};

            \node[right, font=\tiny, text=longCoTColor] at (6,0) {\textit{(\textbf{CoT})}};
            
            \node[font=\tiny, anchor=west, text width=3cm] at (-7,-0.5) {\textbf{Training Strategy}};
            
            \node[diamond, fill=black, inner sep=1.5pt] at (-3.9,-0.5) {};
            \node[right, font=\tiny] at (-3.85,-0.52) {SFT};
            
            \node[regular polygon, regular polygon sides=3, fill=black, inner sep=1.5pt] at (-1,-0.5) {};
            \node[right, font=\tiny] at (-0.95,-0.52) {DPO};
            
            \node[regular polygon, regular polygon sides=4, fill=black, inner sep=1.5pt] at (2,-0.5) {};
            \node[right, font=\tiny] at (2.05,-0.55) {RL};

            \node[circle, regular polygon sides=4, fill=black, inner sep=1.5pt] at (5,-0.5) {};
            \node[right, font=\tiny] at (5.05,-0.55) {Inference-only};
            
        \end{tikzpicture}
    };
    
    \draw[thick, ->] (0,8) -- (\endx+0.5,8);

    \filldraw[correctionColor] (\startx,8) circle (2pt) 
        node[above, text width=1.8cm, align=center, font=\tiny] {Inner Monologue \\\citenumber{huang2022inner}}
        node[below, font=\tiny] {2022/07};
    
    \filldraw[correctionColor] (\startx+\paperspacing,8) circle (2pt) 
        node[above, text width=1.8cm, align=center, font=\tiny] {REFLECT \\\citenumber{liu2023reflect}}
        node[below, font=\tiny] {2023/06};
    
    \filldraw[correctionColor] (\startx+2*\paperspacing,8) circle (2pt) 
        node[above, text width=1.8cm, align=center, font=\tiny] {KnowNo \\\citenumber{renrobots}}
        node[below, font=\tiny] {2023/07};
    
    \filldraw[longCoTColor] (\startx+3*\paperspacing,8) node[diamond, fill=longCoTColor, inner sep=1.5pt] {} 
        node[above, text width=1.8cm, align=center, font=\tiny] {\textit{Embodied-CoT} \\\citenumber{michal2024robotic}}
        node[below, font=\tiny] {2024/07};
    
    \filldraw[longCoTColor] (\startx+4*\paperspacing,8) node[diamond, fill=longCoTColor, inner sep=1.5pt] {} 
        node[above, text width=1.8cm, align=center, font=\tiny] {\textit{CoA} \\\citenumber{li2024improving}}
        node[below, font=\tiny] {2024/12};
    
    \filldraw[longCoTColor] (\startx+5*\paperspacing,8) node[diamond, fill=longCoTColor, inner sep=1.5pt] {}
        node[above, text width=1.8cm, align=center, font=\tiny] {\textit{SpatialCoT} \\\citenumber{liu2025spatialcot}}
        node[below, font=\tiny] {2025/01};
    
    \filldraw[longCoTColor] (\startx+6*\paperspacing,8) node[diamond, fill=longCoTColor, inner sep=1.5pt] {}
        node[above, text width=1.8cm, align=center, font=\tiny] {\textit{RAD} \\\citenumber{clark2025action}}
        node[below, font=\tiny] {2025/02};
    
    \filldraw[longCoTColor] (\startx+7*\paperspacing,8) node[regular polygon, regular polygon sides=4, fill=longCoTColor, inner sep=1.5pt]{}
        node[above, text width=1.8cm, align=center, font=\tiny] {Cosmos-Reason1 \\\citenumber{azzolini2025cosmos}}
        node[below, font=\tiny] {2025/03};
    
    \filldraw[longCoTColor] (\startx+8.1*\paperspacing,8) node[diamond, fill=longCoTColor, inner sep=1.5pt] {} 
        node[above, text width=1.8cm, align=center, font=\tiny] {\textit{Gemini Robotics} \\\citenumber{team2025gemini}}
        node[below, font=\tiny] {2025/03};
    
    \filldraw[longCoTColor] (\startx+9*\paperspacing,8) node[diamond, fill=longCoTColor, inner sep=1.5pt] {} 
        node[above, text width=1.8cm, align=center, font=\tiny] {CoT-VLA \\\citenumber{zhao2025cot}}
        node[below, font=\tiny] {2025/03};
    
    \filldraw[longCoTColor] (\startx+10*\paperspacing,8) node[diamond, fill=longCoTColor, inner sep=1.5pt] {}  
        node[above, text width=1.8cm, align=center, font=\tiny] {Embodied-Reasoner \\\citenumber{zhang2025embodied}}
        node[below, font=\tiny] {2025/03};

\end{tikzpicture}
\caption{Works of applying test-time scaling methods in the Embodied AI field.}
\label{fig:timeline-wrapped-embodied}
\end{figure}
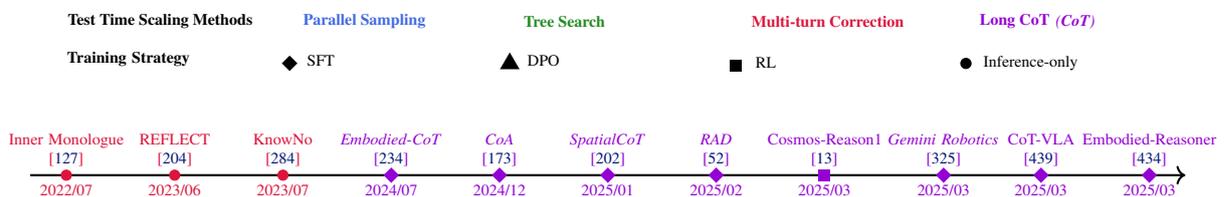
Embodied AI is essential to advancing AGI, as it establishes the foundational link between cognitive representation and interaction with the physical world. By enabling robots to engage with their surrounding environments and manipulate objects, embodied AI allows for the execution of real-world and complex tasks, necessitating sophisticated levels of cognition and reasoning abilities. Cognition engineering further supports the enhancement of these cognitive and reasoning abilities within embodied AI systems. Embodied AI systems typically adopt a hierarchical architecture comprising two distinct yet interconnected operational phases: high-level planning and low-level control policy execution~\cite{ahn2022can}.

First, high-level planning is the process of creating a sequence of sub-tasks that a robot can follow to achieve a specific goal within its environment. This involves decision-making based on both the robot's current state and the predicted outcomes of its actions, aiming to optimize efficiency, safety, and goal attainment within often dynamic and complex surroundings. This process frequently requires advanced cognitive abilities for analytical thinking and reasoning.~\citet{huang2022inner} pioneer the use of inner thoughts as a feedback mechanism to enhance reasoning capabilities, enabling multi-turn self-correction through internal monologue. Building on this idea,~\citet{liu2023reflect} leverage LLMs to generate explicit failure explanations, which are then utilized to refine reasoning processes, thereby achieving substantial improvements in planning and problem-solving performance.~\citet{renrobots} propose a method to quantify the uncertainty of LLM-based planners and trigger assistance requests when the uncertainty surpasses a predefined threshold, which enhances reasoning ability through uncertainty alignment. Inspired by o1~\cite{openai_o1_system_card},~\citet{liu2025spatialcot} employ bi-directional spatial coordinate alignment and chain-of-thought spatial grounding to facilitate the generation of long thoughts, thereby improving planning performance. Furthermore,~\citet{chaiempowering} propose a novel approach based on Q-learning, which enables the model to make optimal decisions. Inspired by Deepseek-R1~\cite{deepseekai2025deepseekr1incentivizingreasoningcapability}, \citet{azzolini2025cosmos} introduce a novel VLM designed from the ground up to enhance reasoning capabilities in physical environments. The model incorporates an innovative hybrid architecture that combines Mamba, MLP, and Transformer components. This approach is complemented by comprehensive vision pre-training, specialized physical AI SFT, and physical AI reinforcement learning. The resulting system effectively processes video input paired with linguistic instructions, generating long reasoning thoughts before predicting appropriate subsequent actions. Meanwhile, \citet{zhang2025embodied} propose a dataset comprising 9.3K synthesized instances to enhance embodied reasoning capabilities, utilizing GPT-4o to generate Observation-Thought-Action trajectories that serve as extended reasoning chains for long-horizon action prediction tasks.

Second, low-level control policy aims to translate tasks into executable actions, such as those performed by a 7-DOF robotic arm in joint space. Currently, the most prevalent approach leverages large models, particularly through vision-language-action (VLA) frameworks. These models fine-tune pretrained vision-language models on trajectory data to generate actionable sequences. Importantly, reasoning plays a key role in this process. For instance, if the task involves placing apples on one plate and bananas on another, the model must first distinguish between apples and bananas, rather than relying on simple ``muscle memory'' from prior learning. Embodied CoT~\cite{michal2024robotic} proposes constructing CoT that bridges perceptual information and task objectives by incorporating external knowledge from other models or algorithms. This approach has been demonstrated to substantially improve performance. In contrast, \citet{zhao2025cot} introduce a visual CoT framework that integrates explicit visual reasoning processes into VLA models. Their approach generates future image frames in an autoregressive manner, establishing them as visual objectives before producing concise action sequences designed to achieve these predetermined goals.~\citet{zhang2024grape} introduce a novel preference alignment approach for robotic policy learning, allowing VLA models to learn not only from successful trajectories but also from failure trajectories. Additionally,~\citet{li2024improving} integrate diverse robot affordance information to enhance the model's generalization in reasoning during testing and leverage the generated reasoning to foster long-horizon reasoning capabilities. However, SFT-based approaches rely on high-quality expert datasets, which are both expensive and challenging to obtain in the robotics domain. Moreover, these datasets may not fully align VLA models with real-world physical environments due to distribution shift issues. To address this limitation,~\citet{guo2025improving} propose a novel method for low-level action generation that alternates between online reinforcement learning and supervised learning stages, significantly enhancing the generalization capability of VLA models. Additionally, \citet{clark2025action} harness extensive human video data to augment reasoning capabilities. This approach employs Gemini to synthesize reasoning steps from human demonstrations. Subsequently, the method leverages limited robot data to train the model in mapping abstract reasoning to low-level actions, while the action-free datasets serve to enhance the model's overall reasoning proficiency. Moreover, \citet{team2025gemini} train Gemini Robotics-ER based on Gemini 2.0, a VLM that demonstrates enhanced embodied reasoning capabilities. They subsequently extend this work to develop Gemini Robotics, a VLA model that integrates robot action data. This integration enables high-frequency dexterous control, robust generalization, and rapid adaptation across diverse robotic tasks and embodiments.

While numerous studies have successfully elicited long CoT reasoning to enhance performance, there remains significant room for improvement in embodied AI. First, although pure reinforcement learning has proven effective in LLMs for fostering self-reflection through self-exploration~\cite{deepseekai2025deepseekr1incentivizingreasoningcapability}, a viable approach for embodied AI has yet to be established. Second, existing frameworks typically decouple high-level planning from low-level policy. However, a more effective paradigm would integrate planning and execution into a unified process, enabling continuous optimization through internal reasoning and iterative feedback.

\begin{AIbox}{Future Direction for Embodied AI}
\begin{itemize}
\item Develop unified frameworks that seamlessly integrate high-level planning and low-level policy execution into a continuous process, enabling agents to optimize performance through internal reasoning loops and real-time iterative feedback during physical environment interactions.
\item Establish fine-grained evaluation frameworks that systematically isolate and assess embodied agents' effectiveness across thinking, planning, and execution processes, providing deeper insights into how individual components contribute to overall performance in physical-world interactions. 
\end{itemize}
\end{AIbox}

\subsection{Safety}
\definecolor{parallelColor}{RGB}{65, 105, 225}  
\definecolor{treeSearchColor}{RGB}{34, 139, 34}  
\definecolor{correctionColor}{RGB}{220, 20, 60}  
\definecolor{longCoTColor}{RGB}{148, 0, 211}  
\definecolor{islColor}{RGB}{0, 0, 0}  

\begin{figure}[h]
\centering
\begin{tikzpicture}[scale=0.95]
    \def\rowspacing{1.5}
    \def\paperspacing{1.5}
    \def\startx{0.5}

    \node[anchor=north east] at (\startx+10*\paperspacing-0.5,10.5) {
      \begin{tikzpicture}
        \node[font=\tiny, anchor=west] at (-7,0) {\textbf{Test Time Scaling Methods}};
        \node[right, font=\tiny, text=parallelColor] at (-3.9,0) {\textbf{Parallel Sampling}};
        \node[right, font=\tiny, text=treeSearchColor] at (-1,0) {\textbf{Tree Search}};
        \node[right, font=\tiny, text=correctionColor] at (2,0) {\textbf{Multi-turn Correction}};
        \node[right, font=\tiny, text=longCoTColor] at (5,0) {\textbf{Long CoT}};
        \node[right, font=\tiny, text=longCoTColor] at (6,0) {\textit{(\textbf{CoT})}};
        \node[font=\tiny, anchor=west, text width=3cm] at (-7,-0.5) {\textbf{Training Strategy}};
        \node[diamond, fill=black, inner sep=1.5pt] at (-3.9,-0.5) {}; \node[right, font=\tiny] at (-3.85,-0.52) {SFT};
        \node[regular polygon, regular polygon sides=3, fill=black, inner sep=1.5pt] at (-1,-0.5) {}; \node[right, font=\tiny] at (-0.95,-0.52) {DPO};
        \node[regular polygon, regular polygon sides=4, fill=black, inner sep=1.5pt] at (2,-0.5) {}; \node[right, font=\tiny] at (2.05,-0.55) {RL};
        \node[circle, regular polygon sides=4, fill=black, inner sep=1.5pt] at (5,-0.5) {}; \node[right, font=\tiny] at (5.05,-0.55) {Inference-only};
      \end{tikzpicture}
    };

\draw[thick] (0,8) -- (\startx+10.5*\paperspacing,8);
\draw[thick] (\startx+10.5*\paperspacing,8) -- (\startx+10.5*\paperspacing,8-\rowspacing);
\draw[thick,->] (\startx+10.5*\paperspacing,8-\rowspacing) -- (\startx+4*\paperspacing,8-\rowspacing);

    \filldraw[parallelColor] (\startx+0*\paperspacing,8) circle (2pt)
        node[above, text width=1.8cm, align=center, font=\tiny] {SelfCheckGPT \\\citenumber{manakul2023selfcheckgpt}}
        node[below=2pt, font=\tiny] {2023/03};

    \filldraw[correctionColor] (\startx+1*\paperspacing,8) circle (2pt)
        node[above, text width=1.8cm, align=center, font=\tiny] {Improve\\ Factuality \\\citenumber{du2023improvingfactualityreasoninglanguage}}
        node[below=2pt, font=\tiny] {2023/05};

    \filldraw[treeSearchColor] (\startx+2*\paperspacing,8) node[regular polygon, regular polygon sides=4, fill=treeSearchColor, inner sep=1.5pt] {}
        node[above, text width=1.8cm, align=center, font=\tiny] {C-MCTS \\\citenumber{dinesh2023cmcts}}
        node[below=2pt, font=\tiny] {2023/05};

    \filldraw[correctionColor] (\startx+3*\paperspacing,8) circle (2pt)
        node[above, text width=1.8cm, align=center, font=\tiny] {Chain-of-\\ Verification \\\citenumber{dhuliawala2023chainofverificationreduceshallucinationlarge}}
        node[below=2pt, font=\tiny] {2023/09};

    \filldraw[correctionColor] (\startx+4*\paperspacing,8) node[diamond, fill=correctionColor, inner sep=1.5pt] {}
        node[above, text width=1.8cm, align=center, font=\tiny] {DebateGPT \\\citenumber{subramaniam2024debategpt}}
        node[below=2pt, font=\tiny] {2023/09};

    \filldraw[treeSearchColor] (\startx+5*\paperspacing,8) circle (2pt)
        node[above, text width=1.8cm, align=center, font=\tiny]{PPO-MCTS \\\citenumber{liuDontThrowAway2024}}
        node[below=2pt, font=\tiny] {2023/09};

    \filldraw[correctionColor] (\startx+6*\paperspacing,8) circle (2pt)
        node[above, text width=1.8cm, align=center, font=\tiny] {MART \\\citenumber{ge2024mart}}
        node[below=2pt, font=\tiny] {2023/11};

    \filldraw[correctionColor] (\startx+7*\paperspacing,8) circle (2pt)
        node[above, text width=1.8cm, align=center, font=\tiny] {Combat\\ Adversarial \\\citenumber{chern2024combatadversarial}}
        node[below=2pt, font=\tiny] {2024/01};

    \filldraw[treeSearchColor] (\startx+8*\paperspacing,8) circle (2pt)
        node[above, text width=1.8cm, align=center, font=\tiny] {ARGS \\\citenumber{khanov2024args}}
        node[below=2pt, font=\tiny] {2024/02};

    \filldraw[longCoTColor] (\startx+9*\paperspacing,8) node[diamond, fill=longCoTColor, inner sep=1.5pt] {}
        node[above, text width=1.8cm, align=center, font=\tiny] {\textit{MoTE} \\\citenumber{liu2024mixture}}
        node[below=2pt, font=\tiny] {2024/05};

    \filldraw[correctionColor] (\startx+10*\paperspacing,8) circle (2pt)
        node[above, text width=1.8cm, align=center, font=\tiny] {Multi-expert\\ Prompting \\\citenumber{long2024multiexpert}}
        node[below=2pt, font=\tiny] {2024/11};

    \filldraw[longCoTColor] (\startx+10*\paperspacing,8-\rowspacing) node[regular polygon, regular polygon sides=4, fill=longCoTColor, inner sep=1.5pt] {}
        node[above, text width=1.8cm, align=center, font=\tiny] {Deliberate\\ Alignment \\\citenumber{guan2025deliberate}}
        node[below=2pt, font=\tiny] {2024/12};

    \filldraw[treeSearchColor] (\startx+9*\paperspacing,8-\rowspacing) node[diamond, fill=treeSearchColor, inner sep=1.5pt] {}
        node[above, text width=1.8cm, align=center, font=\tiny] {HaluSearch \\\citenumber{cheng2025halusearch}}
        node[below=2pt, font=\tiny] {2025/01};

    \filldraw[treeSearchColor] (\startx+8*\paperspacing,8-\rowspacing) circle (2pt)
        node[above, text width=1.8cm, align=center, font=\tiny] {InferenceGuard \\\citenumber{ji2025inferenceguard}}
        node[below=2pt, font=\tiny] {2025/02};

        \filldraw[treeSearchColor] (\startx+7*\paperspacing,8-\rowspacing) node[regular polygon, regular polygon sides=3, fill=treeSearchColor, inner sep=1.5pt] {}
        node[above, text width=1.8cm, align=center, font=\tiny] {STAIR \\\citenumber{zhang2025stair}}
        node[below=2pt, font=\tiny] {2025/02};

        \filldraw[parallelColor] (\startx+6*\paperspacing,8-\rowspacing) node[diamond, fill=parallelColor, inner sep=1.5pt] {}
        node[above, text width=1.8cm, align=center, font=\tiny] {SRG \\\citenumber{wang2025SRG}}
        node[below=2pt, font=\tiny] {2025/02};

    \filldraw[longCoTColor] (\startx+5*\paperspacing,8-\rowspacing) node[diamond, fill=longCoTColor, inner sep=1.5pt] {}
        node[above, text width=1.8cm, align=center, font=\tiny] {SafeChain \\\citenumber{jiang2025safechainsafetylanguagemodels}}
        node[below=2pt, font=\tiny] {2025/02};

\end{tikzpicture}
\caption{Works of applying test-time scaling methods in the safety field.}
\label{fig:timeline-wrapped-safety}
\end{figure}
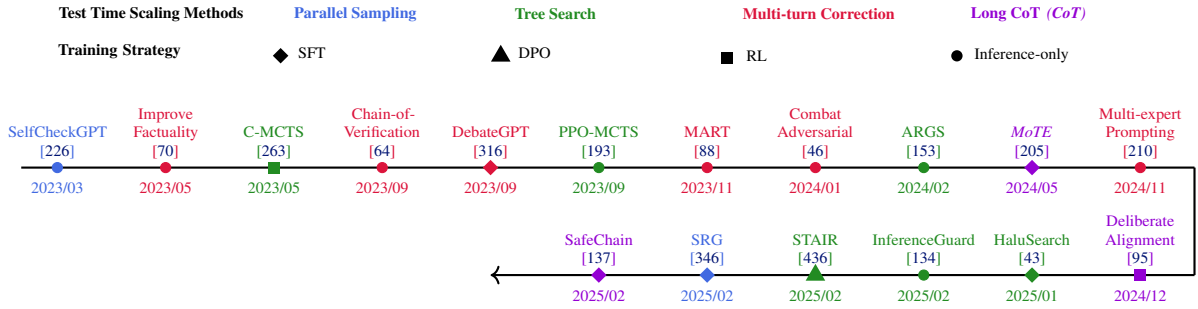

The advancement of AI systems with the ability to conduct complex, long-horizon reasoning as a result of test-time scaling carries significant implications for AI safety~\cite{bengio2023FAQ, park2024deception}. The consequences are two-fold: AI systems that evolve to successfully solve highly complex problems can help address and identify emerging safety concerns themselves, potentially detecting and mitigating risks that would otherwise remain unnoticed by humans, which allows for more thorough exploration of edge cases and vulnerabilities \cite{openai_o1_system_card}. However, these same AI systems may also explore decisions that exceed human cognitive limits, potentially pursuing strategies misaligned with human values that could cause catastrophic consequences due to unmanageable, prolonged exchanges \cite{hendrycks2023catastrophic}. Below, we address how scaling test-time thinking has helped with identifying and mitigating safety risks (such as hallucination, jailbreaks, adversarial attacks, and more) in LLMs.

Recent research has increasingly explored parallel sampling as a mechanism for improving safety reasoning in LLMs, particularly in settings where retraining or internal access is impractical. The core insight is that sampling multiple responses to the same input enables more reliable estimation of factuality, uncertainty, and safety alignment. SelfCheckGPT \citep{manakul2023selfcheckgpt} introduces a zero-resource hallucination detection method that uses sampling to detect factual inconsistencies across generations, under the assumption that reliable knowledge yields consistent responses. Similarly, \citet{lin2024generatingwithconfidence} formalize this principle through the lens of semantic dispersion, showing that uncertainty estimates based on the diversity of sampled responses can reliably predict the trustworthiness of a model's output, even under black-box access. While these methods focus on sampling in their frameworks, SRG \citep{wang2025SRG} proposes injecting explicit reasoning steps guided by predefined safety policies into the models. Conducting evaluation with Best-of-N sampling, they demonstrate improvements in generalization performance against OOD attacks. These approaches exemplify a powerful paradigm of test-time scaling: the ability to amplify alignment, robustness, and factual reliability by operating over multiple outputs during inference, rather than modifying the underlying model weights. Parallel sampling thus serves as a latent knowledge elicitation strategy that decouples safety improvements from expensive and infrastructure-heavy training processes, suggesting a scalable path forward for enhancing trustworthiness in real-world LLM deployments.

Apart from parallel sampling strategies, tree search-based methods offer a structured approach to test-time safety by enabling deliberate exploration of alternative outputs or reasoning paths. Techniques such as InferenceGuard \cite{ji2025inferenceguard} and ARGS \cite{khanov2024args} frame generation as a reward-constrained search process, steering outputs toward safety-aligned completions using constrained Markov decision process within the
LLM’s latent space or learned reward models. Other methods like HaluSearch \cite{cheng2025halusearch} improve factual reliability and reasoning robustness by explicitly searching over intermediate thought steps and scoring them using self-evaluation or heuristics (deciding when models should ``think slower'' versus fallback to fast generation). Planning-focused approaches such as C-MCTS \cite{dinesh2023cmcts} and STAIR \cite{zhang2025stair} integrate safety critics or introspective mechanisms into MCTS to avoid unsafe action sequences during decision making. These methods enhance test-time alignment by enabling backtracking and surfacing interpretable intermediate decision steps—key advantages for maintaining safety as models scale.

Building on this foundation, recent research has also explored how increasing interaction steps during inference—through multi-turn correction or multi-agent collaboration—can further improve safety and robustness.
Several multi-agent interaction frameworks~\cite{du2023improvingfactualityreasoninglanguage,ge2024mart, chern2024combatadversarial, long2024multiexpert} have proven effective in reducing harmful outputs by enabling models to critique and refine responses collaboratively to identify and mitigate potential safety risks, such as reducing hallucination, toxicity, and adversarial attacks. Further, encouraging divergent thinking through structured debate~\cite{liang2024encouragingdivergentthinkinglarge} enhances reliability by promoting nuanced reasoning and cross-validation of outputs. Additionally, recent work leverages long CoT reasoning to enhance safety in LLMs by extending the depth of model deliberation at test time. SafeChain \cite{jiang2025safechainsafetylanguagemodels} introduces a training dataset of long-form safety-aligned reasoning and shows that LLMs fine-tuned on these trajectories can maintain high reasoning ability while improving refusal rates and reduces harmful content. Deliberative Alignment \cite{guan2025deliberate} trains models to explicitly consult and reason over safety policies via multi-step CoT before responding, demonstrating models that ``think before speaking'' become safer as reasoning depth increases—linking CoT length directly to improved alignment under test-time scaling. Similarly, Chain-of-Verification \cite{dhuliawala2023chainofverificationreduceshallucinationlarge} enhances factual safety by prompting the model to generate and answer verification questions about its own output, turning the reasoning process into a multi-phase, self-auditing chain—which becomes more reliable as models grow and can handle longer CoT. Lastly, MoTE \cite{liu2024mixture} decomposes safety reasoning into specialized CoT stages (question analysis, guidance, answer, and checking), assigning expert modules to each; this modular approach benefits from larger models and longer reasoning chains, enabling more scalable and interpretable self-alignment.

These studies collectively suggest that strategically increasing interaction and reasoning steps at inference time offers a scalable path to safer and more robust model behavior that doesn't require retraining.

\begin{AIbox}{Future Direction for Safety}
\begin{itemize}
    \item Investigate how test-time scaling methods can be integrated with existing alignment strategies, such as RLHF and process-based supervision, to ensure they complement each other and understand how they can collectively enhance model safety.
    \item  Investigate the extent to which test-time scaling methods improve a model’s generalization to real-world safety challenges, including deception, adversarial attacks, and other out-of-distribution threats.
    \item Conduct rigorous testing to determine the resilience of long-CoT models against various jailbreaks and attacks. This includes analyzing whether increased test-time computation inadvertently introduces new vulnerabilities or mitigates existing ones. 
\end{itemize}
\end{AIbox}

\subsection{RAG}

\definecolor{parallelColor}{RGB}{65, 105, 225}  
\definecolor{treeSearchColor}{RGB}{34, 139, 34}  
\definecolor{correctionColor}{RGB}{220, 20, 60}  
\definecolor{longCoTColor}{RGB}{148, 0, 211}  
\definecolor{islColor}{RGB}{0, 0, 0}  

\begin{figure}[h]
\centering
\begin{tikzpicture}[scale=0.95]
    \def\rowspacing{1.5} 
    \def\paperspacing{1.5} 
    \def\papersperrow{11} 
    
    \def\startx{0.5} 
    \def\endx{\startx + 9.2*\paperspacing} 
    
    \node[anchor=north east] at (\endx-0.5,10.5) {
     \begin{tikzpicture}
            \node[font=\tiny, anchor=west] at (-7,0) {\textbf{Test Time Scaling Methods}};
            
            \node[right, font=\tiny, text=parallelColor] at (-3.9,0) {\textbf{Parallel Sampling}};
            
            \node[right, font=\tiny, text=treeSearchColor] at (-1,0) {\textbf{Tree Search}};
            
            \node[right, font=\tiny, text=correctionColor] at (2,0) {\textbf{Multi-turn Correction}};
            
            \node[right, font=\tiny, text=longCoTColor] at (5,0) {\textbf{Long CoT}};

            \node[right, font=\tiny, text=longCoTColor] at (6,0) {\textit{(\textbf{CoT})}};
            
            \node[font=\tiny, anchor=west, text width=3cm] at (-7,-0.5) {\textbf{Training Strategy}};
            
            \node[diamond, fill=black, inner sep=1.5pt] at (-3.9,-0.5) {};
            \node[right, font=\tiny] at (-3.85,-0.52) {SFT};
            
            \node[regular polygon, regular polygon sides=3, fill=black, inner sep=1.5pt] at (-1,-0.5) {};
            \node[right, font=\tiny] at (-0.95,-0.52) {DPO};
            
            \node[regular polygon, regular polygon sides=4, fill=black, inner sep=1.5pt] at (2,-0.5) {};
            \node[right, font=\tiny] at (2.05,-0.55) {RL};

            \node[circle, regular polygon sides=4, fill=black, inner sep=1.5pt] at (5,-0.5) {};
            \node[right, font=\tiny] at (5.05,-0.55) {Inference-only};
            
        \end{tikzpicture}    };
    
    \draw[thick] (0,8) -- (\endx+0.1,8);
    
    \draw[thick, ->] (\endx+0.1,8) -- (\endx+0.5,8);
    
    
    \filldraw[longCoTColor] (\startx,8) circle (2pt) 
        node[above, text width=1.8cm, align=center, font=\tiny] {\textit{IterDRAG} \\\citenumber{yue2024inferencescalinglongcontextretrieval}}
        node[below, font=\tiny] {2024/10};

    \filldraw[longCoTColor] (\startx+1*\paperspacing,8) circle (2pt) 
        node[above, text width=1.8cm, align=center, font=\tiny] {\textit{Plan*RAG} \\\citenumber{verma2025planragefficienttesttimeplanning}}
        node[below, font=\tiny] {2024/10};

    \filldraw[longCoTColor] (\startx+2*\paperspacing,8) node[diamond, fill=longCoTColor, inner sep=1.5pt] {}  
        node[above, text width=1.8cm, align=center, font=\tiny] {\textit{Auto-RAG} \\\citenumber{yu2024autoragautonomousretrievalaugmentedgeneration}}
        node[below, font=\tiny] {2024/11};

    \filldraw[longCoTColor] (\startx+3*\paperspacing,8) circle (2pt) 
        node[above, text width=1.8cm, align=center, font=\tiny] {Search-o1 \\\citenumber{li2025searcho1agenticsearchenhancedlarge}}
        node[below, font=\tiny] {2025/01};

    \filldraw[treeSearchColor] (\startx+4*\paperspacing,8) circle (2pt) 
        node[above, text width=1.8cm, align=center, font=\tiny] {AirRAG \\\citenumber{feng2025airragactivatingintrinsicreasoning}}
        node[below, font=\tiny] {2025/01};
    
    \filldraw[parallelColor] (\startx+5*\paperspacing,8) node[diamond, fill=parallelColor, inner sep=1.5pt] {} 
        node[above, text width=1.8cm, align=center, font=\tiny] {CoRAG \\\citenumber{wang2025chainofretrievalaugmentedgeneration}}
        node[below, font=\tiny] {2025/01};
    
    \filldraw[longCoTColor] (\startx+6*\paperspacing,8) node[regular polygon, regular polygon sides=4, fill=longCoTColor, inner sep=1.5pt] {} 
        node[above, text width=1.8cm, align=center, font=\tiny] {DeepRAG \\\citenumber{guan2025deepragthinkingretrievalstep}}
        node[below, font=\tiny] {2025/02};

    \filldraw[longCoTColor] (\startx+7*\paperspacing,8) node[regular polygon, regular polygon sides=4, fill=longCoTColor, inner sep=1.5pt] {} 
        node[above, text width=1.8cm, align=center, font=\tiny] {R1-Searcher \\\citenumber{song2025r1}}
        node[below, font=\tiny] {2025/03};
    
    \filldraw[longCoTColor] (\startx+8*\paperspacing,8) node[regular polygon, regular polygon sides=4, fill=longCoTColor, inner sep=1.5pt] {} 
        node[above, text width=1.8cm, align=center, font=\tiny] {Search-R1 \\\citenumber{jin2025searchr1trainingllmsreason}}
        node[below, font=\tiny] {2025/03};

    \filldraw[longCoTColor] (\startx+9*\paperspacing,8) node[regular polygon, regular polygon sides=4, fill=longCoTColor, inner sep=1.5pt] {} 
        node[above, text width=1.8cm, align=center, font=\tiny] {ReSearch \\\citenumber{chen2025research}}
        node[below, font=\tiny] {2025/03};
\end{tikzpicture}
\caption{Works of applying test-time scaling methods in the RAG field.}
\label{fig:timeline-wrapped-rag}
\end{figure}
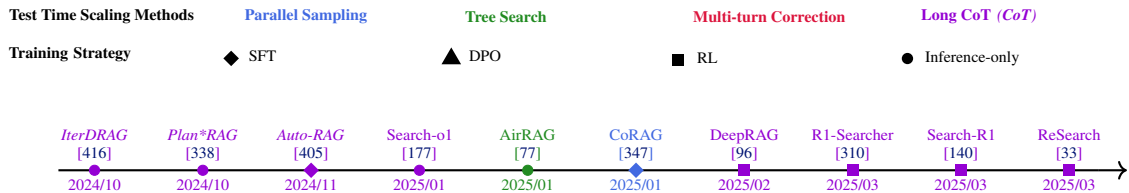
Retrieval-Augmented Generation (RAG) systems enhance LLMs by incorporating external knowledge sources, producing responses that are more factual and contextually grounded. Despite their effectiveness, these systems often struggle when confronted with complex queries requiring multi-hop reasoning across multiple documents. Test-time scaling has emerged as a promising approach to strengthen the reasoning capabilities of RAG systems by strategically allocating additional computational resources during inference.

\citet{yue2024inferencescalinglongcontextretrieval} introduce IterDRAG, which methodically decomposes complex questions into sequential sub-queries and conducts iterative search and reasoning processes to construct comprehensive answers. Their research demonstrates a near-linear relationship between RAG performance and effective context length, establishing a clear test-time scaling law for RAG systems. The effectiveness of this agentic workflow has been further validated in several follow-up studies \cite{verma2025planragefficienttesttimeplanning, guan2025deepragthinkingretrievalstep, li2025searcho1agenticsearchenhancedlarge, feng2025airragactivatingintrinsicreasoning, yu2024autoragautonomousretrievalaugmentedgeneration, wang2025chainofretrievalaugmentedgeneration}.
Beyond prompt-based agents, researchers have developed approaches to fine-tune LLMs for end-to-end interleaved reasoning and search capabilities. One approach for training data collection involves synthesizing reasoning and search trajectories through rejection sampling, followed by model training via supervised fine-tuning~\cite{wang2025chainofretrievalaugmentedgeneration, yu2024autoragautonomousretrievalaugmentedgeneration} or preference fine-tuning~\cite{guan2025deepragthinkingretrievalstep}. More recently, RL has been applied for end-to-end training of RAG, with works such as R1-Searcher~\cite{song2025r1}, Search-R1~\cite{jin2025searchr1trainingllmsreason}, and ReSearch~\cite{chen2025research}. These approaches enable models to learn more efficient search strategies through trial and error rather than imitating human-designed search patterns. However, a significant limitation is that these works primarily focus on open-domain question answering tasks and rely on rule-based rewards designed for short, factual answers, which may not generalize well to more complex reasoning tasks requiring detailed explanations.

For future directions, an important aspect lies in developing more sophisticated reward functions that can evaluate long-form generation rather than just short factual answers for RL training. Moreover, current benchmarks do not separately assess the benefits of internal reasoning and external search in RAG systems, highlighting the importance of developing specialized evaluation frameworks for RAG in this new era.

\begin{AIbox}{Future Direction for RAG}
\begin{itemize}
    \item Develop more sophisticated reward functions for RL that can effectively evaluate long-form generation rather than focusing solely on short factual answers.
    \item Create specialized evaluation frameworks that isolate and measure the distinct contributions of internal reasoning versus external search in RAG systems.
\end{itemize}
\end{AIbox}

\subsection{Evaluation}

\definecolor{parallelColor}{RGB}{65, 105, 225}  
\definecolor{treeSearchColor}{RGB}{34, 139, 34}  
\definecolor{correctionColor}{RGB}{220, 20, 60}  
\definecolor{longCoTColor}{RGB}{148, 0, 211}  
\definecolor{islColor}{RGB}{0, 0, 0}  

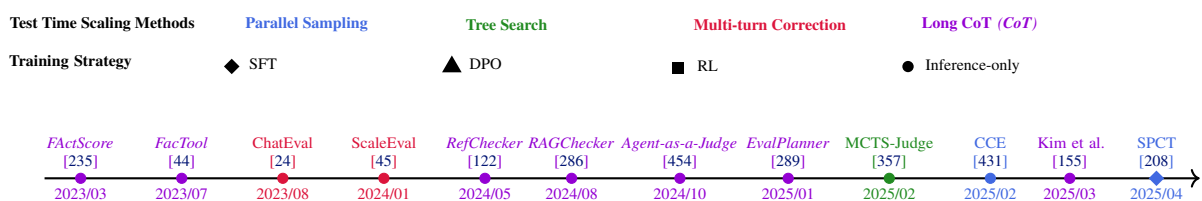
\begin{figure}[h]
\centering
\begin{tikzpicture}[scale=0.95]
    \def\rowspacing{1.5} 
    \def\paperspacing{1.4} 
    \def\papersperrow{11} 
    
    \def\startx{0.5} 
    \def\endx{\startx + 10*\paperspacing} 
    
    \node[anchor=north east] at (\endx-0.5,10.5) {
       \begin{tikzpicture}
            \node[font=\tiny, anchor=west] at (-7,0) {\textbf{Test Time Scaling Methods}};
            
            \node[right, font=\tiny, text=parallelColor] at (-3.9,0) {\textbf{Parallel Sampling}};
            
            \node[right, font=\tiny, text=treeSearchColor] at (-1,0) {\textbf{Tree Search}};
            
            \node[right, font=\tiny, text=correctionColor] at (2,0) {\textbf{Multi-turn Correction}};
            
            \node[right, font=\tiny, text=longCoTColor] at (5,0) {\textbf{Long CoT}};

            \node[right, font=\tiny, text=longCoTColor] at (6,0) {\textit{(\textbf{CoT})}};
            
            \node[font=\tiny, anchor=west, text width=3cm] at (-7,-0.5) {\textbf{Training Strategy}};
            
            \node[diamond, fill=black, inner sep=1.5pt] at (-3.9,-0.5) {};
            \node[right, font=\tiny] at (-3.85,-0.52) {SFT};
            
            \node[regular polygon, regular polygon sides=3, fill=black, inner sep=1.5pt] at (-1,-0.5) {};
            \node[right, font=\tiny] at (-0.95,-0.52) {DPO};
            
            \node[regular polygon, regular polygon sides=4, fill=black, inner sep=1.5pt] at (2,-0.5) {};
            \node[right, font=\tiny] at (2.05,-0.55) {RL};

            \node[circle, regular polygon sides=4, fill=black, inner sep=1.5pt] at (5,-0.5) {};
            \node[right, font=\tiny] at (5.05,-0.55) {Inference-only};
            
        \end{tikzpicture}
    };
    
    \draw[thick, ->] (0,8) -- (\endx+1.5,8);
    
    
    
    \filldraw[longCoTColor] (\startx,8) circle (2pt) 
        node[above, text width=1.8cm, align=center, font=\tiny] {\textit{FActScore} \\\citenumber{min2023factscorefinegrainedatomicevaluation}}
        node[below, font=\tiny] {2023/03};

    \filldraw[longCoTColor] (\startx+\paperspacing,8) circle (2pt) 
        node[above, text width=1.8cm, align=center, font=\tiny] {\textit{FacTool} \\\citenumber{chern2023factoolfactualitydetectiongenerative}}
        node[below, font=\tiny] {2023/07};

    \filldraw[correctionColor] (\startx+2*\paperspacing,8) circle (2pt) 
        node[above, text width=1.8cm, align=center, font=\tiny] {ChatEval \\\citenumber{chan2023chateval}}
        node[below, font=\tiny] {2023/08};

    \filldraw[correctionColor] (\startx+3*\paperspacing,8) circle (2pt) 
        node[above, text width=1.8cm, align=center, font=\tiny] {ScaleEval \\\citenumber{chern2024scaleeval}}
        node[below, font=\tiny] {2024/01};

    \filldraw[longCoTColor] (\startx+4*\paperspacing,8) circle (2pt) 
        node[above, text width=1.8cm, align=center, font=\tiny] {\textit{RefChecker} \\\citenumber{hu-etal-2024-knowledge}}
        node[below, font=\tiny] {2024/05};

    \filldraw[longCoTColor] (\startx+5*\paperspacing - 0.2,8) circle (2pt) 
        node[above, text width=1.8cm, align=center, font=\tiny] {\textit{RAGChecker} \\\citenumber{ru2024ragchecker}}
        node[below, font=\tiny] {2024/08};
    
    \filldraw[longCoTColor] (\startx+6*\paperspacing - 0.1,8) circle (2pt) 
        node[above, text width=1.8cm, align=center, font=\tiny] {\textit{Agent-as-a-Judge} \\\citenumber{zhuge2024agentasajudgeevaluateagentsagents}}
        node[below, font=\tiny] {2024/10};
    
    \filldraw[longCoTColor] (\startx+7*\paperspacing,8) circle (2pt) 
        node[above, text width=1.8cm, align=center, font=\tiny] {\textit{EvalPlanner} \\\citenumber{saha2025learningplanreason}}
        node[below, font=\tiny] {2025/01};

    \filldraw[treeSearchColor] (\startx+8*\paperspacing,8) circle (2pt) 
        node[above, text width=1.8cm, align=center, font=\tiny] {MCTS-Judge \\\citenumber{wang2025mctsjudgetesttimescalingllmasajudge}}
        node[below, font=\tiny] {2025/02};

    \filldraw[parallelColor] (\startx+9*\paperspacing,8) circle (2pt) 
        node[above, text width=1.8cm, align=center, font=\tiny] {CCE \\\citenumber{zhang2025crowdcomparativereasoningunlocking}}
        node[below, font=\tiny] {2025/02};

    \filldraw[longCoTColor] (\startx+10*\paperspacing - 0.3,8) circle (2pt) 
        node[above, text width=1.8cm, align=center, font=\tiny] {Kim et al. \\\citenumber{kim2025scalingevaluationtimecomputereasoning}}
        node[below, font=\tiny] {2025/03};

    \filldraw[parallelColor] (\startx+11*\paperspacing - 0.5,8) node[diamond, fill=parallelColor, inner sep=1.5pt] {}
        node[above, text width=1.8cm, align=center, font=\tiny] {SPCT \\\citenumber{liu2025inferencetimescalinggeneralistreward}}
        node[below, font=\tiny] {2025/04};
        
\end{tikzpicture}
\caption{Works of applying test-time scaling methods in the evaluation field.}
\label{fig:timeline-wrapped-evaluation}
\end{figure}
The LLM-as-a-Judge paradigm~\cite{zheng2023judging, gu2025surveyllmasajudge} has transformed the evaluation of language model outputs, shifting away from rule-based metrics like BLEU and ROUGE toward more human-like assessment of generated content. Recent research demonstrates that allocating additional computational resources during inference significantly enhances evaluation quality through several approaches. These include fine-grained evaluation that decomposes the LLM's response and examines it step by step~\cite{chern2023factoolfactualitydetectiongenerative,min2023factscorefinegrainedatomicevaluation,hu-etal-2024-knowledge,ru2024ragchecker}, structuring the CoT in LLM-as-a-Judge into distinct planning and execution phases~\cite{saha2025learningplanreason}, employing multi-agent systems to provide intermediate feedback~\cite{zhuge2024agentasajudgeevaluateagentsagents}, and utilizing parallel sampled crowd responses for more reliable pairwise comparisons~\cite{zhang2025crowdcomparativereasoningunlocking}. Furthermore, MCTS-Judge~\cite{wang2025mctsjudgetesttimescalingllmasajudge} applies MCTS to systematically explore different evaluation perspectives for code assessment, demonstrating the scalability of this method where increasing search depth and rollouts consistently improves accuracy.~\citet{kim2025scalingevaluationtimecomputereasoning} also observe that generating more reasoning tokens leads to better performance of the evaluators for long-CoT models.

As AI development progresses toward more realistic real-world tasks such as software development which comprises multiple subtasks, current evaluation frameworks increasingly focus on complex agents and workflows~\cite{jimenez2024swebench,xie2024osworldbenchmarkingmultimodalagents}. Future research directions should emphasize enhancing the reliability of evaluation for these complex and long-horizon tasks, which require strategies to balance the benefits of test-time scaling and evaluation speed.

\begin{AIbox}{Future Direction for Evaluation}
\begin{itemize}
    \item Enhance the reliability of evaluation for complex and long-horizon tasks.
    \item Integrate evaluation frameworks with reward design in RL training to fully unleash the potential of RL.
\end{itemize}
\end{AIbox}

\section{So What? -- From Scaling to Cognitive Intelligence} \label{sec:implications}

The emergence of cognition engineering through test-time scaling marks a fundamental paradigm shift in artificial intelligence. Far beyond mere technical implementation, this transformation carries profound implications for how we develop AI systems, reimagine human-AI collaboration, and conduct scientific research. 

\subsection{Data Engineering 2.0: Cognition Data Engineering}
Traditional AI has primarily focused on knowledge acquisition—training systems on the outputs of human thinking. Cognition engineering, however, demands something fundamentally different: a shift from products of thought to thought processes themselves. This transition gives rise to a new discipline—\textbf{\emph{cognitive data engineering}}—which revolutionizes our understanding of what constitutes valuable training data.

Cognitive data flows from three distinct yet complementary sources, each bringing unique advantages and challenges to the development process:

\paragraph{Source 1: Human cognition projections} Despite lacking direct brain-computer interfaces to capture human thought processes directly, we can access projections of human cognition in the physical world:

\begin{itemize*}
\item \textbf{Directly recorded artifacts.} Video recordings of expert problem-solving sessions, think-aloud protocols, and detailed research journals capture cognitive processes as they unfold. These records preserve not just solutions but the messy reality of expert thinking—false starts, revisions, and breakthroughs.
\item \textbf{Tool-mediated cognitive traces.} Complex cognitive activities leave traces in specialized tools—laboratory notebooks, collaborative whiteboarding sessions, version control systems in software development, and the progressive refinement of scientific papers through drafts and revisions. These tools serve as proxies that make implicit cognitive processes explicit and observable.

\item \textbf{Frontier expertise extraction.} The most valuable cognitive patterns often reside in the minds of domain experts at the cutting edge of their fields. These patterns require carefully designed elicitation methods—specialized interviewing techniques, tailored problem scenarios, and high-quality interactions that can distill tacit knowledge into explicit reasoning trajectories.
\end{itemize*}

\paragraph{Source 2: AI-generated cognition} Through sophisticated reinforcement learning approaches with proper reward mechanisms, AI systems can now generate valuable cognitive data or trajectories in an environment independently:

\begin{itemize*}
\item \textbf{Environment-reward synergy.} When provided with well-designed environments, appropriate reward functions, and strong initialization models, AI systems can discover novel cognitive strategies through extended exploration. These strategies may differ substantially from human approaches while achieving equal or superior results—similar to AlphaGo's famous ``\texttt{Move 37}'' that initially puzzled human experts but ultimately proved highly effective.

\item \textbf{Self-play and adversarial discovery.} Systems can generate increasingly sophisticated cognitive data by competing against themselves or confronting progressively more challenging scenarios, developing reasoning strategies that might never emerge through imitation of human examples alone.

\item \textbf{Scaling effects in cognitive discovery.} As computational resources increase, AI systems can explore cognitive pathways inaccessible to humans due to biological limitations in memory, attention span, or processing speed—potentially discovering novel problem-solving approaches in domains ranging from mathematics to drug design.

\end{itemize*}

\paragraph{Source 3: Human-AI collaborative generation} Perhaps most promising is the co-creation of cognitive data through human-AI partnership:

\begin{itemize*}
\item \textbf{Trajectory sampling and human filtering.} AI agents can generate diverse solution paths that human experts then evaluate and refine, combining machine-generated diversity with human judgment about quality and relevance.

\item \textbf{Human seeding with AI expansion.} Human experts can provide initial examples of sophisticated reasoning in challenging domains, which AI systems then conduct \emph{cognitive completion} (i.e., expand, systematically vary, and complete)—creating training datasets far larger than what human annotation alone could achieve~\cite{he2024pcagentsleepai}.

\item \textbf{Iterative refinement cycles.} Human and AI contributions can alternate in progressive cycles, with each building upon and enhancing the other's work—humans providing creative leaps or conceptual reframings, AI supplying systematic exploration of implications and edge cases.

\end{itemize*}

This cognition data establishes an entirely new category of digital resource with the potential to drive AI capabilities beyond what either natural data collection or synthetic generation alone could achieve. The resulting cognitive data repositories will likely become as strategically valuable as large-scale computing resources in determining leadership in AI advancement.

\subsection{Reward \& Environment Engineering}
The transition to cognition engineering fundamentally reshapes how we design the environments in which AI systems operate and the reward signals that guide their development.

\subsubsection{Reward Models Design}

A critical trend in cognition engineering is the increasing difficulty of reward verification as tasks grow more complex.
Mathematical Olympiad problems represent relatively straightforward verification—proofs either follow logically or don't. Moving beyond this, tasks like Deep Research~\cite{deepresearch} and PaperBench~\cite{starace2025paperbenchevaluatingaisability} demand sophisticated planning while producing outputs that resist objective assessment.
Scientific discovery and literary creation occupy the frontier of verification complexity, requiring creativity and unique insights whose value assessment defies complete objectification. These domains involve aesthetic judgments, originality considerations, and contextual relevance that vary across cultures and perspectives.
To address these challenges, we propose two complementary approaches:
\textbf{Reference-based assessment}: A method inspired by text summarization and machine translation evaluation approaches that compares outputs to high-quality exemplars, utilizing human judgment~\cite{bhandari-etal-2020-evaluating} capabilities without necessitating formal quality definitions~\cite{qin2022t5score,yuan2021bartscore,zheng2025deepresearcherscalingdeepresearch}.
\textbf{Criterion-based assessment}: Establishing structured frameworks that decompose subjective judgments into concrete components with specific criteria~\cite{fu2023gptscore,yuan2024llmcrit}.
Together, these methodologies facilitate the creation of effective reward signals in domains where simple correctness metrics prove insufficient.

\subsubsection{Cognitive Environment Design}

Environments for AI cognitive development span a complexity spectrum—from text-only interactions~\cite{song2025r1} to code interpretation~\cite{li2025torlscalingtoolintegratedrl}, browser environments~\cite{zheng2025deepresearcherscalingdeepresearch}, full computer system access~\cite{anthropic2024models}, and physical world interactions.
Within this framework, specialized cognitive simulators develop domain-specific reasoning: scientific discovery environments mimicking the hypothesis-experiment-analysis cycle; legal reasoning arenas requiring statute interpretation and precedent application; medical simulators demanding differential diagnosis under uncertainty.
Adversarial frameworks strengthen cognitive abilities through structured debate platforms, red-teaming simulations, and Socratic dialogues. These environments should form cognitive curricula that systematically develop capabilities from foundations to frontiers.

\subsection{Human-AI Cognitive Partnership}

The emergence of test-time scaling and cognition engineering fundamentally transforms the relationship between human and artificial intelligence, creating possibilities for collaboration that transcend traditional tool-user dynamics. This partnership represents a genuinely new kind of cognitive ecosystem with profound implications.

\paragraph{Bidirectional Cognitive Exchange} Humans transmit thinking strategies through expert demonstrations, metacognitive guidance, and value-aligned reasoning examples that shape how AI systems approach problems, while AI systems reciprocally illuminate blind spots, expand strategy repertoires, and offer novel conceptual frameworks that enhance human thinking. These partnerships leverage the distinctive strengths of each intelligence type, creating a genuine cognitive complementarity that transcends both \emph{\textbf{AI-as-tool}} and \emph{\textbf{AI-as-replacement}} paradigms.

\paragraph{Cognitive Amplification} AI systems extend human working memory by managing details, maintaining consistency, and preserving context throughout complex reasoning processes, while expanding exploration breadth through parallel hypothesis evaluation, exhaustive implication tracing, and counterfactual scenario generation beyond human capacity. This partnership also enables cognitive process externalization through reasoning visualization, assumption surfacing, and inference chain verification that make thinking observable and manipulable, overcoming fundamental human cognitive limitations.

\paragraph{New Interaction Paradigms} Shared workspaces allow humans and AI to visualize problems together using multiple formats (diagrams, text, symbols) and comment on each other's work in real-time, creating a recorded history of their thinking process. Communication becomes adaptive - AI explanations adjust to the user's needs, revealing more details only when necessary and using concepts familiar to the specific user. Tools are designed not just for solving problems but for mutual improvement: humans can ask for clearer explanations from AI, provide targeted feedback on the AI's reasoning, and both partners can review completed work to learn from successes and mistakes. These approaches transform thinking from a solitary activity into a truly collaborative process where ideas are continuously shared, refined, and built upon together.

This evolution of human-AI cognitive partnership represents far more than incremental improvement in AI assistants—it constitutes a fundamentally new kind of intellectual relationship with the potential to dramatically enhance humanity's collective problem-solving capacity. The most profound impacts may emerge in domains where problems exceed the cognitive capacity of individual humans but require the contextual understanding and value alignment that purely artificial approaches struggle to achieve.

\subsection{Research Acceleration}

The application of cognition engineering to scientific research promises to fundamentally transform the pace and nature of discovery across disciplines. By complementing human scientific thinking with machine reasoning capabilities, we stand at the threshold of potentially unprecedented acceleration in knowledge creation.

Traditional science progresses slowly because humans can only generate and test limited hypotheses. Cognition systems overcome these cognitive limitations by systematically mapping knowledge gaps, identifying unexplored connections, and generating thousands of possible explanations simultaneously~\cite{lu2024aiscientist}. They detect anomalies in massive datasets that might signal breakthrough opportunities and determine which experiments would most efficiently test competing theories. These capabilities dramatically expand both the questions scientists can ask and how quickly they can find answers. Additionally, specialized research has created isolated knowledge silos that impede progress. Cognition engineering bridges these divides by connecting findings across disciplines, time periods, and information formats. It identifies similarities between seemingly unrelated phenomena and translates specialized terminology to enable cross-field collaboration. This integration facilitates the emergence of hybrid disciplines and allows insights to transfer between domains, transforming isolated knowledge islands into a coherent scientific network where ideas flow freely.

This acceleration of scientific research through cognition engineering has implications far beyond mere efficiency gains. By removing \emph{cognitive bottlenecks} in hypothesis generation, literature integration, experimental design, and theory refinement, these approaches may enable humanity to address pressing challenges—from climate change to disease prevention to sustainable development—at unprecedented speed and scale. More profoundly, by democratizing participation in scientific discovery, cognition engineering could help realize the full creative potential of global human intelligence, bringing diverse perspectives to bear on our most significant questions.

\section{Infrastructure}\label{sec:infra}
In this section, we discuss the infrastructure of RL and MCTS briefly, two important technologies in cognition engineering. Besides these, the acceleration of long text generation is also important. We refer interested readers to the survey~\cite{liu2025comprehensivesurveylongcontext} for a comprehensive review.

\subsection{RL} \label{sec:infra:rl}

Taking the PPO algorithm as the example, a typical RL workflow consists of two primary steps:
\begin{itemize}
\item \textbf{Rollout:} Prepared prompts are fed into the policy model, which generates responses (i.e., rollout). This process requires the inference backend of the policy model. Although conventional training libraries support inference, they often exhibit slow decoding speeds when generating new sequences. To address this limitation, a dedicated inference backend (e.g., vLLM~\citep{kwon2023efficient} or SGLang~\citep{zheng2024sglangefficientexecutionstructured}) is typically deployed to accelerate rollout for the policy model.
\item \textbf{Model Update:} Given the generated sequences, policy log probabilities, value estimates, reference log probabilities, and rewards are computed using the policy, critic, reference model, and reward function, respectively. Additional values, such as KL divergence loss, return, and advantage, are then derived from these computations. These computed values are then used  to calculate the respective losses for the policy and critic models, which are then applied to update their parameters. The training backend for model updates typically employs frameworks such as DeepSpeed~\citep{rasley2020deepspeed} or Fully Sharded Data Parallel (FSDP)~\citep{zhao2023pytorchfsdpexperiencesscaling}.
\end{itemize}

Popular open-source frameworks for RL training include DeepSpeed-Chat, NeMo-Aligner~\cite{shen2024nemoaligner}, OpenRLHF~\cite{hu2024openrlhf}, and veRL~\cite{verl}. Among these, OpenRLHF and veRL are actively maintained frameworks that support RL training, with their key features summarized in Table~\ref{tabel:rlframework}. Due to the complex workflow involving multiple models described above, these RL frameworks implement different strategies for resource allocation and process scheduling. OpenRLHF, for instance, can allocate dedicated resources to each backend, ensuring operations within a given module remain confined to its designated resource pool. In contrast, veRL primarily employs a shared-resource approach, dynamically reclaiming resources from inactive modules when another requires them. OpenRLHF offers a more concise code implementation, making it more accessible for beginners to understand the algorithm, while veRL provides a relatively centralized programming interface with enhanced programmability.

Although these open-source frameworks facilitate RL training, large-scale RL training can still present significant challenges. For example, \citet{yeo2025demystifyinglongchainofthoughtreasoning} encountered substantial difficulties when attempting to scale the policy model to 32B parameters, ultimately determining that the required number of GPUs was prohibitively large. They observed low hardware utilization during the training process, an issue that is particularly exacerbated in long CoT scenarios due to the higher variance in CoT length, which leads to stragglers during inference. This highlights the ongoing need for optimization of RL frameworks specifically for these scenarios.

\begin{table}[tb!]
\tiny
\caption{Comparison between OpenRLHF and veRL frameworks.}
\label{tabel:rlframework}
\resizebox{0.99\textwidth}{!}{ 
\begin{tabular}{lcccccc}
    \toprule
    \textbf{Framework} & \textbf{Supported Algorithm} & \textbf{Hybrid Engine} & \textbf{Training Backend} & \textbf{Inference Engine} & \textbf{Sequence Parallelism} & \textbf{Multi-Modality} \\
    \midrule
    \textbf{OpenRLHF} & \makecell[c]{PPO, GRPO,\\ RLOO, REINFORCE++} & Supported & Deepspeed ZERO & vLLM, HF Transformers & Ring Attention & Supported \\
    \midrule
    \textbf{veRL} & \makecell[c]{PPO, GRPO, RLOO,\\ REINFORCE++, DAPO, PRIME} & Supported & Megatron-LM, FSDP & vLLM, SGLang, HF Transformers & DeepSpeed Ulysses
 & Supported \\
    \bottomrule
\end{tabular}
}
\end{table}

\subsection{MCTS} \label{sec:infra:mcts}

For the infrastructure of MCTS, several of the aforementioned works have released their code to provide a foundation for applying MCTS in LLMs~\cite{hao2024llmreasonersnewevaluation,haoReasoningLanguageModel2023,chen2024alphamathzeroprocesssupervision,fengAlphazerolikeTreeSearchCan2024}. Although these code repositories facilitate subsequent research, they often lack optimized acceleration strategies that consider hardware and software optimization, which limits large-scale MCTS deployment. We outline the key acceleration strategies as follows.

\paragraph{Speculative Decoding} Speculative Decoding employs a small draft model to generate tokens sequentially, with a larger target model validating these tokens, which is widely implemented to accelerate the rollout speed~\cite{gaoInterpretableContrastiveMonte2024,wang2024seedacceleratingreasoningtree}. SEED~\cite{wang2024seedacceleratingreasoningtree} implements scheduled speculative decoding, which efficiently manages both runtime speed and GPU memory usage simultaneously. The framework leverages a rounds-scheduled strategy that manages the execution flow using a First-Come-First-Serve queue to control verification of the target model without conflicts. SC-MCTS~\cite{gaoInterpretableContrastiveMonte2024} utilizes speculative decoding to speed up MCTS reasoning by an average of 52\% as a ``free lunch.''

\paragraph{KV Cache Management} LLM inference is typically memory bandwidth-bound~\cite{hooper2024squeezedattentionacceleratinglong}. In tree search, each unique trajectory requires a separate KV cache state, creating a significant memory bottleneck. DEFT~\cite{yao2025deftdecodingflashtreeattention} introduces efficient kernel implementations that compute attention with tree-structured KV sharing. Hydragen~\cite{juravsky2024hydragenhighthroughputllminference} and vLLM~\cite{kwon2023efficient} offer support for shared prefix workloads, effectively eliminating KV cache duplication. SGLang~\cite{zheng2024sglangefficientexecutionstructured} implements Radix Attention, which stores and dynamically references reused KV cache segments. Additionally, ETS~\cite{hooper2025etsefficienttreesearch} employs a linear programming cost model that encourages KV cache sharing by penalizing node retention while incorporating semantic coverage parameters to maintain diversity among retained trajectories.

\paragraph{Parallel Processing} The acceleration of tree expansion and simulation phases can be achieved through parallel processing techniques. However, the frequent switches among paths complicates the parallelism.~\citet{ding2025dynamicparalleltreesearch} develop a flexible and adaptive parallelism system for arbitrary paths by implementing fine-grained cache management and alignment during the generation phase. The system adjusts the number of parallel paths processed based on real-time GPU memory availability, optimizing resource utilization.

\section{Tutorial} \label{sec:tutorial}

In this section, we provide a tutorial on utilizing reinforcement learning to unlock the long CoT ability of LLMs.

\subsection{Preparatory Work}
Before starting training, thorough preparation is essential. Though these foundational steps may seem simple, they are critical to the success of the experiment. We briefly introduce the code framework, models, datasets, and other necessary components.

\paragraph{\faHandPointRight\ ~Understanding Reinforcement Learning Framework} For ease of use, stability, and efficiency, this tutorial selects veRL as the reinforcement learning framework. The core code about the algorithm of veRL is organized as follows:
\begin{description}[font=\ttfamily\bfseries, leftmargin=2em, itemindent=1em]
  \item[\faFolder\ trainer/] Contains core training code, including training control flow, advantage estimation functions, loss functions, etc.
  \item[\faFolder\ utils/] Includes data reading, reward functions, checkpoint saving, and other utilities.
  \item[\faFolder\ workers/] Defines various workers, including actor, critic, reward model, vllm, etc.
  \item[\faFile\ protocal.py] Defines the data structures in veRL, used for data exchange between different workers.
\end{description}

\paragraph{\faHandPointRight\ ~Selecting the Base Model}
The choice of the base model has a decisive impact on experimental results. We recommend Qwen2.5-1.5B, Qwen2.5-3B, and Qwen2.5-7B as starting points. We strongly recommend using the base versions of these models rather than versions that have undergone SFT. Since base models have not been fine-tuned for specific tasks, they retain stronger exploratory capabilities, which is advantageous for reinforcement learning as it helps the model discover more diverse problem-solving strategies.

\paragraph{\faHandPointRight\ ~Preparing the Dataset}
Data quality is crucial for training effectiveness. We provide a specially designed dataset of mathematical problems including NuminaMath, DeepScaleR, and MATH. The difficulty of problems in these datasets has been carefully selected to challenge the model's abilities without being so difficult as to cause learning stagnation. Training on this dataset allows for observation of stable performance improvements and growth in the length of solution processes. 

\paragraph{\faHandPointRight\ ~Selecting the Algorithm}
Due to its popularity and effectiveness, we adopt the GRPO algorithm as the primary method for this tutorial.

\subsection{Start RL Training}
You can follow the commands below to start the RL training for the Qwen2.5-1.5B\footnote{\url{https://huggingface.co/Qwen/Qwen2.5-1.5B}} model.
\begin{lstlisting}[language=bash]
git clone https://github.com/GAIR-NLP/cognition-engineering.git
cd cognition-engineering/simple_tts
# Create the conda environment
conda create -n verl python=3.10
conda activate verl
pip install -r requirements.txt
pip3 install vllm==0.7.3
pip3 install flash-attn --no-build-isolation
# launch training
bash examples/simple_tts.sh
\end{lstlisting}

If you want to use another model, you can replace the value of \texttt{policy\_path} in \texttt{examples/simple\_tts.sh}. We provide a set of default parameters that have been verified to be effective, but you can also change other hyperparameters, which will be explained in the next section. Additionally, you can build other training and testing datasets by following the structure of the files in \texttt{data/train/}, and modify the corresponding paths in \texttt{data.train\_files} and \texttt{data.val\_files} within \texttt{examples/simple\_tts.sh}.

\subsection{Understanding Algorithm with Code Analysis}
Let us understand the implementation of the reinforcement learning algorithm through specific code.

\paragraph{\faHandPointRight\ ~Launch Script}
First, let's examine veRL's launch script, focusing on the core algorithm parameters:
\begin{lstlisting}[language=bash]
python3 -m verl.trainer.main_ppo \
    algorithm.adv_estimator=grpo \
    data.train_batch_size=$rollout_batch_size \
    actor_rollout_ref.actor.ppo_mini_batch_size=$mini_batch_size \
    actor_rollout_ref.actor.kl_loss_coef=$kl_loss_coef \
    actor_rollout_ref.actor.entropy_coeff=$entropy_coeff \
    actor_rollout_ref.rollout.temperature=$temperature \
    actor_rollout_ref.rollout.n=$n_samples_per_prompts
\end{lstlisting}
Key parameters:
\begin{itemize}
    \item \texttt{algorithm.adv\_estimator}: Advantage estimation method (GRPO vs PPO/REINFORCE)
    \item \texttt{train\_batch\_size}: Number of prompts per sampling batch
    \item \texttt{mini\_batch\_size}: Number of prompts per update batch
    \item \texttt{n\_samples\_per\_prompts}: Numbers of generated responses per prompt
    \item \texttt{temperature}: Generation randomness control
    \item \texttt{kl\_loss\_coef} and \texttt{entropy\_coeff}: Coefficients for KL loss and entropy loss
\end{itemize}

\paragraph{\faHandPointRight\ ~Algorithm Flow Analysis}
The \texttt{fit} method in \texttt{verl.trainer.ppo.ray\_trainer}'s \texttt{RayPPOTrainer} is the control flow for the entire RL algorithm. Here is the simplified core process:
\begin{lstlisting}[language=python]
# Loop through the training set for total_epochs times
for epoch in range(self.config.trainer.total_epochs):
    # Take train_batch_size prompts from the dataset
    for batch_dict in self.train_dataloader:
        # Prompts are organized into specific data structures
        batch: DataProto = DataProto.from_single_dict(batch_dict)
        # Rollout process, using inference engines to generate answers for prompts
        gen_batch_output = self.actor_rollout_wg.generate_sequences(gen_batch)

        # Performs forward propagation to obtain log probability of old policy
        old_log_prob = self.actor_rollout_wg.compute_log_prob(batch)

        # Perform forward propagation to get log probability of ref model
        ref_log_prob = self.ref_policy_wg.compute_ref_log_prob(batch)

        # Calculate value with critic model
        # But our algorithm GRPO doesn't require critic
        values = self.critic_wg.compute_values(batch)

        # Use reward function to get reward for each sequence
        reward_tensor = self.reward_fn(batch)

        # Calculate the advantage value for each sequence based on the reward
        batch = compute_advantage(batch, adv_estimator, gamma, lam, num_repeat)

        # Update model using calculated advantage values
        critic_output = self.critic_wg.update_critic(batch)
        actor_output = self.actor_rollout_wg.update_actor(batch)
\end{lstlisting}

\paragraph{\faHandPointRight\ ~Data Loading and Prompt Templates}
The dataset class in \texttt{verl.utils.dataset.rl\_dataset} is used to read and process data. Each data entry contains a question and an answer. In the dataset class, the question is combined with a specific template to generate the prompt. The template we adopt is:

\begin{tcolorbox}[title=Prompt, width=16cm, boxrule=0.45mm, fonttitle=\bfseries]
A conversation between User and Assistant. The user asks a question, and the Assistant solves it. The assistant first thinks about the reasoning process in the mind and then provides the user with the answer. \textbackslash nUser: You must put your answer inside \textbackslash boxed\{\} and Your final answer will be extracted automatically by the \textbackslash boxed\{\} tag.\textbackslash n\textcolor{red}{prompt}\textbackslash nAssistant:
\end{tcolorbox}
This template encourages the model to first engage in reasoning, then provide the final answer, and mark the answer with \textbackslash boxed\{\} for subsequent evaluation.

\paragraph{\faHandPointRight\ ~Reward Design}
The \texttt{compute\_score} method in \texttt{verl.utils.reward\_score.math\_verifier} demonstrates how to calculate rewards based on the model's solution and the standard answer:
\begin{lstlisting}[language=python]
def compute_score(solution_str, ground_truth, reward_type) -> float:
     return correctness_score_default(solution_str, ground_truth)            

def correctness_score_default(response, gt):
    # Use regular expressions to extract boxed content from the answer
    pred = boxed_pattern.findall(response)[-1][:-1]
    # Judge whether the answer is correct
    return 1.0 if is_equiv(pred, gt) else -1.0  

def is_equiv(str1, str2, verbose=False):
    # Parse and verify whether two answers are mathematically equivalent
    return verify(parse(str1), parse(str2))
\end{lstlisting}
This reward function is very straightforward: correct answers receive 1 point, incorrect answers receive -1 point. 

\paragraph{\faHandPointRight\ ~GRPO Advantage Estimation}
The implementation of GRPO advantage estimation is provided in

\texttt{verl.trainer.ppo.core\_alogs}:

\begin{lstlisting}[language=python]
# This implementation only considers outcome supervision, i.e., the reward is a scalar
def compute_grpo_outcome_advantage(token_level_rewards: torch.Tensor, eos_mask, index, epsilon: float = 1e-6):
        for idx in id2score:
            id2mean[idx] = torch.mean(torch.tensor(id2score[idx]))
            id2std[idx] = torch.std(torch.tensor([id2score[idx]]))
        # GRPO advantage estimation
        for i in range(bsz):
            scores[i] = (scores[i] - id2mean[index[i]]) / (id2std[index[i]] + epsilon)
        # Expand scalar advantage to sequence length and mask with eos_mask
        scores = scores.unsqueeze(-1).tile([1, response_length]) * eos_mask
    return scores, scores
\end{lstlisting}

\paragraph{\faHandPointRight\ ~Policy Loss Calculation}
Finally, here is how the policy loss is calculated, which is also provided in \texttt{verl.trainer.ppo.core\_alogs}:

\begin{lstlisting}[language=python]
def compute_policy_loss(old_log_prob, log_prob, advantages, eos_mask, cliprange):
    # Calculate the log probability difference between new and old policies
    negative_approx_kl = log_prob - old_log_prob
    # Calculate probability ratio
    ratio = torch.exp(negative_approx_kl)
    # Calculate KL divergence
    ppo_kl = verl_F.masked_mean(-negative_approx_kl, eos_mask)
    # Original policy gradient loss
    pg_losses = -advantages * ratio
    # Clipped policy gradient loss
    pg_losses2 = -advantages * torch.clamp(ratio, 1.0 - cliprange, 1.0 + cliprange)
    # Take the larger of the two as the final loss
    pg_loss = verl_F.masked_mean(torch.max(pg_losses, pg_losses2), eos_mask)
    # Calculate clipping ratio
    pg_clipfrac = verl_F.masked_mean(torch.gt(pg_losses2, pg_losses).float(), eos_mask)
    return pg_loss, pg_clipfrac, ppo_kl
\end{lstlisting}
\subsection{Results}
\begin{figure}[htbp]
\centering
\begin{subfigure}[b]{0.48\textwidth}
\includegraphics[width=\textwidth]{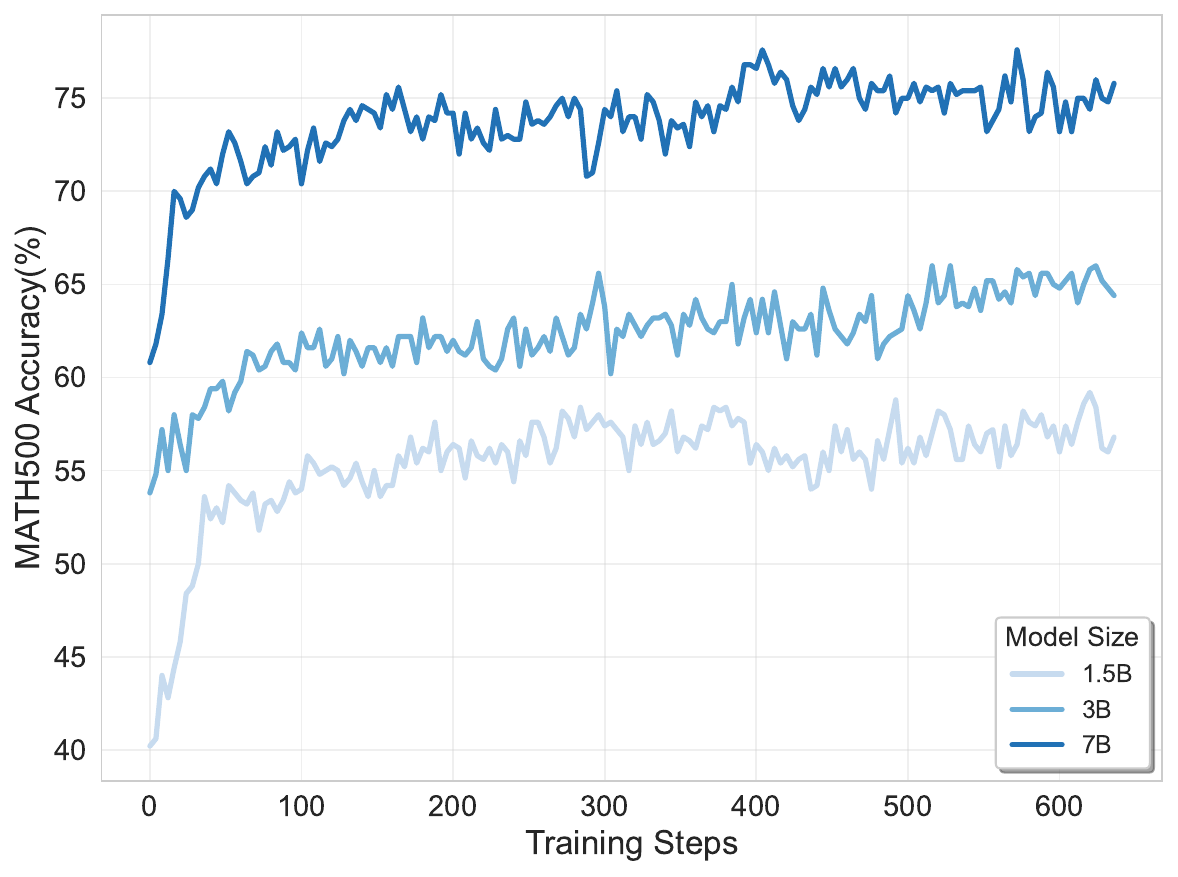}
\caption{}
\end{subfigure}
\hfill
\begin{subfigure}[b]{0.48\textwidth}
\includegraphics[width=\textwidth]{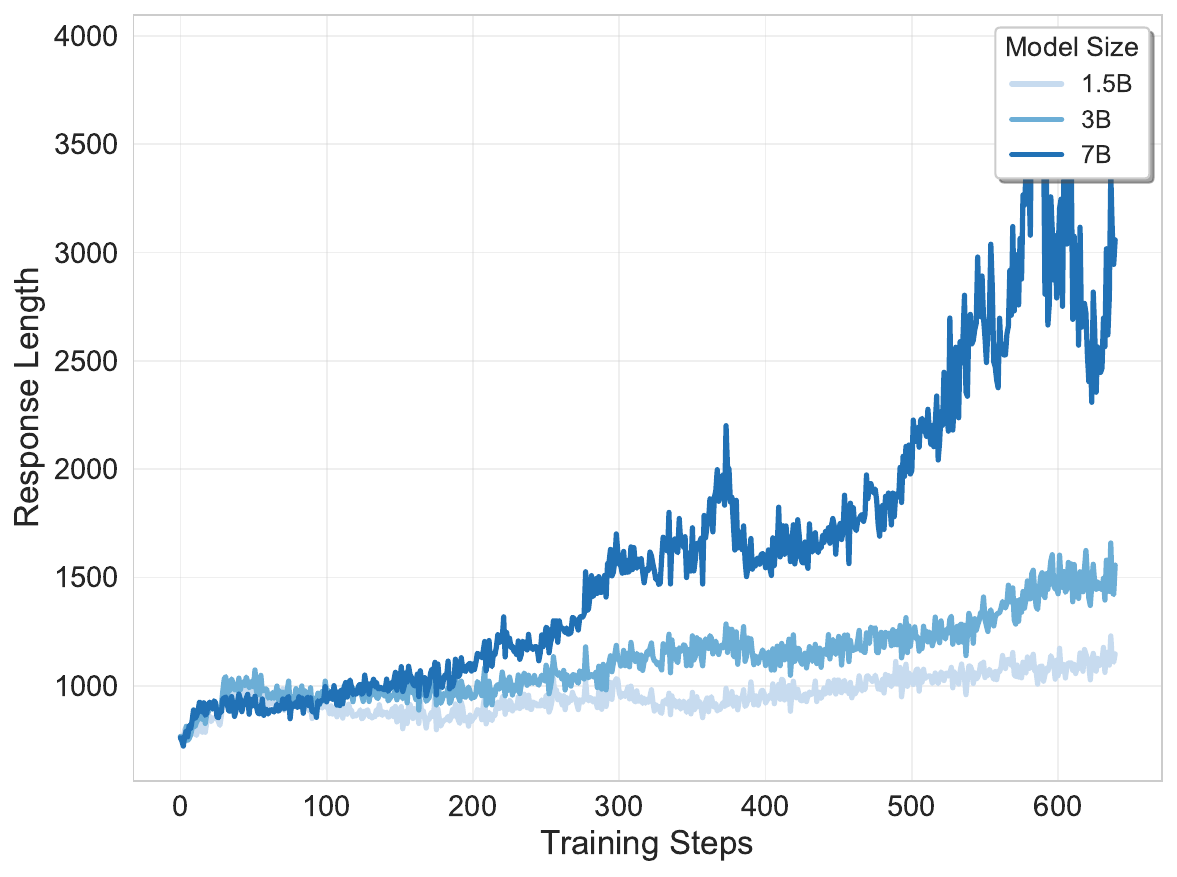}
\caption{}
\end{subfigure}
\caption{Trends in Accuracy on MATH500 and length on the training set of the Qwen-Base model during the RL process.}
\label{fig:tutorial}
\end{figure}

As illustrated in Figure~\ref{fig:tutorial}, we observe significant improvements in:
\begin{itemize}
\item Accuracy: The model's accuracy in solving mathematical problems increases markedly.
\item Response Length: The solution processes generated by the model become more detailed.
\end{itemize}
We also recommend analyzing whether the model exhibited advanced cognitive capabilities such as reflection and self-correction through qualitative case studies and by tracking specific keywords  in the responses.

\section{Future Directions} \label{sec:future_direction}
While we have outlined domain-specific future directions within each application area, cognition engineering also faces several fundamental challenges that cut across these domains. In this section, we identify these critical cross-cutting future directions that could substantially accelerate the entire field of cognition engineering.

\paragraph{New architecture}
Transformer-based architectures face fundamental limitations due to their linear memory scaling and memory-bound nature during generation. This presents a critical constraint for test-time scaling, particularly when generating long context. While the methods described in the improving scaling efficiency section can alleviate this problem, a more fundamental solution requires exploring new architectures. Promising alternatives include state space models like Mamba~\cite{gu2024mambalineartimesequencemodeling,dao2024transformersssmsgeneralizedmodels}, which offers linear-time complexity for sequence modeling, linear transformers that reduce the quadratic attention bottleneck~\cite{katharopoulos2020transformersrnnsfastautoregressive}, and even language diffusion models~\cite{nie2025largelanguagediffusionmodels}.
This architectural transformation requires comprehensive system engineering efforts across multiple dimensions: developing robust theoretical frameworks for new architectures, building infrastructure support for efficient training and inference, and creating large-scale pretrained foundation models based on these architecture. The integration of these architectures with cognition engineering technology could significantly enhance both cognitive capabilities and computational efficiency.

\paragraph{Pretraining on latent thought} Current pretraining data primarily consists of human-written texts but lacks the latent thought processes behind them. The success of RL scaling on models pretrained with data containing cognitive behaviors demonstrates the potential of including human thinking processes~\citep{gandhi2025cognitivebehaviorsenableselfimproving,liu2025understanding,liu2025oatzero}. Recent works have shown the benefits of incorporating hidden thinking processes beyond explicit text~\citep{zelikman2024quietstarlanguagemodelsteach,jiang2024rationalystpretrainingprocesssupervisionimproving,ruan2025reasoninglearnlatentthoughts}, though these are still limited to small-scale experiments. Future research should focus on acquiring large-scale data rich in cognitive behaviors  through techniques such as inferring latent thoughts by utilizing existing reasoning models and examining how pretraining on such data could benefit test-time scaling methods.

\paragraph{RL scaling} While we have examined the common design principles of RL scaling based on the latest research, the field remains in its early development stage, holding significant potential to unlock the cognitive abilities of AI. Given the complex components of RL and the numerous hyperparameters requiring tuning, future research should adopt a more rigorous approach when drawing conclusions that account for all these elements~\citep{jordan2024positionbenchmarkinglimitedreinforcement,hochlehnert2025soberlookprogresslanguage}. Moreover, reproducible open-source work is currently limited to small models and datasets, which constrains the scope of possible conclusions. Large-scale experiments require both infrastructure improvements and algorithm optimizations to become accessible to researchers with modest computational resources. Furthermore, current RL scaling primarily focuses on verifiable tasks such as mathematics and code. Expanding to broader domains necessitates deeper investigation into reward hacking phenomena and establishing clearer relationships between reward reliability and RL scaling. This advancement demands not only empirical investigation but also theoretical analysis.

\paragraph{Evaluation} As an engineering approach, cognition engineering relies on iterative feedback and enhancement, which requires evaluation methods that go beyond simple benchmark performance metrics. Although some work has begun focusing on cognitive behavioral changes, these efforts usually rely on matching specific words (e.g., ``wait'' or ``alternatively'') or case studies~\cite{deepseekai2025deepseekr1incentivizingreasoningcapability,gandhi2025cognitivebehaviorsenableselfimproving}, failing to fully capture the quality of cognitive behaviors or the depth of reasoning processes. Future work should develop comprehensive evaluation frameworks that assess not only task performance but also the quality, efficiency, and generalizability of cognitive processes. This includes creating metrics for reasoning depth, backtracking efficiency, verification quality, and metacognitive awareness. Additionally, dynamic evaluation protocols that can adapt to evolving cognitive capabilities would better capture progress in this rapidly developing field. Developing these evaluation tools will require interdisciplinary collaboration between AI researchers, cognitive scientists, and domain experts to ensure they accurately reflect the cognitive dimensions most relevant to human-like reasoning.

\paragraph{Scientific discovery} Cognition engineering opens up unprecedented possibilities for AI systems to serve as scientific discovery partners. Test-time scaling methods show promise in enabling models to connect disparate knowledge domains and generate innovative insights through extended deliberation. Future work should explore how to optimize these cognitive capabilities specifically for scientific discovery tasks, such as hypothesis generation, experimental design, and theory formation. This will require developing specialized prompting techniques or training methodologies that encourage creative yet scientifically rigorous exploration of solution spaces. Additionally, research should investigate how to integrate domain-specific scientific tools and experimental platforms with reasoning models, allowing them to not only generate hypotheses but also design and potentially execute experiments to validate them~\cite{lu2024aiscientist}.

\begin{table}[tb!]
\centering
\caption{Comparison to existing surveys. $\triangle$ denotes limited discussion of the topic. \textbf{TTS} denotes Test-time Scaling. \textbf{Law} denotes Scaling Laws. \textbf{Efficiency} denotes Scaling Efficiency. }
\label{tab:Comparisons_existing_works}
\resizebox{0.99\textwidth}{!}{ 
\begin{tabular}{lcccccccc}
\toprule
\multirow{2}{*}{\textbf{Work}} & \multicolumn{2}{c}{\textbf{TTS Types}} & \multirow{2}{*}{\textbf{Law}} & \multirow{2}{*}{\textbf{Efficiency}}  &  \multirow{2}{*}{\textbf{Comparison}} & \multirow{2}{*}{\textbf{Ensemble}} & \multirow{2}{*}{\textbf{Long-CoT RL}} & \multirow{2}{*}{\textbf{Long-CoT SFT}} \\
\cmidrule(lr){2-3}
 & \textbf{Long CoT} & \textbf{Others} \\
\midrule

\citet{welleck2024decodingmetagenerationinferencetimealgorithms} & \blackcross & \blackcheck &  $\triangle$ & $\triangle$ & $\triangle$ & \blackcross & \blackcross &  \blackcross  \\

\midrule

\citet{zeng2024scalingsearchlearningroadmap} & \blackcheck & \blackcheck & \blackcross & \blackcross & \blackcross &  $\triangle$ & \blackcross & \blackcross  \\

\midrule

\citet{ji2025testtimecomputesystem1thinking} & \blackcross & \blackcheck & $\triangle$ & $\triangle$ & \blackcross &  \blackcross & \blackcross & \blackcross  \\

\midrule

\citet{li202512surveyreasoning} & \blackcheck & \blackcheck & \blackcross & \blackcross & \blackcross &  \blackcross & $\triangle$ & \blackcross   \\

\midrule

\citet{kumar2025llmposttrainingdeepdive} & \blackcheck & \blackcheck &  \blackcross & \blackcross & \blackcross & \blackcross & $\triangle$ & $\triangle$  \\

\midrule

\citet{chen2025reasoningerasurveylong} & \blackcheck & \blackcheck & \blackcross & \blackcross & \blackcross &  $\triangle$ & \blackcross & $\triangle$  \\

\midrule

\citet{zhang2025whathowwherewell} & \blackcheck & \blackcheck & \blackcross &  \blackcross  &  $\triangle$ &  \blackcross&  \blackcross & $\triangle$  \\

\midrule
Ours & \blackcheck & \blackcheck & \blackcheck & \blackcheck & \blackcheck & \blackcheck & \blackcheck & \blackcheck   \\

\bottomrule
\end{tabular}
}
\end{table}

\section{Comparison to Existing Work} \label{sec:related_work}

Our paper combines aspects of both a comprehensive survey and a position paper. From the survey perspective, we summarize and compare existing works in Table~\ref{tab:Comparisons_existing_works}. Our survey differs from previous works in three key aspects. First, we organize the content from a test-time scaling perspective, which differs from works focused on either system-2 reasoning~\citep{li202512surveyreasoning} or long CoT~\citep{chen2025reasoningerasurveylong,ji2025testtimecomputesystem1thinking}. Building on this focus, we offer detailed discussions of scaling laws, scaling efficiency for four primary test-time scaling methods, comparisons between these methods, and their ensemble applications. While~\citet{welleck2024decodingmetagenerationinferencetimealgorithms} discuss test-time scaling from the meta-generation perspective, their coverage of these critical aspects for utilizing test-time scaling methods remains limited and they do not include the long-CoT technique. Concurrent to our work,~\citet{zhang2025whathowwherewell} focus on test-time scaling but offer insufficient discussion of these vital topics. Other concurrent works such as~\citet{sui2025stopoverthinkingsurveyefficient,qu2025surveyefficientreasoninglarge} focus on scaling efficiency but restrict their analysis to partial test-time scaling methods, primarily long CoT.
Second, we provide detailed, multi-faceted discussions of RL for long-CoT techniques and SFT for long-CoT based on recent research, offering more concrete and specific insights than previous works' vague discussions~\citep{tie2025surveyposttraininglargelanguage,besta2025reasoninglanguagemodelsblueprint,xu2025largereasoningmodelssurvey}.
Third, we cover a broader range of applications for test-time scaling and provide thorough discussion and future directions for each.

From the position paper perspective, we propose the concept of cognition engineering and argue that generative AI has evolved into an era focused on developing deep cognitive capabilities in models. This conceptual framework helps unify various technologies that enhance model cognitive abilities and includes potential future directions like pretraining on latent thought. This feature also distinguishes our work from other survey papers.

\section{Conclusion}

Cognition engineering represents a paradigm shift in AI development, fundamentally transforming our approach from knowledge accumulation to the systematic development of thinking capabilities. This second act of generative AI leverages test-time scaling methodologies alongside specialized training strategies to enable models to engage in deep thinking, complex reasoning, and creative problem-solving. As demonstrated across domains from mathematics to multimodal understanding, these capabilities are already yielding substantial improvements in model performance. The emergence of cognition engineering marks not only technological progress but also the
beginning of a new relationship between human thought and artificial intelligence—a symbiotic relationship based
on deep understanding and cognitive exchange, in which humans and AI can empower each other to jointly explore
new frontiers of cognition.

\section*{Acknowledgment}
We would like to thank Yixin Liu for his constructive comments on this work. We would also like to extend our appreciation to Ethan Chern for his involvement in the early stage of the work.

\newpage
\bibliographystyle{acl_natbib}
\bibliography{related}

\begin{thebibliography}{454}
\expandafter\ifx\csname natexlab\endcsname\relax\def\natexlab#1{#1}\fi

\bibitem[{Ackoff(1989)}]{ackoff1989data}
Russell~L Ackoff. 1989.
\newblock From data to wisdom.
\newblock \emph{Journal of applied systems analysis}, 16(1):3--9.

\bibitem[{Aggarwal et~al.(2023)Aggarwal, Madaan, Yang, and {Mausam}}]{aggarwalLetsSampleStep2023}
Pranjal Aggarwal, Aman Madaan, Yiming Yang, and {Mausam}. 2023.
\newblock \href {https://doi.org/10.18653/v1/2023.emnlp-main.761} {Let{'}s sample step by step: Adaptive-consistency for efficient reasoning and coding with {LLM}s}.
\newblock In \emph{Proceedings of the 2023 Conference on Empirical Methods in Natural Language Processing}, pages 12375--12396, Singapore. Association for Computational Linguistics.

\bibitem[{Aggarwal and Welleck(2025)}]{aggarwal2025l1controllinglongreasoning}
Pranjal Aggarwal and Sean Welleck. 2025.
\newblock \href {https://arxiv.org/abs/2503.04697} {L1: Controlling how long a reasoning model thinks with reinforcement learning}.

\bibitem[{Ahmad et~al.(2025)Ahmad, Narenthiran, Majumdar, Ficek, Jain, Huang, Noroozi, and Ginsburg}]{ahmad2025opencodereasoning}
Wasi~Uddin Ahmad, Sean Narenthiran, Somshubra Majumdar, Aleksander Ficek, Siddhartha Jain, Jocelyn Huang, Vahid Noroozi, and Boris Ginsburg. 2025.
\newblock \href {https://arxiv.org/abs/2504.01943} {Opencodereasoning: Advancing data distillation for competitive coding}.
\newblock \emph{ArXiv preprint}, abs/2504.01943.

\bibitem[{Ahn et~al.(2022)Ahn, Brohan, Brown, Chebotar, Cortes, David, Finn, Fu, Gopalakrishnan, Hausman et~al.}]{ahn2022can}
Michael Ahn, Anthony Brohan, Noah Brown, Yevgen Chebotar, Omar Cortes, Byron David, Chelsea Finn, Chuyuan Fu, Keerthana Gopalakrishnan, Karol Hausman, et~al. 2022.
\newblock \href {https://arxiv.org/abs/2204.01691} {Do as i can, not as i say: Grounding language in robotic affordances}.
\newblock \emph{ArXiv preprint}, abs/2204.01691.

\bibitem[{AlphaProof and teams(2024)}]{alphaproof}
AlphaProof and AlphaGeometry teams. 2024.
\newblock {AI} achieves silver-medal standard solving international mathematical olympiad problems.
\newblock \url{https://deepmind.google/discover/blog/ai-solves-imo-problems-at-silver-medal-level/}.

\bibitem[{Anthropic(2024{\natexlab{a}})}]{anthropic2024claude3}
Anthropic. 2024{\natexlab{a}}.
\newblock \href {https://www-cdn.anthropic.com/de8ba9b01c9ab7cbabf5c33b80b7bbc618857627/Model_Card_Claude_3.pdf} {The claude 3 model family: Opus, sonnet, haiku}.
\newblock Technical Report.

\bibitem[{Anthropic(2024{\natexlab{b}})}]{anthropic2024models}
Anthropic. 2024{\natexlab{b}}.
\newblock \href {https://www.anthropic.com/news/3-5-models-and-computer-use} {Introducing computer use, a new claude 3.5 sonnet, and claude 3.5 haiku}.
\newblock anthropic.com.

\bibitem[{Anthropic(2025)}]{claude3.7sonnet}
Anthropic. 2025.
\newblock \href {https://www.anthropic.com/news/claude-3-7-sonnet} {Introducing deep research}.
\newblock anthropic.com.

\bibitem[{team~swe arena(2025)}]{swe-arena2025}
team~swe arena. 2025.
\newblock Swe arena: An open evaluation platform for automated software engineering.

\bibitem[{Arora and Zanette(2025)}]{arora2025traininglanguagemodelsreason}
Daman Arora and Andrea Zanette. 2025.
\newblock \href {https://arxiv.org/abs/2502.04463} {Training language models to reason efficiently}.

\bibitem[{Aytes et~al.(2025)Aytes, Baek, and Hwang}]{aytes2025sketchofthoughtefficientllmreasoning}
Simon~A. Aytes, Jinheon Baek, and Sung~Ju Hwang. 2025.
\newblock \href {https://arxiv.org/abs/2503.05179} {Sketch-of-thought: Efficient llm reasoning with adaptive cognitive-inspired sketching}.

\bibitem[{Azzolini et~al.(2025)Azzolini, Brandon, Chattopadhyay, Chen, Chu, Cui, Diamond, Ding, Ferroni, Govindaraju et~al.}]{azzolini2025cosmos}
Alisson Azzolini, Hannah Brandon, Prithvijit Chattopadhyay, Huayu Chen, Jinju Chu, Yin Cui, Jenna Diamond, Yifan Ding, Francesco Ferroni, Rama Govindaraju, et~al. 2025.
\newblock \href {https://arxiv.org/abs/2503.15558} {Cosmos-reason1: From physical common sense to embodied reasoning}.
\newblock \emph{ArXiv preprint}, abs/2503.15558.

\bibitem[{Bae et~al.(2025)Bae, Hong, Lee, Kim, Nam, and Kwak}]{bae2025onlinedifficultyfilteringreasoning}
Sanghwan Bae, Jiwoo Hong, Min~Young Lee, Hanbyul Kim, JeongYeon Nam, and Donghyun Kwak. 2025.
\newblock \href {https://arxiv.org/abs/2504.03380} {Online difficulty filtering for reasoning oriented reinforcement learning}.

\bibitem[{Ballon et~al.(2025)Ballon, Algaba, and Ginis}]{ballon2025relationshipreasoningperformancelarge}
Marthe Ballon, Andres Algaba, and Vincent Ginis. 2025.
\newblock \href {https://arxiv.org/abs/2502.15631} {The relationship between reasoning and performance in large language models -- o3 (mini) thinks harder, not longer}.

\bibitem[{Beeching et~al.(2024)Beeching, Tunstall, and Rush}]{beeching2024scalingtesttimecompute}
Edward Beeching, Lewis Tunstall, and Sasha Rush. 2024.
\newblock \href {https://huggingface.co/spaces/HuggingFaceH4/blogpost-scaling-test-time-compute} {Scaling test-time compute with open models}.

\bibitem[{Bengio(2023)}]{bengio2023FAQ}
Yoshua Bengio. 2023.
\newblock \href {https://yoshuabengio.org/2023/06/24/faq-on-catastrophic-ai-risks/} {Faq on catastrophic ai risks}.

\bibitem[{Besta et~al.(2025)Besta, Barth, Schreiber, Kubicek, Catarino, Gerstenberger, Nyczyk, Iff, Li, Houliston, Sternal, Copik, Kwaśniewski, Müller, Łukasz Flis, Eberhard, Niewiadomski, and Hoefler}]{besta2025reasoninglanguagemodelsblueprint}
Maciej Besta, Julia Barth, Eric Schreiber, Ales Kubicek, Afonso Catarino, Robert Gerstenberger, Piotr Nyczyk, Patrick Iff, Yueling Li, Sam Houliston, Tomasz Sternal, Marcin Copik, Grzegorz Kwaśniewski, Jürgen Müller, Łukasz Flis, Hannes Eberhard, Hubert Niewiadomski, and Torsten Hoefler. 2025.
\newblock \href {http://arxiv.org/abs/2501.11223} {Reasoning language models: A blueprint}.

\bibitem[{Bhandari et~al.(2020)Bhandari, Gour, Ashfaq, Liu, and Neubig}]{bhandari-etal-2020-evaluating}
Manik Bhandari, Pranav~Narayan Gour, Atabak Ashfaq, Pengfei Liu, and Graham Neubig. 2020.
\newblock \href {https://doi.org/10.18653/v1/2020.emnlp-main.751} {Re-evaluating evaluation in text summarization}.
\newblock In \emph{Proceedings of the 2020 Conference on Empirical Methods in Natural Language Processing (EMNLP)}, pages 9347--9359, Online. Association for Computational Linguistics.

\bibitem[{Bi et~al.(2024)Bi, Han, Liu, Tang, and Wang}]{bi2025forestofthoughtscalingtesttimecompute}
Zhenni Bi, Kai Han, Chuanjian Liu, Yehui Tang, and Yunhe Wang. 2024.
\newblock \href {https://arxiv.org/abs/2412.09078} {Forest-of-thought: Scaling test-time compute for enhancing llm reasoning}.

\bibitem[{Brown et~al.(2024)Brown, Juravsky, Ehrlich, Clark, Le, R{\'e}, and Mirhoseini}]{brownLargeLanguageMonkeys2024}
Bradley Brown, Jordan Juravsky, Ryan Ehrlich, Ronald Clark, Quoc~V. Le, Christopher R{\'e}, and Azalia Mirhoseini. 2024.
\newblock \href {https://arxiv.org/abs/2407.21787} {Large language monkeys: Scaling inference compute with repeated sampling}.

\bibitem[{Chai et~al.(2024)Chai, Li, Fu, Zhao, and Zhu}]{chaiempowering}
Jiajun Chai, Sicheng Li, Yuqian Fu, Dongbin Zhao, and Yuanheng Zhu. 2024.
\newblock Empowering llm agents with zero-shot optimal decision-making through q-learning.
\newblock In \emph{Adaptive Foundation Models: Evolving AI for Personalized and Efficient Learning}.

\bibitem[{Chameleon(2024)}]{team2024chameleon}
Team Chameleon. 2024.
\newblock \href {https://arxiv.org/abs/2405.09818} {Chameleon: Mixed-modal early-fusion foundation models}.
\newblock \emph{ArXiv preprint}, abs/2405.09818.

\bibitem[{Chan et~al.(2023)Chan, Chen, Su, Yu, Xue, Zhang, Fu, and Liu}]{chan2023chateval}
Chi-Min Chan, Weize Chen, Yusheng Su, Jianxuan Yu, Wei Xue, Shanghang Zhang, Jie Fu, and Zhiyuan Liu. 2023.
\newblock \href {https://arxiv.org/abs/2308.07201} {Chateval: Towards better llm-based evaluators through multi-agent debate}.
\newblock \emph{ArXiv preprint}, abs/2308.07201.

\bibitem[{Chen et~al.(2023{\natexlab{a}})Chen, Zhang, Nguyen, Zan, Lin, Lou, and Chen}]{chenCodeTCodeGeneration2022}
Bei Chen, Fengji Zhang, Anh Nguyen, Daoguang Zan, Zeqi Lin, Jian{-}Guang Lou, and Weizhu Chen. 2023{\natexlab{a}}.
\newblock \href {https://openreview.net/pdf?id=ktrw68Cmu9c} {Codet: Code generation with generated tests}.
\newblock In \emph{The Eleventh International Conference on Learning Representations, {ICLR} 2023, Kigali, Rwanda, May 1-5, 2023}. OpenReview.net.

\bibitem[{Chen et~al.(2024{\natexlab{a}})Chen, Liao, Li, and Fan}]{chen2024alphamathzeroprocesssupervision}
Guoxin Chen, Minpeng Liao, Chengxi Li, and Kai Fan. 2024{\natexlab{a}}.
\newblock \href {https://arxiv.org/abs/2405.03553} {Alphamath almost zero: Process supervision without process}.

\bibitem[{Chen et~al.(2025{\natexlab{a}})Chen, Tu, Liu, Tang, Du, Zhou, and Xie}]{vl-thinking2025}
Hardy Chen, Haoqin Tu, Hui Liu, Xianfeng Tang, Xinya Du, Yuyin Zhou, and Cihang Xie. 2025{\natexlab{a}}.
\newblock Vl-thinking: An r1-derived visual instruction tuning dataset for thinkable lvlms.
\newblock \url{https://github.com/UCSC-VLAA/VL-Thinking}.

\bibitem[{Chen et~al.(2025{\natexlab{b}})Chen, Ren, Chen, Yang, Sun, and Arık}]{chen2025setsleveragingselfverificationselfcorrection}
Jiefeng Chen, Jie Ren, Xinyun Chen, Chengrun Yang, Ruoxi Sun, and Sercan~Ö Arık. 2025{\natexlab{b}}.
\newblock \href {https://arxiv.org/abs/2501.19306} {Sets: Leveraging self-verification and self-correction for improved test-time scaling}.

\bibitem[{Chen et~al.(2024{\natexlab{b}})Chen, Cai, Ji, Wang, Liu, Wang, Hou, and Wang}]{chen2024huatuogpto1medicalcomplexreasoning}
Junying Chen, Zhenyang Cai, Ke~Ji, Xidong Wang, Wanlong Liu, Rongsheng Wang, Jianye Hou, and Benyou Wang. 2024{\natexlab{b}}.
\newblock \href {https://arxiv.org/abs/2412.18925} {Huatuogpt-o1, towards medical complex reasoning with llms}.

\bibitem[{Chen et~al.(2025{\natexlab{c}})Chen, Li, Zhao, Song, and Vinci}]{chen2025r1v}
Liang Chen, Lei Li, Haozhe Zhao, Yifan Song, and Vinci. 2025{\natexlab{c}}.
\newblock R1-v: Reinforcing super generalization ability in vision-language models with less than \$3.
\newblock \url{https://github.com/Deep-Agent/R1-V}.

\bibitem[{Chen et~al.(2024{\natexlab{c}})Chen, Davis, Hanin, Bailis, Stoica, Zaharia, and Zou}]{chenAreMoreLLM2024}
Lingjiao Chen, Jared~Quincy Davis, Boris Hanin, Peter Bailis, Ion Stoica, Matei Zaharia, and James Zou. 2024{\natexlab{c}}.
\newblock \href {https://arxiv.org/abs/2403.02419} {Are more llm calls all you need? towards scaling laws of compound inference systems}.

\bibitem[{Chen et~al.(2021)Chen, Tworek, Jun, Yuan, Pinto, Kaplan, Edwards, Burda, Joseph, Brockman et~al.}]{chen2021evaluating}
Mark Chen, Jerry Tworek, Heewoo Jun, Qiming Yuan, Henrique Ponde De~Oliveira Pinto, Jared Kaplan, Harri Edwards, Yuri Burda, Nicholas Joseph, Greg Brockman, et~al. 2021.
\newblock \href {https://arxiv.org/abs/2107.03374} {Evaluating large language models trained on code}.
\newblock \emph{ArXiv preprint}, abs/2107.03374.

\bibitem[{Chen et~al.(2025{\natexlab{d}})Chen, Li, Sun, Zhou, Zhu, Wang, Pan, Zhang, Chen, Yang, Zhou, and Chen}]{chen2025research}
Mingyang Chen, Tianpeng Li, Haoze Sun, Yijie Zhou, Chenzheng Zhu, Haofen Wang, Jeff~Z. Pan, Wen Zhang, Huajun Chen, Fan Yang, Zenan Zhou, and Weipeng Chen. 2025{\natexlab{d}}.
\newblock \href {https://arxiv.org/abs/2503.19470} {Research: Learning to reason with search for llms via reinforcement learning}.

\bibitem[{Chen et~al.(2025{\natexlab{e}})Chen, Qin, Liu, Peng, Guan, Wang, Hu, Zhou, Gao, and Che}]{chen2025reasoningerasurveylong}
Qiguang Chen, Libo Qin, Jinhao Liu, Dengyun Peng, Jiannan Guan, Peng Wang, Mengkang Hu, Yuhang Zhou, Te~Gao, and Wanxiang Che. 2025{\natexlab{e}}.
\newblock \href {https://arxiv.org/abs/2503.09567} {Towards reasoning era: A survey of long chain-of-thought for reasoning large language models}.

\bibitem[{Chen et~al.(2025{\natexlab{f}})Chen, Wu, Liu, Pan, Liu, Xie, Yu, and Ruan}]{chen2025janus}
Xiaokang Chen, Zhiyu Wu, Xingchao Liu, Zizheng Pan, Wen Liu, Zhenda Xie, Xingkai Yu, and Chong Ruan. 2025{\natexlab{f}}.
\newblock \href {https://arxiv.org/abs/2501.17811} {Janus-pro: Unified multimodal understanding and generation with data and model scaling}.
\newblock \emph{ArXiv preprint}, abs/2501.17811.

\bibitem[{Chen et~al.(2024{\natexlab{d}})Chen, Xu, Liang, He, Pang, Yu, Song, Liu, Zhou, Zhang, Wang, Tu, Mi, and Yu}]{chen2025think23overthinkingo1like}
Xingyu Chen, Jiahao Xu, Tian Liang, Zhiwei He, Jianhui Pang, Dian Yu, Linfeng Song, Qiuzhi Liu, Mengfei Zhou, Zhuosheng Zhang, Rui Wang, Zhaopeng Tu, Haitao Mi, and Dong Yu. 2024{\natexlab{d}}.
\newblock \href {https://arxiv.org/abs/2412.21187} {Do not think that much for 2+3=? on the overthinking of o1-like llms}.

\bibitem[{Chen et~al.(2023{\natexlab{b}})Chen, Aksitov, Alon, Ren, Xiao, Yin, Prakash, Sutton, Wang, and Zhou}]{chen2023universalselfconsistencylargelanguage}
Xinyun Chen, Renat Aksitov, Uri Alon, Jie Ren, Kefan Xiao, Pengcheng Yin, Sushant Prakash, Charles Sutton, Xuezhi Wang, and Denny Zhou. 2023{\natexlab{b}}.
\newblock \href {https://arxiv.org/abs/2311.17311} {Universal self-consistency for large language model generation}.

\bibitem[{Chen et~al.(2023{\natexlab{c}})Chen, Lin, Schärli, and Zhou}]{chen2023teachinglargelanguagemodels}
Xinyun Chen, Maxwell Lin, Nathanael Schärli, and Denny Zhou. 2023{\natexlab{c}}.
\newblock \href {https://arxiv.org/abs/2304.05128} {Teaching large language models to self-debug}.

\bibitem[{Chen et~al.(2025{\natexlab{g}})Chen, Min, Zhang, Chen, Jiang, Cheng, Zhao, Liu, Miao, Lu, Fang, Wang, and Wen}]{chen2025empiricalstudyelicitingimproving}
Zhipeng Chen, Yingqian Min, Beichen Zhang, Jie Chen, Jinhao Jiang, Daixuan Cheng, Wayne~Xin Zhao, Zheng Liu, Xu~Miao, Yang Lu, Lei Fang, Zhongyuan Wang, and Ji-Rong Wen. 2025{\natexlab{g}}.
\newblock \href {https://arxiv.org/abs/2503.04548} {An empirical study on eliciting and improving r1-like reasoning models}.

\bibitem[{Cheng and Durme(2024)}]{cheng2024compressedchainthoughtefficient}
Jeffrey Cheng and Benjamin~Van Durme. 2024.
\newblock \href {https://arxiv.org/abs/2412.13171} {Compressed chain of thought: Efficient reasoning through dense representations}.

\bibitem[{Cheng et~al.(2024{\natexlab{a}})Cheng, Liu, Wang, Gu, Lu, Zhang, Dong, Tang, Wang, and Huang}]{cheng2025sparselfplaytreesearchrefinement}
Jiale Cheng, Xiao Liu, Cunxiang Wang, Xiaotao Gu, Yida Lu, Dan Zhang, Yuxiao Dong, Jie Tang, Hongning Wang, and Minlie Huang. 2024{\natexlab{a}}.
\newblock \href {https://arxiv.org/abs/2412.11605} {Spar: Self-play with tree-search refinement to improve instruction-following in large language models}.

\bibitem[{Cheng et~al.(2024{\natexlab{b}})Cheng, Li, Xu, Zhang, Zhou, and Liu}]{r3v}
Kanzhi Cheng, Yantao Li, Fangzhi Xu, Jianbing Zhang, Hao Zhou, and Yang Liu. 2024{\natexlab{b}}.
\newblock \href {https://arxiv.org/abs/2411.00855} {Vision-language models can self-improve reasoning via reflection}.
\newblock \emph{ArXiv preprint}, abs/2411.00855.

\bibitem[{Cheng et~al.(2025)Cheng, Li, Zhao, and Wen}]{cheng2025halusearch}
Xiaoxue Cheng, Junyi Li, Wayne~Xin Zhao, and Ji{-}Rong Wen. 2025.
\newblock \href {https://arxiv.org/abs/2501.01306} {Think more, hallucinate less: Mitigating hallucinations via dual process of fast and slow thinking}.
\newblock \emph{ArXiv preprint}, abs/2501.01306.

\bibitem[{Chern et~al.(2023)Chern, Chern, Chen, Yuan, Feng, Zhou, He, Neubig, and Liu}]{chern2023factoolfactualitydetectiongenerative}
I-Chun Chern, Steffi Chern, Shiqi Chen, Weizhe Yuan, Kehua Feng, Chunting Zhou, Junxian He, Graham Neubig, and Pengfei Liu. 2023.
\newblock \href {https://arxiv.org/abs/2307.13528} {Factool: Factuality detection in generative ai -- a tool augmented framework for multi-task and multi-domain scenarios}.

\bibitem[{Chern et~al.(2024{\natexlab{a}})Chern, Chern, Neubig, and Liu}]{chern2024scaleeval}
Steffi Chern, Ethan Chern, Graham Neubig, and Pengfei Liu. 2024{\natexlab{a}}.
\newblock \href {https://arxiv.org/abs/2401.16788} {Can large language models be trusted for evaluation? scalable meta-evaluation of llms as evaluators via agent debate}.
\newblock \emph{ArXiv preprint}, abs/2401.16788.

\bibitem[{Chern et~al.(2024{\natexlab{b}})Chern, Fan, and Liu}]{chern2024combatadversarial}
Steffi Chern, Zhen Fan, and Andy Liu. 2024{\natexlab{b}}.
\newblock \href {https://arxiv.org/abs/2401.05998} {Combating adversarial attacks with multi-agent debate}.
\newblock \emph{ArXiv preprint}, abs/2401.05998.

\bibitem[{Chi et~al.(2025)Chi, Chen, Angelopoulos, Chiang, Mittal, Jain, Zhang, Stoica, Donahue, and Talwalkar}]{chi2025copilot}
Wayne Chi, Valerie Chen, Anastasios~Nikolas Angelopoulos, Wei-Lin Chiang, Aditya Mittal, Naman Jain, Tianjun Zhang, Ion Stoica, Chris Donahue, and Ameet Talwalkar. 2025.
\newblock \href {https://arxiv.org/abs/2502.09328} {Copilot arena: A platform for code llm evaluation in the wild}.
\newblock \emph{ArXiv preprint}, abs/2502.09328.

\bibitem[{Chiang and Lee(2023)}]{chiang2023largelanguagemodelsalternative}
Cheng-Han Chiang and Hung-yi Lee. 2023.
\newblock \href {https://doi.org/10.18653/v1/2023.acl-long.870} {Can large language models be an alternative to human evaluations?}
\newblock In \emph{Proceedings of the 61st Annual Meeting of the Association for Computational Linguistics (Volume 1: Long Papers)}, pages 15607--15631, Toronto, Canada. Association for Computational Linguistics.

\bibitem[{Chow et~al.(2024)Chow, Tennenholtz, Gur, Zhuang, Dai, Thiagarajan, Boutilier, Agarwal, Kumar, and Faust}]{chowInferenceAwareFineTuningBestofN2024}
Yinlam Chow, Guy Tennenholtz, Izzeddin Gur, Vincent Zhuang, Bo~Dai, Sridhar Thiagarajan, Craig Boutilier, Rishabh Agarwal, Aviral Kumar, and Aleksandra Faust. 2024.
\newblock \href {https://arxiv.org/abs/2412.15287} {Inference-aware fine-tuning for best-of-n sampling in large language models}.

\bibitem[{Chu et~al.(2025{\natexlab{a}})Chu, Zhai, Yang, Tong, Xie, Schuurmans, Le, Levine, and Ma}]{chu2025sftmemorizesrlgeneralizes}
Tianzhe Chu, Yuexiang Zhai, Jihan Yang, Shengbang Tong, Saining Xie, Dale Schuurmans, Quoc~V. Le, Sergey Levine, and Yi~Ma. 2025{\natexlab{a}}.
\newblock \href {https://arxiv.org/abs/2501.17161} {Sft memorizes, rl generalizes: A comparative study of foundation model post-training}.

\bibitem[{Chu et~al.(2025{\natexlab{b}})Chu, Huang, Zhang, Wei, and Wang}]{chu2025gpgsimplestrongreinforcement}
Xiangxiang Chu, Hailang Huang, Xiao Zhang, Fei Wei, and Yong Wang. 2025{\natexlab{b}}.
\newblock \href {https://arxiv.org/abs/2504.02546} {Gpg: A simple and strong reinforcement learning baseline for model reasoning}.

\bibitem[{Clark et~al.(2025)Clark, Mirchandani, Sadigh, and Belkhale}]{clark2025action}
Jaden Clark, Suvir Mirchandani, Dorsa Sadigh, and Suneel Belkhale. 2025.
\newblock \href {https://arxiv.org/abs/2502.03729} {Action-free reasoning for policy generalization}.
\newblock \emph{ArXiv preprint}, abs/2502.03729.

\bibitem[{Cobbe et~al.(2021)Cobbe, Kosaraju, Bavarian, Chen, Jun, Kaiser, Plappert, Tworek, Hilton, Nakano, Hesse, and Schulman}]{cobbe2021trainingverifierssolvemath}
Karl Cobbe, Vineet Kosaraju, Mohammad Bavarian, Mark Chen, Heewoo Jun, Lukasz Kaiser, Matthias Plappert, Jerry Tworek, Jacob Hilton, Reiichiro Nakano, Christopher Hesse, and John Schulman. 2021.
\newblock \href {https://arxiv.org/abs/2110.14168} {Training verifiers to solve math word problems}.

\bibitem[{Cuadron et~al.(2025)Cuadron, Li, Ma, Wang, Wang, Zhuang, Liu, Schroeder, Xia, Mao, Thumiger, Desai, Stoica, Klimovic, Neubig, and Gonzalez}]{cuadron2025dangeroverthinkingexaminingreasoningaction}
Alejandro Cuadron, Dacheng Li, Wenjie Ma, Xingyao Wang, Yichuan Wang, Siyuan Zhuang, Shu Liu, Luis~Gaspar Schroeder, Tian Xia, Huanzhi Mao, Nicholas Thumiger, Aditya Desai, Ion Stoica, Ana Klimovic, Graham Neubig, and Joseph~E. Gonzalez. 2025.
\newblock \href {https://arxiv.org/abs/2502.08235} {The danger of overthinking: Examining the reasoning-action dilemma in agentic tasks}.

\bibitem[{Cui et~al.(2025{\natexlab{a}})Cui, Yuan, Wang, Wang, Li, He, Fan, Yu, Xu, Chen, Yuan, Chen, Zhang, Lv, Wang, Yao, Peng, Cheng, Liu, Sun, Zhou, and Ding}]{cui2024process}
Ganqu Cui, Lifan Yuan, Zefan Wang, Hanbin Wang, Wendi Li, Bingxiang He, Yuchen Fan, Tianyu Yu, Qixin Xu, Weize Chen, Jiarui Yuan, Huayu Chen, Kaiyan Zhang, Xingtai Lv, Shuo Wang, Yuan Yao, Hao Peng, Yu~Cheng, Zhiyuan Liu, Maosong Sun, Bowen Zhou, and Ning Ding. 2025{\natexlab{a}}.
\newblock Process reinforcement through implicit rewards.

\bibitem[{Cui et~al.(2025{\natexlab{b}})Cui, He, Zeng, Liu, Tang, Dai, Han, Luo, Huang, Li, Wang, Xing, Tang, and He}]{cui2025stepwiseperplexityguidedrefinementefficient}
Yingqian Cui, Pengfei He, Jingying Zeng, Hui Liu, Xianfeng Tang, Zhenwei Dai, Yan Han, Chen Luo, Jing Huang, Zhen Li, Suhang Wang, Yue Xing, Jiliang Tang, and Qi~He. 2025{\natexlab{b}}.
\newblock \href {https://arxiv.org/abs/2502.13260} {Stepwise perplexity-guided refinement for efficient chain-of-thought reasoning in large language models}.

\bibitem[{Dang and Ngo(2025)}]{dang2025reinforcementlearningreasoningsmall}
Quy-Anh Dang and Chris Ngo. 2025.
\newblock \href {https://arxiv.org/abs/2503.16219} {Reinforcement learning for reasoning in small llms: What works and what doesn't}.

\bibitem[{Dao and Gu(2024)}]{dao2024transformersssmsgeneralizedmodels}
Tri Dao and Albert Gu. 2024.
\newblock \href {https://arxiv.org/abs/2405.21060} {Transformers are ssms: Generalized models and efficient algorithms through structured state space duality}.

\bibitem[{De~Moura et~al.(2015)De~Moura, Kong, Avigad, Van~Doorn, and von Raumer}]{de2015lean}
Leonardo De~Moura, Soonho Kong, Jeremy Avigad, Floris Van~Doorn, and Jakob von Raumer. 2015.
\newblock The lean theorem prover (system description).
\newblock In \emph{Automated Deduction-CADE-25: 25th International Conference on Automated Deduction, Berlin, Germany, August 1-7, 2015, Proceedings 25}, pages 378--388. Springer.

\bibitem[{DeepSeek-AI(2024)}]{deepseekv2}
DeepSeek-AI. 2024.
\newblock \href {http://arxiv.org/abs/2405.04434} {Deepseek-v2: A strong, economical, and efficient mixture-of-experts language model}.

\bibitem[{DeepSeek-AI et~al.(2025)DeepSeek-AI, Guo, Yang, Zhang, Song, Zhang, Xu, Zhu, Ma, Wang, Bi, Zhang, Yu, Wu, Wu, Gou, Shao, Li, Gao, Liu, Xue, Wang, Wu, Feng, Lu, Zhao, Deng, Zhang, Ruan, Dai, Chen, Ji, Li, Lin, Dai, Luo, Hao, Chen, Li, Zhang, Bao, Xu, Wang, Ding, Xin, Gao, Qu, Li, Guo, Li, Wang, Chen, Yuan, Qiu, Li, Cai, Ni, Liang, Chen, Dong, Hu, Gao, Guan, Huang, Yu, Wang, Zhang, Zhao, Wang, Zhang, Xu, Xia, Zhang, Zhang, Tang, Li, Wang, Li, Tian, Huang, Zhang, Wang, Chen, Du, Ge, Zhang, Pan, Wang, Chen, Jin, Chen, Lu, Zhou, Chen, Ye, Wang, Yu, Zhou, Pan, Li, Zhou, Wu, Ye, Yun, Pei, Sun, Wang, Zeng, Zhao, Liu, Liang, Gao, Yu, Zhang, Xiao, An, Liu, Wang, Chen, Nie, Cheng, Liu, Xie, Liu, Yang, Li, Su, Lin, Li, Jin, Shen, Chen, Sun, Wang, Song, Zhou, Wang, Shan, Li, Wang, Wei, Zhang, Xu, Li, Zhao, Sun, Wang, Yu, Zhang, Shi, Xiong, He, Piao, Wang, Tan, Ma, Liu, Guo, Ou, Wang, Gong, Zou, He, Xiong, Luo, You, Liu, Zhou, Zhu, Xu, Huang, Li, Zheng, Zhu, Ma, Tang, Zha, Yan, Ren, Ren, Sha, Fu, Xu, Xie, Zhang,
  Hao, Ma, Yan, Wu, Gu, Zhu, Liu, Li, Xie, Song, Pan, Huang, Xu, Zhang, and Zhang}]{deepseekai2025deepseekr1incentivizingreasoningcapability}
DeepSeek-AI, Daya Guo, Dejian Yang, Haowei Zhang, Junxiao Song, Ruoyu Zhang, Runxin Xu, Qihao Zhu, Shirong Ma, Peiyi Wang, Xiao Bi, Xiaokang Zhang, Xingkai Yu, Yu~Wu, Z.~F. Wu, Zhibin Gou, Zhihong Shao, Zhuoshu Li, Ziyi Gao, Aixin Liu, Bing Xue, Bingxuan Wang, Bochao Wu, Bei Feng, Chengda Lu, Chenggang Zhao, Chengqi Deng, Chenyu Zhang, Chong Ruan, Damai Dai, Deli Chen, Dongjie Ji, Erhang Li, Fangyun Lin, Fucong Dai, Fuli Luo, Guangbo Hao, Guanting Chen, Guowei Li, H.~Zhang, Han Bao, Hanwei Xu, Haocheng Wang, Honghui Ding, Huajian Xin, Huazuo Gao, Hui Qu, Hui Li, Jianzhong Guo, Jiashi Li, Jiawei Wang, Jingchang Chen, Jingyang Yuan, Junjie Qiu, Junlong Li, J.~L. Cai, Jiaqi Ni, Jian Liang, Jin Chen, Kai Dong, Kai Hu, Kaige Gao, Kang Guan, Kexin Huang, Kuai Yu, Lean Wang, Lecong Zhang, Liang Zhao, Litong Wang, Liyue Zhang, Lei Xu, Leyi Xia, Mingchuan Zhang, Minghua Zhang, Minghui Tang, Meng Li, Miaojun Wang, Mingming Li, Ning Tian, Panpan Huang, Peng Zhang, Qiancheng Wang, Qinyu Chen, Qiushi Du, Ruiqi Ge, Ruisong
  Zhang, Ruizhe Pan, Runji Wang, R.~J. Chen, R.~L. Jin, Ruyi Chen, Shanghao Lu, Shangyan Zhou, Shanhuang Chen, Shengfeng Ye, Shiyu Wang, Shuiping Yu, Shunfeng Zhou, Shuting Pan, S.~S. Li, Shuang Zhou, Shaoqing Wu, Shengfeng Ye, Tao Yun, Tian Pei, Tianyu Sun, T.~Wang, Wangding Zeng, Wanjia Zhao, Wen Liu, Wenfeng Liang, Wenjun Gao, Wenqin Yu, Wentao Zhang, W.~L. Xiao, Wei An, Xiaodong Liu, Xiaohan Wang, Xiaokang Chen, Xiaotao Nie, Xin Cheng, Xin Liu, Xin Xie, Xingchao Liu, Xinyu Yang, Xinyuan Li, Xuecheng Su, Xuheng Lin, X.~Q. Li, Xiangyue Jin, Xiaojin Shen, Xiaosha Chen, Xiaowen Sun, Xiaoxiang Wang, Xinnan Song, Xinyi Zhou, Xianzu Wang, Xinxia Shan, Y.~K. Li, Y.~Q. Wang, Y.~X. Wei, Yang Zhang, Yanhong Xu, Yao Li, Yao Zhao, Yaofeng Sun, Yaohui Wang, Yi~Yu, Yichao Zhang, Yifan Shi, Yiliang Xiong, Ying He, Yishi Piao, Yisong Wang, Yixuan Tan, Yiyang Ma, Yiyuan Liu, Yongqiang Guo, Yuan Ou, Yuduan Wang, Yue Gong, Yuheng Zou, Yujia He, Yunfan Xiong, Yuxiang Luo, Yuxiang You, Yuxuan Liu, Yuyang Zhou, Y.~X. Zhu,
  Yanhong Xu, Yanping Huang, Yaohui Li, Yi~Zheng, Yuchen Zhu, Yunxian Ma, Ying Tang, Yukun Zha, Yuting Yan, Z.~Z. Ren, Zehui Ren, Zhangli Sha, Zhe Fu, Zhean Xu, Zhenda Xie, Zhengyan Zhang, Zhewen Hao, Zhicheng Ma, Zhigang Yan, Zhiyu Wu, Zihui Gu, Zijia Zhu, Zijun Liu, Zilin Li, Ziwei Xie, Ziyang Song, Zizheng Pan, Zhen Huang, Zhipeng Xu, Zhongyu Zhang, and Zhen Zhang. 2025.
\newblock \href {https://arxiv.org/abs/2501.12948} {Deepseek-r1: Incentivizing reasoning capability in llms via reinforcement learning}.

\bibitem[{Deng et~al.(2024)Deng, Choi, and Shieber}]{deng2024explicitcotimplicitcot}
Yuntian Deng, Yejin Choi, and Stuart Shieber. 2024.
\newblock \href {https://arxiv.org/abs/2405.14838} {From explicit cot to implicit cot: Learning to internalize cot step by step}.

\bibitem[{Deng et~al.(2023)Deng, Prasad, Fernandez, Smolensky, Chaudhary, and Shieber}]{deng2023implicitchainthoughtreasoning}
Yuntian Deng, Kiran Prasad, Roland Fernandez, Paul Smolensky, Vishrav Chaudhary, and Stuart Shieber. 2023.
\newblock \href {https://arxiv.org/abs/2311.01460} {Implicit chain of thought reasoning via knowledge distillation}.

\bibitem[{Dhuliawala et~al.(2023)Dhuliawala, Komeili, Xu, Raileanu, Li, Celikyilmaz, and Weston}]{dhuliawala2023chainofverificationreduceshallucinationlarge}
Shehzaad Dhuliawala, Mojtaba Komeili, Jing Xu, Roberta Raileanu, Xian Li, Asli Celikyilmaz, and Jason Weston. 2023.
\newblock \href {https://arxiv.org/abs/2309.11495} {Chain-of-verification reduces hallucination in large language models}.

\bibitem[{Ding et~al.(2024)Ding, Liu, Fu, Song, Xie, and Zhang}]{ding2024breakchainlargelanguage}
Mengru Ding, Hanmeng Liu, Zhizhang Fu, Jian Song, Wenbo Xie, and Yue Zhang. 2024.
\newblock \href {https://arxiv.org/abs/2406.06580} {Break the chain: Large language models can be shortcut reasoners}.

\bibitem[{Ding et~al.(2025)Ding, Jiang, Liu, Jing, Guo, Wang, Zhang, Wang, Liu, Du, Liu, and Tao}]{ding2025dynamicparalleltreesearch}
Yifu Ding, Wentao Jiang, Shunyu Liu, Yongcheng Jing, Jinyang Guo, Yingjie Wang, Jing Zhang, Zengmao Wang, Ziwei Liu, Bo~Du, Xianglong Liu, and Dacheng Tao. 2025.
\newblock \href {https://arxiv.org/abs/2502.16235} {Dynamic parallel tree search for efficient llm reasoning}.

\bibitem[{Dong et~al.(2023)Dong, Xiong, Goyal, Zhang, Chow, Pan, Diao, Zhang, Shum, and Zhang}]{dongRAFTRewardRAnked2023}
Hanze Dong, Wei Xiong, Deepanshu Goyal, Yihan Zhang, Winnie Chow, Rui Pan, Shizhe Diao, Jipeng Zhang, Kashun Shum, and Tong Zhang. 2023.
\newblock \href {https://arxiv.org/abs/2304.06767} {Raft: Reward ranked finetuning for generative foundation model alignment}.

\bibitem[{Dong et~al.(2024)Dong, Liu, Sun, Yang, Hu, Rao, and Liu}]{insightv}
Yuhao Dong, Zuyan Liu, Hai-Long Sun, Jingkang Yang, Winston Hu, Yongming Rao, and Ziwei Liu. 2024.
\newblock \href {https://arxiv.org/abs/2411.14432} {Insight-v: Exploring long-chain visual reasoning with multimodal large language models}.
\newblock \emph{ArXiv preprint}, abs/2411.14432.

\bibitem[{Du et~al.(2025)Du, Liu, Li, Zhao, Huo, Wang, Chen, Liu, Wang, and Wen}]{du2025virgo}
Yifan Du, Zikang Liu, Yifan Li, Wayne~Xin Zhao, Yuqi Huo, Bingning Wang, Weipeng Chen, Zheng Liu, Zhongyuan Wang, and Ji-Rong Wen. 2025.
\newblock \href {https://arxiv.org/abs/2501.01904} {Virgo: A preliminary exploration on reproducing o1-like mllm}.
\newblock \emph{ArXiv preprint}, abs/2501.01904.

\bibitem[{Du et~al.(2023)Du, Li, Torralba, Tenenbaum, and Mordatch}]{du2023improvingfactualityreasoninglanguage}
Yilun Du, Shuang Li, Antonio Torralba, Joshua~B. Tenenbaum, and Igor Mordatch. 2023.
\newblock \href {https://arxiv.org/abs/2305.14325} {Improving factuality and reasoning in language models through multiagent debate}.

\bibitem[{El-Kishky et~al.(2025)El-Kishky, Wei, Saraiva, Minaev, Selsam, Dohan, Song, Lightman, Clavera, Pachocki et~al.}]{el2025competitive}
Ahmed El-Kishky, Alexander Wei, Andre Saraiva, Borys Minaev, Daniel Selsam, David Dohan, Francis Song, Hunter Lightman, Ignasi Clavera, Jakub Pachocki, et~al. 2025.
\newblock \href {https://arxiv.org/abs/2502.06807} {Competitive programming with large reasoning models}.
\newblock \emph{ArXiv preprint}, abs/2502.06807.

\bibitem[{EvolvingLMMs-Lab(2025)}]{open-r1-multimodal}
EvolvingLMMs-Lab. 2025.
\newblock open-r1-multimodal: A fork to add multimodal model training to open-r1.
\newblock \url{https://github.com/EvolvingLMMs-Lab/open-r1-multimodal}.

\bibitem[{Face(2025)}]{openr1}
Hugging Face. 2025.
\newblock \href {https://github.com/huggingface/open-r1} {Open r1: A fully open reproduction of deepseek-r1}.

\bibitem[{Fang et~al.(2025)Fang, Duan, Wang, Huang, Li, Yan, Tian, Zeng, Zhao, Dai et~al.}]{fang2025got}
Rongyao Fang, Chengqi Duan, Kun Wang, Linjiang Huang, Hao Li, Shilin Yan, Hao Tian, Xingyu Zeng, Rui Zhao, Jifeng Dai, et~al. 2025.
\newblock \href {https://arxiv.org/abs/2503.10639} {Got: Unleashing reasoning capability of multimodal large language model for visual generation and editing}.
\newblock \emph{ArXiv preprint}, abs/2503.10639.

\bibitem[{Fatemi et~al.(2025)Fatemi, Rafiee, Tang, and Talamadupula}]{fatemi2025concisereasoningreinforcementlearning}
Mehdi Fatemi, Banafsheh Rafiee, Mingjie Tang, and Kartik Talamadupula. 2025.
\newblock \href {https://arxiv.org/abs/2504.05185} {Concise reasoning via reinforcement learning}.

\bibitem[{Feng et~al.(2025{\natexlab{a}})Feng, Gong, Li, Guo, Wang, Peng, Wang, and Yue}]{feng2025video}
Kaituo Feng, Kaixiong Gong, Bohao Li, Zonghao Guo, Yibing Wang, Tianshuo Peng, Benyou Wang, and Xiangyu Yue. 2025{\natexlab{a}}.
\newblock \href {https://arxiv.org/abs/2503.21776} {Video-r1: Reinforcing video reasoning in mllms}.
\newblock \emph{ArXiv preprint}, abs/2503.21776.

\bibitem[{Feng et~al.(2025{\natexlab{b}})Feng, Hao, Zhang, Song, and Wang}]{feng2025airragactivatingintrinsicreasoning}
Wenfeng Feng, Chuzhan Hao, Yuewei Zhang, Jingyi Song, and Hao Wang. 2025{\natexlab{b}}.
\newblock \href {https://arxiv.org/abs/2501.10053} {Airrag: Activating intrinsic reasoning for retrieval augmented generation using tree-based search}.

\bibitem[{Feng et~al.(2023)Feng, Wan, Wen, McAleer, Wen, Zhang, and Wang}]{fengAlphazerolikeTreeSearchCan2024}
Xidong Feng, Ziyu Wan, Muning Wen, Stephen~Marcus McAleer, Ying Wen, Weinan Zhang, and Jun Wang. 2023.
\newblock \href {https://arxiv.org/abs/2309.17179} {Alphazero-like tree-search can guide large language model decoding and training}.

\bibitem[{First et~al.(2023)First, Rabe, Ringer, and Brun}]{first2023baldurwholeproofgenerationrepair}
Emily First, Markus~N. Rabe, Talia Ringer, and Yuriy Brun. 2023.
\newblock \href {https://arxiv.org/abs/2303.04910} {Baldur: Whole-proof generation and repair with large language models}.

\bibitem[{Fu and Khot(2022)}]{fu2022gptroadmap}
Hao Fu, Yao;~Peng and Tushar Khot. 2022.
\newblock \href {https://yaofu.notion.site/How-does-GPT-Obtain-its-Ability-Tracing-Emergent-Abilities-of-Language-Models-to-their-Sources-b9a57ac0fcf74f30a1ab9e3e36fa1dc1} {How does gpt obtain its ability? tracing emergent abilities of language models to their sources}.
\newblock \emph{Yao Fu’s Notion}.

\bibitem[{Fu et~al.(2024{\natexlab{a}})Fu, Ng, Jiang, and Liu}]{fu2023gptscore}
Jinlan Fu, See-Kiong Ng, Zhengbao Jiang, and Pengfei Liu. 2024{\natexlab{a}}.
\newblock \href {https://aclanthology.org/2024.naacl-long.365} {{GPTS}core: Evaluate as you desire}.
\newblock In \emph{Proceedings of the 2024 Conference of the North American Chapter of the Association for Computational Linguistics: Human Language Technologies (Volume 1: Long Papers)}, pages 6556--6576, Mexico City, Mexico. Association for Computational Linguistics.

\bibitem[{Fu et~al.(2024{\natexlab{b}})Fu, Chen, Zhu, Fu, Dai, Qiao, and Zhang}]{fu2024efficientlyservingllmreasoning}
Yichao Fu, Junda Chen, Siqi Zhu, Zheyu Fu, Zhongdongming Dai, Aurick Qiao, and Hao Zhang. 2024{\natexlab{b}}.
\newblock \href {https://arxiv.org/abs/2412.20993} {Efficiently serving llm reasoning programs with certaindex}.

\bibitem[{Gandhi et~al.(2025)Gandhi, Chakravarthy, Singh, Lile, and Goodman}]{gandhi2025cognitivebehaviorsenableselfimproving}
Kanishk Gandhi, Ayush Chakravarthy, Anikait Singh, Nathan Lile, and Noah~D. Goodman. 2025.
\newblock \href {https://arxiv.org/abs/2503.01307} {Cognitive behaviors that enable self-improving reasoners, or, four habits of highly effective stars}.

\bibitem[{Gandhi et~al.(2024)Gandhi, Lee, Grand, Liu, Cheng, Sharma, and Goodman}]{gandhi2024streamsearchsoslearning}
Kanishk Gandhi, Denise Lee, Gabriel Grand, Muxin Liu, Winson Cheng, Archit Sharma, and Noah~D. Goodman. 2024.
\newblock \href {https://arxiv.org/abs/2404.03683} {Stream of search (sos): Learning to search in language}.

\bibitem[{Gao et~al.(2024{\natexlab{a}})Gao, Xu, Ye, Liu, He, Fu, Mei, Wang, and Wu}]{gao2024designingeffectiverlreward}
Jiaxuan Gao, Shusheng Xu, Wenjie Ye, Weilin Liu, Chuyi He, Wei Fu, Zhiyu Mei, Guangju Wang, and Yi~Wu. 2024{\natexlab{a}}.
\newblock \href {https://arxiv.org/abs/2410.15115} {On designing effective rl reward at training time for llm reasoning}.

\bibitem[{Gao et~al.(2023)Gao, Schulman, and Hilton}]{gao2022scalinglawsrewardmodel}
Leo Gao, John Schulman, and Jacob Hilton. 2023.
\newblock \href {https://proceedings.mlr.press/v202/gao23h.html} {Scaling laws for reward model overoptimization}.
\newblock In \emph{International Conference on Machine Learning, {ICML} 2023, 23-29 July 2023, Honolulu, Hawaii, {USA}}, volume 202 of \emph{Proceedings of Machine Learning Research}, pages 10835--10866. {PMLR}.

\bibitem[{Gao et~al.(2024{\natexlab{b}})Gao, Niu, He, Xu, Liu, Liu, Hu, and Wen}]{gaoInterpretableContrastiveMonte2024}
Zitian Gao, Boye Niu, Xuzheng He, Haotian Xu, Hongzhang Liu, Aiwei Liu, Xuming Hu, and Lijie Wen. 2024{\natexlab{b}}.
\newblock \href {https://arxiv.org/abs/2410.01707} {Interpretable contrastive monte carlo tree search reasoning}.

\bibitem[{Ge et~al.(2024)Ge, Zhou, Hou, Khabsa, Wang, Wang, Han, and Mao}]{ge2024mart}
Suyu Ge, Chunting Zhou, Rui Hou, Madian Khabsa, Yi-Chia Wang, Qifan Wang, Jiawei Han, and Yuning Mao. 2024.
\newblock \href {https://aclanthology.org/2024.naacl-long.107} {{MART}: Improving {LLM} safety with multi-round automatic red-teaming}.
\newblock In \emph{Proceedings of the 2024 Conference of the North American Chapter of the Association for Computational Linguistics: Human Language Technologies (Volume 1: Long Papers)}, pages 1927--1937, Mexico City, Mexico. Association for Computational Linguistics.

\bibitem[{Geiping et~al.(2025)Geiping, McLeish, Jain, Kirchenbauer, Singh, Bartoldson, Kailkhura, Bhatele, and Goldstein}]{geiping2025scalingtesttimecomputelatent}
Jonas Geiping, Sean McLeish, Neel Jain, John Kirchenbauer, Siddharth Singh, Brian~R. Bartoldson, Bhavya Kailkhura, Abhinav Bhatele, and Tom Goldstein. 2025.
\newblock \href {https://arxiv.org/abs/2502.05171} {Scaling up test-time compute with latent reasoning: A recurrent depth approach}.

\bibitem[{Gou et~al.(2023)Gou, Shao, Gong, Shen, Yang, Duan, and Chen}]{gou2024criticlargelanguagemodels}
Zhibin Gou, Zhihong Shao, Yeyun Gong, Yelong Shen, Yujiu Yang, Nan Duan, and Weizhu Chen. 2023.
\newblock \href {https://arxiv.org/abs/2305.11738} {Critic: Large language models can self-correct with tool-interactive critiquing}.

\bibitem[{Gu and Dao(2023)}]{gu2024mambalineartimesequencemodeling}
Albert Gu and Tri Dao. 2023.
\newblock \href {https://arxiv.org/abs/2312.00752} {Mamba: Linear-time sequence modeling with selective state spaces}.

\bibitem[{Gu et~al.(2024{\natexlab{a}})Gu, Rozi{\`e}re, Leather, Solar-Lezama, Synnaeve, and Wang}]{gu2024cruxeval}
Alex Gu, Baptiste Rozi{\`e}re, Hugh Leather, Armando Solar-Lezama, Gabriel Synnaeve, and Sida~I Wang. 2024{\natexlab{a}}.
\newblock \href {https://arxiv.org/abs/2401.03065} {Cruxeval: A benchmark for code reasoning, understanding and execution}.
\newblock \emph{ArXiv preprint}, abs/2401.03065.

\bibitem[{Gu et~al.(2024{\natexlab{b}})Gu, Jiang, Shi, Tan, Zhai, Xu, Li, Shen, Ma, Liu, Wang, Zhang, Wang, Gao, Ni, and Guo}]{gu2025surveyllmasajudge}
Jiawei Gu, Xuhui Jiang, Zhichao Shi, Hexiang Tan, Xuehao Zhai, Chengjin Xu, Wei Li, Yinghan Shen, Shengjie Ma, Honghao Liu, Saizhuo Wang, Kun Zhang, Yuanzhuo Wang, Wen Gao, Lionel Ni, and Jian Guo. 2024{\natexlab{b}}.
\newblock \href {https://arxiv.org/abs/2411.15594} {A survey on llm-as-a-judge}.

\bibitem[{Gu et~al.(2023)Gu, Deng, and Su}]{guDontGenerateDiscriminate2023}
Yu~Gu, Xiang Deng, and Yu~Su. 2023.
\newblock \href {https://doi.org/10.18653/v1/2023.acl-long.270} {Don{'}t generate, discriminate: A proposal for grounding language models to real-world environments}.
\newblock In \emph{Proceedings of the 61st Annual Meeting of the Association for Computational Linguistics (Volume 1: Long Papers)}, pages 4928--4949, Toronto, Canada. Association for Computational Linguistics.

\bibitem[{Guan et~al.(2024)Guan, Joglekar, Wallace, Jain, Barak, Helyar, Dias, Vallone, Ren, Wei, Chung, Toyer, Heidecke, Beutel, and Glaese}]{guan2025deliberate}
Melody~Y. Guan, Manas Joglekar, Eric Wallace, Saachi Jain, Boaz Barak, Alec Helyar, Rachel Dias, Andrea Vallone, Hongyu Ren, Jason Wei, Hyung~Won Chung, Sam Toyer, Johannes Heidecke, Alex Beutel, and Amelia Glaese. 2024.
\newblock \href {https://arxiv.org/abs/2412.16339} {Deliberative alignment: Reasoning enables safer language models}.
\newblock \emph{ArXiv preprint}, abs/2412.16339.

\bibitem[{Guan et~al.(2025{\natexlab{a}})Guan, Zeng, Meng, Xin, Lu, Lin, Han, Sun, and Zhou}]{guan2025deepragthinkingretrievalstep}
Xinyan Guan, Jiali Zeng, Fandong Meng, Chunlei Xin, Yaojie Lu, Hongyu Lin, Xianpei Han, Le~Sun, and Jie Zhou. 2025{\natexlab{a}}.
\newblock \href {https://arxiv.org/abs/2502.01142} {Deeprag: Thinking to retrieval step by step for large language models}.

\bibitem[{Guan et~al.(2025{\natexlab{b}})Guan, Zhang, Liu, Shang, Sun, Zhu, Yang, and Yang}]{guanRStarMathSmallLLMs2025}
Xinyu Guan, Li~Lyna Zhang, Yifei Liu, Ning Shang, Youran Sun, Yi~Zhu, Fan Yang, and Mao Yang. 2025{\natexlab{b}}.
\newblock \href {https://arxiv.org/abs/2501.04519} {rstar-math: Small llms can master math reasoning with self-evolved deep thinking}.

\bibitem[{Gulcehre et~al.(2023)Gulcehre, Paine, Srinivasan, Konyushkova, Weerts, Sharma, Siddhant, Ahern, Wang, Gu et~al.}]{gulcehreReinforcedSelfTrainingReST2023}
Caglar Gulcehre, Tom~Le Paine, Srivatsan Srinivasan, Ksenia Konyushkova, Lotte Weerts, Abhishek Sharma, Aditya Siddhant, Alex Ahern, Miaosen Wang, Chenjie Gu, et~al. 2023.
\newblock \href {https://arxiv.org/abs/2308.08998} {Reinforced self-training (rest) for language modeling}.
\newblock \emph{ArXiv preprint}, abs/2308.08998.

\bibitem[{Guo et~al.(2024)Guo, Zheng, Bai, Li, Wang, Zhu, Li, Neubig, Chen, and Yue}]{guo2024mammoth}
Jarvis Guo, Tuney Zheng, Yuelin Bai, Bo~Li, Yubo Wang, King Zhu, Yizhi Li, Graham Neubig, Wenhu Chen, and Xiang Yue. 2024.
\newblock \href {https://arxiv.org/abs/2412.05237} {Mammoth-vl: Eliciting multimodal reasoning with instruction tuning at scale}.
\newblock \emph{ArXiv preprint}, abs/2412.05237.

\bibitem[{Guo et~al.(2025{\natexlab{a}})Guo, Zhang, Chen, Ji, Wang, Hu, and Chen}]{guo2025improving}
Yanjiang Guo, Jianke Zhang, Xiaoyu Chen, Xiang Ji, Yen-Jen Wang, Yucheng Hu, and Jianyu Chen. 2025{\natexlab{a}}.
\newblock \href {https://arxiv.org/abs/2501.16664} {Improving vision-language-action model with online reinforcement learning}.
\newblock \emph{ArXiv preprint}, abs/2501.16664.

\bibitem[{Guo et~al.(2025{\natexlab{b}})Guo, Zhang, Tong, Zhao, Gao, Li, and Heng}]{guo2025can}
Ziyu Guo, Renrui Zhang, Chengzhuo Tong, Zhizheng Zhao, Peng Gao, Hongsheng Li, and Pheng-Ann Heng. 2025{\natexlab{b}}.
\newblock \href {https://arxiv.org/abs/2501.13926} {Can we generate images with cot? let's verify and reinforce image generation step by step}.
\newblock \emph{ArXiv preprint}, abs/2501.13926.

\bibitem[{Haluptzok et~al.(2023)Haluptzok, Bowers, and Kalai}]{haluptzok2023languagemodelsteachprogram}
Patrick Haluptzok, Matthew Bowers, and Adam~Tauman Kalai. 2023.
\newblock \href {https://openreview.net/pdf?id=SaRj2ka1XZ3} {Language models can teach themselves to program better}.
\newblock In \emph{The Eleventh International Conference on Learning Representations, {ICLR} 2023, Kigali, Rwanda, May 1-5, 2023}. OpenReview.net.

\bibitem[{Han et~al.(2024)Han, Wang, Fang, Zhao, Ma, and Chen}]{han2025tokenbudgetawarellmreasoning}
Tingxu Han, Zhenting Wang, Chunrong Fang, Shiyu Zhao, Shiqing Ma, and Zhenyu Chen. 2024.
\newblock \href {https://arxiv.org/abs/2412.18547} {Token-budget-aware llm reasoning}.

\bibitem[{Hao et~al.(2024{\natexlab{a}})Hao, Gu, Luo, Liu, Shao, Wang, Xie, Ma, Samavedhi, Gao, Wang, and Hu}]{hao2024llmreasonersnewevaluation}
Shibo Hao, Yi~Gu, Haotian Luo, Tianyang Liu, Xiyan Shao, Xinyuan Wang, Shuhua Xie, Haodi Ma, Adithya Samavedhi, Qiyue Gao, Zhen Wang, and Zhiting Hu. 2024{\natexlab{a}}.
\newblock \href {https://arxiv.org/abs/2404.05221} {Llm reasoners: New evaluation, library, and analysis of step-by-step reasoning with large language models}.

\bibitem[{Hao et~al.(2023)Hao, Gu, Ma, Hong, Wang, Wang, and Hu}]{haoReasoningLanguageModel2023}
Shibo Hao, Yi~Gu, Haodi Ma, Joshua Hong, Zhen Wang, Daisy Wang, and Zhiting Hu. 2023.
\newblock \href {https://doi.org/10.18653/v1/2023.emnlp-main.507} {Reasoning with language model is planning with world model}.
\newblock In \emph{Proceedings of the 2023 Conference on Empirical Methods in Natural Language Processing}, pages 8154--8173, Singapore. Association for Computational Linguistics.

\bibitem[{Hao et~al.(2024{\natexlab{b}})Hao, Sukhbaatar, Su, Li, Hu, Weston, and Tian}]{hao2024traininglargelanguagemodels}
Shibo Hao, Sainbayar Sukhbaatar, DiJia Su, Xian Li, Zhiting Hu, Jason Weston, and Yuandong Tian. 2024{\natexlab{b}}.
\newblock \href {https://arxiv.org/abs/2412.06769} {Training large language models to reason in a continuous latent space}.

\bibitem[{Hart et~al.(1968)Hart, Nilsson, and Raphael}]{hart1968formal}
Peter~E Hart, Nils~J Nilsson, and Bertram Raphael. 1968.
\newblock A formal basis for the heuristic determination of minimum cost paths.
\newblock \emph{IEEE transactions on Systems Science and Cybernetics}, 4(2):100--107.

\bibitem[{Havrilla et~al.(2024)Havrilla, Raparthy, Nalmpantis, Dwivedi-Yu, Zhuravinskyi, Hambro, and Raileanu}]{havrilla2024glorewhenwhereimprove}
Alex Havrilla, Sharath Raparthy, Christoforus Nalmpantis, Jane Dwivedi-Yu, Maksym Zhuravinskyi, Eric Hambro, and Roberta Raileanu. 2024.
\newblock \href {https://arxiv.org/abs/2402.10963} {Glore: When, where, and how to improve llm reasoning via global and local refinements}.

\bibitem[{He et~al.(2024)He, Jin, Xia, Su, Fan, Zou, Hu, and Liu}]{he2024pcagentsleepai}
Yanheng He, Jiahe Jin, Shijie Xia, Jiadi Su, Runze Fan, Haoyang Zou, Xiangkun Hu, and Pengfei Liu. 2024.
\newblock \href {https://arxiv.org/abs/2412.17589} {Pc agent: While you sleep, ai works -- a cognitive journey into digital world}.

\bibitem[{Hendrycks et~al.(2021)Hendrycks, Burns, Kadavath, Arora, Basart, Tang, Song, and Steinhardt}]{hendrycks2021measuringmathematicalproblemsolving}
Dan Hendrycks, Collin Burns, Saurav Kadavath, Akul Arora, Steven Basart, Eric Tang, Dawn Song, and Jacob Steinhardt. 2021.
\newblock \href {https://arxiv.org/abs/2103.03874} {Measuring mathematical problem solving with the math dataset}.

\bibitem[{Hendrycks et~al.(2023)Hendrycks, Mazeika, and Woodside}]{hendrycks2023catastrophic}
Dan Hendrycks, Mantas Mazeika, and Thomas Woodside. 2023.
\newblock \href {https://arxiv.org/abs/2306.12001} {An overview of catastrophic {AI} risks}.
\newblock \emph{ArXiv preprint}, abs/2306.12001.

\bibitem[{Hochlehnert et~al.(2025)Hochlehnert, Bhatnagar, Udandarao, Albanie, Prabhu, and Bethge}]{hochlehnert2025soberlookprogresslanguage}
Andreas Hochlehnert, Hardik Bhatnagar, Vishaal Udandarao, Samuel Albanie, Ameya Prabhu, and Matthias Bethge. 2025.
\newblock \href {https://arxiv.org/abs/2504.07086} {A sober look at progress in language model reasoning: Pitfalls and paths to reproducibility}.

\bibitem[{Hooper et~al.(2024)Hooper, Kim, Mohammadzadeh, Maheswaran, Paik, Mahoney, Keutzer, and Gholami}]{hooper2024squeezedattentionacceleratinglong}
Coleman Hooper, Sehoon Kim, Hiva Mohammadzadeh, Monishwaran Maheswaran, June Paik, Michael~W. Mahoney, Kurt Keutzer, and Amir Gholami. 2024.
\newblock \href {https://arxiv.org/abs/2411.09688} {Squeezed attention: Accelerating long context length llm inference}.

\bibitem[{Hooper et~al.(2025)Hooper, Kim, Moon, Dilmen, Maheswaran, Lee, Mahoney, Shao, Keutzer, and Gholami}]{hooper2025etsefficienttreesearch}
Coleman Hooper, Sehoon Kim, Suhong Moon, Kerem Dilmen, Monishwaran Maheswaran, Nicholas Lee, Michael~W. Mahoney, Sophia Shao, Kurt Keutzer, and Amir Gholami. 2025.
\newblock \href {https://arxiv.org/abs/2502.13575} {Ets: Efficient tree search for inference-time scaling}.

\bibitem[{Hosseini et~al.(2024)Hosseini, Yuan, Malkin, Courville, Sordoni, and Agarwal}]{hosseiniVSTaRTrainingVerifiers2024b}
Arian Hosseini, Xingdi Yuan, Nikolay Malkin, Aaron Courville, Alessandro Sordoni, and Rishabh Agarwal. 2024.
\newblock \href {https://arxiv.org/abs/2402.06457} {V-star: Training verifiers for self-taught reasoners}.

\bibitem[{Hou et~al.(2024)Hou, Du, Niu, Du, Zeng, Liu, Huang, Wang, Tang, and Dong}]{hou2024doesrlhfscaleexploring}
Zhenyu Hou, Pengfan Du, Yilin Niu, Zhengxiao Du, Aohan Zeng, Xiao Liu, Minlie Huang, Hongning Wang, Jie Tang, and Yuxiao Dong. 2024.
\newblock \href {https://arxiv.org/abs/2412.06000} {Does rlhf scale? exploring the impacts from data, model, and method}.

\bibitem[{Hou et~al.(2025)Hou, Lv, Lu, Zhang, Li, Yao, Li, Tang, and Dong}]{hou2025advancinglanguagemodelreasoning}
Zhenyu Hou, Xin Lv, Rui Lu, Jiajie Zhang, Yujiang Li, Zijun Yao, Juanzi Li, Jie Tang, and Yuxiao Dong. 2025.
\newblock \href {https://arxiv.org/abs/2501.11651} {Advancing language model reasoning through reinforcement learning and inference scaling}.

\bibitem[{Hu et~al.(2022)Hu, Shen, Wallis, Allen{-}Zhu, Li, Wang, Wang, and Chen}]{hu2021loralowrankadaptationlarge}
Edward~J. Hu, Yelong Shen, Phillip Wallis, Zeyuan Allen{-}Zhu, Yuanzhi Li, Shean Wang, Lu~Wang, and Weizhu Chen. 2022.
\newblock \href {https://openreview.net/forum?id=nZeVKeeFYf9} {Lora: Low-rank adaptation of large language models}.
\newblock In \emph{The Tenth International Conference on Learning Representations, {ICLR} 2022, Virtual Event, April 25-29, 2022}. OpenReview.net.

\bibitem[{Hu(2025)}]{hu2025reinforcesimpleefficientapproach}
Jian Hu. 2025.
\newblock \href {https://arxiv.org/abs/2501.03262} {Reinforce++: A simple and efficient approach for aligning large language models}.

\bibitem[{Hu et~al.(2024{\natexlab{a}})Hu, Wu, Zhu, Wang, Zhang, Cao et~al.}]{hu2024openrlhf}
Jian Hu, Xibin Wu, Zilin Zhu, Weixun Wang, Dehao Zhang, Yu~Cao, et~al. 2024{\natexlab{a}}.
\newblock \href {https://arxiv.org/abs/2405.11143} {Openrlhf: An easy-to-use, scalable and high-performance rlhf framework}.
\newblock \emph{ArXiv preprint}, abs/2405.11143.

\bibitem[{Hu et~al.(2025)Hu, Zhang, Han, Jiang, and Xiangyu~Zhang}]{OpenReasonerZero2025}
Jingcheng Hu, Yinmin Zhang, Qi~Han, Daxin Jiang, and Heung-Yeung~Shum Xiangyu~Zhang. 2025.
\newblock Open-reasoner-zero: An open source approach to scaling reinforcement learning on the base model.
\newblock \url{https://github.com/Open-Reasoner-Zero/Open-Reasoner-Zero}.

\bibitem[{Hu et~al.(2024{\natexlab{b}})Hu, Ru, Qiu, Guo, Zhang, Xu, Luo, Liu, Zhang, and Zhang}]{hu-etal-2024-knowledge}
Xiangkun Hu, Dongyu Ru, Lin Qiu, Qipeng Guo, Tianhang Zhang, Yang Xu, Yun Luo, Pengfei Liu, Yue Zhang, and Zheng Zhang. 2024{\natexlab{b}}.
\newblock \href {https://doi.org/10.18653/v1/2024.emnlp-main.395} {Knowledge-centric hallucination detection}.
\newblock In \emph{Proceedings of the 2024 Conference on Empirical Methods in Natural Language Processing}, pages 6953--6975, Miami, Florida, USA. Association for Computational Linguistics.

\bibitem[{Huang et~al.(2025{\natexlab{a}})Huang, Huang, Leng, Liu, and Huang}]{huang2025efficienttesttimescalingselfcalibration}
Chengsong Huang, Langlin Huang, Jixuan Leng, Jiacheng Liu, and Jiaxin Huang. 2025{\natexlab{a}}.
\newblock \href {https://arxiv.org/abs/2503.00031} {Efficient test-time scaling via self-calibration}.

\bibitem[{Huang et~al.(2023{\natexlab{a}})Huang, Gu, Hou, Wu, Wang, Yu, and Han}]{huangLargeLanguageModels2023}
Jiaxin Huang, Shixiang Gu, Le~Hou, Yuexin Wu, Xuezhi Wang, Hongkun Yu, and Jiawei Han. 2023{\natexlab{a}}.
\newblock \href {https://doi.org/10.18653/v1/2023.emnlp-main.67} {Large language models can self-improve}.
\newblock In \emph{Proceedings of the 2023 Conference on Empirical Methods in Natural Language Processing}, pages 1051--1068, Singapore. Association for Computational Linguistics.

\bibitem[{Huang et~al.(2023{\natexlab{b}})Huang, Chen, Mishra, Zheng, Yu, Song, and Zhou}]{huang2024largelanguagemodelsselfcorrect}
Jie Huang, Xinyun Chen, Swaroop Mishra, Huaixiu~Steven Zheng, Adams~Wei Yu, Xinying Song, and Denny Zhou. 2023{\natexlab{b}}.
\newblock \href {https://arxiv.org/abs/2310.01798} {Large language models cannot self-correct reasoning yet}.

\bibitem[{Huang et~al.(2025{\natexlab{b}})Huang, Yu, Ma, Zhong, Feng, Wang, Chen, Peng, Feng, Qin, and Liu}]{Huang_2025}
Lei Huang, Weijiang Yu, Weitao Ma, Weihong Zhong, Zhangyin Feng, Haotian Wang, Qianglong Chen, Weihua Peng, Xiaocheng Feng, Bing Qin, and Ting Liu. 2025{\natexlab{b}}.
\newblock \href {https://doi.org/10.1145/3703155} {A survey on hallucination in large language models: Principles, taxonomy, challenges, and open questions}.
\newblock \emph{ACM Transactions on Information Systems}, 43(2):1–55.

\bibitem[{Huang et~al.(2022)Huang, Xia, Xiao, Chan, Liang, Florence, Zeng, Tompson, Mordatch, Chebotar, Sermanet, Brown, Jackson, Luu, Levine, Hausman, and Ichter}]{huang2022inner}
Wenlong Huang, Fei Xia, Ted Xiao, Harris Chan, Jacky Liang, Pete Florence, Andy Zeng, Jonathan Tompson, Igor Mordatch, Yevgen Chebotar, Pierre Sermanet, Noah Brown, Tomas Jackson, Linda Luu, Sergey Levine, Karol Hausman, and Brian Ichter. 2022.
\newblock \href {https://arxiv.org/abs/2207.05608} {Inner monologue: Embodied reasoning through planning with language models}.

\bibitem[{Huang et~al.(2025{\natexlab{c}})Huang, Jia, Zhai, Cao, Ye, Zhao, Hu, and Lin}]{huang2025vision}
Wenxuan Huang, Bohan Jia, Zijie Zhai, Shaosheng Cao, Zheyu Ye, Fei Zhao, Yao Hu, and Shaohui Lin. 2025{\natexlab{c}}.
\newblock \href {https://arxiv.org/abs/2503.06749} {Vision-r1: Incentivizing reasoning capability in multimodal large language models}.
\newblock \emph{ArXiv preprint}, abs/2503.06749.

\bibitem[{Huang et~al.(2024)Huang, Zou, Li, Liu, Zheng, Chern, Xia, Qin, Yuan, and Liu}]{o1journeypart2}
Zhen Huang, Haoyang Zou, Xuefeng Li, Yixiu Liu, Yuxiang Zheng, Ethan Chern, Shijie Xia, Yiwei Qin, Weizhe Yuan, and Pengfei Liu. 2024.
\newblock \href {https://github.com/GAIR-NLP/O1-Journey} {O1 replication journey – part 2: Surpassing o1-preview through simple distillation}.
\newblock \emph{Github}.

\bibitem[{Huang et~al.(2025{\natexlab{d}})Huang, Geng, Hua, Huang, Zou, Zhang, Liu, and Zhang}]{huang2025o1replicationjourney}
Zhongzhen Huang, Gui Geng, Shengyi Hua, Zhen Huang, Haoyang Zou, Shaoting Zhang, Pengfei Liu, and Xiaofan Zhang. 2025{\natexlab{d}}.
\newblock \href {https://arxiv.org/abs/2501.06458} {O1 replication journey -- part 3: Inference-time scaling for medical reasoning}.

\bibitem[{Hui et~al.(2024)Hui, Yang, Cui, Yang, Liu, Zhang, Liu, Zhang, Yu, Lu et~al.}]{hui2024qwen25coder}
Binyuan Hui, Jian Yang, Zeyu Cui, Jiaxi Yang, Dayiheng Liu, Lei Zhang, Tianyu Liu, Jiajun Zhang, Bowen Yu, Keming Lu, et~al. 2024.
\newblock \href {https://arxiv.org/abs/2409.12186} {Qwen2. 5-coder technical report}.
\newblock \emph{ArXiv preprint}, abs/2409.12186.

\bibitem[{Hwang et~al.(2024)Hwang, Kim, Kim, Ye, and Seo}]{hwang2024selfexploreenhancingmathematicalreasoning}
Hyeonbin Hwang, Doyoung Kim, Seungone Kim, Seonghyeon Ye, and Minjoon Seo. 2024.
\newblock \href {https://arxiv.org/abs/2404.10346} {Self-explore: Enhancing mathematical reasoning in language models with fine-grained rewards}.

\bibitem[{Jain et~al.(2025)Jain, Gonzalez-Pumariega, Chen, Rush, Zhao, and Choudhury}]{jain2025multiturncodegenerationsinglestep}
Arnav~Kumar Jain, Gonzalo Gonzalez-Pumariega, Wayne Chen, Alexander~M Rush, Wenting Zhao, and Sanjiban Choudhury. 2025.
\newblock \href {https://arxiv.org/abs/2502.20380} {Multi-turn code generation through single-step rewards}.

\bibitem[{Ji et~al.(2025{\natexlab{a}})Ji, Ramesh, Zimmer, Bogunovic, Wang, and Bou{-}Ammar}]{ji2025inferenceguard}
Xiaotong Ji, Shyam~Sundhar Ramesh, Matthieu Zimmer, Ilija Bogunovic, Jun Wang, and Haitham Bou{-}Ammar. 2025{\natexlab{a}}.
\newblock \href {https://arxiv.org/abs/2502.01208} {Almost surely safe alignment of large language models at inference-time}.
\newblock \emph{ArXiv preprint}, abs/2502.01208.

\bibitem[{Ji et~al.(2025{\natexlab{b}})Ji, Li, Ye, Wu, Yao, Xu, Mo, and Zhang}]{ji2025testtimecomputesystem1thinking}
Yixin Ji, Juntao Li, Hai Ye, Kaixin Wu, Kai Yao, Jia Xu, Linjian Mo, and Min Zhang. 2025{\natexlab{b}}.
\newblock \href {http://arxiv.org/abs/2501.02497} {Test-time compute: from system-1 thinking to system-2 thinking}.

\bibitem[{Jiang et~al.(2024)Jiang, Wang, Lu, Wang, Zhang, Liu, Durme, and Khashabi}]{jiang2024rationalystpretrainingprocesssupervisionimproving}
Dongwei Jiang, Guoxuan Wang, Yining Lu, Andrew Wang, Jingyu Zhang, Chuyu Liu, Benjamin~Van Durme, and Daniel Khashabi. 2024.
\newblock \href {https://arxiv.org/abs/2410.01044} {Rationalyst: Pre-training process-supervision for improving reasoning}.

\bibitem[{Jiang et~al.(2025{\natexlab{a}})Jiang, Xu, Li, Niu, Xiang, Li, Lin, and Poovendran}]{jiang2025safechainsafetylanguagemodels}
Fengqing Jiang, Zhangchen Xu, Yuetai Li, Luyao Niu, Zhen Xiang, Bo~Li, Bill~Yuchen Lin, and Radha Poovendran. 2025{\natexlab{a}}.
\newblock \href {https://arxiv.org/abs/2502.12025} {Safechain: Safety of language models with long chain-of-thought reasoning capabilities}.

\bibitem[{Jiang et~al.(2025{\natexlab{b}})Jiang, Lin, Cao, Tian, Kang, Wang, Sun, and Han}]{jiang2025deepretrievalhackingrealsearch}
Pengcheng Jiang, Jiacheng Lin, Lang Cao, Runchu Tian, SeongKu Kang, Zifeng Wang, Jimeng Sun, and Jiawei Han. 2025{\natexlab{b}}.
\newblock \href {http://arxiv.org/abs/2503.00223} {Deepretrieval: Hacking real search engines and retrievers with large language models via reinforcement learning}.

\bibitem[{Jimenez et~al.(2024)Jimenez, Yang, Wettig, Yao, Pei, Press, and Narasimhan}]{jimenez2024swebench}
Carlos~E Jimenez, John Yang, Alexander Wettig, Shunyu Yao, Kexin Pei, Ofir Press, and Karthik~R Narasimhan. 2024.
\newblock \href {https://openreview.net/forum?id=VTF8yNQM66} {{SWE}-bench: Can language models resolve real-world github issues?}
\newblock In \emph{The Twelfth International Conference on Learning Representations}.

\bibitem[{Jin et~al.(2025)Jin, Zeng, Yue, Wang, Zamani, and Han}]{jin2025searchr1trainingllmsreason}
Bowen Jin, Hansi Zeng, Zhenrui Yue, Dong Wang, Hamed Zamani, and Jiawei Han. 2025.
\newblock \href {https://arxiv.org/abs/2503.09516} {Search-r1: Training llms to reason and leverage search engines with reinforcement learning}.

\bibitem[{Jin et~al.(2024)Jin, Yu, Shu, Zhao, Hua, Meng, Zhang, and Du}]{jin2024impactreasoningsteplength}
Mingyu Jin, Qinkai Yu, Dong Shu, Haiyan Zhao, Wenyue Hua, Yanda Meng, Yongfeng Zhang, and Mengnan Du. 2024.
\newblock \href {https://arxiv.org/abs/2401.04925} {The impact of reasoning step length on large language models}.

\bibitem[{Jordan et~al.(2024)Jordan, White, da~Silva, White, and Thomas}]{jordan2024positionbenchmarkinglimitedreinforcement}
Scott~M. Jordan, Adam White, Bruno~Castro da~Silva, Martha White, and Philip~S. Thomas. 2024.
\newblock \href {https://arxiv.org/abs/2406.16241} {Position: Benchmarking is limited in reinforcement learning research}.

\bibitem[{Juravsky et~al.(2024)Juravsky, Brown, Ehrlich, Fu, Ré, and Mirhoseini}]{juravsky2024hydragenhighthroughputllminference}
Jordan Juravsky, Bradley Brown, Ryan Ehrlich, Daniel~Y. Fu, Christopher Ré, and Azalia Mirhoseini. 2024.
\newblock \href {https://arxiv.org/abs/2402.05099} {Hydragen: High-throughput llm inference with shared prefixes}.

\bibitem[{Kambhampati(2024)}]{Kambhampati_2024}
Subbarao Kambhampati. 2024.
\newblock \href {https://doi.org/10.1111/nyas.15125} {Can large language models reason and plan?}
\newblock \emph{Annals of the New York Academy of Sciences}, 1534(1):15–18.

\bibitem[{Kamoi et~al.(2024)Kamoi, Zhang, Zhang, Han, and Zhang}]{kamoi-etal-2024-llms}
Ryo Kamoi, Yusen Zhang, Nan Zhang, Jiawei Han, and Rui Zhang. 2024.
\newblock \href {https://doi.org/10.1162/tacl_a_00713} {When can {LLM}s actually correct their own mistakes? a critical survey of self-correction of {LLM}s}.
\newblock \emph{Transactions of the Association for Computational Linguistics}, 12:1417--1440.

\bibitem[{Kang et~al.(2024{\natexlab{a}})Kang, Li, Chen, Kazemi, Sun, Chen, Li, He, He, Wen, Hao, and Yao}]{kangMindStarEnhancingMath2024}
Jikun Kang, Xin~Zhe Li, Xi~Chen, Amirreza Kazemi, Qianyi Sun, Boxing Chen, Dong Li, Xu~He, Quan He, Feng Wen, Jianye Hao, and Jun Yao. 2024{\natexlab{a}}.
\newblock \href {https://arxiv.org/abs/2405.16265} {Mindstar: Enhancing math reasoning in pre-trained llms at inference time}.

\bibitem[{Kang et~al.(2024{\natexlab{b}})Kang, Sun, Chen, and Zou}]{kang2024c3otgeneratingshorterchainofthought}
Yu~Kang, Xianghui Sun, Liangyu Chen, and Wei Zou. 2024{\natexlab{b}}.
\newblock \href {https://arxiv.org/abs/2412.11664} {C3ot: Generating shorter chain-of-thought without compromising effectiveness}.

\bibitem[{Kang et~al.(2025)Kang, Zhao, and Song}]{kang2025scalablebestofnselectionlarge}
Zhewei Kang, Xuandong Zhao, and Dawn Song. 2025.
\newblock \href {https://arxiv.org/abs/2502.18581} {Scalable best-of-n selection for large language models via self-certainty}.

\bibitem[{Kaplan et~al.(2020)Kaplan, McCandlish, Henighan, Brown, Chess, Child, Gray, Radford, Wu, and Amodei}]{kaplan2020scalinglawsneurallanguage}
Jared Kaplan, Sam McCandlish, Tom Henighan, Tom~B. Brown, Benjamin Chess, Rewon Child, Scott Gray, Alec Radford, Jeffrey Wu, and Dario Amodei. 2020.
\newblock \href {https://arxiv.org/abs/2001.08361} {Scaling laws for neural language models}.

\bibitem[{Katharopoulos et~al.(2020)Katharopoulos, Vyas, Pappas, and Fleuret}]{katharopoulos2020transformersrnnsfastautoregressive}
Angelos Katharopoulos, Apoorv Vyas, Nikolaos Pappas, and Fran{\c{c}}ois Fleuret. 2020.
\newblock \href {http://proceedings.mlr.press/v119/katharopoulos20a.html} {Transformers are rnns: Fast autoregressive transformers with linear attention}.
\newblock In \emph{Proceedings of the 37th International Conference on Machine Learning, {ICML} 2020, 13-18 July 2020, Virtual Event}, volume 119 of \emph{Proceedings of Machine Learning Research}, pages 5156--5165. {PMLR}.

\bibitem[{Kazemnejad et~al.(2024)Kazemnejad, Aghajohari, Portelance, Sordoni, Reddy, Courville, and Roux}]{kazemnejad2024vineppounlockingrlpotential}
Amirhossein Kazemnejad, Milad Aghajohari, Eva Portelance, Alessandro Sordoni, Siva Reddy, Aaron Courville, and Nicolas~Le Roux. 2024.
\newblock \href {https://arxiv.org/abs/2410.01679} {Vineppo: Unlocking rl potential for llm reasoning through refined credit assignment}.

\bibitem[{Khan et~al.(2024)Khan, Hughes, Valentine, Ruis, Sachan, Radhakrishnan, Grefenstette, Bowman, Rockt{\"a}schel, and Perez}]{khanDebatingMorePersuasive2024}
Akbir Khan, John Hughes, Dan Valentine, Laura Ruis, Kshitij Sachan, Ansh Radhakrishnan, Edward Grefenstette, Samuel~R. Bowman, Tim Rockt{\"a}schel, and Ethan Perez. 2024.
\newblock \href {https://arxiv.org/abs/2402.06782} {Debating with more persuasive llms leads to more truthful answers}.

\bibitem[{Khanov et~al.(2024)Khanov, Burapacheep, and Li}]{khanov2024args}
Maxim Khanov, Jirayu Burapacheep, and Yixuan Li. 2024.
\newblock \href {https://openreview.net/forum?id=shgx0eqdw6} {{ARGS:} alignment as reward-guided search}.
\newblock In \emph{The Twelfth International Conference on Learning Representations, {ICLR} 2024, Vienna, Austria, May 7-11, 2024}. OpenReview.net.

\bibitem[{Kim et~al.(2023)Kim, Baldi, and McAleer}]{kim2023languagemodelssolvecomputer}
Geunwoo Kim, Pierre Baldi, and Stephen McAleer. 2023.
\newblock \href {http://papers.nips.cc/paper\_files/paper/2023/hash/7cc1005ec73cfbaac9fa21192b622507-Abstract-Conference.html} {Language models can solve computer tasks}.
\newblock In \emph{Advances in Neural Information Processing Systems 36: Annual Conference on Neural Information Processing Systems 2023, NeurIPS 2023, New Orleans, LA, USA, December 10 - 16, 2023}.

\bibitem[{Kim et~al.(2025)Kim, Wu, Lee, Yue, Lee, Moon, Gashteovski, Lawrence, Hockenmaier, Neubig, and Welleck}]{kim2025scalingevaluationtimecomputereasoning}
Seungone Kim, Ian Wu, Jinu Lee, Xiang Yue, Seongyun Lee, Mingyeong Moon, Kiril Gashteovski, Carolin Lawrence, Julia Hockenmaier, Graham Neubig, and Sean Welleck. 2025.
\newblock \href {https://arxiv.org/abs/2503.19877} {Scaling evaluation-time compute with reasoning models as process evaluators}.

\bibitem[{Kimi et~al.(2025)Kimi, Du, Gao, Xing, Jiang, Chen, Li, Xiao, Du, Liao et~al.}]{MoonshotAI}
Team Kimi, Angang Du, Bofei Gao, Bowei Xing, Changjiu Jiang, Cheng Chen, Cheng Li, Chenjun Xiao, Chenzhuang Du, Chonghua Liao, et~al. 2025.
\newblock \href {https://arxiv.org/abs/2501.12599} {Kimi k1. 5: Scaling reinforcement learning with llms}.
\newblock \emph{ArXiv preprint}, abs/2501.12599.

\bibitem[{{Kimi Team}(2025)}]{kimi2024vl}
{Kimi Team}. 2025.
\newblock Kimi-vl technical report.
\newblock \url{https://github.com/MoonshotAI/Kimi-VL/blob/main/Kimi-VL.pdf}.

\bibitem[{Kocsis and Szepesv{\'a}ri(2006)}]{kocsis2006bandit}
Levente Kocsis and Csaba Szepesv{\'a}ri. 2006.
\newblock Bandit based monte-carlo planning.
\newblock In \emph{European conference on machine learning}, pages 282--293. Springer.

\bibitem[{Koh et~al.(2024)Koh, McAleer, Fried, and Salakhutdinov}]{koh2024treesearchlanguagemodel}
Jing~Yu Koh, Stephen McAleer, Daniel Fried, and Ruslan Salakhutdinov. 2024.
\newblock \href {https://arxiv.org/abs/2407.01476} {Tree search for language model agents}.

\bibitem[{Kumar et~al.(2024)Kumar, Zhuang, Agarwal, Su, Co-Reyes, Singh, Baumli, Iqbal, Bishop, Roelofs, Zhang, McKinney, Shrivastava, Paduraru, Tucker, Precup, Behbahani, and Faust}]{kumar2024traininglanguagemodelsselfcorrect}
Aviral Kumar, Vincent Zhuang, Rishabh Agarwal, Yi~Su, John~D Co-Reyes, Avi Singh, Kate Baumli, Shariq Iqbal, Colton Bishop, Rebecca Roelofs, Lei~M Zhang, Kay McKinney, Disha Shrivastava, Cosmin Paduraru, George Tucker, Doina Precup, Feryal Behbahani, and Aleksandra Faust. 2024.
\newblock \href {https://arxiv.org/abs/2409.12917} {Training language models to self-correct via reinforcement learning}.

\bibitem[{Kumar et~al.(2025)Kumar, Ashraf, Thawakar, Anwer, Cholakkal, Shah, Yang, Torr, Khan, and Khan}]{kumar2025llmposttrainingdeepdive}
Komal Kumar, Tajamul Ashraf, Omkar Thawakar, Rao~Muhammad Anwer, Hisham Cholakkal, Mubarak Shah, Ming-Hsuan Yang, Phillip H.~S. Torr, Fahad~Shahbaz Khan, and Salman Khan. 2025.
\newblock \href {http://arxiv.org/abs/2502.21321} {Llm post-training: A deep dive into reasoning large language models}.

\bibitem[{Kwon et~al.(2023)Kwon, Li, Zhuang, Sheng, Zheng, Yu, Gonzalez, Zhang, and Stoica}]{kwon2023efficient}
Woosuk Kwon, Zhuohan Li, Siyuan Zhuang, Ying Sheng, Lianmin Zheng, Cody~Hao Yu, Joseph~E. Gonzalez, Hao Zhang, and Ion Stoica. 2023.
\newblock Efficient memory management for large language model serving with pagedattention.
\newblock In \emph{Proceedings of the ACM SIGOPS 29th Symposium on Operating Systems Principles}.

\bibitem[{Lample et~al.(2022)Lample, Lacroix, Lachaux, Rodriguez, Hayat, Lavril, Ebner, and Martinet}]{lample2022hypertreeproofsearchneural}
Guillaume Lample, Timoth{\'{e}}e Lacroix, Marie{-}Anne Lachaux, Aur{\'{e}}lien Rodriguez, Amaury Hayat, Thibaut Lavril, Gabriel Ebner, and Xavier Martinet. 2022.
\newblock \href {http://papers.nips.cc/paper\_files/paper/2022/hash/a8901c5e85fb8e1823bbf0f755053672-Abstract-Conference.html} {Hypertree proof search for neural theorem proving}.
\newblock In \emph{Advances in Neural Information Processing Systems 35: Annual Conference on Neural Information Processing Systems 2022, NeurIPS 2022, New Orleans, LA, USA, November 28 - December 9, 2022}.

\bibitem[{Le et~al.(2022)Le, Wang, Gotmare, Savarese, and Hoi}]{le2022coderl}
Hung Le, Yue Wang, Akhilesh~Deepak Gotmare, Silvio Savarese, and Steven~Chu{-}Hong Hoi. 2022.
\newblock \href {http://papers.nips.cc/paper\_files/paper/2022/hash/8636419dea1aa9fbd25fc4248e702da4-Abstract-Conference.html} {Coderl: Mastering code generation through pretrained models and deep reinforcement learning}.
\newblock In \emph{Advances in Neural Information Processing Systems 35: Annual Conference on Neural Information Processing Systems 2022, NeurIPS 2022, New Orleans, LA, USA, November 28 - December 9, 2022}.

\bibitem[{Lee et~al.(2025)Lee, Che, and Peng}]{lee2025llmscompresschainofthoughttoken}
Ayeong Lee, Ethan Che, and Tianyi Peng. 2025.
\newblock \href {https://arxiv.org/abs/2503.01141} {How well do llms compress their own chain-of-thought? a token complexity approach}.

\bibitem[{Lee et~al.(2024)Lee, Yang, Heo, Han, and Yoo}]{lee2024tokensupervisedvaluemodelsenhancing}
Jung~Hyun Lee, June~Yong Yang, Byeongho Heo, Dongyoon Han, and Kang~Min Yoo. 2024.
\newblock \href {https://arxiv.org/abs/2407.12863} {Token-supervised value models for enhancing mathematical reasoning capabilities of large language models}.

\bibitem[{Lehnert et~al.(2024)Lehnert, Sukhbaatar, Su, Zheng, Mcvay, Rabbat, and Tian}]{lehnert2024abetterplanningtransformers}
Lucas Lehnert, Sainbayar Sukhbaatar, DiJia Su, Qinqing Zheng, Paul Mcvay, Michael Rabbat, and Yuandong Tian. 2024.
\newblock \href {https://arxiv.org/abs/2402.14083} {Beyond a*: Better planning with transformers via search dynamics bootstrapping}.

\bibitem[{Li et~al.(2024{\natexlab{a}})Li, Zhang, Guo, Zhang, Li, Zhang, Zhang, Zhang, Li, Liu et~al.}]{llavaonevision}
Bo~Li, Yuanhan Zhang, Dong Guo, Renrui Zhang, Feng Li, Hao Zhang, Kaichen Zhang, Peiyuan Zhang, Yanwei Li, Ziwei Liu, et~al. 2024{\natexlab{a}}.
\newblock \href {https://arxiv.org/abs/2408.03326} {Llava-onevision: Easy visual task transfer}.
\newblock \emph{ArXiv preprint}, abs/2408.03326.

\bibitem[{Li et~al.(2025{\natexlab{a}})Li, Wu, Zhang, Xia, Mao, Dong, Vuli{\'c}, and Wei}]{li2025imagine}
Chengzu Li, Wenshan Wu, Huanyu Zhang, Yan Xia, Shaoguang Mao, Li~Dong, Ivan Vuli{\'c}, and Furu Wei. 2025{\natexlab{a}}.
\newblock \href {https://arxiv.org/abs/2501.07542} {Imagine while reasoning in space: Multimodal visualization-of-thought}.
\newblock \emph{ArXiv preprint}, abs/2501.07542.

\bibitem[{Li et~al.(2025{\natexlab{b}})Li, Cao, Cao, Li, Tan, Keutzer, Xing, Gonzalez, and Stoica}]{li2025s}
Dacheng Li, Shiyi Cao, Chengkun Cao, Xiuyu Li, Shangyin Tan, Kurt Keutzer, Jiarong Xing, Joseph~E Gonzalez, and Ion Stoica. 2025{\natexlab{b}}.
\newblock \href {https://arxiv.org/abs/2502.14382} {S*: Test time scaling for code generation}.
\newblock \emph{ArXiv preprint}, abs/2502.14382.

\bibitem[{Li et~al.(2025{\natexlab{c}})Li, Cao, Griggs, Liu, Mo, Patil, Zaharia, Gonzalez, and Stoica}]{li2025llmseasilylearnreason}
Dacheng Li, Shiyi Cao, Tyler Griggs, Shu Liu, Xiangxi Mo, Shishir~G. Patil, Matei Zaharia, Joseph~E. Gonzalez, and Ion Stoica. 2025{\natexlab{c}}.
\newblock \href {https://arxiv.org/abs/2502.07374} {Llms can easily learn to reason from demonstrations structure, not content, is what matters!}

\bibitem[{Li et~al.(2025{\natexlab{d}})Li, Zhou, Lu, Tyen, Gui, Aloisi, and He}]{li2025headsbetteronedualmodel}
Jiazheng Li, Yuxiang Zhou, Junru Lu, Gladys Tyen, Lin Gui, Cesare Aloisi, and Yulan He. 2025{\natexlab{d}}.
\newblock \href {https://arxiv.org/abs/2502.19230} {Two heads are better than one: Dual-model verbal reflection at inference-time}.

\bibitem[{Li et~al.(2024{\natexlab{b}})Li, Zhu, Tang, Wen, Zhu, Liu, Li, Cheng, Peng, and Feng}]{li2024improving}
Jinming Li, Yichen Zhu, Zhibin Tang, Junjie Wen, Minjie Zhu, Xiaoyu Liu, Chengmeng Li, Ran Cheng, Yaxin Peng, and Feifei Feng. 2024{\natexlab{b}}.
\newblock \href {https://arxiv.org/abs/2412.20451} {Improving vision-language-action models via chain-of-affordance}.
\newblock \emph{ArXiv preprint}, abs/2412.20451.

\bibitem[{Li et~al.(2024{\natexlab{c}})Li, Zhang, Yu, Fu, and Ye}]{li2024agentsneed}
Junyou Li, Qin Zhang, Yangbin Yu, Qiang Fu, and Deheng Ye. 2024{\natexlab{c}}.
\newblock \href {https://arxiv.org/abs/2402.05120} {More agents is all you need}.

\bibitem[{Li et~al.(2024{\natexlab{d}})Li, Chen, Chen, Zhang, Su, Xing, and Zhang}]{li2024confidencemattersrevisitingintrinsic}
Loka Li, Zhenhao Chen, Guangyi Chen, Yixuan Zhang, Yusheng Su, Eric Xing, and Kun Zhang. 2024{\natexlab{d}}.
\newblock \href {https://arxiv.org/abs/2402.12563} {Confidence matters: Revisiting intrinsic self-correction capabilities of large language models}.

\bibitem[{Li et~al.(2024{\natexlab{e}})Li, Xia, Du, Dai, Tang, Wang, Yu, and Zhang}]{li2024rethinkmctsrefiningerroneousthoughts}
Qingyao Li, Wei Xia, Kounianhua Du, Xinyi Dai, Ruiming Tang, Yasheng Wang, Yong Yu, and Weinan Zhang. 2024{\natexlab{e}}.
\newblock \href {https://arxiv.org/abs/2409.09584} {Rethinkmcts: Refining erroneous thoughts in monte carlo tree search for code generation}.

\bibitem[{Li et~al.(2025{\natexlab{e}})Li, Dong, Jin, Zhang, Zhou, Zhu, Zhang, and Dou}]{li2025searcho1agenticsearchenhancedlarge}
Xiaoxi Li, Guanting Dong, Jiajie Jin, Yuyao Zhang, Yujia Zhou, Yutao Zhu, Peitian Zhang, and Zhicheng Dou. 2025{\natexlab{e}}.
\newblock \href {https://arxiv.org/abs/2501.05366} {Search-o1: Agentic search-enhanced large reasoning models}.

\bibitem[{Li et~al.(2025{\natexlab{f}})Li, Zou, and Liu}]{li2025limrrlscaling}
Xuefeng Li, Haoyang Zou, and Pengfei Liu. 2025{\natexlab{f}}.
\newblock \href {https://arxiv.org/abs/2502.11886} {Limr: Less is more for rl scaling}.

\bibitem[{Li et~al.(2025{\natexlab{g}})Li, Zou, and Liu}]{li2025torlscalingtoolintegratedrl}
Xuefeng Li, Haoyang Zou, and Pengfei Liu. 2025{\natexlab{g}}.
\newblock \href {https://arxiv.org/abs/2503.23383} {Torl: Scaling tool-integrated rl}.

\bibitem[{Li et~al.(2024{\natexlab{f}})Li, Yuan, Feng, Pan, Wang, Sun, Wang, and Li}]{liEscapeSkyhighCost2024}
Yiwei Li, Peiwen Yuan, Shaoxiong Feng, Boyuan Pan, Xinglin Wang, Bin Sun, Heda Wang, and Kan Li. 2024{\natexlab{f}}.
\newblock \href {https://arxiv.org/abs/2401.10480} {Escape sky-high cost: Early-stopping self-consistency for multi-step reasoning}.

\bibitem[{Li et~al.(2025{\natexlab{h}})Li, Yue, Xu, Jiang, Niu, Lin, Ramasubramanian, and Poovendran}]{li2025smallmodelsstrugglelearn}
Yuetai Li, Xiang Yue, Zhangchen Xu, Fengqing Jiang, Luyao Niu, Bill~Yuchen Lin, Bhaskar Ramasubramanian, and Radha Poovendran. 2025{\natexlab{h}}.
\newblock \href {https://arxiv.org/abs/2502.12143} {Small models struggle to learn from strong reasoners}.

\bibitem[{Li et~al.(2022)Li, Choi, Chung, Kushman, Schrittwieser, Leblond, Eccles, Keeling, Gimeno, Lago, Hubert, Choy, {d'Autume}, Babuschkin, Chen, Huang, Welbl, Gowal, Cherepanov, Molloy, Mankowitz, Robson, Kohli, de~Freitas, Kavukcuoglu, and Vinyals}]{liCompetitionLevelCodeGeneration2022}
Yujia Li, David Choi, Junyoung Chung, Nate Kushman, Julian Schrittwieser, R{\'e}mi Leblond, Tom Eccles, James Keeling, Felix Gimeno, Agustin~Dal Lago, Thomas Hubert, Peter Choy, Cyprien de~Masson {d'Autume}, Igor Babuschkin, Xinyun Chen, Po-Sen Huang, Johannes Welbl, Sven Gowal, Alexey Cherepanov, James Molloy, Daniel~J. Mankowitz, Esme~Sutherland Robson, Pushmeet Kohli, Nando de~Freitas, Koray Kavukcuoglu, and Oriol Vinyals. 2022.
\newblock \href {https://doi.org/10.1126/science.abq1158} {Competition-level code generation with alphacode}.
\newblock \emph{Science}, 378(6624):1092--1097.

\bibitem[{Li et~al.(2025{\natexlab{i}})Li, Zhang, Zhang, Zhang, Liu, Yao, Xu, Zheng, Wang, Chen, Zhang, Yin, Dong, Guo, Song, and Liu}]{li202512surveyreasoning}
Zhong-Zhi Li, Duzhen Zhang, Ming-Liang Zhang, Jiaxin Zhang, Zengyan Liu, Yuxuan Yao, Haotian Xu, Junhao Zheng, Pei-Jie Wang, Xiuyi Chen, Yingying Zhang, Fei Yin, Jiahua Dong, Zhijiang Guo, Le~Song, and Cheng-Lin Liu. 2025{\natexlab{i}}.
\newblock \href {http://arxiv.org/abs/2502.17419} {From system 1 to system 2: A survey of reasoning large language models}.

\bibitem[{Li(2025)}]{li2025cold}
Ziniu Li. 2025.
\newblock Can better cold-start strategies improve rl training for llms?
\newblock https://tangible-polo-203.notion.site/Can-Better-Cold-Start-Strategies-Improve-RL-Training-for-LLMs-17aa0742a51680828616c867ed53bc6b.
\newblock Notion Blog.

\bibitem[{Liang et~al.(2023)Liang, He, Jiao, Wang, Wang, Wang, Yang, Shi, and Tu}]{liang2024encouragingdivergentthinkinglarge}
Tian Liang, Zhiwei He, Wenxiang Jiao, Xing Wang, Yan Wang, Rui Wang, Yujiu Yang, Shuming Shi, and Zhaopeng Tu. 2023.
\newblock \href {https://arxiv.org/abs/2305.19118} {Encouraging divergent thinking in large language models through multi-agent debate}.

\bibitem[{Liao et~al.(2025)Liao, Xie, Zhang, Kong, Lu, Yang, and Deng}]{liao2025improved}
Zhenyi Liao, Qingsong Xie, Yanhao Zhang, Zijian Kong, Haonan Lu, Zhenyu Yang, and Zhijie Deng. 2025.
\newblock \href {https://arxiv.org/abs/2504.00883} {Improved visual-spatial reasoning via r1-zero-like training}.
\newblock \emph{ArXiv preprint}, abs/2504.00883.

\bibitem[{Lifshitz et~al.(2025)Lifshitz, McIlraith, and Du}]{lifshitz2025multiagentverificationscalingtesttime}
Shalev Lifshitz, Sheila~A. McIlraith, and Yilun Du. 2025.
\newblock \href {https://arxiv.org/abs/2502.20379} {Multi-agent verification: Scaling test-time compute with multiple verifiers}.

\bibitem[{Lightman et~al.(2023)Lightman, Kosaraju, Burda, Edwards, Baker, Lee, Leike, Schulman, Sutskever, and Cobbe}]{lightman2023letsverifystepstep}
Hunter Lightman, Vineet Kosaraju, Yura Burda, Harri Edwards, Bowen Baker, Teddy Lee, Jan Leike, John Schulman, Ilya Sutskever, and Karl Cobbe. 2023.
\newblock \href {https://arxiv.org/abs/2305.20050} {Let's verify step by step}.

\bibitem[{Lin et~al.(2024)Lin, Trivedi, and Sun}]{lin2024generatingwithconfidence}
Zhen Lin, Shubhendu Trivedi, and Jimeng Sun. 2024.
\newblock \href {https://openreview.net/forum?id=DWkJCSxKU5} {Generating with confidence: Uncertainty quantification for black-box large language models}.
\newblock \emph{Trans. Mach. Learn. Res.}, 2024.

\bibitem[{Liu et~al.(2025{\natexlab{a}})Liu, Wang, Cai, Zhang, Zhan, and Duan}]{liu2025video}
Fangfu Liu, Hanyang Wang, Yimo Cai, Kaiyan Zhang, Xiaohang Zhan, and Yueqi Duan. 2025{\natexlab{a}}.
\newblock \href {https://arxiv.org/abs/2503.18942} {Video-t1: Test-time scaling for video generation}.
\newblock \emph{ArXiv preprint}, abs/2503.18942.

\bibitem[{Liu et~al.(2024{\natexlab{a}})Liu, Li, Li, and Lee}]{llava_15}
Haotian Liu, Chunyuan Li, Yuheng Li, and Yong~Jae Lee. 2024{\natexlab{a}}.
\newblock Improved baselines with visual instruction tuning.
\newblock In \emph{Proceedings of the IEEE/CVF Conference on Computer Vision and Pattern Recognition}, pages 26296--26306.

\bibitem[{Liu et~al.(2023{\natexlab{a}})Liu, Li, Wu, and Lee}]{llava}
Haotian Liu, Chunyuan Li, Qingyang Wu, and Yong~Jae Lee. 2023{\natexlab{a}}.
\newblock \href {http://papers.nips.cc/paper\_files/paper/2023/hash/6dcf277ea32ce3288914faf369fe6de0-Abstract-Conference.html} {Visual instruction tuning}.
\newblock In \emph{Advances in Neural Information Processing Systems 36: Annual Conference on Neural Information Processing Systems 2023, NeurIPS 2023, New Orleans, LA, USA, December 10 - 16, 2023}.

\bibitem[{Liu et~al.(2023{\natexlab{b}})Liu, Cohen, Pasunuru, Choi, Hajishirzi, and Celikyilmaz}]{liuDontThrowAway2024}
Jiacheng Liu, Andrew Cohen, Ramakanth Pasunuru, Yejin Choi, Hannaneh Hajishirzi, and Asli Celikyilmaz. 2023{\natexlab{b}}.
\newblock \href {https://arxiv.org/abs/2309.15028} {Don't throw away your value model! generating more preferable text with value-guided monte-carlo tree search decoding}.

\bibitem[{Liu et~al.(2025{\natexlab{b}})Liu, Zhu, Bai, He, Liao, Que, Wang, Zhang, Zhang, Zhang, Zhang, Chen, Guo, Li, Liu, Shan, Song, Tian, Wu, Zhou, Zhu, Feng, Gao, He, Li, Liu, Meng, Su, Tan, Wang, Yang, Ye, Zheng, Zhou, Huang, Li, and Zhang}]{liu2025comprehensivesurveylongcontext}
Jiaheng Liu, Dawei Zhu, Zhiqi Bai, Yancheng He, Huanxuan Liao, Haoran Que, Zekun Wang, Chenchen Zhang, Ge~Zhang, Jiebin Zhang, Yuanxing Zhang, Zhuo Chen, Hangyu Guo, Shilong Li, Ziqiang Liu, Yong Shan, Yifan Song, Jiayi Tian, Wenhao Wu, Zhejian Zhou, Ruijie Zhu, Junlan Feng, Yang Gao, Shizhu He, Zhoujun Li, Tianyu Liu, Fanyu Meng, Wenbo Su, Yingshui Tan, Zili Wang, Jian Yang, Wei Ye, Bo~Zheng, Wangchunshu Zhou, Wenhao Huang, Sujian Li, and Zhaoxiang Zhang. 2025{\natexlab{b}}.
\newblock \href {https://arxiv.org/abs/2503.17407} {A comprehensive survey on long context language modeling}.

\bibitem[{Liu et~al.(2021)Liu, Yuan, Fu, Jiang, Hayashi, and Neubig}]{liu2021pretrainpromptpredictsystematic}
Pengfei Liu, Weizhe Yuan, Jinlan Fu, Zhengbao Jiang, Hiroaki Hayashi, and Graham Neubig. 2021.
\newblock \href {https://arxiv.org/abs/2107.13586} {Pre-train, prompt, and predict: A systematic survey of prompting methods in natural language processing}.

\bibitem[{Liu et~al.(2025{\natexlab{c}})Liu, Gao, Zhao, Zhang, Li, Qi, Ouyang, and Zhou}]{liu20251bllmsurpass405b}
Runze Liu, Junqi Gao, Jian Zhao, Kaiyan Zhang, Xiu Li, Biqing Qi, Wanli Ouyang, and Bowen Zhou. 2025{\natexlab{c}}.
\newblock \href {https://arxiv.org/abs/2502.06703} {Can 1b llm surpass 405b llm? rethinking compute-optimal test-time scaling}.

\bibitem[{Liu et~al.(2024{\natexlab{b}})Liu, Ye, Xing, and Zou}]{reducing_vlm_hallucination}
Sheng Liu, Haotian Ye, Lei Xing, and James Zou. 2024{\natexlab{b}}.
\newblock \href {https://arxiv.org/abs/2410.15778} {Reducing hallucinations in vision-language models via latent space steering}.
\newblock \emph{ArXiv preprint}, abs/2410.15778.

\bibitem[{Liu et~al.(2024{\natexlab{c}})Liu, Zhu, Liu, Xin, Li, Long, Chen, Yang, Xia, Peng et~al.}]{liu2024fullstack}
Siyao Liu, He~Zhu, Jerry Liu, Shulin Xin, Aoyan Li, Rui Long, Li~Chen, Jack Yang, Jinxiang Xia, ZY~Peng, et~al. 2024{\natexlab{c}}.
\newblock \href {https://arxiv.org/abs/2412.00535} {Fullstack bench: Evaluating llms as full stack coder}.
\newblock \emph{ArXiv preprint}, abs/2412.00535.

\bibitem[{Liu et~al.(2024{\natexlab{d}})Liu, Guo, Hu, Jiayang, Zhang, Qiu, and Zhang}]{liu2024languagemodelslearnskip}
Tengxiao Liu, Qipeng Guo, Xiangkun Hu, Cheng Jiayang, Yue Zhang, Xipeng Qiu, and Zheng Zhang. 2024{\natexlab{d}}.
\newblock \href {https://arxiv.org/abs/2411.01855} {Can language models learn to skip steps?}

\bibitem[{Liu et~al.(2023{\natexlab{c}})Liu, Iter, Xu, Wang, Xu, and Zhu}]{liu2023gevalnlgevaluationusing}
Yang Liu, Dan Iter, Yichong Xu, Shuohang Wang, Ruochen Xu, and Chenguang Zhu. 2023{\natexlab{c}}.
\newblock \href {https://doi.org/10.18653/v1/2023.emnlp-main.153} {{G}-eval: {NLG} evaluation using gpt-4 with better human alignment}.
\newblock In \emph{Proceedings of the 2023 Conference on Empirical Methods in Natural Language Processing}, pages 2511--2522, Singapore. Association for Computational Linguistics.

\bibitem[{Liu et~al.(2023{\natexlab{d}})Liu, Singh, Freeman, Co-Reyes, and Liu}]{liu2023improvinglargelanguagemodel}
Yixin Liu, Avi Singh, C.~Daniel Freeman, John~D. Co-Reyes, and Peter~J. Liu. 2023{\natexlab{d}}.
\newblock \href {https://arxiv.org/abs/2310.10047} {Improving large language model fine-tuning for solving math problems}.

\bibitem[{Liu et~al.(2025{\natexlab{d}})Liu, Chi, Wu, Zhang, Hu, Zhang, Zhang, Wu, Cao, Huang et~al.}]{liu2025spatialcot}
Yuecheng Liu, Dafeng Chi, Shiguang Wu, Zhanguang Zhang, Yaochen Hu, Lingfeng Zhang, Yingxue Zhang, Shuang Wu, Tongtong Cao, Guowei Huang, et~al. 2025{\natexlab{d}}.
\newblock \href {https://arxiv.org/abs/2501.10074} {Spatialcot: Advancing spatial reasoning through coordinate alignment and chain-of-thought for embodied task planning}.
\newblock \emph{ArXiv preprint}, abs/2501.10074.

\bibitem[{Liu et~al.(2025{\natexlab{e}})Liu, Peng, Zhong, Yue, Lu, Yu, and Jia}]{liu2025seg}
Yuqi Liu, Bohao Peng, Zhisheng Zhong, Zihao Yue, Fanbin Lu, Bei Yu, and Jiaya Jia. 2025{\natexlab{e}}.
\newblock \href {https://arxiv.org/abs/2503.06520} {Seg-zero: Reasoning-chain guided segmentation via cognitive reinforcement}.
\newblock \emph{ArXiv preprint}, abs/2503.06520.

\bibitem[{Liu et~al.(2023{\natexlab{e}})Liu, Bahety, and Song}]{liu2023reflect}
Zeyi Liu, Arpit Bahety, and Shuran Song. 2023{\natexlab{e}}.
\newblock Reflect: Summarizing robot experiences for failure explanation and correction.
\newblock In \emph{Conference on Robot Learning}, pages 3468--3484. PMLR.

\bibitem[{Liu et~al.(2024{\natexlab{e}})Liu, Gou, Chen, Hong, Gao, Mi, Zhang, Li, Jiang, Liu et~al.}]{liu2024mixture}
Zhili Liu, Yunhao Gou, Kai Chen, Lanqing Hong, Jiahui Gao, Fei Mi, Yu~Zhang, Zhenguo Li, Xin Jiang, Qun Liu, et~al. 2024{\natexlab{e}}.
\newblock \href {https://arxiv.org/abs/2405.00557} {Mixture of insightful experts (mote): The synergy of thought chains and expert mixtures in self-alignment}.
\newblock \emph{ArXiv preprint}, abs/2405.00557.

\bibitem[{Liu et~al.(2025{\natexlab{f}})Liu, Chen, Li, Pang, Du, and Lin}]{liu2025oatzero}
Zichen Liu, Changyu Chen, Wenjun Li, Tianyu Pang, Chao Du, and Min Lin. 2025{\natexlab{f}}.
\newblock There may not be aha moment in r1-zero-like training — a pilot study.
\newblock \url{https://oatllm.notion.site/oat-zero}.
\newblock Notion Blog.

\bibitem[{Liu et~al.(2025{\natexlab{g}})Liu, Chen, Li, Qi, Pang, Du, Lee, and Lin}]{liu2025understanding}
Zichen Liu, Changyu Chen, Wenjun Li, Penghui Qi, Tianyu Pang, Chao Du, Wee~Sun Lee, and Min Lin. 2025{\natexlab{g}}.
\newblock Understanding r1-zero-like training: A critical perspective.
\newblock \url{https://github.com/sail-sg/understand-r1-zero}.

\bibitem[{Liu et~al.(2025{\natexlab{h}})Liu, Wang, Xu, Ma, Ruan, Li, Liu, and Wu}]{liu2025inferencetimescalinggeneralistreward}
Zijun Liu, Peiyi Wang, Runxin Xu, Shirong Ma, Chong Ruan, Peng Li, Yang Liu, and Yu~Wu. 2025{\natexlab{h}}.
\newblock \href {https://arxiv.org/abs/2504.02495} {Inference-time scaling for generalist reward modeling}.

\bibitem[{Liu et~al.(2025{\natexlab{i}})Liu, Sun, Zang, Dong, Cao, Duan, Lin, and Wang}]{liu2025visual}
Ziyu Liu, Zeyi Sun, Yuhang Zang, Xiaoyi Dong, Yuhang Cao, Haodong Duan, Dahua Lin, and Jiaqi Wang. 2025{\natexlab{i}}.
\newblock \href {https://arxiv.org/abs/2503.01785} {Visual-rft: Visual reinforcement fine-tuning}.
\newblock \emph{ArXiv preprint}, abs/2503.01785.

\bibitem[{Long et~al.(2024)Long, Yen, Luu, Kawaguchi, Kan, and Chen}]{long2024multiexpert}
Do~Xuan Long, Duong~Ngoc Yen, Anh~Tuan Luu, Kenji Kawaguchi, Min{-}Yen Kan, and Nancy~F. Chen. 2024.
\newblock \href {https://aclanthology.org/2024.emnlp-main.1135} {Multi-expert prompting improves reliability, safety and usefulness of large language models}.
\newblock In \emph{Proceedings of the 2024 Conference on Empirical Methods in Natural Language Processing, {EMNLP} 2024, Miami, FL, USA, November 12-16, 2024}, pages 20370--20401. Association for Computational Linguistics.

\bibitem[{Long(2023)}]{longLargeLanguageModel2023}
Jieyi Long. 2023.
\newblock \href {https://arxiv.org/abs/2305.08291} {Large language model guided tree-of-thought}.

\bibitem[{Lu et~al.(2024{\natexlab{a}})Lu, Lu, Lange, Foerster, Clune, and Ha}]{lu2024aiscientist}
Chris Lu, Cong Lu, Robert~Tjarko Lange, Jakob Foerster, Jeff Clune, and David Ha. 2024{\natexlab{a}}.
\newblock \href {https://arxiv.org/abs/2408.06292} {The {AI} {S}cientist: Towards fully automated open-ended scientific discovery}.
\newblock \emph{ArXiv preprint}, abs/2408.06292.

\bibitem[{Lu et~al.(2025)Lu, Tan, Xu, Yao, Qu, Chu, Xu, and Qi}]{lu2025scp116khighqualityproblemsolutiondataset}
Dakuan Lu, Xiaoyu Tan, Rui Xu, Tianchu Yao, Chao Qu, Wei Chu, Yinghui Xu, and Yuan Qi. 2025.
\newblock \href {https://arxiv.org/abs/2501.15587} {Scp-116k: A high-quality problem-solution dataset and a generalized pipeline for automated extraction in the higher education science domain}.

\bibitem[{Lu et~al.(2024{\natexlab{b}})Lu, Dou, Wang, Cao, Dai, Wan, and Guo}]{lu2024autopsvautomatedprocesssupervisedverifier}
Jianqiao Lu, Zhiyang Dou, Hongru Wang, Zeyu Cao, Jianbo Dai, Yingjia Wan, and Zhijiang Guo. 2024{\natexlab{b}}.
\newblock \href {https://arxiv.org/abs/2405.16802} {Autopsv: Automated process-supervised verifier}.

\bibitem[{Luo et~al.(2025{\natexlab{a}})Luo, Shen, He, Wang, Liu, Li, Tan, Cao, and Tao}]{luo2025o1prunerlengthharmonizingfinetuningo1like}
Haotian Luo, Li~Shen, Haiying He, Yibo Wang, Shiwei Liu, Wei Li, Naiqiang Tan, Xiaochun Cao, and Dacheng Tao. 2025{\natexlab{a}}.
\newblock \href {https://arxiv.org/abs/2501.12570} {O1-pruner: Length-harmonizing fine-tuning for o1-like reasoning pruning}.

\bibitem[{Luo et~al.(2024)Luo, Liu, Liu, Phatale, Guo, Lara, Li, Shu, Zhu, Meng, Sun, and Rastogi}]{luo2024improvemathematicalreasoninglanguage}
Liangchen Luo, Yinxiao Liu, Rosanne Liu, Samrat Phatale, Meiqi Guo, Harsh Lara, Yunxuan Li, Lei Shu, Yun Zhu, Lei Meng, Jiao Sun, and Abhinav Rastogi. 2024.
\newblock \href {https://arxiv.org/abs/2406.06592} {Improve mathematical reasoning in language models by automated process supervision}.

\bibitem[{Luo et~al.(2025{\natexlab{b}})Luo, Tan, Huang, Shi, Xin, Cai, Patel, Ariyak, Wu, Zhang, Li, Popa, and Stoica}]{deepcoder2025}
Michael Luo, Sijun Tan, Roy Huang, Xiaoxiang Shi, Rachel Xin, Colin Cai, Ameen Patel, Alpay Ariyak, Qingyang Wu, Ce~Zhang, Li~Erran Li, Raluca~Ada Popa, and Ion Stoica. 2025{\natexlab{b}}.
\newblock Deepcoder: A fully open-source 14b coder at o3-mini level.
\newblock https://pretty-radio-b75.notion.site/DeepCoder-A-Fully-Open-Source-14B-Coder-at-O3-mini-Level-1cf81902c14680b3bee5eb349a512a51.
\newblock Notion Blog.

\bibitem[{Luo et~al.(2025{\natexlab{c}})Luo, Tan, Wong, Shi, Tang, Roongta, Cai, Luo, Zhang, Li, Popa, and Stoica}]{deepscaler2025}
Michael Luo, Sijun Tan, Justin Wong, Xiaoxiang Shi, William~Y. Tang, Manan Roongta, Colin Cai, Jeffrey Luo, Tianjun Zhang, Li~Erran Li, Raluca~Ada Popa, and Ion Stoica. 2025{\natexlab{c}}.
\newblock Deepscaler: Surpassing o1-preview with a 1.5b model by scaling rl.
\newblock https://pretty-radio-b75.notion.site/DeepScaleR-Surpassing-O1-Preview-with-a-1-5B-Model-by-Scaling-RL-19681902c1468005bed8ca303013a4e2.
\newblock Notion Blog.

\bibitem[{Luo et~al.(2025{\natexlab{d}})Luo, Zheng, Wang, Yu, Ni, Lin, Zeng, and Yang}]{luo2025ursa}
Ruilin Luo, Zhuofan Zheng, Yifan Wang, Yiyao Yu, Xinzhe Ni, Zicheng Lin, Jin Zeng, and Yujiu Yang. 2025{\natexlab{d}}.
\newblock \href {https://arxiv.org/abs/2501.04686} {Ursa: Understanding and verifying chain-of-thought reasoning in multimodal mathematics}.
\newblock \emph{ArXiv preprint}, abs/2501.04686.

\bibitem[{Ma et~al.(2025{\natexlab{a}})Ma, Wang, Liu, Liu, Chen, Zhang, Zhou, Du, and Li}]{ma2025s2rteachingllmsselfverify}
Ruotian Ma, Peisong Wang, Cheng Liu, Xingyan Liu, Jiaqi Chen, Bang Zhang, Xin Zhou, Nan Du, and Jia Li. 2025{\natexlab{a}}.
\newblock \href {https://arxiv.org/abs/2502.12853} {S$^2$r: Teaching llms to self-verify and self-correct via reinforcement learning}.

\bibitem[{Ma et~al.(2025{\natexlab{b}})Ma, Wan, Yu, Fang, and Wang}]{ma2025cotvalvelengthcompressiblechainofthoughttuning}
Xinyin Ma, Guangnian Wan, Runpeng Yu, Gongfan Fang, and Xinchao Wang. 2025{\natexlab{b}}.
\newblock \href {https://arxiv.org/abs/2502.09601} {Cot-valve: Length-compressible chain-of-thought tuning}.

\bibitem[{Ma et~al.(2025{\natexlab{c}})Ma, Chern, Shen, Zhong, and Liu}]{MAYE}
Yan Ma, Steffi Chern, Xuyang Shen, Yiran Zhong, and Pengfei Liu. 2025{\natexlab{c}}.
\newblock \href {https://arxiv.org/abs/2504.02587} {Rethinking rl scaling for vision language models: A transparent, from-scratch framework and comprehensive evaluation scheme}.
\newblock \emph{ArXiv preprint}, abs/2504.02587.

\bibitem[{Ma et~al.(2025{\natexlab{d}})Ma, Li, Dong, Jiang, Cao, Chen, Huang, and Li}]{ma2025thinkinglongerlargerenhancing}
Yingwei Ma, Yongbin Li, Yihong Dong, Xue Jiang, Rongyu Cao, Jue Chen, Fei Huang, and Binhua Li. 2025{\natexlab{d}}.
\newblock \href {https://arxiv.org/abs/2503.23803} {Thinking longer, not larger: Enhancing software engineering agents via scaling test-time compute}.

\bibitem[{Ma et~al.(2023)Ma, Liu, Yu, Zhang, Jiang, Wang, and Li}]{ma2023training}
Yingwei Ma, Yue Liu, Yue Yu, Yuanliang Zhang, Yu~Jiang, Changjian Wang, and Shanshan Li. 2023.
\newblock \href {https://arxiv.org/abs/2309.16298} {At which training stage does code data help llms reasoning?}
\newblock \emph{ArXiv preprint}, abs/2309.16298.

\bibitem[{Madaan et~al.(2023)Madaan, Tandon, Gupta, Hallinan, Gao, Wiegreffe, Alon, Dziri, Prabhumoye, Yang, Gupta, Majumder, Hermann, Welleck, Yazdanbakhsh, and Clark}]{madaanSELFREFINEIterativeRefinement}
Aman Madaan, Niket Tandon, Prakhar Gupta, Skyler Hallinan, Luyu Gao, Sarah Wiegreffe, Uri Alon, Nouha Dziri, Shrimai Prabhumoye, Yiming Yang, Shashank Gupta, Bodhisattwa~Prasad Majumder, Katherine Hermann, Sean Welleck, Amir Yazdanbakhsh, and Peter Clark. 2023.
\newblock \href {http://papers.nips.cc/paper\_files/paper/2023/hash/91edff07232fb1b55a505a9e9f6c0ff3-Abstract-Conference.html} {Self-refine: Iterative refinement with self-feedback}.
\newblock In \emph{Advances in Neural Information Processing Systems 36: Annual Conference on Neural Information Processing Systems 2023, NeurIPS 2023, New Orleans, LA, USA, December 10 - 16, 2023}.

\bibitem[{Manakul et~al.(2023)Manakul, Liusie, and Gales}]{manakul2023selfcheckgpt}
Potsawee Manakul, Adian Liusie, and Mark Gales. 2023.
\newblock \href {https://doi.org/10.18653/v1/2023.emnlp-main.557} {{S}elf{C}heck{GPT}: Zero-resource black-box hallucination detection for generative large language models}.
\newblock In \emph{Proceedings of the 2023 Conference on Empirical Methods in Natural Language Processing}, pages 9004--9017, Singapore. Association for Computational Linguistics.

\bibitem[{Meng et~al.(2025{\natexlab{a}})Meng, Du, and Liu}]{r1-multimodal-journey}
Fanqing Meng, Lingxiao Du, and Xiangyan. Liu. 2025{\natexlab{a}}.
\newblock R1-multimodal-journey: A journey to real multimodal r1.
\newblock \url{https://github.com/FanqingM/R1-Multimodal-Journey}.

\bibitem[{Meng et~al.(2025{\natexlab{b}})Meng, Du, Liu, Zhou, Lu, Fu, Shi, Wang, He, Zhang et~al.}]{meng2025mm}
Fanqing Meng, Lingxiao Du, Zongkai Liu, Zhixiang Zhou, Quanfeng Lu, Daocheng Fu, Botian Shi, Wenhai Wang, Junjun He, Kaipeng Zhang, et~al. 2025{\natexlab{b}}.
\newblock \href {https://arxiv.org/abs/2503.07365} {Mm-eureka: Exploring visual aha moment with rule-based large-scale reinforcement learning}.
\newblock \emph{ArXiv preprint}, abs/2503.07365.

\bibitem[{Meng et~al.(2024)Meng, Xia, and Chen}]{meng2024simposimplepreferenceoptimization}
Yu~Meng, Mengzhou Xia, and Danqi Chen. 2024.
\newblock \href {https://arxiv.org/abs/2405.14734} {Simpo: Simple preference optimization with a reference-free reward}.

\bibitem[{Merrill and Sabharwal(2023)}]{merrill2024expressivepowertransformerschain}
William Merrill and Ashish Sabharwal. 2023.
\newblock \href {https://arxiv.org/abs/2310.07923} {The expressive power of transformers with chain of thought}.

\bibitem[{Meta(2023)}]{touvron2023llama2openfoundation}
Meta. 2023.
\newblock \href {https://arxiv.org/abs/2307.09288} {Llama 2: Open foundation and fine-tuned chat models}.

\bibitem[{Meta(2024)}]{grattafiori2024llama3herdmodels}
Meta. 2024.
\newblock \href {https://arxiv.org/abs/2407.21783} {The llama 3 herd of models}.

\bibitem[{Metcalfe and Shimamura(1994)}]{metcalfe1994metacognition}
Janet Metcalfe and Arthur~P Shimamura. 1994.
\newblock \emph{Metacognition: Knowing about knowing}.
\newblock MIT press.

\bibitem[{Micha{\l} et~al.(2024)Micha{\l}, William, Karl, Oier, Chelsea, and Sergey}]{michal2024robotic}
Zawalski Micha{\l}, Chen William, Pertsch Karl, Mees Oier, Finn Chelsea, and Levine Sergey. 2024.
\newblock \href {https://arxiv.org/abs/2407.08693} {Robotic control via embodied chain-of-thought reasoning}.
\newblock \emph{ArXiv preprint}, abs/2407.08693.

\bibitem[{Min et~al.(2023)Min, Krishna, Lyu, Lewis, Yih, Koh, Iyyer, Zettlemoyer, and Hajishirzi}]{min2023factscorefinegrainedatomicevaluation}
Sewon Min, Kalpesh Krishna, Xinxi Lyu, Mike Lewis, Wen-tau Yih, Pang Koh, Mohit Iyyer, Luke Zettlemoyer, and Hannaneh Hajishirzi. 2023.
\newblock \href {https://doi.org/10.18653/v1/2023.emnlp-main.741} {{FA}ct{S}core: Fine-grained atomic evaluation of factual precision in long form text generation}.
\newblock In \emph{Proceedings of the 2023 Conference on Empirical Methods in Natural Language Processing}, pages 12076--12100, Singapore. Association for Computational Linguistics.

\bibitem[{Min et~al.(2024)Min, Chen, Jiang, Chen, Deng, Hu, Tang, Wang, Cheng, Song, Zhao, Liu, Wang, and Wen}]{min2024imitateexploreselfimprovereproduction}
Yingqian Min, Zhipeng Chen, Jinhao Jiang, Jie Chen, Jia Deng, Yiwen Hu, Yiru Tang, Jiapeng Wang, Xiaoxue Cheng, Huatong Song, Wayne~Xin Zhao, Zheng Liu, Zhongyuan Wang, and Ji-Rong Wen. 2024.
\newblock \href {https://arxiv.org/abs/2412.09413} {Imitate, explore, and self-improve: A reproduction report on slow-thinking reasoning systems}.

\bibitem[{Mirzadeh et~al.(2024)Mirzadeh, Alizadeh, Shahrokhi, Tuzel, Bengio, and Farajtabar}]{mirzadeh2024gsmsymbolicunderstandinglimitationsmathematical}
Iman Mirzadeh, Keivan Alizadeh, Hooman Shahrokhi, Oncel Tuzel, Samy Bengio, and Mehrdad Farajtabar. 2024.
\newblock \href {https://arxiv.org/abs/2410.05229} {Gsm-symbolic: Understanding the limitations of mathematical reasoning in large language models}.

\bibitem[{ModelScope(2024)}]{mmmu_reasoning_distill_validation}
ModelScope. 2024.
\newblock {MMMU-Reasoning-Distill-Validation}.
\newblock \url{https://modelscope.cn/datasets/modelscope/MMMU-Reasoning-Distill-Validation}.
\newblock ModelScope.

\bibitem[{Muennighoff et~al.(2025)Muennighoff, Yang, Shi, Li, Fei-Fei, Hajishirzi, Zettlemoyer, Liang, Candès, and Hashimoto}]{muennighoff2025s1simpletesttimescaling}
Niklas Muennighoff, Zitong Yang, Weijia Shi, Xiang~Lisa Li, Li~Fei-Fei, Hannaneh Hajishirzi, Luke Zettlemoyer, Percy Liang, Emmanuel Candès, and Tatsunori Hashimoto. 2025.
\newblock \href {https://arxiv.org/abs/2501.19393} {s1: Simple test-time scaling}.

\bibitem[{Munkhbat et~al.(2025)Munkhbat, Ho, Kim, Yang, Kim, and Yun}]{munkhbat2025selftrainingelicitsconcisereasoning}
Tergel Munkhbat, Namgyu Ho, Seo~Hyun Kim, Yongjin Yang, Yujin Kim, and Se-Young Yun. 2025.
\newblock \href {https://arxiv.org/abs/2502.20122} {Self-training elicits concise reasoning in large language models}.

\bibitem[{Nayab et~al.(2024)Nayab, Rossolini, Simoni, Saracino, Buttazzo, Manes, and Giacomelli}]{nayab2025concisethoughtsimpactoutput}
Sania Nayab, Giulio Rossolini, Marco Simoni, Andrea Saracino, Giorgio Buttazzo, Nicolamaria Manes, and Fabrizio Giacomelli. 2024.
\newblock \href {https://arxiv.org/abs/2407.19825} {Concise thoughts: Impact of output length on llm reasoning and cost}.

\bibitem[{Newell et~al.(1959)Newell, Shaw, and Simon}]{newell1959report}
Allen Newell, John~C Shaw, and Herbert~A Simon. 1959.
\newblock Report on a general problem solving program.
\newblock In \emph{IFIP congress}, page~64. Pittsburgh, PA.

\bibitem[{Newell et~al.(1972)Newell, Simon et~al.}]{newell1972human}
Allen Newell, Herbert~Alexander Simon, et~al. 1972.
\newblock \emph{Human problem solving}.
\newblock Prentice-hall Englewood Cliffs, NJ.

\bibitem[{Nie et~al.(2025)Nie, Zhu, You, Zhang, Ou, Hu, Zhou, Lin, Wen, and Li}]{nie2025largelanguagediffusionmodels}
Shen Nie, Fengqi Zhu, Zebin You, Xiaolu Zhang, Jingyang Ou, Jun Hu, Jun Zhou, Yankai Lin, Ji-Rong Wen, and Chongxuan Li. 2025.
\newblock \href {https://arxiv.org/abs/2502.09992} {Large language diffusion models}.

\bibitem[{Nipkow et~al.(2002)Nipkow, Wenzel, and Paulson}]{nipkow2002isabelle}
Tobias Nipkow, Markus Wenzel, and Lawrence~C Paulson. 2002.
\newblock \emph{Isabelle/HOL: a proof assistant for higher-order logic}.
\newblock Springer.

\bibitem[{Nowak et~al.(2024)Nowak, Svete, Butoi, and Cotterell}]{nowak2025representationalcapacityneurallanguage}
Franz Nowak, Anej Svete, Alexandra Butoi, and Ryan Cotterell. 2024.
\newblock \href {https://arxiv.org/abs/2406.14197} {On the representational capacity of neural language models with chain-of-thought reasoning}.

\bibitem[{N{\'u}{\~n}ez et~al.(2019)N{\'u}{\~n}ez, Allen, Gao, Miller~Rigoli, Relaford-Doyle, and Semenuks}]{nunez2019happened}
Rafael N{\'u}{\~n}ez, Michael Allen, Richard Gao, Carson Miller~Rigoli, Josephine Relaford-Doyle, and Arturs Semenuks. 2019.
\newblock What happened to cognitive science?
\newblock \emph{Nature human behaviour}, 3(8):782--791.

\bibitem[{Nye et~al.(2021)Nye, Andreassen, Gur-Ari, Michalewski, Austin, Bieber, Dohan, Lewkowycz, Bosma, Luan, Sutton, and Odena}]{nye2021workscratchpadsintermediatecomputation}
Maxwell Nye, Anders~Johan Andreassen, Guy Gur-Ari, Henryk Michalewski, Jacob Austin, David Bieber, David Dohan, Aitor Lewkowycz, Maarten Bosma, David Luan, Charles Sutton, and Augustus Odena. 2021.
\newblock \href {https://arxiv.org/abs/2112.00114} {Show your work: Scratchpads for intermediate computation with language models}.

\bibitem[{Olausson et~al.(2023)Olausson, Inala, Wang, Gao, and Solar-Lezama}]{olausson2024selfrepairsilverbulletcode}
Theo~X. Olausson, Jeevana~Priya Inala, Chenglong Wang, Jianfeng Gao, and Armando Solar-Lezama. 2023.
\newblock \href {https://arxiv.org/abs/2306.09896} {Is self-repair a silver bullet for code generation?}

\bibitem[{OpenAI(2023)}]{openai2024gpt4technicalreport}
OpenAI. 2023.
\newblock \href {https://arxiv.org/abs/2303.08774} {Gpt-4 technical report}.

\bibitem[{OpenAI(2024)}]{openai_o1_system_card}
OpenAI. 2024.
\newblock \href {https://openai.com/index/openai-o1-system-card/} {Openai o1 system card}.
\newblock Accessed: 2024-11-07.

\bibitem[{OpenAI(2025{\natexlab{a}})}]{cuaopenai}
OpenAI. 2025{\natexlab{a}}.
\newblock \href {https://openai.com/index/computer-using-agent/} {Computer-using agent}.
\newblock openai.com.

\bibitem[{OpenAI(2025{\natexlab{b}})}]{deepresearch}
OpenAI. 2025{\natexlab{b}}.
\newblock \href {https://openai.com/index/introducing-deep-research/} {Introducing deep research}.
\newblock openai.com.

\bibitem[{{OpenAI}(2025{\natexlab{a}})}]{o3_o4_mini}
{OpenAI}. 2025{\natexlab{a}}.
\newblock \href {https://openai.com/index/introducing-o3-and-o4-mini/} {Introducing openai o3 and o4-mini}.
\newblock Accessed: 2025-04-17.

\bibitem[{{OpenAI}(2025{\natexlab{b}})}]{thinking_with_images}
{OpenAI}. 2025{\natexlab{b}}.
\newblock \href {https://openai.com/index/thinking-with-images/} {Thinking with images}.
\newblock Accessed: 2025-04-17.

\bibitem[{Ouyang et~al.(2022)Ouyang, Wu, Jiang, Almeida, Wainwright, Mishkin, Zhang, Agarwal, Slama, Ray, Schulman, Hilton, Kelton, Miller, Simens, Askell, Welinder, Christiano, Leike, and Lowe}]{ouyang2022traininglanguagemodelsfollow}
Long Ouyang, Jeffrey Wu, Xu~Jiang, Diogo Almeida, Carroll~L. Wainwright, Pamela Mishkin, Chong Zhang, Sandhini Agarwal, Katarina Slama, Alex Ray, John Schulman, Jacob Hilton, Fraser Kelton, Luke Miller, Maddie Simens, Amanda Askell, Peter Welinder, Paul~F. Christiano, Jan Leike, and Ryan Lowe. 2022.
\newblock \href {http://papers.nips.cc/paper\_files/paper/2022/hash/b1efde53be364a73914f58805a001731-Abstract-Conference.html} {Training language models to follow instructions with human feedback}.
\newblock In \emph{Advances in Neural Information Processing Systems 35: Annual Conference on Neural Information Processing Systems 2022, NeurIPS 2022, New Orleans, LA, USA, November 28 - December 9, 2022}.

\bibitem[{Pan et~al.(2024{\natexlab{a}})Pan, Wang, Neubig, Jaitly, Ji, Suhr, and Zhang}]{pan2024training}
Jiayi Pan, Xingyao Wang, Graham Neubig, Navdeep Jaitly, Heng Ji, Alane Suhr, and Yizhe Zhang. 2024{\natexlab{a}}.
\newblock \href {https://arxiv.org/abs/2412.21139} {Training software engineering agents and verifiers with swe-gym}.
\newblock \emph{ArXiv preprint}, abs/2412.21139.

\bibitem[{Pan et~al.(2024{\natexlab{b}})Pan, Zhang, Tomlin, Zhou, Levine, and Suhr}]{pan2024autonomousevaluationrefinementdigital}
Jiayi Pan, Yichi Zhang, Nicholas Tomlin, Yifei Zhou, Sergey Levine, and Alane Suhr. 2024{\natexlab{b}}.
\newblock \href {https://arxiv.org/abs/2404.06474} {Autonomous evaluation and refinement of digital agents}.

\bibitem[{Pan et~al.(2023)Pan, Saxon, Xu, Nathani, Wang, and Wang}]{pan2023automaticallycorrectinglargelanguage}
Liangming Pan, Michael Saxon, Wenda Xu, Deepak Nathani, Xinyi Wang, and William~Yang Wang. 2023.
\newblock \href {https://arxiv.org/abs/2308.03188} {Automatically correcting large language models: Surveying the landscape of diverse self-correction strategies}.

\bibitem[{Pang et~al.(2024{\natexlab{a}})Pang, Ye, Wong, He, Chen, and Wang}]{pang2024anchorbasedlargelanguagemodels}
Jianhui Pang, Fanghua Ye, Derek~Fai Wong, Xin He, Wanshun Chen, and Longyue Wang. 2024{\natexlab{a}}.
\newblock \href {https://arxiv.org/abs/2402.07616} {Anchor-based large language models}.

\bibitem[{Pang et~al.(2024{\natexlab{b}})Pang, Yuan, Cho, He, Sukhbaatar, and Weston}]{pangIterativeReasoningPreference2024a}
Richard~Yuanzhe Pang, Weizhe Yuan, Kyunghyun Cho, He~He, Sainbayar Sukhbaatar, and Jason Weston. 2024{\natexlab{b}}.
\newblock \href {https://arxiv.org/abs/2404.19733} {Iterative reasoning preference optimization}.

\bibitem[{Park et~al.(2024)Park, Goldstein, O'Gara, Chen, and Hendrycks}]{park2024deception}
Peter~S. Park, Simon Goldstein, Aidan O'Gara, Michael Chen, and Dan Hendrycks. 2024.
\newblock \href {https://doi.org/10.1016/J.PATTER.2024.100988} {{AI} deception: {A} survey of examples, risks, and potential solutions}.
\newblock \emph{Patterns}, 5(6):100988.

\bibitem[{Parthasarathy et~al.(2023)Parthasarathy, Kontes, Plinge, and Mutschler}]{dinesh2023cmcts}
Dinesh Parthasarathy, Georgios~D. Kontes, Axel Plinge, and Christopher Mutschler. 2023.
\newblock \href {https://arxiv.org/abs/2305.16209} {{C-MCTS:} safe planning with monte carlo tree search}.
\newblock \emph{ArXiv preprint}, abs/2305.16209.

\bibitem[{Paul et~al.(2024)Paul, Ismayilzada, Peyrard, Borges, Bosselut, West, and Faltings}]{paulREFINERReasoningFeedback2024}
Debjit Paul, Mete Ismayilzada, Maxime Peyrard, Beatriz Borges, Antoine Bosselut, Robert West, and Boi Faltings. 2024.
\newblock \href {https://aclanthology.org/2024.eacl-long.67} {{REFINER}: Reasoning feedback on intermediate representations}.
\newblock In \emph{Proceedings of the 18th Conference of the European Chapter of the Association for Computational Linguistics (Volume 1: Long Papers)}, pages 1100--1126, St. Julian{'}s, Malta. Association for Computational Linguistics.

\bibitem[{Peng et~al.(2025)Peng, Zhang, Zhang, You, Liu, Zhu, Yang, Xu, Geng, and Yang}]{peng2025lmmr1}
Yingzhe Peng, Gongrui Zhang, Miaosen Zhang, Zhiyuan You, Jie Liu, Qipeng Zhu, Kai Yang, Xingzhong Xu, Xin Geng, and Xu~Yang. 2025.
\newblock \href {https://arxiv.org/abs/2503.07536} {Lmm-r1: Empowering 3b lmms with strong reasoning abilities through two-stage rule-based rl}.
\newblock \emph{ArXiv preprint}, abs/2503.07536.

\bibitem[{Phan et~al.(2025)Phan, Gatti, Han, Li, and et~al.}]{phan2025humanitysexam}
Long Phan, Alice Gatti, Ziwen Han, Nathaniel Li, and Josephina~Hu et~al. 2025.
\newblock \href {https://arxiv.org/abs/2501.14249} {Humanity's last exam}.

\bibitem[{Polu and Sutskever(2020)}]{polu2020generativelanguagemodelingautomated}
Stanislas Polu and Ilya Sutskever. 2020.
\newblock \href {https://arxiv.org/abs/2009.03393} {Generative language modeling for automated theorem proving}.

\bibitem[{Putta et~al.(2024)Putta, Mills, Garg, Motwani, Finn, Garg, and Rafailov}]{putta2024agentqadvancedreasoning}
Pranav Putta, Edmund Mills, Naman Garg, Sumeet Motwani, Chelsea Finn, Divyansh Garg, and Rafael Rafailov. 2024.
\newblock \href {https://arxiv.org/abs/2408.07199} {Agent q: Advanced reasoning and learning for autonomous ai agents}.

\bibitem[{Qi et~al.(2024)Qi, Ma, Xu, Zhang, Yang, and Yang}]{qiMutualReasoningMakes2024a}
Zhenting Qi, Mingyuan Ma, Jiahang Xu, Li~Lyna Zhang, Fan Yang, and Mao Yang. 2024.
\newblock \href {https://arxiv.org/abs/2408.06195} {Mutual reasoning makes smaller llms stronger problem-solvers}.

\bibitem[{Qin et~al.(2024)Qin, Li, Zou, Liu, Xia, Huang, Ye, Yuan, Liu, Li, and Liu}]{o1journey}
Yiwei Qin, Xuefeng Li, Haoyang Zou, Yixiu Liu, Shijie Xia, Zhen Huang, Yixin Ye, Weizhe Yuan, Zhengzhong Liu, Yuanzhi Li, and Pengfei Liu. 2024.
\newblock \href {https://arxiv.org/abs/2410.18982} {O1 replication journey: A strategic progress report – part 1}.
\newblock \emph{ArXiv preprint}, abs/2410.18982.

\bibitem[{Qin et~al.(2023)Qin, Yuan, Neubig, and Liu}]{qin2022t5score}
Yiwei Qin, Weizhe Yuan, Graham Neubig, and Pengfei Liu. 2023.
\newblock \href {https://doi.org/10.18653/v1/2023.findings-emnlp.1014} {{T}5{S}core: Discriminative fine-tuning of generative evaluation metrics}.
\newblock In \emph{Findings of the Association for Computational Linguistics: EMNLP 2023}, pages 15185--15202, Singapore. Association for Computational Linguistics.

\bibitem[{Qin et~al.(2025)Qin, Ye, Fang, Wang, Liang, Tian, Zhang, Li, Li, Huang, Zhong, Li, Yang, Miao, Lin, Liu, Jiang, Ma, Li, Xiao, Cai, Li, Zheng, Jin, Li, Zhou, Wang, Chen, Li, Yang, Liu, Lin, Peng, Liu, and Shi}]{qin2025uitarspioneeringautomatedgui}
Yujia Qin, Yining Ye, Junjie Fang, Haoming Wang, Shihao Liang, Shizuo Tian, Junda Zhang, Jiahao Li, Yunxin Li, Shijue Huang, Wanjun Zhong, Kuanye Li, Jiale Yang, Yu~Miao, Woyu Lin, Longxiang Liu, Xu~Jiang, Qianli Ma, Jingyu Li, Xiaojun Xiao, Kai Cai, Chuang Li, Yaowei Zheng, Chaolin Jin, Chen Li, Xiao Zhou, Minchao Wang, Haoli Chen, Zhaojian Li, Haihua Yang, Haifeng Liu, Feng Lin, Tao Peng, Xin Liu, and Guang Shi. 2025.
\newblock \href {https://arxiv.org/abs/2501.12326} {Ui-tars: Pioneering automated gui interaction with native agents}.

\bibitem[{Qiu et~al.(2024)Qiu, Lu, Zeng, Guo, Geng, Wang, Huang, Wu, and Wang}]{qiu2024treebonenhancinginferencetimealignment}
Jiahao Qiu, Yifu Lu, Yifan Zeng, Jiacheng Guo, Jiayi Geng, Huazheng Wang, Kaixuan Huang, Yue Wu, and Mengdi Wang. 2024.
\newblock \href {https://arxiv.org/abs/2410.16033} {Treebon: Enhancing inference-time alignment with speculative tree-search and best-of-n sampling}.

\bibitem[{Qiu et~al.(2023)Qiu, Jiang, Lu, Sclar, Pyatkin, Bhagavatula, Wang, Kim, Choi, Dziri, and Ren}]{qiuPHENOMENALPUZZLINGTESTING2024}
Linlu Qiu, Liwei Jiang, Ximing Lu, Melanie Sclar, Valentina Pyatkin, Chandra Bhagavatula, Bailin Wang, Yoon Kim, Yejin Choi, Nouha Dziri, and Xiang Ren. 2023.
\newblock \href {https://arxiv.org/abs/2310.08559} {Phenomenal yet puzzling: Testing inductive reasoning capabilities of language models with hypothesis refinement}.

\bibitem[{Qu et~al.(2025{\natexlab{a}})Qu, Li, Su, Sun, Yan, Liu, Cui, Liu, Liang, He, Li, Wei, Shao, Lu, Zhang, Hua, Zhou, and Cheng}]{qu2025surveyefficientreasoninglarge}
Xiaoye Qu, Yafu Li, Zhaochen Su, Weigao Sun, Jianhao Yan, Dongrui Liu, Ganqu Cui, Daizong Liu, Shuxian Liang, Junxian He, Peng Li, Wei Wei, Jing Shao, Chaochao Lu, Yue Zhang, Xian-Sheng Hua, Bowen Zhou, and Yu~Cheng. 2025{\natexlab{a}}.
\newblock \href {http://arxiv.org/abs/2503.21614} {A survey of efficient reasoning for large reasoning models: Language, multimodality, and beyond}.

\bibitem[{Qu et~al.(2025{\natexlab{b}})Qu, Yang, Setlur, Tunstall, Beeching, Salakhutdinov, and Kumar}]{qu2025optimizingtesttimecomputemeta}
Yuxiao Qu, Matthew Y.~R. Yang, Amrith Setlur, Lewis Tunstall, Edward~Emanuel Beeching, Ruslan Salakhutdinov, and Aviral Kumar. 2025{\natexlab{b}}.
\newblock \href {https://arxiv.org/abs/2503.07572} {Optimizing test-time compute via meta reinforcement fine-tuning}.

\bibitem[{Qu et~al.(2024)Qu, Zhang, Garg, and Kumar}]{qu2024recursiveintrospectionteachinglanguage}
Yuxiao Qu, Tianjun Zhang, Naman Garg, and Aviral Kumar. 2024.
\newblock \href {https://arxiv.org/abs/2407.18219} {Recursive introspection: Teaching language model agents how to self-improve}.

\bibitem[{Qwen(2024)}]{qvq-72b-preview}
Team Qwen. 2024.
\newblock \href {https://qwenlm.github.io/blog/qvq-72b-preview/} {Qvq: To see the world with wisdom}.

\bibitem[{Rabby et~al.(2024)Rabby, Keya, Zamil, and Auer}]{rabby2024mcnestenhancingmathematical}
Gollam Rabby, Farhana Keya, Parvez Zamil, and Sören Auer. 2024.
\newblock \href {https://arxiv.org/abs/2411.15645} {Mc-nest -- enhancing mathematical reasoning in large language models with a monte carlo nash equilibrium self-refine tree}.

\bibitem[{Rafailov et~al.(2024)Rafailov, Chittepu, Park, Sikchi, Hejna, Knox, Finn, and Niekum}]{rafailov2024scalinglawsrewardmodel}
Rafael Rafailov, Yaswanth Chittepu, Ryan Park, Harshit Sikchi, Joey Hejna, Bradley Knox, Chelsea Finn, and Scott Niekum. 2024.
\newblock \href {https://arxiv.org/abs/2406.02900} {Scaling laws for reward model overoptimization in direct alignment algorithms}.

\bibitem[{Rafailov et~al.(2023)Rafailov, Sharma, Mitchell, Manning, Ermon, and Finn}]{rafailov2024directpreferenceoptimizationlanguage}
Rafael Rafailov, Archit Sharma, Eric Mitchell, Christopher~D. Manning, Stefano Ermon, and Chelsea Finn. 2023.
\newblock \href {http://papers.nips.cc/paper\_files/paper/2023/hash/a85b405ed65c6477a4fe8302b5e06ce7-Abstract-Conference.html} {Direct preference optimization: Your language model is secretly a reward model}.
\newblock In \emph{Advances in Neural Information Processing Systems 36: Annual Conference on Neural Information Processing Systems 2023, NeurIPS 2023, New Orleans, LA, USA, December 10 - 16, 2023}.

\bibitem[{Rasley et~al.(2020)Rasley, Rajbhandari, Ruwase, and He}]{rasley2020deepspeed}
Jeff Rasley, Samyam Rajbhandari, Olatunji Ruwase, and Yuxiong He. 2020.
\newblock \href {https://dl.acm.org/doi/10.1145/3394486.3406703} {Deepspeed: System optimizations enable training deep learning models with over 100 billion parameters}.
\newblock In \emph{{KDD} '20: The 26th {ACM} {SIGKDD} Conference on Knowledge Discovery and Data Mining, Virtual Event, CA, USA, August 23-27, 2020}, pages 3505--3506. {ACM}.

\bibitem[{Rein et~al.(2023)Rein, Hou, Stickland, Petty, Pang, Dirani, Michael, and Bowman}]{rein2023gpqagraduatelevelgoogleproofqa}
David Rein, Betty~Li Hou, Asa~Cooper Stickland, Jackson Petty, Richard~Yuanzhe Pang, Julien Dirani, Julian Michael, and Samuel~R. Bowman. 2023.
\newblock \href {https://arxiv.org/abs/2311.12022} {Gpqa: A graduate-level google-proof q\&a benchmark}.

\bibitem[{Ren et~al.(2023)Ren, Dixit, Bodrova, Singh, Tu, Brown, Xu, Takayama, Xia, Varley et~al.}]{renrobots}
Allen~Z Ren, Anushri Dixit, Alexandra Bodrova, Sumeet Singh, Stephen Tu, Noah Brown, Peng Xu, Leila Takayama, Fei Xia, Jake Varley, et~al. 2023.
\newblock Robots that ask for help: Uncertainty alignment for large language model planners.
\newblock In \emph{7th Annual Conference on Robot Learning}.

\bibitem[{Rosin(2011)}]{rosin2011multi}
Christopher~D Rosin. 2011.
\newblock Multi-armed bandits with episode context.
\newblock \emph{Annals of Mathematics and Artificial Intelligence}, 61(3):203--230.

\bibitem[{Ru et~al.(2024)Ru, Qiu, Hu, Zhang, Shi, Chang, Jiayang, Wang, Sun, Li, Zhang, Wang, Jiang, He, Wang, Liu, Zhang, and Zhang}]{ru2024ragchecker}
Dongyu Ru, Lin Qiu, Xiangkun Hu, Tianhang Zhang, Peng Shi, Shuaichen Chang, Cheng Jiayang, Cunxiang Wang, Shichao Sun, Huanyu Li, Zizhao Zhang, Binjie Wang, Jiarong Jiang, Tong He, Zhiguo Wang, Pengfei Liu, Yue Zhang, and Zheng Zhang. 2024.
\newblock \href {https://openreview.net/forum?id=J9oefdGUuM} {{RAGC}hecker: A fine-grained framework for diagnosing retrieval-augmented generation}.
\newblock In \emph{The Thirty-eight Conference on Neural Information Processing Systems Datasets and Benchmarks Track}.

\bibitem[{Ruan et~al.(2025)Ruan, Band, Maddison, and Hashimoto}]{ruan2025reasoninglearnlatentthoughts}
Yangjun Ruan, Neil Band, Chris~J. Maddison, and Tatsunori Hashimoto. 2025.
\newblock \href {https://arxiv.org/abs/2503.18866} {Reasoning to learn from latent thoughts}.

\bibitem[{Saha et~al.(2024{\natexlab{a}})Saha, Levy, Celikyilmaz, Bansal, Weston, and Li}]{saha2024branchsolvemergeimproveslargelanguage}
Swarnadeep Saha, Omer Levy, Asli Celikyilmaz, Mohit Bansal, Jason Weston, and Xian Li. 2024{\natexlab{a}}.
\newblock \href {https://aclanthology.org/2024.naacl-long.462} {Branch-solve-merge improves large language model evaluation and generation}.
\newblock In \emph{Proceedings of the 2024 Conference of the North American Chapter of the Association for Computational Linguistics: Human Language Technologies (Volume 1: Long Papers)}, pages 8352--8370, Mexico City, Mexico. Association for Computational Linguistics.

\bibitem[{Saha et~al.(2025)Saha, Li, Ghazvininejad, Weston, and Wang}]{saha2025learningplanreason}
Swarnadeep Saha, Xian Li, Marjan Ghazvininejad, Jason Weston, and Tianlu Wang. 2025.
\newblock \href {https://arxiv.org/abs/2501.18099} {Learning to plan \& reason for evaluation with thinking-llm-as-a-judge}.

\bibitem[{Saha et~al.(2024{\natexlab{b}})Saha, Prasad, Chen, Hase, Stengel-Eskin, and Bansal}]{saha2024system1xlearningbalancefast}
Swarnadeep Saha, Archiki Prasad, Justin Chih-Yao Chen, Peter Hase, Elias Stengel-Eskin, and Mohit Bansal. 2024{\natexlab{b}}.
\newblock \href {https://arxiv.org/abs/2407.14414} {System-1.x: Learning to balance fast and slow planning with language models}.

\bibitem[{Sahoo et~al.(2024)Sahoo, Singh, Saha, Jain, Mondal, and Chadha}]{sahoo2025systematicsurveypromptengineering}
Pranab Sahoo, Ayush~Kumar Singh, Sriparna Saha, Vinija Jain, Samrat Mondal, and Aman Chadha. 2024.
\newblock \href {https://arxiv.org/abs/2402.07927} {A systematic survey of prompt engineering in large language models: Techniques and applications}.

\bibitem[{Schulman et~al.(2016)Schulman, Moritz, Levine, Jordan, and Abbeel}]{schulman2018highdimensionalcontinuouscontrolusing}
John Schulman, Philipp Moritz, Sergey Levine, Michael~I. Jordan, and Pieter Abbeel. 2016.
\newblock \href {http://arxiv.org/abs/1506.02438} {High-dimensional continuous control using generalized advantage estimation}.
\newblock In \emph{4th International Conference on Learning Representations, {ICLR} 2016, San Juan, Puerto Rico, May 2-4, 2016, Conference Track Proceedings}.

\bibitem[{Schulman et~al.(2017)Schulman, Wolski, Dhariwal, Radford, and Klimov}]{schulman2017proximalpolicyoptimizationalgorithms}
John Schulman, Filip Wolski, Prafulla Dhariwal, Alec Radford, and Oleg Klimov. 2017.
\newblock \href {https://arxiv.org/abs/1707.06347} {Proximal policy optimization algorithms}.

\bibitem[{Seed(2025)}]{seedthinking2025}
Bytedance Seed. 2025.
\newblock Seed-thinking-v1.5: Advancing superb reasoning models with reinforcement learning.
\newblock \url{https://github.com/ByteDance-Seed/Seed-Thinking-v1.5}.

\bibitem[{Setlur et~al.(2024{\natexlab{a}})Setlur, Garg, Geng, Garg, Smith, and Kumar}]{setlur2024rlincorrectsyntheticdata}
Amrith Setlur, Saurabh Garg, Xinyang Geng, Naman Garg, Virginia Smith, and Aviral Kumar. 2024{\natexlab{a}}.
\newblock \href {https://arxiv.org/abs/2406.14532} {Rl on incorrect synthetic data scales the efficiency of llm math reasoning by eight-fold}.

\bibitem[{Setlur et~al.(2024{\natexlab{b}})Setlur, Nagpal, Fisch, Geng, Eisenstein, Agarwal, Agarwal, Berant, and Kumar}]{setlur2024rewardingprogressscalingautomated}
Amrith Setlur, Chirag Nagpal, Adam Fisch, Xinyang Geng, Jacob Eisenstein, Rishabh Agarwal, Alekh Agarwal, Jonathan Berant, and Aviral Kumar. 2024{\natexlab{b}}.
\newblock \href {https://arxiv.org/abs/2410.08146} {Rewarding progress: Scaling automated process verifiers for llm reasoning}.

\bibitem[{Setlur et~al.(2025)Setlur, Rajaraman, Levine, and Kumar}]{setlur2025scalingtesttimecomputeverification}
Amrith Setlur, Nived Rajaraman, Sergey Levine, and Aviral Kumar. 2025.
\newblock \href {https://arxiv.org/abs/2502.12118} {Scaling test-time compute without verification or rl is suboptimal}.

\bibitem[{Shao et~al.(2024)Shao, Wang, Zhu, Xu, Song, Bi, Zhang, Zhang, Li, Wu, and Guo}]{shao2024deepseekmathpushinglimitsmathematical}
Zhihong Shao, Peiyi Wang, Qihao Zhu, Runxin Xu, Junxiao Song, Xiao Bi, Haowei Zhang, Mingchuan Zhang, Y.~K. Li, Y.~Wu, and Daya Guo. 2024.
\newblock \href {https://arxiv.org/abs/2402.03300} {Deepseekmath: Pushing the limits of mathematical reasoning in open language models}.

\bibitem[{Shen et~al.(2024)Shen, Wang, Delalleau, Zeng, Dong, Egert, Sun, Zhang, Jain, Taghibakhshi, Ausin, Aithal, and Kuchaiev}]{shen2024nemoaligner}
Gerald Shen, Zhilin Wang, Olivier Delalleau, Jiaqi Zeng, Yi~Dong, Daniel Egert, Shengyang Sun, Jimmy Zhang, Sahil Jain, Ali Taghibakhshi, Markel~Sanz Ausin, Ashwath Aithal, and Oleksii Kuchaiev. 2024.
\newblock \href {http://arxiv.org/abs/2405.01481} {Nemo-aligner: Scalable toolkit for efficient model alignment}.

\bibitem[{Shen et~al.(2025{\natexlab{a}})Shen, Zhang, Zhang, Xu, and Zhao}]{shen2025vlmr1}
Haozhan Shen, Zilun Zhang, Qianqian Zhang, Ruochen Xu, and Tiancheng Zhao. 2025{\natexlab{a}}.
\newblock Vlm-r1: A stable and generalizable r1-style large vision-language model.
\newblock \url{https://github.com/om-ai-lab/VLM-R1}.

\bibitem[{Shen et~al.(2025{\natexlab{b}})Shen, Zhang, Huang, Shi, Zhang, Yan, Wang, Wang, and Lian}]{shen2025dastdifficultyadaptiveslowthinkinglarge}
Yi~Shen, Jian Zhang, Jieyun Huang, Shuming Shi, Wenjing Zhang, Jiangze Yan, Ning Wang, Kai Wang, and Shiguo Lian. 2025{\natexlab{b}}.
\newblock \href {https://arxiv.org/abs/2503.04472} {Dast: Difficulty-adaptive slow-thinking for large reasoning models}.

\bibitem[{Shen et~al.(2025{\natexlab{c}})Shen, Yan, Zhang, Hu, Du, and He}]{shen2025codicompressingchainofthoughtcontinuous}
Zhenyi Shen, Hanqi Yan, Linhai Zhang, Zhanghao Hu, Yali Du, and Yulan He. 2025{\natexlab{c}}.
\newblock \href {https://arxiv.org/abs/2502.21074} {Codi: Compressing chain-of-thought into continuous space via self-distillation}.

\bibitem[{Sheng et~al.(2024)Sheng, Zhang, Ye, Wu, Zhang, Zhang, Peng, Lin, and Wu}]{verl}
Guangming Sheng, Chi Zhang, Zilingfeng Ye, Xibin Wu, Wang Zhang, Ru~Zhang, Yanghua Peng, Haibin Lin, and Chuan Wu. 2024.
\newblock \href {https://arxiv.org/abs/2409.19256} {Hybridflow: A flexible and efficient rlhf framework}.
\newblock \emph{ArXiv preprint}, abs/2409.19256.

\bibitem[{Shi et~al.(2022)Shi, Fried, Ghazvininejad, Zettlemoyer, and Wang}]{shi2022naturallanguagecodetranslation}
Freda Shi, Daniel Fried, Marjan Ghazvininejad, Luke Zettlemoyer, and Sida~I. Wang. 2022.
\newblock \href {https://doi.org/10.18653/v1/2022.emnlp-main.231} {Natural language to code translation with execution}.
\newblock In \emph{Proceedings of the 2022 Conference on Empirical Methods in Natural Language Processing}, pages 3533--3546, Abu Dhabi, United Arab Emirates. Association for Computational Linguistics.

\bibitem[{Shinn et~al.(2023)Shinn, Cassano, Gopinath, Narasimhan, and Yao}]{shinnReflexionLanguageAgents2023a}
Noah Shinn, Federico Cassano, Ashwin Gopinath, Karthik Narasimhan, and Shunyu Yao. 2023.
\newblock \href {http://papers.nips.cc/paper\_files/paper/2023/hash/1b44b878bb782e6954cd888628510e90-Abstract-Conference.html} {Reflexion: language agents with verbal reinforcement learning}.
\newblock In \emph{Advances in Neural Information Processing Systems 36: Annual Conference on Neural Information Processing Systems 2023, NeurIPS 2023, New Orleans, LA, USA, December 10 - 16, 2023}.

\bibitem[{Silver et~al.(2016)Silver, Huang, Maddison, Guez, Sifre, Van Den~Driessche, Schrittwieser, Antonoglou, Panneershelvam, Lanctot et~al.}]{MasteringGameGo}
David Silver, Aja Huang, Chris~J Maddison, Arthur Guez, Laurent Sifre, George Van Den~Driessche, Julian Schrittwieser, Ioannis Antonoglou, Veda Panneershelvam, Marc Lanctot, et~al. 2016.
\newblock Mastering the game of go with deep neural networks and tree search.
\newblock \emph{nature}, 529(7587):484--489.

\bibitem[{Singh et~al.(2023)Singh, Co-Reyes, Agarwal, Anand, Patil, Garcia, Liu, Harrison, Lee, Xu, Parisi, Kumar, Alemi, Rizkowsky, Nova, Adlam, Bohnet, Elsayed, Sedghi, Mordatch, Simpson, Gur, Snoek, Pennington, Hron, Kenealy, Swersky, Mahajan, Culp, Xiao, Bileschi, Constant, Novak, Liu, Warkentin, Qian, Bansal, Dyer, Neyshabur, Sohl-Dickstein, and Fiedel}]{singhHumanDataScaling2024}
Avi Singh, John~D. Co-Reyes, Rishabh Agarwal, Ankesh Anand, Piyush Patil, Xavier Garcia, Peter~J. Liu, James Harrison, Jaehoon Lee, Kelvin Xu, Aaron Parisi, Abhishek Kumar, Alex Alemi, Alex Rizkowsky, Azade Nova, Ben Adlam, Bernd Bohnet, Gamaleldin Elsayed, Hanie Sedghi, Igor Mordatch, Isabelle Simpson, Izzeddin Gur, Jasper Snoek, Jeffrey Pennington, Jiri Hron, Kathleen Kenealy, Kevin Swersky, Kshiteej Mahajan, Laura Culp, Lechao Xiao, Maxwell~L. Bileschi, Noah Constant, Roman Novak, Rosanne Liu, Tris Warkentin, Yundi Qian, Yamini Bansal, Ethan Dyer, Behnam Neyshabur, Jascha Sohl-Dickstein, and Noah Fiedel. 2023.
\newblock \href {https://arxiv.org/abs/2312.06585} {Beyond human data: Scaling self-training for problem-solving with language models}.

\bibitem[{Singhi et~al.(2025)Singhi, Bansal, Hosseini, Grover, Chang, Rohrbach, and Rohrbach}]{singhi2025solveverifycomputeoptimalproblem}
Nishad Singhi, Hritik Bansal, Arian Hosseini, Aditya Grover, Kai-Wei Chang, Marcus Rohrbach, and Anna Rohrbach. 2025.
\newblock \href {https://arxiv.org/abs/2504.01005} {When to solve, when to verify: Compute-optimal problem solving and generative verification for llm reasoning}.

\bibitem[{Snell et~al.(2024)Snell, Lee, Xu, and Kumar}]{snell2024scalingllmtesttimecompute}
Charlie Snell, Jaehoon Lee, Kelvin Xu, and Aviral Kumar. 2024.
\newblock \href {https://arxiv.org/abs/2408.03314} {Scaling llm test-time compute optimally can be more effective than scaling model parameters}.

\bibitem[{Song et~al.(2025)Song, Jiang, Min, Chen, Chen, Zhao, Fang, and Wen}]{song2025r1}
Huatong Song, Jinhao Jiang, Yingqian Min, Jie Chen, Zhipeng Chen, Wayne~Xin Zhao, Lei Fang, and Ji-Rong Wen. 2025.
\newblock \href {https://arxiv.org/abs/2503.05592} {R1-searcher: Incentivizing the search capability in llms via reinforcement learning}.
\newblock \emph{ArXiv preprint}, abs/2503.05592.

\bibitem[{Sprague et~al.(2024)Sprague, Yin, Rodriguez, Jiang, Wadhwa, Singhal, Zhao, Ye, Mahowald, and Durrett}]{sprague2024cotcotchainofthoughthelps}
Zayne Sprague, Fangcong Yin, Juan~Diego Rodriguez, Dongwei Jiang, Manya Wadhwa, Prasann Singhal, Xinyu Zhao, Xi~Ye, Kyle Mahowald, and Greg Durrett. 2024.
\newblock \href {https://arxiv.org/abs/2409.12183} {To cot or not to cot? chain-of-thought helps mainly on math and symbolic reasoning}.

\bibitem[{Starace et~al.(2025)Starace, Jaffe, Sherburn, Aung, Chan, Maksin, Dias, Mays, Kinsella, Thompson, Heidecke, Glaese, and Patwardhan}]{starace2025paperbenchevaluatingaisability}
Giulio Starace, Oliver Jaffe, Dane Sherburn, James Aung, Jun~Shern Chan, Leon Maksin, Rachel Dias, Evan Mays, Benjamin Kinsella, Wyatt Thompson, Johannes Heidecke, Amelia Glaese, and Tejal Patwardhan. 2025.
\newblock \href {https://arxiv.org/abs/2504.01848} {Paperbench: Evaluating ai's ability to replicate ai research}.

\bibitem[{Stechly et~al.(2023)Stechly, Marquez, and Kambhampati}]{stechlyGPT4DoesntKnow2023}
Kaya Stechly, Matthew Marquez, and Subbarao Kambhampati. 2023.
\newblock \href {https://arxiv.org/abs/2310.12397} {Gpt-4 doesn't know it's wrong: An analysis of iterative prompting for reasoning problems}.

\bibitem[{Stroebl et~al.(2024)Stroebl, Kapoor, and Narayanan}]{stroeblInferenceScalingFLaws2024}
Benedikt Stroebl, Sayash Kapoor, and Arvind Narayanan. 2024.
\newblock \href {https://arxiv.org/abs/2411.17501} {Inference scaling flaws: The limits of llm resampling with imperfect verifiers}.

\bibitem[{Su et~al.(2024)Su, Sukhbaatar, Rabbat, Tian, and Zheng}]{su2024dualformercontrollablefastslow}
DiJia Su, Sainbayar Sukhbaatar, Michael Rabbat, Yuandong Tian, and Qinqing Zheng. 2024.
\newblock \href {https://arxiv.org/abs/2410.09918} {Dualformer: Controllable fast and slow thinking by learning with randomized reasoning traces}.

\bibitem[{Subramaniam et~al.(2024)Subramaniam, Torralba, and Li}]{subramaniam2024debategpt}
Vighnesh Subramaniam, Antonio Torralba, and Shuang Li. 2024.
\newblock Debategpt: Fine-tuning large language models with multi-agent debate supervision.

\bibitem[{Sui et~al.(2025)Sui, Chuang, Wang, Zhang, Zhang, Yuan, Liu, Wen, Zhong, Chen, and Hu}]{sui2025stopoverthinkingsurveyefficient}
Yang Sui, Yu-Neng Chuang, Guanchu Wang, Jiamu Zhang, Tianyi Zhang, Jiayi Yuan, Hongyi Liu, Andrew Wen, Shaochen Zhong, Hanjie Chen, and Xia Hu. 2025.
\newblock \href {http://arxiv.org/abs/2503.16419} {Stop overthinking: A survey on efficient reasoning for large language models}.

\bibitem[{Sumers et~al.(2023)Sumers, Yao, Narasimhan, and Griffiths}]{sumers2024cognitivearchitectureslanguageagents}
Theodore~R. Sumers, Shunyu Yao, Karthik Narasimhan, and Thomas~L. Griffiths. 2023.
\newblock \href {https://arxiv.org/abs/2309.02427} {Cognitive architectures for language agents}.

\bibitem[{Sun et~al.(2024{\natexlab{a}})Sun, Haider, Zhang, Yang, Qiu, Yin, Wang, Bartlett, and Zanette}]{sun2024fastbestofndecodingspeculative}
Hanshi Sun, Momin Haider, Ruiqi Zhang, Huitao Yang, Jiahao Qiu, Ming Yin, Mengdi Wang, Peter Bartlett, and Andrea Zanette. 2024{\natexlab{a}}.
\newblock \href {https://arxiv.org/abs/2410.20290} {Fast best-of-n decoding via speculative rejection}.

\bibitem[{Sun et~al.(2024{\natexlab{b}})Sun, Yu, Shen, Liu, Yang, Welleck, and Gan}]{sun2024easytohardgeneralizationscalablealignment}
Zhiqing Sun, Longhui Yu, Yikang Shen, Weiyang Liu, Yiming Yang, Sean Welleck, and Chuang Gan. 2024{\natexlab{b}}.
\newblock \href {https://arxiv.org/abs/2403.09472} {Easy-to-hard generalization: Scalable alignment beyond human supervision}.

\bibitem[{Sutton et~al.(1999)Sutton, McAllester, Singh, and Mansour}]{sutton1999policy}
Richard~S Sutton, David McAllester, Satinder Singh, and Yishay Mansour. 1999.
\newblock Policy gradient methods for reinforcement learning with function approximation.
\newblock \emph{Advances in neural information processing systems}, 12.

\bibitem[{Tajwar et~al.(2024)Tajwar, Singh, Sharma, Rafailov, Schneider, Xie, Ermon, Finn, and Kumar}]{tajwar2024preferencefinetuningllmsleverage}
Fahim Tajwar, Anikait Singh, Archit Sharma, Rafael Rafailov, Jeff Schneider, Tengyang Xie, Stefano Ermon, Chelsea Finn, and Aviral Kumar. 2024.
\newblock \href {https://arxiv.org/abs/2404.14367} {Preference fine-tuning of llms should leverage suboptimal, on-policy data}.

\bibitem[{Tang et~al.(2025)Tang, Li, Xiao, Ding, Sun, Wang, Liu, Huang, Liu, Yu, and Lin}]{tang2025realcriticeffectivenessdrivenevaluationlanguage}
Zhengyang Tang, Ziniu Li, Zhenyang Xiao, Tian Ding, Ruoyu Sun, Benyou Wang, Dayiheng Liu, Fei Huang, Tianyu Liu, Bowen Yu, and Junyang Lin. 2025.
\newblock \href {https://arxiv.org/abs/2501.14492} {Realcritic: Towards effectiveness-driven evaluation of language model critiques}.

\bibitem[{Team(2024{\natexlab{a}})}]{AlphaCode2Technical}
AlphaCode2 Team. 2024{\natexlab{a}}.
\newblock \href {https://www.semanticscholar.org/paper/AlphaCode-2-Technical-Report/67be4a0c75c2b2da3e834eb12b8305050a64a525} {Alphacode 2 technical report}.

\bibitem[{Team et~al.(2025)Team, Abeyruwan, Ainslie, Alayrac, Arenas, Armstrong, Balakrishna, Baruch, Bauza, Blokzijl et~al.}]{team2025gemini}
Gemini~Robotics Team, Saminda Abeyruwan, Joshua Ainslie, Jean-Baptiste Alayrac, Montserrat~Gonzalez Arenas, Travis Armstrong, Ashwin Balakrishna, Robert Baruch, Maria Bauza, Michiel Blokzijl, et~al. 2025.
\newblock \href {https://arxiv.org/abs/2503.20020} {Gemini robotics: Bringing ai into the physical world}.
\newblock \emph{ArXiv preprint}, abs/2503.20020.

\bibitem[{Team(2024{\natexlab{b}})}]{openo1}
Open~O1 Team. 2024{\natexlab{b}}.
\newblock \href {https://github.com/Open-Source-O1/Open-O1} {Open o1}.

\bibitem[{Team(2025{\natexlab{a}})}]{openthoughts}
OpenThoughts Team. 2025{\natexlab{a}}.
\newblock {Open Thoughts}.
\newblock https://open-thoughts.ai.

\bibitem[{Team(2024{\natexlab{c}})}]{qwq-32b-preview}
Qwen Team. 2024{\natexlab{c}}.
\newblock \href {https://qwenlm.github.io/blog/qwq-32b-preview/} {Qwq: Reflect deeply on the boundaries of the unknown}.

\bibitem[{Team(2025{\natexlab{b}})}]{qwq32b}
Qwen Team. 2025{\natexlab{b}}.
\newblock \href {https://qwenlm.github.io/blog/qwq-32b/} {Qwq-32b: Embracing the power of reinforcement learning}.

\bibitem[{Thawakar et~al.(2025)Thawakar, Dissanayake, More, Thawkar, Heakl, Ahsan, Li, Zumri, Lahoud, Anwer et~al.}]{llamavo1}
Omkar Thawakar, Dinura Dissanayake, Ketan More, Ritesh Thawkar, Ahmed Heakl, Noor Ahsan, Yuhao Li, Mohammed Zumri, Jean Lahoud, Rao~Muhammad Anwer, et~al. 2025.
\newblock \href {https://arxiv.org/abs/2501.06186} {Llamav-o1: Rethinking step-by-step visual reasoning in llms}.
\newblock \emph{ArXiv preprint}, abs/2501.06186.

\bibitem[{Tian et~al.(2024)Tian, Peng, Song, Jin, Yu, Mi, and Yu}]{tianSelfImprovementLLMsImagination2024b}
Ye~Tian, Baolin Peng, Linfeng Song, Lifeng Jin, Dian Yu, Haitao Mi, and Dong Yu. 2024.
\newblock \href {https://arxiv.org/abs/2404.12253} {Toward self-improvement of llms via imagination, searching, and criticizing}.

\bibitem[{Tie et~al.(2025)Tie, Zhao, Song, Wei, Zhou, Dai, Yin, Yang, Yan, Su, Dai, Xie, Cao, Sun, Zhou, He, Chen, Zhang, Wen, Liu, Gong, Tang, Xiong, Ji, Yu, and Gao}]{tie2025surveyposttraininglargelanguage}
Guiyao Tie, Zeli Zhao, Dingjie Song, Fuyang Wei, Rong Zhou, Yurou Dai, Wen Yin, Zhejian Yang, Jiangyue Yan, Yao Su, Zhenhan Dai, Yifeng Xie, Yihan Cao, Lichao Sun, Pan Zhou, Lifang He, Hechang Chen, Yu~Zhang, Qingsong Wen, Tianming Liu, Neil~Zhenqiang Gong, Jiliang Tang, Caiming Xiong, Heng Ji, Philip~S. Yu, and Jianfeng Gao. 2025.
\newblock \href {http://arxiv.org/abs/2503.06072} {A survey on post-training of large language models}.

\bibitem[{Trinh et~al.(2024)Trinh, Wu, Le, He, and Luong}]{alphageometry}
Trieu~H Trinh, Yuhuai Wu, Quoc~V Le, He~He, and Thang Luong. 2024.
\newblock Solving olympiad geometry without human demonstrations.
\newblock \emph{Nature}.

\bibitem[{Tyen et~al.(2023)Tyen, Mansoor, Cărbune, Chen, and Mak}]{tyen2024llmsreasoningerrorscorrect}
Gladys Tyen, Hassan Mansoor, Victor Cărbune, Peter Chen, and Tony Mak. 2023.
\newblock \href {https://arxiv.org/abs/2311.08516} {Llms cannot find reasoning errors, but can correct them given the error location}.

\bibitem[{Uesato et~al.(2022)Uesato, Kushman, Kumar, Song, Siegel, Wang, Creswell, Irving, and Higgins}]{uesato2022solvingmathwordproblems}
Jonathan Uesato, Nate Kushman, Ramana Kumar, Francis Song, Noah Siegel, Lisa Wang, Antonia Creswell, Geoffrey Irving, and Irina Higgins. 2022.
\newblock \href {https://arxiv.org/abs/2211.14275} {Solving math word problems with process- and outcome-based feedback}.

\bibitem[{Valmeekam et~al.(2023)Valmeekam, Marquez, and Kambhampati}]{valmeekamCanLargeLanguage2023}
Karthik Valmeekam, Matthew Marquez, and Subbarao Kambhampati. 2023.
\newblock \href {https://arxiv.org/abs/2310.08118} {Can large language models really improve by self-critiquing their own plans?}

\bibitem[{Varshney et~al.(2023)Varshney, Yao, Zhang, Chen, and Yu}]{varshney2023stitchtimesavesnine}
Neeraj Varshney, Wenlin Yao, Hongming Zhang, Jianshu Chen, and Dong Yu. 2023.
\newblock \href {https://arxiv.org/abs/2307.03987} {A stitch in time saves nine: Detecting and mitigating hallucinations of llms by validating low-confidence generation}.

\bibitem[{Verma et~al.(2024)Verma, Midigeshi, Sinha, Solin, Natarajan, and Sharma}]{verma2025planragefficienttesttimeplanning}
Prakhar Verma, Sukruta~Prakash Midigeshi, Gaurav Sinha, Arno Solin, Nagarajan Natarajan, and Amit Sharma. 2024.
\newblock \href {https://arxiv.org/abs/2410.20753} {Plan*rag: Efficient test-time planning for retrieval augmented generation}.

\bibitem[{Von~Eckardt(1995)}]{von1995cognitive}
Barbara Von~Eckardt. 1995.
\newblock \emph{What is cognitive science?}
\newblock MIT press.

\bibitem[{Wan et~al.(2024)Wan, Wu, Chen, and Li}]{wanDynamicSelfConsistencyLeveraging2024}
Guangya Wan, Yuqi Wu, Jie Chen, and Sheng Li. 2024.
\newblock \href {https://arxiv.org/abs/2408.17017} {Dynamic self-consistency: Leveraging reasoning paths for efficient llm sampling}.

\bibitem[{Wang et~al.(2024{\natexlab{a}})Wang, Song, Tian, Peng, Yu, Mi, Su, and Yu}]{wang2024litesearchefficacioustreesearch}
Ante Wang, Linfeng Song, Ye~Tian, Baolin Peng, Dian Yu, Haitao Mi, Jinsong Su, and Dong Yu. 2024{\natexlab{a}}.
\newblock \href {https://arxiv.org/abs/2407.00320} {Litesearch: Efficacious tree search for llm}.

\bibitem[{Wang et~al.(2025{\natexlab{a}})Wang, Song, Tian, Yu, Mi, Duan, Tu, Su, and Yu}]{wang2025dontlosttreesstreamlining}
Ante Wang, Linfeng Song, Ye~Tian, Dian Yu, Haitao Mi, Xiangyu Duan, Zhaopeng Tu, Jinsong Su, and Dong Yu. 2025{\natexlab{a}}.
\newblock \href {https://arxiv.org/abs/2502.11183} {Don't get lost in the trees: Streamlining llm reasoning by overcoming tree search exploration pitfalls}.

\bibitem[{Wang et~al.(2024{\natexlab{b}})Wang, Deng, Lyu, Zeng, He, Yan, and An}]{wangImprovingMultistepReasoning2024b}
Chaojie Wang, Yanchen Deng, Zhiyi Lyu, Liang Zeng, Jujie He, Shuicheng Yan, and Bo~An. 2024{\natexlab{b}}.
\newblock \href {https://arxiv.org/abs/2406.14283} {Q*: Improving multi-step reasoning for llms with deliberative planning}.

\bibitem[{Wang et~al.(2024{\natexlab{c}})Wang, Cassano, Wu, Bai, Song, Nath, Han, Hendryx, Yue, and Zhang}]{wangPlanningNaturalLanguage2024}
Evan Wang, Federico Cassano, Catherine Wu, Yunfeng Bai, Will Song, Vaskar Nath, Ziwen Han, Sean Hendryx, Summer Yue, and Hugh Zhang. 2024{\natexlab{c}}.
\newblock \href {https://arxiv.org/abs/2409.03733} {Planning in natural language improves llm search for code generation}.

\bibitem[{Wang et~al.(2024{\natexlab{d}})Wang, Xiong, Xie, Zhao, and Zhang}]{wang2024interpretablepreferencesmultiobjectivereward}
Haoxiang Wang, Wei Xiong, Tengyang Xie, Han Zhao, and Tong Zhang. 2024{\natexlab{d}}.
\newblock \href {https://arxiv.org/abs/2406.12845} {Interpretable preferences via multi-objective reward modeling and mixture-of-experts}.

\bibitem[{Wang et~al.(2025{\natexlab{b}})Wang, Qin, Shen, Wang, Cheng, and Tao}]{wang2025SRG}
Haoyu Wang, Zeyu Qin, Li~Shen, Xueqian Wang, Minhao Cheng, and Dacheng Tao. 2025{\natexlab{b}}.
\newblock \href {https://arxiv.org/abs/2502.04040} {Leveraging reasoning with guidelines to elicit and utilize knowledge for enhancing safety alignment}.
\newblock \emph{ArXiv preprint}, abs/2502.04040.

\bibitem[{Wang et~al.(2025{\natexlab{c}})Wang, Chen, Yang, Huang, Dou, and Wei}]{wang2025chainofretrievalaugmentedgeneration}
Liang Wang, Haonan Chen, Nan Yang, Xiaolong Huang, Zhicheng Dou, and Furu Wei. 2025{\natexlab{c}}.
\newblock \href {https://arxiv.org/abs/2501.14342} {Chain-of-retrieval augmented generation}.

\bibitem[{Wang et~al.(2023{\natexlab{a}})Wang, Li, Shao, Xu, Dai, Li, Chen, Wu, and Sui}]{wangMathShepherdVerifyReinforce2024b}
Peiyi Wang, Lei Li, Zhihong Shao, R.~X. Xu, Damai Dai, Yifei Li, Deli Chen, Y.~Wu, and Zhifang Sui. 2023{\natexlab{a}}.
\newblock \href {https://arxiv.org/abs/2312.08935} {Math-shepherd: Verify and reinforce llms step-by-step without human annotations}.

\bibitem[{Wang et~al.(2023{\natexlab{b}})Wang, Zelikman, Poesia, Pu, Haber, and Goodman}]{wangHYPOTHESISSEARCHINDUCTIVE2024}
Ruocheng Wang, Eric Zelikman, Gabriel Poesia, Yewen Pu, Nick Haber, and Noah~D. Goodman. 2023{\natexlab{b}}.
\newblock \href {https://arxiv.org/abs/2309.05660} {Hypothesis search: Inductive reasoning with language models}.

\bibitem[{Wang et~al.(2024{\natexlab{e}})Wang, Chen, Han, and Bai}]{wangCPLCriticalPlan2024}
Tianlong Wang, Junzhe Chen, Xueting Han, and Jing Bai. 2024{\natexlab{e}}.
\newblock \href {https://arxiv.org/abs/2409.08642} {Cpl: Critical plan step learning boosts llm generalization in reasoning tasks}.

\bibitem[{Wang and Peng(2025)}]{wang-2025-open-r1-video}
Xiaodong Wang and Peixi Peng. 2025.
\newblock Open-r1-video.
\newblock \url{https://github.com/Wang-Xiaodong1899/Open-R1-Video}.

\bibitem[{Wang et~al.(2024{\natexlab{f}})Wang, Feng, Li, Yuan, Zhang, Pan, Wang, Hu, and Li}]{wangMakeEveryPenny2024}
Xinglin Wang, Shaoxiong Feng, Yiwei Li, Peiwen Yuan, Yueqi Zhang, Boyuan Pan, Heda Wang, Yao Hu, and Kan Li. 2024{\natexlab{f}}.
\newblock \href {https://arxiv.org/abs/2408.13457} {Make every penny count: Difficulty-adaptive self-consistency for cost-efficient reasoning}.

\bibitem[{Wang et~al.(2023{\natexlab{c}})Wang, Wei, Schuurmans, Le, Chi, Narang, Chowdhery, and Zhou}]{wangSelfConsistencyImprovesChain2023}
Xuezhi Wang, Jason Wei, Dale Schuurmans, Quoc~V. Le, Ed~H. Chi, Sharan Narang, Aakanksha Chowdhery, and Denny Zhou. 2023{\natexlab{c}}.
\newblock \href {https://openreview.net/pdf?id=1PL1NIMMrw} {Self-consistency improves chain of thought reasoning in language models}.
\newblock In \emph{The Eleventh International Conference on Learning Representations, {ICLR} 2023, Kigali, Rwanda, May 1-5, 2023}. OpenReview.net.

\bibitem[{Wang et~al.(2025{\natexlab{d}})Wang, Liu, He, Zhang, Huang, Zhang, Shu, Tao, She, Yu, Li, Dai, Song, Song, and Jiang}]{mint2025}
Yi~Wang, Mushui Liu, Wanggui He, Longxiang Zhang, Ziwei Huang, Guanghao Zhang, Fangxun Shu, Zhong Tao, Dong She, Zhelun Yu, Haoyuan Li, Weilong Dai, Mingli Song, Jie Song, and Hao Jiang. 2025{\natexlab{d}}.
\newblock \href {https://arxiv.org/abs/2503.01298} {Mint: Multi-modal chain of thought in unified generative models for enhanced image generation}.
\newblock \emph{ArXiv preprint}, abs/2503.01298.

\bibitem[{Wang et~al.(2025{\natexlab{e}})Wang, Zhang, Huang, Yang, Zhang, Huang, and Wang}]{wang2025samplingefficienttesttimescalingselfestimating}
Yiming Wang, Pei Zhang, Siyuan Huang, Baosong Yang, Zhuosheng Zhang, Fei Huang, and Rui Wang. 2025{\natexlab{e}}.
\newblock \href {https://arxiv.org/abs/2503.01422} {Sampling-efficient test-time scaling: Self-estimating the best-of-n sampling in early decoding}.

\bibitem[{Wang et~al.(2024{\natexlab{g}})Wang, Ma, Zhang, Ni, Chandra, Guo, Ren, Arulraj, He, Jiang, Li, Ku, Wang, Zhuang, Fan, Yue, and Chen}]{wang2024mmluprorobustchallengingmultitask}
Yubo Wang, Xueguang Ma, Ge~Zhang, Yuansheng Ni, Abhranil Chandra, Shiguang Guo, Weiming Ren, Aaran Arulraj, Xuan He, Ziyan Jiang, Tianle Li, Max Ku, Kai Wang, Alex Zhuang, Rongqi Fan, Xiang Yue, and Wenhu Chen. 2024{\natexlab{g}}.
\newblock \href {https://arxiv.org/abs/2406.01574} {Mmlu-pro: A more robust and challenging multi-task language understanding benchmark}.

\bibitem[{Wang et~al.(2025{\natexlab{f}})Wang, Ji, Yang, Li, Hu, Li, and Sartoretti}]{wang2025mctsjudgetesttimescalingllmasajudge}
Yutong Wang, Pengliang Ji, Chaoqun Yang, Kaixin Li, Ming Hu, Jiaoyang Li, and Guillaume Sartoretti. 2025{\natexlab{f}}.
\newblock \href {https://arxiv.org/abs/2502.12468} {Mcts-judge: Test-time scaling in llm-as-a-judge for code correctness evaluation}.

\bibitem[{Wang et~al.(2023{\natexlab{d}})Wang, Li, Xia, and Liu}]{wang2024mathpilebilliontokenscalepretrainingcorpus}
Zengzhi Wang, Xuefeng Li, Rui Xia, and Pengfei Liu. 2023{\natexlab{d}}.
\newblock \href {https://arxiv.org/abs/2312.17120} {Mathpile: A billion-token-scale pretraining corpus for math}.

\bibitem[{Wang et~al.(2024{\natexlab{h}})Wang, Wu, Lai, Zhang, and Zhou}]{wang2024seedacceleratingreasoningtree}
Zhenglin Wang, Jialong Wu, Yilong Lai, Congzhi Zhang, and Deyu Zhou. 2024{\natexlab{h}}.
\newblock \href {https://arxiv.org/abs/2406.18200} {Seed: Accelerating reasoning tree construction via scheduled speculative decoding}.

\bibitem[{Wang et~al.(2024{\natexlab{i}})Wang, Li, Wu, Luo, Hou, Yu, and Shang}]{wang2024multistepproblemsolvingverifier}
Zihan Wang, Yunxuan Li, Yuexin Wu, Liangchen Luo, Le~Hou, Hongkun Yu, and Jingbo Shang. 2024{\natexlab{i}}.
\newblock \href {https://arxiv.org/abs/2402.02658} {Multi-step problem solving through a verifier: An empirical analysis on model-induced process supervision}.

\bibitem[{Wei et~al.(2022)Wei, Wang, Schuurmans, Bosma, Ichter, Xia, Chi, Le, and Zhou}]{wei2023chainofthoughtpromptingelicitsreasoning}
Jason Wei, Xuezhi Wang, Dale Schuurmans, Maarten Bosma, Brian Ichter, Fei Xia, Ed~H. Chi, Quoc~V. Le, and Denny Zhou. 2022.
\newblock \href {http://papers.nips.cc/paper\_files/paper/2022/hash/9d5609613524ecf4f15af0f7b31abca4-Abstract-Conference.html} {Chain-of-thought prompting elicits reasoning in large language models}.
\newblock In \emph{Advances in Neural Information Processing Systems 35: Annual Conference on Neural Information Processing Systems 2022, NeurIPS 2022, New Orleans, LA, USA, November 28 - December 9, 2022}.

\bibitem[{Wei et~al.(2025)Wei, Duchenne, Copet, Carbonneaux, Zhang, Fried, Synnaeve, Singh, and Wang}]{wei2025swerladvancingllmreasoning}
Yuxiang Wei, Olivier Duchenne, Jade Copet, Quentin Carbonneaux, Lingming Zhang, Daniel Fried, Gabriel Synnaeve, Rishabh Singh, and Sida~I. Wang. 2025.
\newblock \href {https://arxiv.org/abs/2502.18449} {Swe-rl: Advancing llm reasoning via reinforcement learning on open software evolution}.

\bibitem[{Welleck et~al.(2024)Welleck, Bertsch, Finlayson, Schoelkopf, Xie, Neubig, Kulikov, and Harchaoui}]{welleck2024decodingmetagenerationinferencetimealgorithms}
Sean Welleck, Amanda Bertsch, Matthew Finlayson, Hailey Schoelkopf, Alex Xie, Graham Neubig, Ilia Kulikov, and Zaid Harchaoui. 2024.
\newblock \href {https://arxiv.org/abs/2406.16838} {From decoding to meta-generation: Inference-time algorithms for large language models}.

\bibitem[{Welleck et~al.(2023)Welleck, Lu, West, Brahman, Shen, Khashabi, and Choi}]{welleckGeneratingSequencesLearning2022}
Sean Welleck, Ximing Lu, Peter West, Faeze Brahman, Tianxiao Shen, Daniel Khashabi, and Yejin Choi. 2023.
\newblock \href {https://openreview.net/pdf?id=hH36JeQZDaO} {Generating sequences by learning to self-correct}.
\newblock In \emph{The Eleventh International Conference on Learning Representations, {ICLR} 2023, Kigali, Rwanda, May 1-5, 2023}. OpenReview.net.

\bibitem[{Wu et~al.(2025)Wu, Yao, Liu, Liu, Fu, Han, Li, Zhen, Zhong, and Yuan}]{wu2025unlockingefficientlongtoshortllm}
Han Wu, Yuxuan Yao, Shuqi Liu, Zehua Liu, Xiaojin Fu, Xiongwei Han, Xing Li, Hui-Ling Zhen, Tao Zhong, and Mingxuan Yuan. 2025.
\newblock \href {https://arxiv.org/abs/2503.20641} {Unlocking efficient long-to-short llm reasoning with model merging}.

\bibitem[{Wu et~al.(2024{\natexlab{a}})Wu, Lan, Yuan, Jiao, Weston, and Sukhbaatar}]{wu2024thinkingllmsgeneralinstruction}
Tianhao Wu, Janice Lan, Weizhe Yuan, Jiantao Jiao, Jason Weston, and Sainbayar Sukhbaatar. 2024{\natexlab{a}}.
\newblock \href {https://arxiv.org/abs/2410.10630} {Thinking llms: General instruction following with thought generation}.

\bibitem[{Wu et~al.(2024{\natexlab{b}})Wu, Li, and Liu}]{wu2024progressregressselfimprovementreversal}
Ting Wu, Xuefeng Li, and Pengfei Liu. 2024{\natexlab{b}}.
\newblock \href {https://arxiv.org/abs/2407.05013} {Progress or regress? self-improvement reversal in post-training}.

\bibitem[{Wu et~al.(2024{\natexlab{c}})Wu, Sun, Li, Welleck, and Yang}]{wuInferenceScalingLaws2024}
Yangzhen Wu, Zhiqing Sun, Shanda Li, Sean Welleck, and Yiming Yang. 2024{\natexlab{c}}.
\newblock \href {https://arxiv.org/abs/2408.00724} {Inference scaling laws: An empirical analysis of compute-optimal inference for problem-solving with language models}.

\bibitem[{Wu et~al.(2024{\natexlab{d}})Wu, Qiu, Ross, Aky{\"u}rek, Chen, Wang, Kim, Andreas, and Kim}]{wu2024reasoningrecitingexploringcapabilities}
Zhaofeng Wu, Linlu Qiu, Alexis Ross, Ekin Aky{\"u}rek, Boyuan Chen, Bailin Wang, Najoung Kim, Jacob Andreas, and Yoon Kim. 2024{\natexlab{d}}.
\newblock \href {https://aclanthology.org/2024.naacl-long.102} {Reasoning or reciting? exploring the capabilities and limitations of language models through counterfactual tasks}.
\newblock In \emph{Proceedings of the 2024 Conference of the North American Chapter of the Association for Computational Linguistics: Human Language Technologies (Volume 1: Long Papers)}, pages 1819--1862, Mexico City, Mexico. Association for Computational Linguistics.

\bibitem[{Xi et~al.(2024)Xi, Yang, Huang, Tang, Li, Ding, He, Hong, Do, Zhan, Wang, Zheng, Ji, Shi, Zhai, Weng, Wang, Cai, Gui, Wu, Zhang, Qiu, Huang, and Jiang}]{xiEnhancingLLMReasoning2024}
Zhiheng Xi, Dingwen Yang, Jixuan Huang, Jiafu Tang, Guanyu Li, Yiwen Ding, Wei He, Boyang Hong, Shihan Do, Wenyu Zhan, Xiao Wang, Rui Zheng, Tao Ji, Xiaowei Shi, Yitao Zhai, Rongxiang Weng, Jingang Wang, Xunliang Cai, Tao Gui, Zuxuan Wu, Qi~Zhang, Xipeng Qiu, Xuanjing Huang, and Yu-Gang Jiang. 2024.
\newblock \href {https://arxiv.org/abs/2411.16579} {Enhancing llm reasoning via critique models with test-time and training-time supervision}.

\bibitem[{Xia et~al.(2025)Xia, Li, Leong, Wang, and Li}]{xia2025tokenskipcontrollablechainofthoughtcompression}
Heming Xia, Yongqi Li, Chak~Tou Leong, Wenjie Wang, and Wenjie Li. 2025.
\newblock \href {https://arxiv.org/abs/2502.12067} {Tokenskip: Controllable chain-of-thought compression in llms}.

\bibitem[{Xia et~al.(2024)Xia, Li, Liu, Wu, and Liu}]{xia2025evaluatingmathematicalreasoningaccuracy}
Shijie Xia, Xuefeng Li, Yixin Liu, Tongshuang Wu, and Pengfei Liu. 2024.
\newblock \href {https://arxiv.org/abs/2404.05692} {Evaluating mathematical reasoning beyond accuracy}.

\bibitem[{Xiang et~al.(2025)Xiang, Snell, Gandhi, Albalak, Singh, Blagden, Phung, Rafailov, Lile, Mahan, Castricato, Franken, Haber, and Finn}]{xiang20252reasoningllmslearning}
Violet Xiang, Charlie Snell, Kanishk Gandhi, Alon Albalak, Anikait Singh, Chase Blagden, Duy Phung, Rafael Rafailov, Nathan Lile, Dakota Mahan, Louis Castricato, Jan-Philipp Franken, Nick Haber, and Chelsea Finn. 2025.
\newblock \href {https://arxiv.org/abs/2501.04682} {Towards system 2 reasoning in llms: Learning how to think with meta chain-of-thought}.

\bibitem[{Xie et~al.(2025)Xie, Gao, Ren, Luo, Hong, Dai, Zhou, Qiu, Wu, and Luo}]{xie2025logicrlunleashingllmreasoning}
Tian Xie, Zitian Gao, Qingnan Ren, Haoming Luo, Yuqian Hong, Bryan Dai, Joey Zhou, Kai Qiu, Zhirong Wu, and Chong Luo. 2025.
\newblock \href {https://arxiv.org/abs/2502.14768} {Logic-rl: Unleashing llm reasoning with rule-based reinforcement learning}.

\bibitem[{Xie et~al.(2024{\natexlab{a}})Xie, Zhang, Chen, Li, Zhao, Cao, Hua, Cheng, Shin, Lei, Liu, Xu, Zhou, Savarese, Xiong, Zhong, and Yu}]{xie2024osworldbenchmarkingmultimodalagents}
Tianbao Xie, Danyang Zhang, Jixuan Chen, Xiaochuan Li, Siheng Zhao, Ruisheng Cao, Toh~Jing Hua, Zhoujun Cheng, Dongchan Shin, Fangyu Lei, Yitao Liu, Yiheng Xu, Shuyan Zhou, Silvio Savarese, Caiming Xiong, Victor Zhong, and Tao Yu. 2024{\natexlab{a}}.
\newblock \href {https://arxiv.org/abs/2404.07972} {Osworld: Benchmarking multimodal agents for open-ended tasks in real computer environments}.

\bibitem[{Xie et~al.(2024{\natexlab{b}})Xie, Goyal, Zheng, Kan, Lillicrap, Kawaguchi, and Shieh}]{xieMonteCarloTree2024}
Yuxi Xie, Anirudh Goyal, Wenyue Zheng, Min-Yen Kan, Timothy~P. Lillicrap, Kenji Kawaguchi, and Michael Shieh. 2024{\natexlab{b}}.
\newblock \href {https://arxiv.org/abs/2405.00451} {Monte carlo tree search boosts reasoning via iterative preference learning}.

\bibitem[{Xie et~al.(2023)Xie, Kawaguchi, Zhao, Zhao, Kan, He, and Xie}]{xieSelfEvaluationGuidedBeam}
Yuxi Xie, Kenji Kawaguchi, Yiran Zhao, James~Xu Zhao, Min{-}Yen Kan, Junxian He, and Michael~Qizhe Xie. 2023.
\newblock \href {http://papers.nips.cc/paper\_files/paper/2023/hash/81fde95c4dc79188a69ce5b24d63010b-Abstract-Conference.html} {Self-evaluation guided beam search for reasoning}.
\newblock In \emph{Advances in Neural Information Processing Systems 36: Annual Conference on Neural Information Processing Systems 2023, NeurIPS 2023, New Orleans, LA, USA, December 10 - 16, 2023}.

\bibitem[{Xin et~al.(2024)Xin, Ren, Song, Shao, Zhao, Wang, Liu, Zhang, Lu, Du, Gao, Zhu, Yang, Gou, Wu, Luo, and Ruan}]{xin2024deepseekproverv15harnessingproofassistant}
Huajian Xin, Z.~Z. Ren, Junxiao Song, Zhihong Shao, Wanjia Zhao, Haocheng Wang, Bo~Liu, Liyue Zhang, Xuan Lu, Qiushi Du, Wenjun Gao, Qihao Zhu, Dejian Yang, Zhibin Gou, Z.~F. Wu, Fuli Luo, and Chong Ruan. 2024.
\newblock \href {https://arxiv.org/abs/2408.08152} {Deepseek-prover-v1.5: Harnessing proof assistant feedback for reinforcement learning and monte-carlo tree search}.

\bibitem[{Xin et~al.(2025)Xin, Xi, Yang, Chen, Wu, Xiao, Sun, Zheng, and Shen}]{xin2025bfsproverscalablebestfirsttree}
Ran Xin, Chenguang Xi, Jie Yang, Feng Chen, Hang Wu, Xia Xiao, Yifan Sun, Shen Zheng, and Kai Shen. 2025.
\newblock \href {https://arxiv.org/abs/2502.03438} {Bfs-prover: Scalable best-first tree search for llm-based automatic theorem proving}.

\bibitem[{Xiong et~al.(2025)Xiong, Zhang, Ye, Chen, Jiang, and Zhang}]{xiong2025selfrewardingcorrectionmathematicalreasoning}
Wei Xiong, Hanning Zhang, Chenlu Ye, Lichang Chen, Nan Jiang, and Tong Zhang. 2025.
\newblock \href {https://arxiv.org/abs/2502.19613} {Self-rewarding correction for mathematical reasoning}.

\bibitem[{Xiyao et~al.(2024)Xiyao, Zhengyuan, Linjie, Hongjin, Yuancheng, Lin, Kevin, Furong, and Lijuan}]{visvm}
Wang Xiyao, Yang Zhengyuan, Li~Linjie, Lu~Hongjin, Xu~Yuancheng, Lin Chung-Ching Lin, Lin Kevin, Huang Furong, and Wang Lijuan. 2024.
\newblock \href {https://arxiv.org/abs/2412.03704} {Scaling inference-time search with vision value model for improved visual comprehension}.
\newblock \emph{ArXiv preprint}, abs/2412.03704.

\bibitem[{Xu et~al.(2024{\natexlab{a}})Xu, Lin, Li, and Gao}]{xu2024sramctsselfdrivenreasoningaugmentation}
Bin Xu, Yiguan Lin, Yinghao Li, and Yang Gao. 2024{\natexlab{a}}.
\newblock \href {https://arxiv.org/abs/2411.11053} {Sra-mcts: Self-driven reasoning augmentation with monte carlo tree search for code generation}.

\bibitem[{Xu et~al.(2025{\natexlab{a}})Xu, Hao, Zong, Wang, Zhang, Wang, Lan, Gong, Ouyang, Meng, Shao, Yan, Yang, Song, Ren, Hu, Li, Feng, Gao, and Li}]{xu2025largereasoningmodelssurvey}
Fengli Xu, Qianyue Hao, Zefang Zong, Jingwei Wang, Yunke Zhang, Jingyi Wang, Xiaochong Lan, Jiahui Gong, Tianjian Ouyang, Fanjin Meng, Chenyang Shao, Yuwei Yan, Qinglong Yang, Yiwen Song, Sijian Ren, Xinyuan Hu, Yu~Li, Jie Feng, Chen Gao, and Yong Li. 2025{\natexlab{a}}.
\newblock \href {http://arxiv.org/abs/2501.09686} {Towards large reasoning models: A survey of reinforced reasoning with large language models}.

\bibitem[{Xu et~al.(2024{\natexlab{b}})Xu, Jin, Hao, Song, Sun, and Yuan}]{llavacot}
Guowei Xu, Peng Jin, Li~Hao, Yibing Song, Lichao Sun, and Li~Yuan. 2024{\natexlab{b}}.
\newblock \href {https://arxiv.org/abs/2411.10440} {Llava-cot: Let vision language models reason step-by-step}.
\newblock \emph{ArXiv preprint}, abs/2411.10440.

\bibitem[{Xu et~al.(2025{\natexlab{b}})Xu, Wu, Wang, Li, Zheng, Chen, Hu, Kang, Ji, Zhang, Guo, Yang, Zhang, and Zhang}]{xu2025redstardoesscalinglongcot}
Haotian Xu, Xing Wu, Weinong Wang, Zhongzhi Li, Da~Zheng, Boyuan Chen, Yi~Hu, Shijia Kang, Jiaming Ji, Yingying Zhang, Zhijiang Guo, Yaodong Yang, Muhan Zhang, and Debing Zhang. 2025{\natexlab{b}}.
\newblock \href {https://arxiv.org/abs/2501.11284} {Redstar: Does scaling long-cot data unlock better slow-reasoning systems?}

\bibitem[{Xu et~al.(2025{\natexlab{c}})Xu, Xie, Zhao, and He}]{xu2025chaindraftthinkingfaster}
Silei Xu, Wenhao Xie, Lingxiao Zhao, and Pengcheng He. 2025{\natexlab{c}}.
\newblock \href {https://arxiv.org/abs/2502.18600} {Chain of draft: Thinking faster by writing less}.

\bibitem[{Xu et~al.(2024{\natexlab{c}})Xu, Zhu, Zhao, Pan, Li, and Wang}]{xuPridePrejudiceLLM2024}
Wenda Xu, Guanglei Zhu, Xuandong Zhao, Liangming Pan, Lei Li, and William~Yang Wang. 2024{\natexlab{c}}.
\newblock \href {https://arxiv.org/abs/2402.11436} {Pride and prejudice: Llm amplifies self-bias in self-refinement}.

\bibitem[{Xu et~al.(2025{\natexlab{d}})Xu, Guo, Zeng, and Miao}]{xu2025softcotsoftchainofthoughtefficient}
Yige Xu, Xu~Guo, Zhiwei Zeng, and Chunyan Miao. 2025{\natexlab{d}}.
\newblock \href {https://arxiv.org/abs/2502.12134} {Softcot: Soft chain-of-thought for efficient reasoning with llms}.

\bibitem[{Xu et~al.(2025{\natexlab{e}})Xu, Liu, Yin, Zhou, and Poovendran}]{xu2025kodcode}
Zhangchen Xu, Yang Liu, Yueqin Yin, Mingyuan Zhou, and Radha Poovendran. 2025{\natexlab{e}}.
\newblock \href {https://arxiv.org/abs/2503.02951} {Kodcode: A diverse, challenging, and verifiable synthetic dataset for coding}.
\newblock \emph{ArXiv preprint}, abs/2503.02951.

\bibitem[{Yan et~al.(2025)Yan, Shen, Liu, Jiang, Zhang, Shao, and Zhuang}]{yan2025inftythinkbreakinglengthlimits}
Yuchen Yan, Yongliang Shen, Yang Liu, Jin Jiang, Mengdi Zhang, Jian Shao, and Yueting Zhuang. 2025.
\newblock \href {https://arxiv.org/abs/2503.06692} {Inftythink: Breaking the length limits of long-context reasoning in large language models}.

\bibitem[{Yang et~al.(2024{\natexlab{a}})Yang, Zhang, Hui, Gao, Yu, Li, Liu, Tu, Zhou, Lin, Lu, Xue, Lin, Liu, Ren, and Zhang}]{yangQwen25MathTechnicalReport2024}
An~Yang, Beichen Zhang, Binyuan Hui, Bofei Gao, Bowen Yu, Chengpeng Li, Dayiheng Liu, Jianhong Tu, Jingren Zhou, Junyang Lin, Keming Lu, Mingfeng Xue, Runji Lin, Tianyu Liu, Xingzhang Ren, and Zhenru Zhang. 2024{\natexlab{a}}.
\newblock \href {https://arxiv.org/abs/2409.12122} {Qwen2.5-math technical report: Toward mathematical expert model via self-improvement}.

\bibitem[{Yang et~al.(2024{\natexlab{b}})Yang, Poesia, He, Li, Lauter, Chaudhuri, and Song}]{yang2024formalmathematicalreasoningnew}
Kaiyu Yang, Gabriel Poesia, Jingxuan He, Wenda Li, Kristin Lauter, Swarat Chaudhuri, and Dawn Song. 2024{\natexlab{b}}.
\newblock \href {https://arxiv.org/abs/2412.16075} {Formal mathematical reasoning: A new frontier in ai}.

\bibitem[{Yang et~al.(2024{\natexlab{c}})Yang, Liao, and Fan}]{yang2025markovchainthoughtefficient}
Wen Yang, Minpeng Liao, and Kai Fan. 2024{\natexlab{c}}.
\newblock \href {https://arxiv.org/abs/2410.17635} {Markov chain of thought for efficient mathematical reasoning}.

\bibitem[{Yang et~al.(2025{\natexlab{a}})Yang, Ma, Lin, and Wei}]{yang2025thinkingoptimalscalingtesttimecompute}
Wenkai Yang, Shuming Ma, Yankai Lin, and Furu Wei. 2025{\natexlab{a}}.
\newblock \href {https://arxiv.org/abs/2502.18080} {Towards thinking-optimal scaling of test-time compute for llm reasoning}.

\bibitem[{Yang et~al.(2025{\natexlab{b}})Yang, He, Pan, Jiang, Deng, Yang, Lu, Yin, Rao, Zhu, Zhang, and Chen}]{yang2025r1onevisionadvancinggeneralizedmultimodal}
Yi~Yang, Xiaoxuan He, Hongkun Pan, Xiyan Jiang, Yan Deng, Xingtao Yang, Haoyu Lu, Dacheng Yin, Fengyun Rao, Minfeng Zhu, Bo~Zhang, and Wei Chen. 2025{\natexlab{b}}.
\newblock \href {https://arxiv.org/abs/2503.10615} {R1-onevision: Advancing generalized multimodal reasoning through cross-modal formalization}.

\bibitem[{Yao et~al.(2024{\natexlab{a}})Yao, Huang, Wu, Zhang, Wang, Liu, Wang, Song, Feng, Shen et~al.}]{yao2024mulberry}
Huanjin Yao, Jiaxing Huang, Wenhao Wu, Jingyi Zhang, Yibo Wang, Shunyu Liu, Yingjie Wang, Yuxin Song, Haocheng Feng, Li~Shen, et~al. 2024{\natexlab{a}}.
\newblock \href {https://arxiv.org/abs/2412.18319} {Mulberry: Empowering mllm with o1-like reasoning and reflection via collective monte carlo tree search}.
\newblock \emph{ArXiv preprint}, abs/2412.18319.

\bibitem[{Yao et~al.(2024{\natexlab{b}})Yao, Chen, Zhang, You, Yuan, Wang, and Lin}]{yao2025deftdecodingflashtreeattention}
Jinwei Yao, Kaiqi Chen, Kexun Zhang, Jiaxuan You, Binhang Yuan, Zeke Wang, and Tao Lin. 2024{\natexlab{b}}.
\newblock \href {https://arxiv.org/abs/2404.00242} {Deft: Decoding with flash tree-attention for efficient tree-structured llm inference}.

\bibitem[{Yao et~al.(2023{\natexlab{a}})Yao, Yu, Zhao, Shafran, Griffiths, Cao, and Narasimhan}]{yao2023treethoughtsdeliberateproblem}
Shunyu Yao, Dian Yu, Jeffrey Zhao, Izhak Shafran, Tom Griffiths, Yuan Cao, and Karthik Narasimhan. 2023{\natexlab{a}}.
\newblock \href {http://papers.nips.cc/paper\_files/paper/2023/hash/271db9922b8d1f4dd7aaef84ed5ac703-Abstract-Conference.html} {Tree of thoughts: Deliberate problem solving with large language models}.
\newblock In \emph{Advances in Neural Information Processing Systems 36: Annual Conference on Neural Information Processing Systems 2023, NeurIPS 2023, New Orleans, LA, USA, December 10 - 16, 2023}.

\bibitem[{Yao et~al.(2023{\natexlab{b}})Yao, Zhao, Yu, Du, Shafran, Narasimhan, and Cao}]{yao2023reactsynergizingreasoningacting}
Shunyu Yao, Jeffrey Zhao, Dian Yu, Nan Du, Izhak Shafran, Karthik~R. Narasimhan, and Yuan Cao. 2023{\natexlab{b}}.
\newblock \href {https://openreview.net/pdf?id=WE\_vluYUL-X} {React: Synergizing reasoning and acting in language models}.
\newblock In \emph{The Eleventh International Conference on Learning Representations, {ICLR} 2023, Kigali, Rwanda, May 1-5, 2023}. OpenReview.net.

\bibitem[{Ye et~al.(2024)Ye, Xu, Li, and Allen-Zhu}]{ye2024physicslanguagemodels22}
Tian Ye, Zicheng Xu, Yuanzhi Li, and Zeyuan Allen-Zhu. 2024.
\newblock \href {https://arxiv.org/abs/2408.16293} {Physics of language models: Part 2.2, how to learn from mistakes on grade-school math problems}.

\bibitem[{Ye et~al.(2025)Ye, Huang, Xiao, Chern, Xia, and Liu}]{ye2025limoreasoning}
Yixin Ye, Zhen Huang, Yang Xiao, Ethan Chern, Shijie Xia, and Pengfei Liu. 2025.
\newblock \href {https://arxiv.org/abs/2502.03387} {Limo: Less is more for reasoning}.

\bibitem[{Yeo et~al.(2025)Yeo, Tong, Niu, Neubig, and Yue}]{yeo2025demystifyinglongchainofthoughtreasoning}
Edward Yeo, Yuxuan Tong, Morry Niu, Graham Neubig, and Xiang Yue. 2025.
\newblock \href {https://arxiv.org/abs/2502.03373} {Demystifying long chain-of-thought reasoning in llms}.

\bibitem[{Yu et~al.(2024{\natexlab{a}})Yu, Xu, Weston, and Kulikov}]{yu2024distilling21}
Ping Yu, Jing Xu, Jason Weston, and Ilia Kulikov. 2024{\natexlab{a}}.
\newblock \href {https://arxiv.org/abs/2407.06023} {Distilling system 2 into system 1}.

\bibitem[{Yu et~al.(2025{\natexlab{a}})Yu, Zhang, Zhu, Yuan, Zuo, Yue, Fan, Liu, Liu, Liu, Lin, Lin, Ma, Sheng, Tong, Zhang, Zhang, Zhang, Zhu, Zhu, Chen, Chen, Wang, Yu, Dai, Song, Wei, Zhou, Liu, Ma, Zhang, Yan, Qiao, Wu, and Wang}]{yu2025dapoopensourcellmreinforcement}
Qiying Yu, Zheng Zhang, Ruofei Zhu, Yufeng Yuan, Xiaochen Zuo, Yu~Yue, Tiantian Fan, Gaohong Liu, Lingjun Liu, Xin Liu, Haibin Lin, Zhiqi Lin, Bole Ma, Guangming Sheng, Yuxuan Tong, Chi Zhang, Mofan Zhang, Wang Zhang, Hang Zhu, Jinhua Zhu, Jiaze Chen, Jiangjie Chen, Chengyi Wang, Hongli Yu, Weinan Dai, Yuxuan Song, Xiangpeng Wei, Hao Zhou, Jingjing Liu, Wei-Ying Ma, Ya-Qin Zhang, Lin Yan, Mu~Qiao, Yonghui Wu, and Mingxuan Wang. 2025{\natexlab{a}}.
\newblock \href {https://arxiv.org/abs/2503.14476} {Dapo: An open-source llm reinforcement learning system at scale}.

\bibitem[{Yu et~al.(2024{\natexlab{b}})Yu, Zhang, and Feng}]{yu2024autoragautonomousretrievalaugmentedgeneration}
Tian Yu, Shaolei Zhang, and Yang Feng. 2024{\natexlab{b}}.
\newblock \href {https://arxiv.org/abs/2411.19443} {Auto-rag: Autonomous retrieval-augmented generation for large language models}.

\bibitem[{Yu et~al.(2025{\natexlab{b}})Yu, Wu, Zhao, Cohan, and Zhang}]{yu2025z1efficienttesttimescaling}
Zhaojian Yu, Yinghao Wu, Yilun Zhao, Arman Cohan, and Xiao-Ping Zhang. 2025{\natexlab{b}}.
\newblock \href {https://arxiv.org/abs/2504.00810} {Z1: Efficient test-time scaling with code}.

\bibitem[{Yu et~al.(2025{\natexlab{c}})Yu, Xu, Jin, Sankararaman, He, Zhou, Zeng, Helenowski, Zhu, Wang, Ma, and Fang}]{yu2025thinksmarterharderadaptive}
Zishun Yu, Tengyu Xu, Di~Jin, Karthik~Abinav Sankararaman, Yun He, Wenxuan Zhou, Zhouhao Zeng, Eryk Helenowski, Chen Zhu, Sinong Wang, Hao Ma, and Han Fang. 2025{\natexlab{c}}.
\newblock \href {https://arxiv.org/abs/2501.17974} {Think smarter not harder: Adaptive reasoning with inference aware optimization}.

\bibitem[{Yuan et~al.(2024{\natexlab{a}})Yuan, Li, Chen, Cui, Ding, Zhang, Zhou, Liu, and Peng}]{yuan2024freeprocessrewardsprocess}
Lifan Yuan, Wendi Li, Huayu Chen, Ganqu Cui, Ning Ding, Kaiyan Zhang, Bowen Zhou, Zhiyuan Liu, and Hao Peng. 2024{\natexlab{a}}.
\newblock \href {https://arxiv.org/abs/2412.01981} {Free process rewards without process labels}.

\bibitem[{Yuan et~al.(2024{\natexlab{b}})Yuan, Liu, and Gall{\'e}}]{yuan2024llmcrit}
Weizhe Yuan, Pengfei Liu, and Matthias Gall{\'e}. 2024{\natexlab{b}}.
\newblock \href {https://arxiv.org/abs/2403.01069} {Llmcrit: Teaching large language models to use criteria}.
\newblock \emph{ArXiv preprint}, abs/2403.01069.

\bibitem[{Yuan et~al.(2021)Yuan, Neubig, and Liu}]{yuan2021bartscore}
Weizhe Yuan, Graham Neubig, and Pengfei Liu. 2021.
\newblock \href {https://proceedings.neurips.cc/paper/2021/hash/e4d2b6e6fdeca3e60e0f1a62fee3d9dd-Abstract.html} {Bartscore: Evaluating generated text as text generation}.
\newblock In \emph{Advances in Neural Information Processing Systems 34: Annual Conference on Neural Information Processing Systems 2021, NeurIPS 2021, December 6-14, 2021, virtual}, pages 27263--27277.

\bibitem[{Yuan et~al.(2024{\natexlab{c}})Yuan, Pang, Cho, Li, Sukhbaatar, Xu, and Weston}]{yuanSelfRewardingLanguageModels2024}
Weizhe Yuan, Richard~Yuanzhe Pang, Kyunghyun Cho, Xian Li, Sainbayar Sukhbaatar, Jing Xu, and Jason Weston. 2024{\natexlab{c}}.
\newblock \href {https://arxiv.org/abs/2401.10020} {Self-rewarding language models}.

\bibitem[{Yuan et~al.(2025)Yuan, Yue, Zhu, Fan, and Yan}]{yuan2025whatspposcollapselongcot}
Yufeng Yuan, Yu~Yue, Ruofei Zhu, Tiantian Fan, and Lin Yan. 2025.
\newblock \href {https://arxiv.org/abs/2503.01491} {What's behind ppo's collapse in long-cot? value optimization holds the secret}.

\bibitem[{Yuan et~al.(2023)Yuan, Yuan, Li, Dong, Lu, Tan, Zhou, and Zhou}]{yuanScalingRelationshipLearning2023}
Zheng Yuan, Hongyi Yuan, Chengpeng Li, Guanting Dong, Keming Lu, Chuanqi Tan, Chang Zhou, and Jingren Zhou. 2023.
\newblock \href {https://arxiv.org/abs/2308.01825} {Scaling relationship on learning mathematical reasoning with large language models}.
\newblock \emph{ArXiv preprint}, abs/2308.01825.

\bibitem[{Yue et~al.(2025{\natexlab{a}})Yue, Chen, Lu, Zhao, Wang, Yue, Song, and Huang}]{yue2025doesreinforcementlearningreally}
Yang Yue, Zhiqi Chen, Rui Lu, Andrew Zhao, Zhaokai Wang, Yang Yue, Shiji Song, and Gao Huang. 2025{\natexlab{a}}.
\newblock \href {http://arxiv.org/abs/2504.13837} {Does reinforcement learning really incentivize reasoning capacity in llms beyond the base model?}

\bibitem[{Yue et~al.(2025{\natexlab{b}})Yue, Yuan, Yu, Zuo, Zhu, Xu, Chen, Wang, Fan, Du, Wei, Yu, Liu, Liu, Liu, Lin, Lin, Ma, Zhang, Zhang, Zhang, Zhu, Zhang, Liu, Wang, Wu, and Yan}]{yue2025vapoefficientreliablereinforcement}
Yu~Yue, Yufeng Yuan, Qiying Yu, Xiaochen Zuo, Ruofei Zhu, Wenyuan Xu, Jiaze Chen, Chengyi Wang, TianTian Fan, Zhengyin Du, Xiangpeng Wei, Xiangyu Yu, Gaohong Liu, Juncai Liu, Lingjun Liu, Haibin Lin, Zhiqi Lin, Bole Ma, Chi Zhang, Mofan Zhang, Wang Zhang, Hang Zhu, Ru~Zhang, Xin Liu, Mingxuan Wang, Yonghui Wu, and Lin Yan. 2025{\natexlab{b}}.
\newblock \href {https://arxiv.org/abs/2504.05118} {Vapo: Efficient and reliable reinforcement learning for advanced reasoning tasks}.

\bibitem[{Yue et~al.(2024)Yue, Zhuang, Bai, Hui, Jagerman, Zeng, Qin, Wang, Wang, and Bendersky}]{yue2024inferencescalinglongcontextretrieval}
Zhenrui Yue, Honglei Zhuang, Aijun Bai, Kai Hui, Rolf Jagerman, Hansi Zeng, Zhen Qin, Dong Wang, Xuanhui Wang, and Michael Bendersky. 2024.
\newblock \href {https://arxiv.org/abs/2410.04343} {Inference scaling for long-context retrieval augmented generation}.

\bibitem[{Zeleny(1987)}]{zeleny1987management}
Milan Zeleny. 1987.
\newblock Management support systems: Towards integrated knowledge management.
\newblock \emph{Human systems management}, 7(1):59--70.

\bibitem[{Zelikman et~al.(2024{\natexlab{a}})Zelikman, Harik, Shao, Jayasiri, Haber, and Goodman}]{zelikman2024quietstarlanguagemodelsteach}
Eric Zelikman, Georges Harik, Yijia Shao, Varuna Jayasiri, Nick Haber, and Noah~D. Goodman. 2024{\natexlab{a}}.
\newblock \href {https://arxiv.org/abs/2403.09629} {Quiet-star: Language models can teach themselves to think before speaking}.

\bibitem[{Zelikman et~al.(2024{\natexlab{b}})Zelikman, Lorch, Mackey, and Kalai}]{zelikman2024self}
Eric Zelikman, Eliana Lorch, Lester Mackey, and Adam~Tauman Kalai. 2024{\natexlab{b}}.
\newblock Self-taught optimizer (stop): Recursively self-improving code generation.
\newblock In \emph{First Conference on Language Modeling}.

\bibitem[{Zelikman et~al.(2022)Zelikman, Wu, Mu, and Goodman}]{zelikmanSTaRBootstrappingReasoning2022}
Eric Zelikman, Yuhuai Wu, Jesse Mu, and Noah~D. Goodman. 2022.
\newblock \href {http://papers.nips.cc/paper\_files/paper/2022/hash/639a9a172c044fbb64175b5fad42e9a5-Abstract-Conference.html} {Star: Bootstrapping reasoning with reasoning}.
\newblock In \emph{Advances in Neural Information Processing Systems 35: Annual Conference on Neural Information Processing Systems 2022, NeurIPS 2022, New Orleans, LA, USA, November 28 - December 9, 2022}.

\bibitem[{Zeng et~al.(2025{\natexlab{a}})Zeng, Huang, Liu, He, Liu, Ma, and He}]{zeng2025simplerl}
Weihao Zeng, Yuzhen Huang, Wei Liu, Keqing He, Qian Liu, Zejun Ma, and Junxian He. 2025{\natexlab{a}}.
\newblock 7b model and 8k examples: Emerging reasoning with reinforcement learning is both effective and efficient.
\newblock \url{https://hkust-nlp.notion.site/simplerl-reason}.
\newblock Notion Blog.

\bibitem[{Zeng et~al.(2024)Zeng, Cheng, Yin, Wang, Li, Zhou, Guo, Huang, and Qiu}]{zeng2024scalingsearchlearningroadmap}
Zhiyuan Zeng, Qinyuan Cheng, Zhangyue Yin, Bo~Wang, Shimin Li, Yunhua Zhou, Qipeng Guo, Xuanjing Huang, and Xipeng Qiu. 2024.
\newblock \href {https://arxiv.org/abs/2412.14135} {Scaling of search and learning: A roadmap to reproduce o1 from reinforcement learning perspective}.

\bibitem[{Zeng et~al.(2025{\natexlab{b}})Zeng, Cheng, Yin, Zhou, and Qiu}]{zeng2025revisitingtesttimescalingo1like}
Zhiyuan Zeng, Qinyuan Cheng, Zhangyue Yin, Yunhua Zhou, and Xipeng Qiu. 2025{\natexlab{b}}.
\newblock \href {https://arxiv.org/abs/2502.12215} {Revisiting the test-time scaling of o1-like models: Do they truly possess test-time scaling capabilities?}

\bibitem[{Zhang et~al.(2024{\natexlab{a}})Zhang, Zhoubian, Hu, Yue, Dong, and Tang}]{zhangReSTMCTSLLMSelfTraining2024a}
Dan Zhang, Sining Zhoubian, Ziniu Hu, Yisong Yue, Yuxiao Dong, and Jie Tang. 2024{\natexlab{a}}.
\newblock \href {https://arxiv.org/abs/2406.03816} {Rest-mcts*: Llm self-training via process reward guided tree search}.

\bibitem[{Zhang et~al.(2024{\natexlab{b}})Zhang, Huang, Zhou, Li, and Ouyang}]{zhangAccessingGPT4Level2024a}
Di~Zhang, Xiaoshui Huang, Dongzhan Zhou, Yuqiang Li, and Wanli Ouyang. 2024{\natexlab{b}}.
\newblock \href {https://arxiv.org/abs/2406.07394} {Accessing gpt-4 level mathematical olympiad solutions via monte carlo tree self-refine with llama-3 8b}.

\bibitem[{Zhang et~al.(2024{\natexlab{c}})Zhang, Wu, Lei, Che, Li, Xie, Huang, Zhang, Pavone, Li, Ouyang, and Zhou}]{zhang2024llamaberrypairwiseoptimizationo1like}
Di~Zhang, Jianbo Wu, Jingdi Lei, Tong Che, Jiatong Li, Tong Xie, Xiaoshui Huang, Shufei Zhang, Marco Pavone, Yuqiang Li, Wanli Ouyang, and Dongzhan Zhou. 2024{\natexlab{c}}.
\newblock \href {https://arxiv.org/abs/2410.02884} {Llama-berry: Pairwise optimization for o1-like olympiad-level mathematical reasoning}.

\bibitem[{Zhang et~al.(2024{\natexlab{d}})Zhang, Da, Lee, Robinson, Wu, Song, Zhao, Raja, Zhuang, Slack, Lyu, Hendryx, Kaplan, Lunati, and Yue}]{zhang2024carefulexaminationlargelanguage}
Hugh Zhang, Jeff Da, Dean Lee, Vaughn Robinson, Catherine Wu, Will Song, Tiffany Zhao, Pranav Raja, Charlotte Zhuang, Dylan Slack, Qin Lyu, Sean Hendryx, Russell Kaplan, Michele Lunati, and Summer Yue. 2024{\natexlab{d}}.
\newblock \href {https://arxiv.org/abs/2405.00332} {A careful examination of large language model performance on grade school arithmetic}.

\bibitem[{Zhang et~al.(2024{\natexlab{e}})Zhang, Huang, Jin, and Lu}]{vlm_survey}
Jingyi Zhang, Jiaxing Huang, Sheng Jin, and Shijian Lu. 2024{\natexlab{e}}.
\newblock Vision-language models for vision tasks: A survey.
\newblock \emph{IEEE Transactions on Pattern Analysis and Machine Intelligence}.

\bibitem[{Zhang et~al.(2025{\natexlab{a}})Zhang, Zhu, Sun, Luo, Qiao, Du, Zheng, Chen, and Zhang}]{zhang2025lightthinkerthinkingstepbystepcompression}
Jintian Zhang, Yuqi Zhu, Mengshu Sun, Yujie Luo, Shuofei Qiao, Lun Du, Da~Zheng, Huajun Chen, and Ningyu Zhang. 2025{\natexlab{a}}.
\newblock \href {https://arxiv.org/abs/2502.15589} {Lightthinker: Thinking step-by-step compression}.

\bibitem[{Zhang et~al.(2025{\natexlab{b}})Zhang, Lyu, Sun, Wang, Zhang, Guo, Wang, Muennighoff, King, Liu, and Ma}]{zhang2025whathowwherewell}
Qiyuan Zhang, Fuyuan Lyu, Zexu Sun, Lei Wang, Weixu Zhang, Zhihan Guo, Yufei Wang, Niklas Muennighoff, Irwin King, Xue Liu, and Chen Ma. 2025{\natexlab{b}}.
\newblock \href {http://arxiv.org/abs/2503.24235} {What, how, where, and how well? a survey on test-time scaling in large language models}.

\bibitem[{Zhang et~al.(2025{\natexlab{c}})Zhang, Wang, Jiang, Li, Wu, Wang, Jiang, Shang, Tang, Lyu, and Ma}]{zhang2025crowdcomparativereasoningunlocking}
Qiyuan Zhang, Yufei Wang, Yuxin Jiang, Liangyou Li, Chuhan Wu, Yasheng Wang, Xin Jiang, Lifeng Shang, Ruiming Tang, Fuyuan Lyu, and Chen Ma. 2025{\natexlab{c}}.
\newblock \href {https://arxiv.org/abs/2502.12501} {Crowd comparative reasoning: Unlocking comprehensive evaluations for llm-as-a-judge}.

\bibitem[{Zhang et~al.(2025{\natexlab{d}})Zhang, Liu, Zhang, Liu, Luo, Huang, and Gong}]{zhang2025processbasedselfrewardinglanguagemodels}
Shimao Zhang, Xiao Liu, Xin Zhang, Junxiao Liu, Zheheng Luo, Shujian Huang, and Yeyun Gong. 2025{\natexlab{d}}.
\newblock \href {https://arxiv.org/abs/2503.03746} {Process-based self-rewarding language models}.

\bibitem[{Zhang et~al.(2023)Zhang, Chen, Shen, Ding, Tenenbaum, and Gan}]{zhangPlanningLargeLanguage2023}
Shun Zhang, Zhenfang Chen, Yikang Shen, Mingyu Ding, Joshua~B. Tenenbaum, and Chuang Gan. 2023.
\newblock \href {https://openreview.net/pdf?id=Lr8cOOtYbfL} {Planning with large language models for code generation}.
\newblock In \emph{The Eleventh International Conference on Learning Representations, {ICLR} 2023, Kigali, Rwanda, May 1-5, 2023}. OpenReview.net.

\bibitem[{Zhang et~al.(2025{\natexlab{e}})Zhang, Wang, Liu, Huixin, Jiang, Shen, Hou, Zheng, Zhang, Li et~al.}]{zhang2025embodied}
Wenqi Zhang, Mengna Wang, Gangao Liu, Xu~Huixin, Yiwei Jiang, Yongliang Shen, Guiyang Hou, Zhe Zheng, Hang Zhang, Xin Li, et~al. 2025{\natexlab{e}}.
\newblock \href {https://arxiv.org/abs/2503.21696} {Embodied-reasoner: Synergizing visual search, reasoning, and action for embodied interactive tasks}.
\newblock \emph{ArXiv preprint}, abs/2503.21696.

\bibitem[{Zhang et~al.(2024{\natexlab{f}})Zhang, Du, Pang, Liu, Gao, and Lin}]{zhangChainPreferenceOptimization2024}
Xuan Zhang, Chao Du, Tianyu Pang, Qian Liu, Wei Gao, and Min Lin. 2024{\natexlab{f}}.
\newblock \href {https://arxiv.org/abs/2406.09136} {Chain of preference optimization: Improving chain-of-thought reasoning in llms}.

\bibitem[{Zhang et~al.(2025{\natexlab{f}})Zhang, Zhang, Huang, Xia, Fang, Yang, Duan, Yan, Dong, and Zhu}]{zhang2025stair}
Yichi Zhang, Siyuan Zhang, Yao Huang, Zeyu Xia, Zhengwei Fang, Xiao Yang, Ranjie Duan, Dong Yan, Yinpeng Dong, and Jun Zhu. 2025{\natexlab{f}}.
\newblock \href {https://arxiv.org/abs/2502.02384} {Stair: Improving safety alignment with introspective reasoning}.
\newblock \emph{ArXiv preprint}, abs/2502.02384.

\bibitem[{Zhang et~al.(2024{\natexlab{g}})Zhang, Wu, Yang, Shu, Xiao, Kong, and Sang}]{zhang2024o1codero1replicationcoding}
Yuxiang Zhang, Shangxi Wu, Yuqi Yang, Jiangming Shu, Jinlin Xiao, Chao Kong, and Jitao Sang. 2024{\natexlab{g}}.
\newblock \href {https://arxiv.org/abs/2412.00154} {O1-coder: An o1 replication for coding}.

\bibitem[{Zhang et~al.(2024{\natexlab{h}})Zhang, Zheng, Chen, Jang, Li, Wang, Ding, Fox, and Yao}]{zhang2024grape}
Zijian Zhang, Kaiyuan Zheng, Zhaorun Chen, Joel Jang, Yi~Li, Chaoqi Wang, Mingyu Ding, Dieter Fox, and Huaxiu Yao. 2024{\natexlab{h}}.
\newblock \href {https://arxiv.org/abs/2411.19309} {Grape: Generalizing robot policy via preference alignment}.
\newblock \emph{ArXiv preprint}, abs/2411.19309.

\bibitem[{Zhao et~al.(2025)Zhao, Lu, Kim, Fu, Zhang, Wu, Li, Ma, Han, Finn et~al.}]{zhao2025cot}
Qingqing Zhao, Yao Lu, Moo~Jin Kim, Zipeng Fu, Zhuoyang Zhang, Yecheng Wu, Zhaoshuo Li, Qianli Ma, Song Han, Chelsea Finn, et~al. 2025.
\newblock \href {https://arxiv.org/abs/2503.22020} {Cot-vla: Visual chain-of-thought reasoning for vision-language-action models}.
\newblock \emph{ArXiv preprint}, abs/2503.22020.

\bibitem[{Zhao et~al.(2023{\natexlab{a}})Zhao, Li, Joty, Qin, and Bing}]{zhao2023verifyandeditknowledgeenhancedchainofthoughtframework}
Ruochen Zhao, Xingxuan Li, Shafiq Joty, Chengwei Qin, and Lidong Bing. 2023{\natexlab{a}}.
\newblock \href {https://doi.org/10.18653/v1/2023.acl-long.320} {Verify-and-edit: A knowledge-enhanced chain-of-thought framework}.
\newblock In \emph{Proceedings of the 61st Annual Meeting of the Association for Computational Linguistics (Volume 1: Long Papers)}, pages 5823--5840, Toronto, Canada. Association for Computational Linguistics.

\bibitem[{Zhao et~al.(2023{\natexlab{b}})Zhao, Zhou, Li, Tang, Wang, Hou, Min, Zhang, Zhang, Dong, Du, Yang, Chen, Chen, Jiang, Ren, Li, Tang, Liu, Liu, Nie, and Wen}]{zhao2025surveylargelanguagemodels}
Wayne~Xin Zhao, Kun Zhou, Junyi Li, Tianyi Tang, Xiaolei Wang, Yupeng Hou, Yingqian Min, Beichen Zhang, Junjie Zhang, Zican Dong, Yifan Du, Chen Yang, Yushuo Chen, Zhipeng Chen, Jinhao Jiang, Ruiyang Ren, Yifan Li, Xinyu Tang, Zikang Liu, Peiyu Liu, Jian-Yun Nie, and Ji-Rong Wen. 2023{\natexlab{b}}.
\newblock \href {https://arxiv.org/abs/2303.18223} {A survey of large language models}.

\bibitem[{Zhao et~al.(2023{\natexlab{c}})Zhao, Gu, Varma, Luo, Huang, Xu, Wright, Shojanazeri, Ott, Shleifer, Desmaison, Balioglu, Damania, Nguyen, Chauhan, Hao, Mathews, and Li}]{zhao2023pytorchfsdpexperiencesscaling}
Yanli Zhao, Andrew Gu, Rohan Varma, Liang Luo, Chien-Chin Huang, Min Xu, Less Wright, Hamid Shojanazeri, Myle Ott, Sam Shleifer, Alban Desmaison, Can Balioglu, Pritam Damania, Bernard Nguyen, Geeta Chauhan, Yuchen Hao, Ajit Mathews, and Shen Li. 2023{\natexlab{c}}.
\newblock \href {https://arxiv.org/abs/2304.11277} {Pytorch fsdp: Experiences on scaling fully sharded data parallel}.

\bibitem[{Zhao et~al.(2024)Zhao, Yin, Zeng, Wang, Shi, Lyu, Wang, Luo, and Zhang}]{zhao2024marcoo1openreasoningmodels}
Yu~Zhao, Huifeng Yin, Bo~Zeng, Hao Wang, Tianqi Shi, Chenyang Lyu, Longyue Wang, Weihua Luo, and Kaifu Zhang. 2024.
\newblock \href {https://arxiv.org/abs/2411.14405} {Marco-o1: Towards open reasoning models for open-ended solutions}.

\bibitem[{Zheng et~al.(2023{\natexlab{a}})Zheng, Chiang, Sheng, Zhuang, Wu, Zhuang, Lin, Li, Li, Xing, Zhang, Gonzalez, and Stoica}]{zheng2023judging}
Lianmin Zheng, Wei{-}Lin Chiang, Ying Sheng, Siyuan Zhuang, Zhanghao Wu, Yonghao Zhuang, Zi~Lin, Zhuohan Li, Dacheng Li, Eric~P. Xing, Hao Zhang, Joseph~E. Gonzalez, and Ion Stoica. 2023{\natexlab{a}}.
\newblock \href {http://papers.nips.cc/paper\_files/paper/2023/hash/91f18a1287b398d378ef22505bf41832-Abstract-Datasets\_and\_Benchmarks.html} {Judging llm-as-a-judge with mt-bench and chatbot arena}.
\newblock In \emph{Advances in Neural Information Processing Systems 36: Annual Conference on Neural Information Processing Systems 2023, NeurIPS 2023, New Orleans, LA, USA, December 10 - 16, 2023}.

\bibitem[{Zheng et~al.(2023{\natexlab{b}})Zheng, Yin, Xie, Sun, Huang, Yu, Cao, Kozyrakis, Stoica, Gonzalez, Barrett, and Sheng}]{zheng2024sglangefficientexecutionstructured}
Lianmin Zheng, Liangsheng Yin, Zhiqiang Xie, Chuyue Sun, Jeff Huang, Cody~Hao Yu, Shiyi Cao, Christos Kozyrakis, Ion Stoica, Joseph~E. Gonzalez, Clark Barrett, and Ying Sheng. 2023{\natexlab{b}}.
\newblock \href {https://arxiv.org/abs/2312.07104} {Sglang: Efficient execution of structured language model programs}.

\bibitem[{Zheng et~al.(2025)Zheng, Fu, Hu, Cai, Ye, Lu, and Liu}]{zheng2025deepresearcherscalingdeepresearch}
Yuxiang Zheng, Dayuan Fu, Xiangkun Hu, Xiaojie Cai, Lyumanshan Ye, Pengrui Lu, and Pengfei Liu. 2025.
\newblock \href {https://arxiv.org/abs/2504.03160} {Deepresearcher: Scaling deep research via reinforcement learning in real-world environments}.

\bibitem[{Zhou et~al.(2023)Zhou, Yan, {Shlapentokh-Rothman}, Wang, and Wang}]{zhouLanguageAgentTree2024}
Andy Zhou, Kai Yan, Michal {Shlapentokh-Rothman}, Haohan Wang, and Yu-Xiong Wang. 2023.
\newblock \href {https://arxiv.org/abs/2310.04406} {Language agent tree search unifies reasoning acting and planning in language models}.

\bibitem[{Zhou et~al.(2024{\natexlab{a}})Zhou, Yu, Babu, Tirumala, Yasunaga, Shamis, Kahn, Ma, Zettlemoyer, and Levy}]{zhou2024transfusion}
Chunting Zhou, Lili Yu, Arun Babu, Kushal Tirumala, Michihiro Yasunaga, Leonid Shamis, Jacob Kahn, Xuezhe Ma, Luke Zettlemoyer, and Omer Levy. 2024{\natexlab{a}}.
\newblock \href {https://arxiv.org/abs/2408.11039} {Transfusion: Predict the next token and diffuse images with one multi-modal model}.
\newblock \emph{ArXiv preprint}, abs/2408.11039.

\bibitem[{Zhou et~al.(2024{\natexlab{b}})Zhou, Wang, Liu, Li, and Liu}]{zhou2025programmingexampleliftingpretraining}
Fan Zhou, Zengzhi Wang, Qian Liu, Junlong Li, and Pengfei Liu. 2024{\natexlab{b}}.
\newblock \href {https://arxiv.org/abs/2409.17115} {Programming every example: Lifting pre-training data quality like experts at scale}.

\bibitem[{Zhou et~al.(2025{\natexlab{a}})Zhou, Li, Wang, Cheng, Zhou, and Hsieh}]{zhou2025r1zerosahamomentvisual}
Hengguang Zhou, Xirui Li, Ruochen Wang, Minhao Cheng, Tianyi Zhou, and Cho-Jui Hsieh. 2025{\natexlab{a}}.
\newblock \href {https://arxiv.org/abs/2503.05132} {R1-zero's "aha moment" in visual reasoning on a 2b non-sft model}.

\bibitem[{Zhou et~al.(2025{\natexlab{b}})Zhou, Jiang, Tian, Weston, Levine, Sukhbaatar, and Li}]{zhou2025sweetrltrainingmultiturnllm}
Yifei Zhou, Song Jiang, Yuandong Tian, Jason Weston, Sergey Levine, Sainbayar Sukhbaatar, and Xian Li. 2025{\natexlab{b}}.
\newblock \href {https://arxiv.org/abs/2503.15478} {Sweet-rl: Training multi-turn llm agents on collaborative reasoning tasks}.

\bibitem[{Zhu et~al.(2024)Zhu, Guo, Shao, Yang, Wang, Xu, Wu, Li, Gao, Ma et~al.}]{zhu2024deepseek}
Qihao Zhu, Daya Guo, Zhihong Shao, Dejian Yang, Peiyi Wang, Runxin Xu, Y~Wu, Yukun Li, Huazuo Gao, Shirong Ma, et~al. 2024.
\newblock \href {https://arxiv.org/abs/2406.11931} {Deepseek-coder-v2: Breaking the barrier of closed-source models in code intelligence}.
\newblock \emph{ArXiv preprint}, abs/2406.11931.

\bibitem[{Zhuang et~al.(2023)Zhuang, Chen, Yu, Mitra, Bursztyn, Rossi, Sarkhel, and Zhang}]{zhuangToolChainEfficientAction2023}
Yuchen Zhuang, Xiang Chen, Tong Yu, Saayan Mitra, Victor Bursztyn, Ryan~A. Rossi, Somdeb Sarkhel, and Chao Zhang. 2023.
\newblock \href {https://arxiv.org/abs/2310.13227} {Toolchain*: Efficient action space navigation in large language models with a* search}.

\bibitem[{Zhuge et~al.(2024)Zhuge, Zhao, Ashley, Wang, Khizbullin, Xiong, Liu, Chang, Krishnamoorthi, Tian, Shi, Chandra, and Schmidhuber}]{zhuge2024agentasajudgeevaluateagentsagents}
Mingchen Zhuge, Changsheng Zhao, Dylan Ashley, Wenyi Wang, Dmitrii Khizbullin, Yunyang Xiong, Zechun Liu, Ernie Chang, Raghuraman Krishnamoorthi, Yuandong Tian, Yangyang Shi, Vikas Chandra, and Jürgen Schmidhuber. 2024.
\newblock \href {https://arxiv.org/abs/2410.10934} {Agent-as-a-judge: Evaluate agents with agents}.

\end{thebibliography}

\end{document}